\documentclass[10pt,journal,compsoc]{IEEEtran}

\ifCLASSOPTIONcompsoc
    \usepackage[nocompress]{cite}
\else
    \usepackage{cite}
\fi

\usepackage{hyperref}
\usepackage{lettrine}
\usepackage{graphicx}
\usepackage{caption}
\usepackage{subcaption}
\usepackage{booktabs}
\usepackage{multicol}
\usepackage{multirow}
\usepackage{mathtools}
\usepackage{makecell}
\usepackage{enumitem}
\usepackage{natbib}
\setcitestyle{numbers,square}
\usepackage{colortbl}
\usepackage{amssymb}
\usepackage{multirow}
\usepackage{diagbox}

\newcommand{\std}[1]{\scriptsize{$\pm$#1}}
\definecolor{mygray}{gray}{0.94}
\begin{document}

\title{\textsc{InstructLayout}: Instruction-Driven 2D and 3D Layout Synthesis with Semantic Graph Prior}

\author{
    Chenguo Lin*,
    Yuchen Lin*, Panwang Pan, Xuanyang Zhang, and
    Yadong Mu,
    \IEEEcompsocitemizethanks{
        \IEEEcompsocthanksitem *: equal contribution. C. Lin, Y. Lin and Y. Mu are with Wangxuan Institute of Computer Technology, Peking University.
        P. Pan and X. Zhang are with PICO AI group, ByteDance.\\
        E-mail: \{chenguolin, linyuchen\}@stu.pku.edu.cn, \{panpanwang, zhangxuanyang\}@bytedance.com, myd@pku.edu.cn.\\
        Corresponding author: Yadong Mu.
    }
}

\markboth{IEEE Transactions on Pattern Analysis and Machine Intelligence}
{Lin \MakeLowercase{\textit{et al.}} \title}

\IEEEtitleabstractindextext{
    \begin{abstract}
        Comprehending natural language instructions is a charming property for both 2D and 3D layout synthesis systems.
Existing methods implicitly model object joint distributions and express object relations, hindering generation's controllability.
We introduce \textsc{InstructLayout}, a novel generative framework that integrates a semantic graph prior and a layout decoder to improve controllability and fidelity for 2D and 3D layout synthesis.
The proposed semantic graph prior learns layout appearances and object distributions simultaneously, demonstrating versatility across various downstream tasks in a zero-shot manner.
To facilitate the benchmarking for text-driven 2D and 3D scene synthesis, we respectively curate two high-quality datasets of layout-instruction pairs from public Internet resources with large language and multimodal models.
Extensive experimental results reveal that the proposed method outperforms existing state-of-the-art approaches by a large margin in both 2D and 3D layout synthesis tasks.
Thorough ablation studies confirm the efficacy of crucial design components.

    \end{abstract}
    \begin{IEEEkeywords}
        layout synthesis, controllable generation, graph diffusion models, scene graphs.
    \end{IEEEkeywords}
}

\maketitle
\IEEEdisplaynontitleabstractindextext
\IEEEpeerreviewmaketitle

\section{Introduction}\label{sec:intro}

\lettrine[lines=2]{A}{utomatically} synthesizing controllable and appealing 2D and 3D layouts has been a challenging task for computer vision and graphics~\citep{qi2018human,ritchie2019fast,zhang2020deep,Gupta_2021_ICCV,Jyothi_2019_ICCV,yang2021indoor,hollein2023text2room,hsu2023posterlayout,chai2023layoutdm,lin2023autoposter,song2023roomdreamer,feng2023layoutgpt,patil2023advances,lin2024instructscene}, which requires building a generative model by multimodal conditional information.
An ideal layout synthesis system should fulfill at least three objectives:
(1) comprehending instructions in natural languages, thus providing an intuitive and user-friendly interface;
(2) designing object compositions that exhibit aesthetic appeal and thematic harmony;
(3) placing objects in appropriate positions adhering to their functions and regular arrangements.

To enhance the controllability of layout synthesis, we need to address natural language instructions as inputs.
However, natural instructions for layout design often rely on abstract object relationships, which has posed a significant challenge for recent advancements in 2D and 3D layout synthesis.
Previous works~\citep{wang2021sceneformer,paschalidou2021atiss,cao2022geometry,Inoue_2023_CVPR,liu2023clip,tang2024diffuscene} primarily exclude instructional prompts or do not prioritize them, leading to a low level of controllability and interpretability in the generated layouts.
Other studies~\citep{luo2020end,dhamo2021graph,zhai2023commonscenes} utilize relation graphs to provide explicit control over object interactions, which are, however, too complicated and fussy for human users to specify.
Moreover, previous works merely represent objects by categories
~\citep{luo2020end,Gupta_2021_ICCV,Inoue_2023_CVPR,paschalidou2021atiss} or low-dimensional features~\citep{wang2019planit,chai2023layoutdm,lin2023autoposter,tang2024diffuscene} which lack visual appearance details, resulting in style inconsistency and constraining customization options in layout synthesis.

\begin{figure*}[t]
    \begin{center}
        \includegraphics[width=\textwidth]{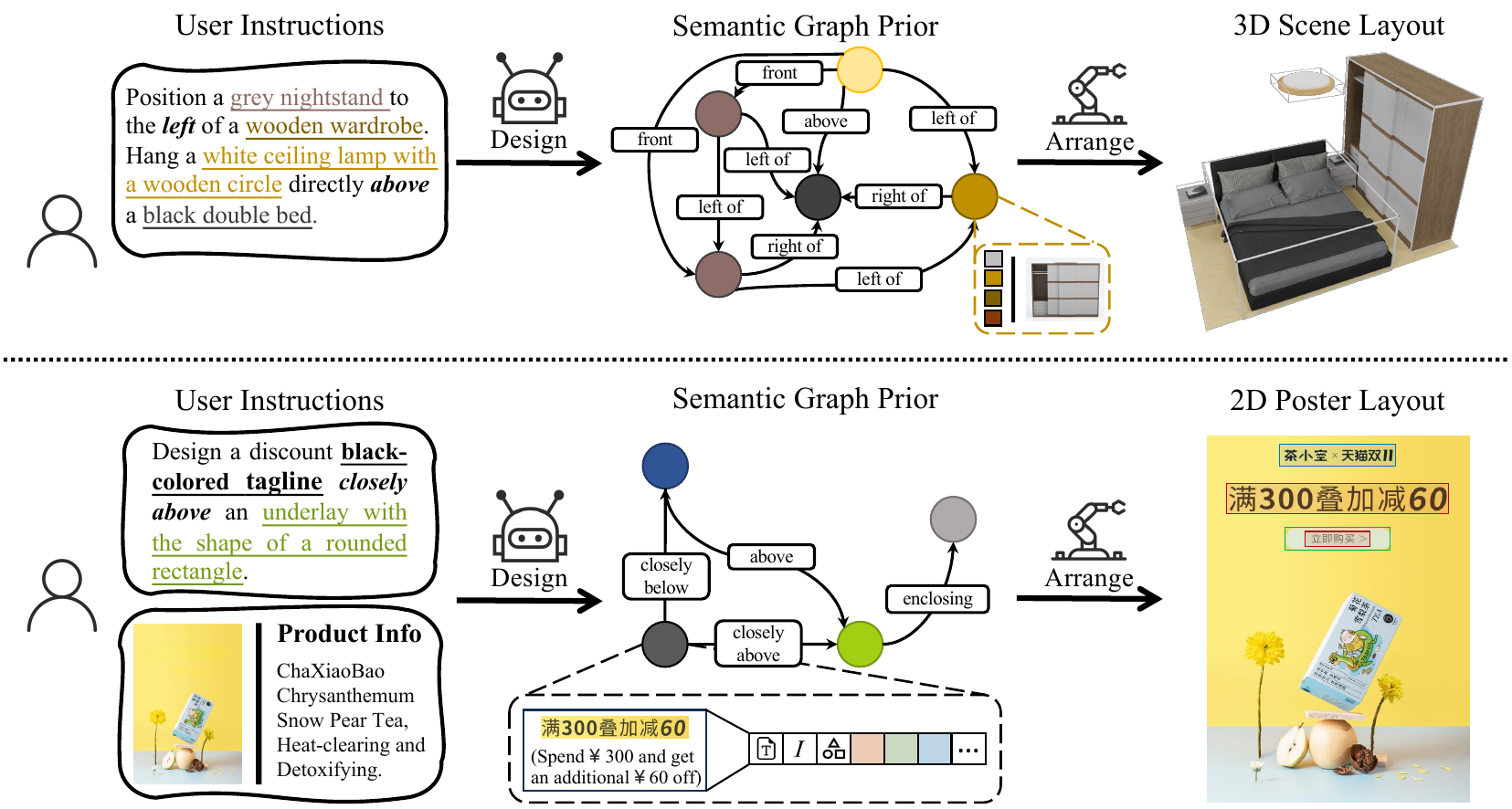}
    \end{center}
    \caption{\textbf{Method overview.} For both 3D scene and 2D poster layout synthesis tasks, (1) \textsc{InstructLayout} first designs a holistic semantic graph based on user instructions.
    Within the graph, each node is an object endowed with semantic features such as categories and appearances, and each edge represents a spatial relationship between objects. (2) It proceeds to arrange objects in a scene or canvas by decoding spatial attributes from the informative graph prior.}
    \label{fig:pipeline}
\end{figure*}

To address these issues, we present \textsc{InstructLayout}, 
a novel generative framework for both 2D and 3D layout synthesis with natural language instructions.
Figure~\ref{fig:pipeline} illustrates the overview of proposed method.
\textsc{InstructLayout} basically comprises two parts: a \textbf{semantic graph prior} and a \textbf{layout decoder}.
In the first stage, it takes instructions about partial object arrangement and attributes as conditions and learns the conditional distribution of semantic graphs for the holistic layout.
In the second stage, harnessing the well-structured and informative graph latents, the layout decoder can easily generate 2D poster and 3D scene layouts that exhibit semantic consistency while closely adhering to the provided instructions.
For 2D poster layout synthesis, we additionally provide a \textbf{tagline generator} to fill the bounding boxes with text content for real-world applications.
With the learned semantic graph prior, \textsc{InstructLayout} also achieves a wide range of instruction-driven generative tasks in a zero-shot manner.

Specific conditional diffusion models are devised for both parts of \textsc{InstructLayout}.
Benefitting from the two-stage scheme, it can separately handle discrete and continuous attributes of layouts, 
drastically reducing the burden of network optimization.
To enhance the capability of aesthetic design, 
\textsc{InstructLayout} utilizes the naturally discrete features of 2D objects, such as fonts, colors, and shapes. In contrast, in terms of 3D objects, it leverages object geometrics and appearances by quantizing semantic features from a multimodal-aligned model~\citep{liu2023openshape}. 
Consequently, our model can handle these discrete features in a unified manner.

To fit practical scenarios and promote the benchmarking of instruction-drive layout synthesis, we curate \textbf{two high-quality datasets} containing paired layouts and instructions with the help of large language and multimodal models~\citep{li2022blip,ouyang2022training,openai2023gpt4} for both 2D E-commerce poster and 3D indoor scene synthesis tasks.
Comprehensive quantitative evaluations reveal that \textsc{InstructLayout} surpasses previous state-of-the-art methods by a large margin in terms of both generation controllability and fidelity.
Each essential component of our method is carefully verified through ablation studies.

Our contributions can be summarized as follows:
\begin{itemize}[topsep=0pt]
    \item We present a unified instruction-driven generative framework for both 3D scene and 2D poster layout synthesis that integrates a semantic graph prior and a layout decoder to improve fidelity and controllability.
    \item The proposed semantic graph prior jointly models object attributes and layout distributions, facilitating various downstream applications in a zero-shot manner.
    \item We curate two high-quality datasets to promote the benchmarking of instruction-driven 2D and 3D layout synthesis, and quantitative experiments demonstrate that the proposed method significantly outperforms existing state-of-the-art techniques.
\end{itemize}

\section{Related Work}\label{sec:related}

\subsection{3D Scene Layout Synthesis}
Graphs have been used to guide complex scene synthesis in the form of scene hierarchies~\citep{li2019grains,gao2023scenehgn}, parse trees~\citep{purkait2020sg}, scene graphs~\citep{zhou2019scenegraphnet,para2021generative}, etc.
\citet{wang2019planit} utilize an image-based module and condition its outputs on the edges of a relation graph within each non-differentiable step.
They also adopt an autoregressive model~\citep{li2018learning} to generate relation graphs, which are unconditional and have limited object attributes.
Other works~\citep{luo2020end,dhamo2021graph,zhai2023commonscenes} adopt conditional VAEs~\citep{kingma2013auto,sohn2015learning} with graph convolutional networks~\citep{johnson2018image} to generate layouts.
While offering high controllability, these methods demand the specification of elaborate graph conditions, which are notably more intricate than those driven by natural languages.
With the advent of attention mechanisms~\citep{vaswani2017attention}, recent approaches~\citep{wang2021sceneformer,paschalidou2021atiss,liu2023clip,tang2024diffuscene} can implicitly acquire object relations by self-attention and condition scene synthesis with texts by cross-attention.
However, text prompts in these works tend to be relatively simple, containing only object categories or lacking layout descriptions, limiting the expressiveness and customization.
Implicit relation modeling also significantly hinders their controllability.

\subsection{2D Poster Layout Synthesis}
Numerous methods have been proposed to automatically synthesize 2D layouts for posters, user interfaces, E-commerce advertisements, etc.
Prior research~\citep{Jyothi_2019_ICCV, Gupta_2021_ICCV, Kikuchi_2021, chai2023layoutdm, Inoue_2023_CVPR} has primarily concentrated on generating layouts from scratch on a blank canvas, while others~\citep{cao2022geometry, zhou2022composition, xu2023unsupervised, hsu2023posterlayout, lin2023autoposter} focus on image content-aware layout synthesis, leading to more practical applications.
However, they are either confined to unconditional generation based solely on the canvas(blank or image prior)~\citep{Jyothi_2019_ICCV, Gupta_2021_ICCV, zhou2022composition, xu2023unsupervised} or relying on object categories~\citep{chai2023layoutdm, Inoue_2023_CVPR}, lacking the capability to control the layout synthesis process using user-friendly instructions.
Furthermore, due to the scarcity of high-quality datasets, 
only a few studies \citep{lin2023autoposter, gao2022caponimage} have delved into predicting 2D object attributes and their textual content, which is vital for generating realistic and informative layouts.

\subsection{Generative Models for Graphs}
There have been lots of endeavors on generative models for undirected graphs, molecules and scene graphs by autoregressive models~\citep{you2018graphrnn,garg2021unconditional}, VAEs~\citep{simonovsky2018graphvae,verma2022varscene}, GANs~\citep{de2018molgan,martinkus2022spectre} and diffusion models~\citep{niu2020permutation,jo2022score,vignac2022digress,kong2023autoregressive}.
However, none of them is capable of understanding text conditions.
\citet{longland2022text} and \citet{lo2023lic} employ VAE and GAN, respectively, for text-driven generation of simple undirected graphs without semantics.
We present a pioneering effort to generate holistic semantic graphs with expressive instructions.

\section{Method}\label{sec:method}

\subsection{Problem Statement}
Denote $\mathcal{S}\coloneqq\{\mathcal{S}_1,\dots,\mathcal{S}_M\}$ as a collection of 2D or 3D layouts.
Each layout $\mathcal{S}_i$ is composed of multiple objects $\mathcal{O}_i\coloneqq\{\mathbf{o}_j^i\}_{j=1}^{N_i}$ with distinct attributes $\mathbf{o}_j^i\coloneqq\{c_j^i,\mathbf{s}_j^i,\mathbf{f}_j^i\}$, including category $c_j^i\in\{1,...,K_c\}$, where $K_c$ is the number of object classes in $\mathcal{S}$, spatial feature $\mathbf{s}_j^i$, such as scale and location, and semantic feature $\mathbf{f}_j^i$ like appearance and style.
To set up a layout, one can generate each object (a 3D model or tagline text) or retrieve it from a database based on $c$ and $\mathbf{f}$.
These objects are then resized and transformed to the expected coordinates by corresponding $\mathbf{s}$.

Given instructions $\mathbf{y}$, our goal is to learn the conditional scene distribution $q(\mathcal{S}|\mathbf{y})$.
Rather than direct modeling~\citep{paschalidou2021atiss,Inoue_2023_CVPR,lin2023autoposter,tang2024diffuscene}, we employ well-structured and informative graphs to serve as general and semantic latent.
Each graph $\mathcal{G}_i$ contains a node set $\mathcal{V}_i\coloneqq\{\mathbf{v}_j^i\}_{j=1}^{N_i}$ and a directed edge set $\mathcal{E}_i\coloneqq\{e_{jk}^i|\mathbf{v}_j^i,\mathbf{v}_k^i\in\mathcal{V}_i\}$.
A node $\mathbf{v}_j^i$ functions as a high-level representation of an object $\mathbf{o}_j^i$, and a directed edge $e_{jk}^i$ explicitly conveys the relations between objects.

To this end, we propose a generative framework, \textsc{InstructLayout}, that consists of two components:
(1) \textbf{semantic graph prior} $p_\phi(\mathcal{G}|\mathbf{y})$ (Sec.~\ref{subsec:graphgen}) that jointly models high-level object and relation distributions conditioned on $\mathbf{y}$;
(2) \textbf{layout decoder} $p_\theta(\mathcal{S}|\mathcal{G})$ (Sec.~\ref{subsec:layoutgen}) that produces precise layout configurations with semantic graphs prior.
Since $\mathcal{G}$ is deterministic by corresponding $\mathcal{S}$, the two networks together yield an instruction-driven generative model for 2D and 3D layouts:
\begin{equation}
    p_{\phi,\theta}(\mathcal{S}|\mathbf{y})=p_{\phi,\theta}(\mathcal{S},\mathcal{G}|\mathbf{y})=p_\phi(\mathcal{G}|\mathbf{y})p_\theta(\mathcal{S}|\mathcal{G}).
\end{equation}

\subsection{Preliminary: General Diffusion Models}\label{subsec:preliminary}
Diffusion generative models~\citep{sohl2015deep} consist of a non-parametric forward process and a learnable reverse process.
The forward process progressively corrupts a data point from $q(\mathbf{x}_0)$ to a sequence of increasingly noisy latent variables: $q(\mathbf{x}_{1:T}|\mathbf{x}_0)=\prod_{t=1}^{T}q(\mathbf{x}_t|\mathbf{x}_{t-1})$.
A neural network is trained to reverse the process by denoising them iteratively: $p_\psi(\mathbf{x}_{0:T}|\mathbf{c})= p(\mathbf{x}_T)\prod_{t=1}^{T}p_\psi(\mathbf{x}_{t-1}|\mathbf{x}_t,\mathbf{c})$, where $\mathbf{c}$ is an optional condition to guide the reverse process as needed.
These two processes are supposed to admit $p(\mathbf{x}_T)\approx q(\mathbf{x}_T|\mathbf{x}_0)$ for a sufficiently large $T$.
The generative model is optimized by minimizing a variational upper bound on $\mathbb{E}_{q(\mathbf{x}_0)}\left[-\log p_\psi(\mathbf{x}_0)\right]$:

\begin{multline}\label{equ:vb}
    \mathcal{L}_{\text{vb}}\coloneqq \mathbb{E}_{q(\mathbf{x}_0)}[\ \ 
    \underbrace{
        D_{\text{KL}}[
        q(\mathbf{x}_T|\mathbf{x}_0)\|p(\mathbf{x}_T)
        ]
    }_{\mathcal{L}_{T}}\\
    + \sum_{t=2}^{T}\mathcal{L}_{t-1}
    \underbrace{
        - \mathbb{E}_{q(\mathbf{x}_1|\mathbf{x}_0)}[\log p_\psi(\mathbf{x}_0|\mathbf{x}_1,\mathbf{c})]
    }_{\mathcal{L}_{0}}
    \ \ ],
\end{multline}
where $\mathcal{L}_{t-1}\coloneqq D_{\text{KL}}[q(\mathbf{x}_{t-1}|\mathbf{x}_t,\mathbf{x}_0)\|p_\psi(\mathbf{x}_{t-1}|\mathbf{x}_t,\mathbf{c})]$ and $\mathcal{L}_T$ is constant during training so can be ignored.
$D_{\text{KL}}[\cdot\|\cdot]$ indicates the KL divergence between two distributions.

Diffusion models can accommodate both continuous and categorical distributions by a meticulous design of the diffusion kernel $q(\mathbf{x}_t|\mathbf{x}_{t-1})$.
A prominent choice for continuous variables is the Gaussian kernel~\citep{ho2020denoising,song2020score,tang2024diffuscene}.
Uniform transition is widely adopted for categorical variables~\citep{hoogeboom2021argmax,austin2021structured,vignac2022digress}.
In this work, we utilize both the discrete and continuous diffusion models for the two stages in \textsc{InstructLayout} respectively to model distinct layout attributes.

\subsection{Semantic Graph Prior}\label{subsec:graphgen}
A semantic graph $\mathcal{G}$ represents an abstract structure of the underlying layout by a node set and an edge set.
Each node includes the category $c$ and semantic features $\mathbf{f}$ of an object.
Spatial features $\mathbf{s}$ can be derived from relations represented by edges, so we leave them to the decoder $p_\theta(\mathcal{S}|\mathcal{G})$.

\begin{figure*}[t]
    \begin{center}
        \includegraphics[width=\textwidth]{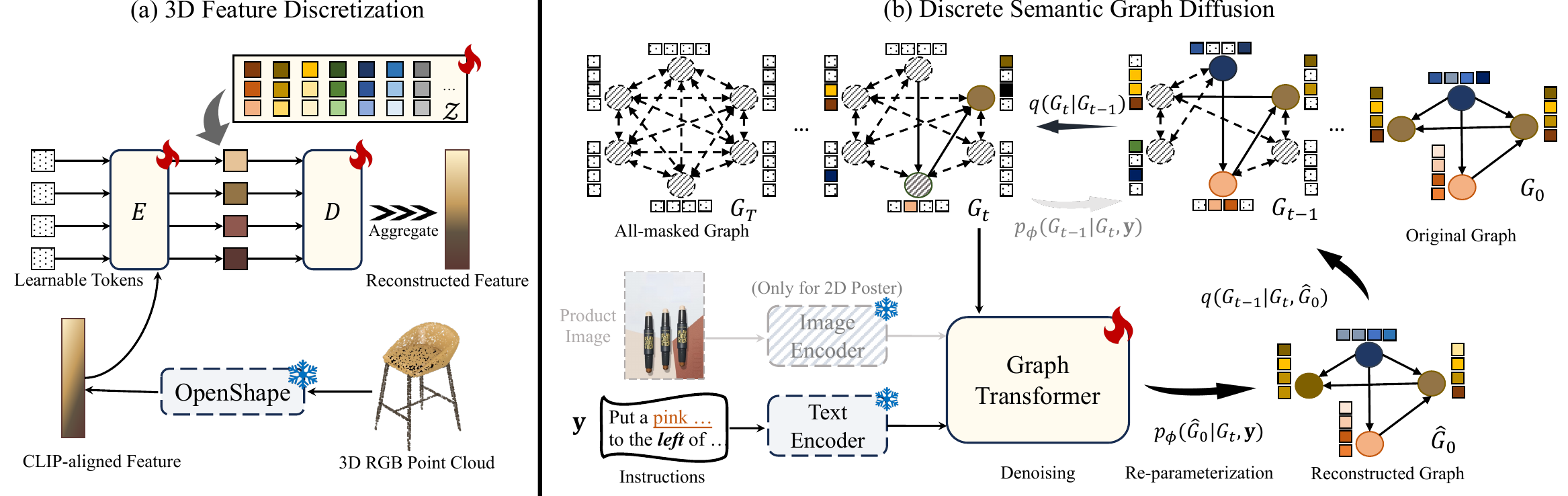}
    \end{center}
    \caption{\textbf{Semantic Graph Prior} (Sec.~\ref{subsec:graphgen}). (a) \textbf{3D Feature Discretization} (Sec.~\ref{subsubsec:fvqvae}). Unlike the naturally discrete features of 2D poster graphics, semantic features for 3D objects are extracted from a frozen multimodal-aligned model OpenShape and then quantized by codebook entries. (b) \textbf{Discrete Semantic Graph Diffusion} (Sec.~\ref{subsubsec:ddiff}). Three categorical variables in $G_0$ are independently diffused; Empty states are not depicted for concision; A graph Transformer with a frozen text encoder and optional image encoder for 2D poster learns the semantic graph prior by iteratively denoising corrupted graphs.}
    \label{fig:vq+prior}
\end{figure*}

\subsubsection{Feature Discretization}\label{subsubsec:fvqvae}
For the 2D poster layout synthesis task, we consider graphic features, including types of text fonts, italicized or not, and border shapes, which are naturally discrete variables like category.
For the colors of text, border, and gradation that are in continuous values, we discretize them in the CIE \textit{Lab} colorspace following \citet{zhang2016colorful} to enhance color saturation and constrain the feature space.

While for the 3D scene layout synthesis task, it's hard to describe geometric shapes and visual appearance of 3D models by explicit categorical attributes, so we utilize a pretrained multimodal-aligned model OpenShape~\citep{liu2023openshape} to extract their semantic features.
However, these high-dimensional features (e.g., $d=1280$ in OpenShape \texttt{pointbert-vitg14}) are too complicated to model.
We circumvent this drawback by introducing a vector-quantized variational autoencoder (VQ-VAE)~\citep{jang2016categorical,ramesh2021zero} for feature vectors and quantize each of them by several learnable tokens from the maintained codebook, as depicted in Figure~\ref{fig:vq+prior}(a).
The intuition behind it is that 3D models share general intrinsic characteristics, encompassing attributes like colors, materials, and basic geometric shapes.
Indexing semantic features in a discrete space achieves a compact feature space for distribution modeling and dramatically reduces the cost of operating in a continuous space.

\subsubsection{Discrete Semantic Graph Diffusion}\label{subsubsec:ddiff}

After feature discretization, all attributes in a semantic graph $\mathcal{G}_i \coloneqq (\mathcal{C}_i, \mathcal{F}_i, \mathcal{E}_i)$ are categorical, where $\mathcal{C}_i \in \{1,\dots,K_c\}^{N_i}$, $\mathcal{F}_i \in \{1,\dots,K_f\}^{N_i \times n_f}$, and $\mathcal{E}_i \in \{1,\dots,K_e\}^{N_i \times N_i}$, where $n_f$ is the number of discrete features.
While continuous embeddings such as one-hot vectors are commonly used, they often obscure the inherent sparsity of categorical data and complicate optimization.
We instead model the semantic graph prior using a \textbf{mask-based discrete diffusion process}.

For a categorical variable $x \in \{1,\dots,K\}$, diffusion noise is defined by transition matrices $\mathbf{Q}t \in \mathbb{R}^{K \times K}$ over timesteps $t$.
The forward process is given by $q(x_t | x_{t-1}) \coloneqq \mathbf{x}_t^\top \mathbf{Q}_t \mathbf{x}_{t-1}$, where $\mathbf{x}_t$ is the column one-hot encoding of $x_t$ and $[\mathbf{Q}_t]_{mn}\coloneqq q(x_t=m|x_{t-1}=n)$ is the probability that $x_{t-1}$ transits to the category $m$ from $n$.
The probabilistic distribution of $x_t$ can be directly derived from $x_0$:
$q(x_t|x_0)\coloneqq\mathbf{x}_t^\top\bar{\mathbf{Q}}_t\mathbf{x}_0$, where $\bar{\mathbf{Q}}_t\coloneqq \mathbf{Q}_t\cdots \mathbf{Q}_1$.

Instead of relying on commonly used Gaussian or uniform transitions for graph generation~\citep{niu2020permutation,hoogeboom2021argmax,jo2022score,vignac2022digress}, we adopt a diffusion process tailored for semantic graphs by independently masking graph attributes (i.e., object class $c$, discrete feature indices $f$, and relation $e$) by introducing an absorbing state, \texttt{[MASK]}~\citep{austin2021structured,gu2022vector}, into each uniform transition matrix.
For the object class $c$, its transition matrix is defined as:
\begin{equation}
    \mathbf{Q}_t^{\mathbf{C}}\coloneqq\left[
    \begin{array}{ccccc}
        \alpha_t^c+\beta_t^c & \beta_t^c & \cdots & \beta_t^c & 0 \\
        \beta_t^c & \alpha_t^c+\beta_t^c & \cdots & \beta_t^c & 0 \\
        \vdots & \vdots & \ddots & \beta_t^c & 0 \\
        \beta_t^c & \beta_t^c & \beta_t^c & \alpha_t^c+\beta_t^c & 0 \\
        \gamma_t^c & \gamma_t^c & \gamma_t^c & \gamma_t^c & 1
    \end{array}
    \right],
\end{equation}
by which $c_t$ has a probability of $\gamma_t^c$ to be masked, a probability of $\alpha_t^c$ to remain unchanged, and a remaining probability of $1 - \gamma_t^c - \alpha_t^c$ to be uniformly resampled from the label space.
The absorbing state \texttt{[MASK]} remains fixed once reached.
Transition matrices for $f$ and $e$, denoted by $\mathbf{Q}_t^{\mathbf{F}}$ and $\mathbf{Q}_t^{\mathbf{E}}$ respectively, follow the same design.
The schedules of $(\alpha_t, \beta_t, \gamma_t)$ are crafted such that semantic graphs start fully masked at the beginning of the diffusion process.

Since the number of objects varies across different layouts, semantic graphs are padded by empty states to maintain a consistent number of $N$ objects.
One-hot encodings for scalar variables $c$, $f$ and $e$ in a layout are denoted as $\mathbf{C}\in\mathbb{R}^{N\times (K_c+2)}$, $\mathbf{F}\in\mathbb{R}^{N\times n_f\times (K_f+2)}$ and $\mathbf{E}\in\mathbb{R}^{N\times N\times (K_e+2)}$ respectively.
Here, ``$+2$'' accounts for each variable's two extra states (i.e., empty state and mask state).
A one-hot encoded semantic graph $G_0\coloneqq(\mathbf{C}_0,\mathbf{F}_0,\mathbf{E}_0)$ at timestep $t$ is formulated as
\begin{equation}
    q(G_t|G_0)=(\bar{\mathbf{Q}}_t^{\mathbf{C}}\mathbf{C}_0,\bar{\mathbf{Q}}_t^{\mathbf{F}}\mathbf{F}_0,\bar{\mathbf{Q}}_t^{\mathbf{E}}\mathbf{E}_0).
\end{equation}
The process for learning the graph prior is illustrated in Figure~\ref{fig:vq+prior}(b).
The independent diffusion with mask states offers two significant advantages:
\begin{itemize}
    \item Perturbed states for one variable (e.g., $\mathbf{C}$) could be recovered by incorporating information from uncorrupted portions of the other variables (e.g., $\mathbf{F}$ and $\mathbf{E}$), compelling the semantic graph prior to learning from the interactions among different scene attributes.
    \item The introduction of mask states facilitates the distinction between corrupted and clean states, thus simplifying the denoising task.
\end{itemize}
These benefits are critical, especially for intricate semantic graphs and diverse downstream generative tasks, compared with simple graph generative tasks~\citep{niu2020permutation,jo2022score,vignac2022digress}.
An ablation study on the choice of $\mathbf{Q}$ is provided in Sec.~\ref{subsec:ablation}.

The output of the graph prior network is re-parameterized to produce the clean scene graphs $\hat{G}_0$, which is then diffused to get the predicted posterior for computing the variational bound in Equation~\ref{equ:vb}: $p_\phi(G_{t-1}|G_t,\mathbf{y})\propto\sum_{\hat{G}_0} q(G_{t-1}|G_{t},\hat{G}_0)p_\phi(\hat{G}_0|G_t,\mathbf{y})$.
Training objective for $p_\phi$ is a weighted summation of variational bounds for three random variables conditioned on $\mathbf{y}$:
\begin{equation}\label{equ:graph_prior_loss}
    \mathcal{L}_{\text{vb}}^{\mathcal{G}|\mathbf{y}}\coloneqq
    \mathcal{L}_{\text{vb}}^{\mathbf{C}|\mathbf{y}} + \lambda_f\cdot\mathcal{L}_{\text{vb}}^{\mathbf{F}|\mathbf{y}} + \lambda_e\cdot\mathcal{L}_{\text{vb}}^{\mathbf{E}|\mathbf{y}},
\end{equation}
where $\lambda_e$ and $\lambda_f$ are hyperparameters to adjust the relative importance of three components in the semantic graph.

\subsection{Layout Decoder}\label{subsec:layoutgen}

\begin{figure*}[t]
    \begin{center}
        \includegraphics[width=\textwidth]{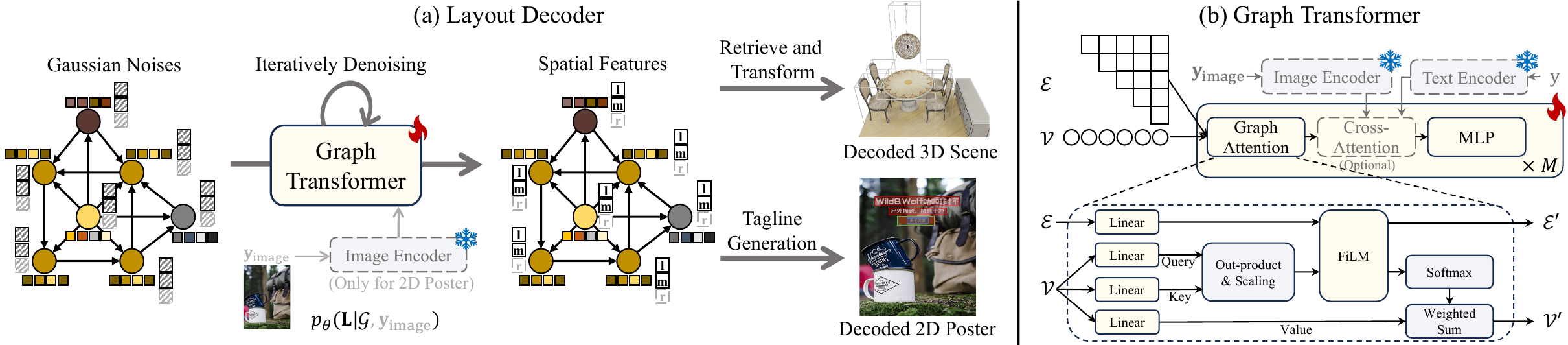}
    \end{center}
    \caption{(a) \textbf{Layout Decoder} (Sec.~\ref{subsec:layoutgen}). Gaussian noises are sampled to attach at every node of semantic graphs; A graph Transformer processes these graphs iteratively to remove noises and generate spatial features (Sec.~\ref{subsubsec:decode}); A \textbf{tagline generator} followed to fill textual contents in each bounding box for 2D poster layout synthesis (Sec.~\ref{subsubsec:tagline}).
    (b) \textbf{Graph Transformer} (Sec.~\ref{subsec:model}). A graph Transformer consists of a stack of $M$ blocks, each comprising graph attention, MLP, and optional cross-attention modules for conditions; AdaLN and multihead scheme are not depicted for concision.}
    \label{fig:decoder+gtf}
\end{figure*}

\subsubsection{Spatial Feature Generation}\label{subsubsec:decode}
Modeling layouts' spatial features become easy with the semantic graph prior. 
Spatial features basically include two attributes $\mathbf{s}\coloneqq\{\mathbf{l},\mathbf{m}\}$, where $\mathbf{l}\in\mathbb{R}^2$ or $\mathbb{R}^3$ is the location of an object in 2D or 3D space and $\mathbf{m}\in\mathbb{R}^2$ or $\mathbb{R}^3$ is its scale.
In 3D scenes, an object has an orientation attribute $r\in\mathbb{R}$, allowing it to rotate along the gravity axis.
We parameterize the $SO(2)$ rotation by $[\cos(r),\sin(r)]^\top$ to continuously represent $r$~\citep{zhou2019continuity}.
Consequently, the spatial features of each layout $\mathcal{S}_i$ can be expressed as matrices $\mathbf{L}_i\in\mathbb{R}^{N_i\times d_l}$, where $d_l=4$ for 2D posters and $d_l=8$ for 3D scenes.
Note that $\mathcal{S}=(\mathbf{L},\mathcal{G})$, so generating layouts $p_\theta(\mathcal{S}|\mathcal{G})$ is equivalent to learning the conditional distributions of layout configurations $p_\theta(\mathbf{L}|\mathcal{G})$.
For the 2D layout synthesis task, although spatial relations can be inferred from edges in the semantic graph, we additionally provide clean product poster images $\mathbf{y}_\text{image}$ as conditions for more precise arrangements and to avoid occluding the principal products.

A continuous diffusion model with variance-preserving Gaussian kernels~\citep{ho2020denoising,song2020score} is adopted to learn $p_\theta(\mathbf{L}|\mathcal{G})$.
Following \citet{ho2020denoising}, the variational bound in Equation~\ref{equ:vb} for the decoder $p_\theta(\mathbf{L}|\mathcal{G})$ is reweighted and simplified:
\begin{equation}
    \mathcal{L}_{\text{simple}}\coloneqq
    \mathbb{E}_{\mathbf{L}_0,t,\boldsymbol{\epsilon}} \|\boldsymbol{\epsilon} -
    \boldsymbol{\epsilon}_\theta\left(\mathbf{L}_t,t,\mathcal{G}\right)\|^2,
\end{equation}
where $t$ is sampled from a uniform distribution $\mathcal{U}(1,T)$, $\boldsymbol{\epsilon}$ is sampled from a standard normal distribution $\mathcal{N}(\mathbf{0},\mathbf{I})$ and $\mathbf{L}_t$ is the noised spatial features.
A diagram of the layout decoder is depicted in Figure~\ref{fig:decoder+gtf}(a).
Intuitively, the network $\boldsymbol{\epsilon}_\theta$ is trained to predict noise $\boldsymbol{\epsilon}$ in $\mathbf{L}_t$.

\vspace{-0.5cm}
\subsubsection{Tagline Generation for 2D Poster}\label{subsubsec:tagline}
After synthesizing semantic features $\mathbf{f}$ and spatial layouts $\mathbf{s}$, the pipeline for 3D scene synthesis is completed by retrieving 3D objects from an off-the-shelf dataset based on the generated $\mathbf{f}$ and transforming them to appropriate positions using the generated $\mathbf{s}$.

For 2D poster synthesis, we further generate textual taglines for each bounding box using all generated $\mathbf{f}$, $\mathbf{s}$ and user-provided product descriptions and images.
We adopt the multimodal captioning framework proposed by \citet{gao2022caponimage}, which uses an autoregressive Transformer decoder conditioned on background images, product descriptions and spatial layout.
The model is trained progressively via caption generation and matching tasks.

However, the absence of clean product images and semantic features of text bounding boxes results in suboptimal tagline generation.
To address this issue, we utilize high-quality cleaned product images processed according to the strategy proposed in Sec.~\ref{subsec:dataset_2d} and incorporate the generated semantic features from Sec~\ref{subsec:graphgen} as an additional conditioning.
This results in more coherent and product-relevant tagline generation.

\subsection{Model Architecture}\label{subsec:model}
The general-purpose Transformer~\citep{vaswani2017attention} architecture is used for all models across tasks in this work.

\noindent
\textbf{Vanilla Transformer.}
As illustrated in Figure~\ref{fig:vq+prior}(a), $n_f$ learnable tokens are employed with a stack of cross-attentions to extract information from object semantic features $\mathbf{f}$ in the encoder $E$ in VQ-VAE.
Regarding the decoder $D$, $n_f$ vectors retrieved from the codebook $\mathcal{Z}$ are fed to another Transformer, and an average pooling on the top of it is applied to aggregate information.

\noindent
\textbf{Graph Transformer.}
The semantic graph prior and spatial decoder share the same model architecture as shown in Figure~\ref{fig:decoder+gtf}(b).
Since relation $e_{jk}$ can be determined by $e_{kj}$, only the upper triangular part of the relation matrix is necessary.
Object categories and features together form input tokens for Transformers.
Message passing on graphs is operated via node self-attention and node-edge fusion with FiLM~\citep{perez2018film}, which linearly modulates edge embeddings and node attention matrices before softmax~\citep{dwivedi2021generalization,vignac2022digress}.
Timestep for diffusion $t$ is injected by AdaLN~\citep{ba2016layer,dhariwal2021diffusion}.
In the graph prior $p_\phi(\mathcal{G}|\mathbf{y})$, instructions and optional product images are embedded by a frozen multimodal encoder CLIP~\citep{radford2021learning} and consistently influence network outputs by cross-attention mechanisms.
Layout decoder $p_\theta(\mathcal{S}|\mathcal{G})$ is conditioned on semantic graphs by appending sampled Gaussian noises on node embeddings, which are then iteratively denoised to produce layout spatial features.

\noindent
\textbf{Permutation Non-invariance.}
Although $\mathcal{G}$ should ideally remain invariant to node permutations, invariant diffusion models could encounter learning challenges for multi-mode modeling.
Thus, each node feature is added with positional encodings~\citep{vaswani2017attention,lei2023nap,tang2024diffuscene} before the permutation-equivariant Transformer.
Exchangeability for graph prior distributions is strived by random permutation augmentation during the training process.
Ablation on the permutation non-invariance is provided in Sec.~\ref{subsec:ablation}.

\section{\textsc{InstructLayout} for 3D Scene Synthesis}\label{sec:exp3d}

\begin{table*}[th]
    \centering
    \caption{Quantitive evaluations for instruction-driven synthesis 3D indoor scene by ATISS~\citep{paschalidou2021atiss}, DiffuScene~\citep{tang2024diffuscene} and our method on three room types. Higher iRecall, lower FID, FID$^\text{CLIP}$, and KID indicate better synthesis quality. For SCA, a score closer to 50\% is better. Standard deviation values are provided as subscripts.}
    \label{tab:scenesyn}
    \renewcommand\arraystretch{1.2}
    \resizebox{0.85\textwidth}{!}{
        \begin{tabular}{ll|c|cccc}
            \toprule[1.2pt]

            \multicolumn{2}{c|}{3D Instruction-driven Synthesis} &
            $\uparrow$ iRecall$_{\%}$ & $\downarrow$ FID & $\downarrow$ FID$^{\text{CLIP}}$ & $\downarrow$ KID$_{\times\text{1e-3}}$ & SCA$_{\%}$ \\

            \midrule[1.2pt]

            \multirow{3}{*}{Bedroom} &
            ATISS &
            48.13\std{2.50} & 119.73\std{1.55} & 6.95\std{0.06} & 0.39\std{0.02} & 59.17\std{1.39} \\
            & DiffuScene &
            56.43\std{2.07} & 123.09\std{0.79} & 7.13\std{0.16} & 0.39\std{0.01} & 60.49\std{2.96} \\
            & \cellcolor{mygray}Ours &
            \cellcolor{mygray}\textbf{73.64}\std{1.37} & \cellcolor{mygray}\textbf{114.78}\std{1.19} & \cellcolor{mygray}\textbf{6.65}\std{0.18} & \cellcolor{mygray}\textbf{0.32}\std{0.03} & \cellcolor{mygray}\textbf{56.02}\std{1.43} \\
            \hline
            \multirow{3}{*}{Living room} &
            ATISS &
            29.50\std{3.67} & 117.67\std{2.32} & 6.08\std{0.13} & 17.60\std{2.65} & 69.38\std{3.38} \\
            & DiffuScene &
            31.15\std{2.49} & 122.20\std{1.09} & 6.10\std{0.11} & 16.49\std{1.24} & 72.92\std{1.29} \\
            & \cellcolor{mygray}Ours &
            \cellcolor{mygray}\textbf{56.81}\std{2.85} & \cellcolor{mygray}\textbf{110.39}\std{0.78} & \cellcolor{mygray}\textbf{5.37}\std{0.07} & \cellcolor{mygray}\textbf{8.16}\std{0.56} & \cellcolor{mygray}\textbf{65.42}\std{2.52} \\
            \hline
            \multirow{3}{*}{Dining room} &
            ATISS &
            37.58\std{1.99} & 137.10\std{0.34} & 8.49\std{0.23} & 23.60\std{2.52} & 67.61\std{3.23} \\
            & DiffuScene &
            37.87\std{2.76} & 145.48\std{1.36} & 8.63\std{0.31} & 24.08\std{1.90} & 70.57\std{2.14} \\
            & \cellcolor{mygray}Ours &
            \cellcolor{mygray}\textbf{61.23}\std{1.67} & \cellcolor{mygray}\textbf{129.76}\std{1.61} & \cellcolor{mygray}\textbf{7.67}\std{0.18} & \cellcolor{mygray}\textbf{13.24}\std{1.79} & \cellcolor{mygray}\textbf{64.20}\std{1.90} \\

            \bottomrule[1.2pt]
        \end{tabular}
    }
\end{table*}

\begin{table*}[th]
    \centering
    \caption{Quantitive evaluations for zero-shot generative applications on three room types. ``Uncond.'' stands for unconditional scene synthesis.}
    \label{tab:app_3d}
    \renewcommand\arraystretch{1.2}
    \resizebox{0.85\textwidth}{!}{
        \begin{tabular}{ll|cc|cc|cc|cc|c}
            \toprule[1.2pt]

            \multicolumn{2}{c|}{\multirow{2}{*}{\makecell[c]{Zero-shot\\Applications}}} &
            \multicolumn{2}{c|}{Stylization} &
            \multicolumn{2}{c|}{Re-arrangement} &
            \multicolumn{2}{c|}{Completion} &
            \multicolumn{1}{c}{Uncond.} \\

            \cmidrule(lr){3-4} \cmidrule(lr){5-6} \cmidrule(lr){7-8} \cmidrule(lr){9-9} &

            &
            $\uparrow$ $\Delta_{\times 1e-3}$ &
            $\downarrow$ FID &
            $\uparrow$ iRecall$_{\%}$ &
            $\downarrow$ FID &
            $\uparrow$ iRecall$_{\%}$ &
            $\downarrow$ FID &
            $\downarrow$ FID \\

            \midrule[1.2pt]

            \multirow{3}{*}{Bedroom} &
            \footnotesize{ATISS} &
            3.44 & 123.91 &
            61.22 & 107.67 &
            64.90 & 89.77 &
            134.51 \\
            & \footnotesize{DiffuScene} &
            1.08 & 127.35 &
            68.57 & 106.15 &
            48.57 & 96.28 &
            135.46 \\
            & \cellcolor{mygray}Ours &
            \cellcolor{mygray}\textbf{6.34} & \cellcolor{mygray}\textbf{122.73} &
            \cellcolor{mygray}\textbf{79.59} & \cellcolor{mygray}\textbf{105.27} &
            \cellcolor{mygray}\textbf{69.80} & \cellcolor{mygray}\textbf{82.98} &
            \cellcolor{mygray}\textbf{124.97} \\
            \hline
            \multirow{3}{*}{Living room} &
            \footnotesize{ATISS} &
            -3.57 & 110.85 &
            31.97 & 117.97 &
            43.20 & 106.48 &
            129.23 \\
            & \footnotesize{DiffuScene} &
            -2.69 & 112.80 &
            41.50 & 115.30 &
            19.73 & 95.94 &
            129.75 \\
            & \cellcolor{mygray}Ours &
            \cellcolor{mygray}\textbf{0.28} & \cellcolor{mygray}\textbf{109.39} &
            \cellcolor{mygray}\textbf{56.12} & \cellcolor{mygray}\textbf{106.85} &
            \cellcolor{mygray}\textbf{46.94} & \cellcolor{mygray}\textbf{92.52} &
            \cellcolor{mygray}\textbf{117.62} \\
            \hline
            \multirow{3}{*}{Dining room} &
            \footnotesize{ATISS} &
            -1.11 & 131.14 &
            36.06 & 134.54 &
            57.99 & 122.44 &
            147.52 \\
            & \footnotesize{DiffuScene} &
            -2.98 & 135.20 &
            46.84 & 133.73 &
            32.34 & 115.08 &
            150.81 \\
            & \cellcolor{mygray}Ours &
            \cellcolor{mygray}\textbf{1.69} & \cellcolor{mygray}\textbf{128.78} &
            \cellcolor{mygray}\textbf{62.08} & \cellcolor{mygray}\textbf{125.07} &
            \cellcolor{mygray}\textbf{60.59} & \cellcolor{mygray}\textbf{107.86} &
            \cellcolor{mygray}\textbf{137.52} \\

            \bottomrule[1.2pt]
        \end{tabular}
    }
\end{table*}

\subsection{3D Indoor Scene-Instruction Paired Dataset}\label{subsec:dataset_3d}
All experiments are conducted on 3D-FRONT~\citep{fu20213d}, a professionally designed collection of synthetic indoor scenes.
However, it does not contain any descriptions of room layouts or object appearances.
To construct a high-quality scene-instruction paired dataset, we initially extract view-dependent spatial relations with predefined rules~\citep{johnson2018image,luo2020end}.
The dataset is further enhanced by captioning objects with BLIP~\citep{li2022blip}.
To ensure the accuracy of descriptions, the generated captions are filtered by ChatGPT~\citep{ouyang2022training,openai2023gpt4} with object ground-truth categories.
The final instructions are derived from randomly selected relation triplets.
Verbs and conjunctions in the sentences are also randomly picked to keep the diversity and fluency of prompts.
Details on 3D dataset curation can be found in Appendix~\ref{apx:dataset_3d}.

\subsection{Experiments Settings}

\subsubsection{Baselines}
We compare our method with two state-of-the-art approaches for the 3D scene synthesis tasks:
\begin{itemize}
    \item  \textbf{ATISS}~\citep{paschalidou2021atiss}, a Transformer-based auto-regressive network that regards scenes as sets of unordered objects and generates objects and their attributes sequentially.
    \item  \textbf{DiffuScene}~\citep{tang2024diffuscene}, a continuous diffusion model with Gaussian kernels that treats object attributes in one scene as a 2D matrix after padding them to a fixed size.
\end{itemize}
Both methods can be conditioned on text prompts by cross-attention with a pretrained text encoder.
Our preliminary experiments suggest that both baselines encounter difficulties in modeling high-dimensional semantic feature distributions, consequently impacting their performance in generating other attributes.
Therefore, we augment them to generate quantized features.

\subsubsection{Evaluation Metrics}\label{subsubsec:metrics3d}
To assess the controllability of layouts, we use a metric named ``instruction recall'' (\textbf{iRecall}), which quantifies the proportion of the required triplets ``(subject, relation, object)'' occurring in synthesized scenes to all provided in instructions.
It is a stringent metric that simultaneously considers all three elements in a layout relation.
Following previous works~\citep{paschalidou2021atiss,liu2023clip,tang2024diffuscene}, we also report Fr\'echet Inception Distance (\textbf{FID})~\citep{heusel2017gans}, \textbf{FID$^\text{CLIP}$}~\citep{kynkaanniemi2022role}, which computes FID scores by CLIP features~\citep{radford2021learning}, Kernel Inception Distance (\textbf{KID})~\citep{binkowski2018demystifying}, scene classification accuracy (\textbf{SCA}).
These metrics evaluate the overall quality of synthesized scenes and rely on rendered images.
We use Blender~\citep{blender} to produce high-quality images for both synthesized and real scenes.

\subsection{Instruction-drive 3D Scene Synthesis}
Table~\ref{tab:scenesyn} presents the quantitive evaluations for synthesizing 3D scenes with instructions.
We report the average scores of five runs with different random seeds.
As demonstrated, even with the enhancement of quantized semantic features, two baseline methods continue to demonstrate inferior performance compared to ours.
ATISS outperforms DiffuScene regarding generation fidelity, owing to its capacity to model in discrete spaces.
DiffuScene shows better controllability to ATISS because it affords global visibility of samples during generation.
Our proposed \textsc{InstructLayout} exhibits the best of both worlds.
Remarkably, we achieve a substantial advancement in controllability, measured in iRecall, for scene generative models, surpassing current state-of-the-art approaches by about \textbf{15\%}$\sim$\textbf{25\%} across various room types, all while maintaining high fidelity.
It is noteworthy that \textsc{InstructLayout} excels in handling more complex scenes, such as living and dining rooms, which typically comprise an average of 20 objects, in contrast to bedrooms, which have only 8 objects on average, revealing the benefits of modeling intricate 3D scenes associated with the semantic graph prior.
Qualitative visualizations are provided in Appendix~\ref{apx:results_3d}.

\subsection{Zero-shot Applications}
Thanks to the discrete design and mask modeling, the learned semantic graph prior can perform diverse downstream tasks without any fine-tuning.
We investigate four zero-shot tasks: \textbf{stylization}, \textbf{re-arrangement}, \textbf{completion}, and \textbf{unconditional generation}.

Stylization and re-arrangement task can be formulated as $p_\phi(\mathbf{f}|c,\mathbf{s}, \mathbf{y})$ and $p_{\phi,\theta}(\mathbf{s}|c,\mathbf{f}, \mathbf{y})$ respectively.
In the completion task, we intend to add new objects $\{\mathbf{o}_k^i\}$ to a partial scene $\mathcal{S}_i$ with instructions.
By filling the partial scene attributes with \texttt{[MASK]} tokens, we treat them as intermediate states during discrete graph denoising, allowing for a straightforward adaptation of the learned semantic graph prior to these tasks in a zero-shot manner.
Unconditional synthesis is implemented by simply setting instruction embeddings $\mathbf{y}$ as zeros.
To assess controllability in the stylization task, we define $\Delta\coloneqq \frac{1}{N}\sum_{i=1}^{N}\text{CosSim}(\mathbf{f}_i,\mathbf{d}^{\text{style}}_i)-\text{CosSim}(\mathbf{f}_i,\mathbf{d}^{\text{class}}_i)$, where $\mathbf{d}^{\text{style}}_i$ represents the CLIP text feature of object class name with the desired style, and $\mathbf{d}^{\text{class}}_i$ is the CLIP text feature with only class information.
$\text{CosSim}(\cdot,\cdot)$ calculates the cosine similarity between two vectors.

Evaluations on zero-shot applications are reported in Table~\ref{tab:app_3d}.
Our method consistently outperforms two strong baselines in both controllability and fidelity.
While ATISS, as an auto-regressive model, is a natural fit for the completion task, its unidirectional dependency chain limits its effectiveness for tasks requiring global scene modeling, such as re-arrangement.
DiffuScene can adapt to these tasks by replacing the known parts with the noised corresponding scene attributes during sampling, similar to image in-painting~\citep{meng2021sdedit,nichol2022glide}.
However, the known attributes are greatly corrupted in the early steps, which could misguide the denoising direction and necessitate fine-tuning.
Additionally, DiffuScene also faces challenges in searching for semantic features in a continuous space for stylization.
In contrast, \textsc{InstructLayout} globally models scene attributes and treats partial scene attributes as intermediate discrete states during training.
These designs effectively eliminate the training-test gap, rendering it highly versatile for a wide range of downstream tasks.
Visualizations of zero-shot applications are available in Appendix~\ref{apx:app_3d}.

\section{\textsc{InstructLayout} for 2D Poster Synthesis}\label{sec:exp2d}

\subsection{2D E-commerce Poster-Instruction Paired Dataset}\label{subsec:dataset_2d}
AutoPoster~\citep{lin2023autoposter} is a recently proposed public poster dataset containing 76,537 E-commerce posters with detailed annotations of layout elements.
Compared with an ideal comprehensive poster design pipeline, however, it lacks three essential components, including (1) design instructions describing the desired layout of a poster, (2) principle product descriptions for tagline generation, and (3) clean product images for use as poster backgrounds.
To the best of our knowledge, no existing dataset offers comprehensive information for the instruction-driven poster synthesis task~\citep{zhou2022composition,hsu2023posterlayout,gao2022caponimage}.

In this work, we bridge this gap by constructing a high-quality 2D E-commerce poster dataset based on the AutoPoster dataset.
Similar to 3D scene dataset curation, we use predefined rules as a systematic instruction generation pipeline and collect \textbf{73,070} product descriptions for AutoPoster posters from public Internet sources using web crawler techniques.
For clean product images, previous works\citep{hsu2023posterlayout, lin2023autoposter} employ an off-the-shelf inpainting model LaMa~\citep{suvorov2022resolution} to remove the graphics elements from the poster.
However, LaMa often struggles with posters featuring frequent and significant color shifts, resulting in blurry and unnatural distortion of the processed product images.
To mitigate this, we first mask out graphic elements with the average color near the masked region and then apply the LaMa inpainting model to obtain clean product images.
More details on the 2D poster dataset curation are provided in Appendix~\ref{apx:dataset_2d}.

\begin{table*}[th]
    \centering
    \caption{Quantitive evaluations for instruction-driven 2D poster synthesis on baselines~\citep{Gupta_2021_ICCV,Jyothi_2019_ICCV,hsu2023posterlayout,Inoue_2023_CVPR} and \textsc{InstructLayout}. Higher iRecall, Val, Und$_l$ and Und$_s$, and lower Ove, N-Ali, and Occ indicate better generation quality.
    The best results among different models are bolded.
    Standard deviation values are provided as subscripts.}
    \label{tab:postersyn}
    \renewcommand\arraystretch{1.3}
    \resizebox{\textwidth}{!}{
        \begin{tabular}{l|c|ccccc|c}
            \toprule[1.2pt]
            2D Instruction-driven Synthesis &
            $\uparrow$ iRecall$_{\%}$ & $\uparrow$ Val$_\%$ & $\downarrow$ Ove$_{\times100}$ & $\downarrow$ N-Ali$_{\times100}$ & $\uparrow$ Und$_{l\%}$ & $\uparrow$ Und$_{s\%}$ & $\downarrow$ Occ$_\%$  \\
            \midrule[1.2pt]
      LayoutTransformer~\citep{Gupta_2021_ICCV}& 65.689\std{0.83} & 98.651\std{0.07} & 0.855\std{0.06} & 0.324\std{0.04} & 95.083\std{0.18} & 84.888\std{0.28} & 7.742\std{0.05}   \\
      LayoutVAE~\citep{Jyothi_2019_ICCV}& 19.002\std{0.57} & 94.877\std{0.10} & 8.437\std{0.06} & 5.261\std{0.10} & 91.195\std{0.25} & 82.620\std{0.35} & 15.918\std{0.03}  \\
      DS-GAN~\citep{hsu2023posterlayout} (No Instruction)& $\backslash$ & 82.385\std{0.42} & 2.072\std{0.11} & 0.225\std{0.01} & 85.933\std{0.64} & 47.115\std{0.75} & 8.537\std{0.41}   \\
      DDPM~\citep{ho2020denoising}& 20.986\std{0.50} & 94.782\std{0.23} & 17.573\std{0.45} & 5.980\std{0.21} & 63.587\std{0.56} & 57.908\std{0.56} & 11.868\std{0.20}   \\
      LayoutDM~\citep{Inoue_2023_CVPR}& 62.999\std{1.62} & 93.008\std{0.24} & 2.500\std{0.11} & 0.240\std{0.02} & 86.965\std{0.36} & 82.454\std{0.42} & 7.242\std{0.04}  \\
      \hline
            \cellcolor{mygray}Ours
            & \cellcolor{mygray}\textbf{78.451}\std{0.294} & \cellcolor{mygray}\textbf{99.977}\std{0.01} & \cellcolor{mygray}\textbf{0.785}\std{0.02} & \cellcolor{mygray}\textbf{0.165}\std{0.01} & \cellcolor{mygray}\textbf{97.123}\std{0.15} & \cellcolor{mygray}\textbf{92.490}\std{0.29} & \cellcolor{mygray}\textbf{6.565}\std{0.05} 
            \\
             \bottomrule[1.2pt]
        \end{tabular}
    }
\end{table*}

\begin{table*}[th]
    \centering
    \caption{Quantitive evaluations for generated semantic features on SAP~\citep{lin2023autoposter} and \textsc{InstructLayout}. ``D'', ``G'', and ``S'' represent dominant, gradient, and stroke colors, respectively. Lower MAE and MSE indicate better results.}
    \label{tab:featuresyn}
    \renewcommand\arraystretch{1.3}
    \resizebox{0.9\textwidth}{!}{
        \begin{tabular}{l|cc|cc|cc}
            \toprule[1.2pt]
            2D Graphic Features
            & $\downarrow$ D-ab(MSE)& $\downarrow$ D-light(MAE)   & $\downarrow$ S-ab(MSE)& $\downarrow$ S-light(MAE) & $\downarrow$ G-ab(MSE)& $\downarrow$ G-light(MAE) \\
            \midrule[1.2pt]
    Random& 78.411 & 5.946 & 76.062 & 6.146 & 80.536 & 6.596  \\
      SAP~\cite{lin2023autoposter} & 74.501 & 4.197 & 55.201 & 2.920 & 72.369 & 3.368   \\
      \hline
      \cellcolor{mygray}Ours &\cellcolor{mygray}\textbf{31.047} &\cellcolor{mygray}\textbf{2.459} &\cellcolor{mygray}\textbf{33.005} &\cellcolor{mygray}\textbf{2.872} &\cellcolor{mygray}\textbf{41.067} &\cellcolor{mygray}\textbf{1.745}  \\
            \bottomrule[1.2pt]
        \end{tabular}
   }
\end{table*}

\subsection{Experiments Settings}

\subsubsection{Baselines}
We re-implemented several state-of-the-art 2D layout synthesis models for a comprehensive comparison:
\begin{itemize}
    \item \textbf{LayoutTransformer}~\citep{Gupta_2021_ICCV}, a Transformer decoder that simultaneously predicts categories and discretizes sizes autoregressively.
    \item \textbf{LayoutVAE}~\cite{Jyothi_2019_ICCV}, a VAE-based model that autoregressively predicts the conditional distribution of elements' counts and bounding boxes.
    \item \textbf{DS-GAN}~\citep{hsu2023posterlayout}, a GAN-based model that builds a Design Sequence GAN and generates layouts with product images in a specially designed order.
    \item \textbf{DDPM}~\citep{ho2020denoising}, a native continuous diffusion model, and we modify it to enable simultaneous prediction of categories and bound boxes.
    \item \textbf{LayoutDM}~\citep{Inoue_2023_CVPR}, a single-stage discrete diffusion model that discretizes layout attributes and diffuses them in a discrete space.
    \item \textbf{SAP}~\citep{lin2023autoposter}, a Transformer encoder designed for graphic semantic feature synthesis with conditions of product images and layout bounding boxes.
\end{itemize}
During the re-implementation process, we further condition these baselines with product images and text prompts.
While DS-GAN~\citep{hsu2023posterlayout} fails to converge with textual instructions, we only feed image conditions to it.
Note that most of these methods~\citep{Gupta_2021_ICCV,Jyothi_2019_ICCV,hsu2023posterlayout,ho2020denoising,Inoue_2023_CVPR} can only generate categories $c$ and spatial features $\mathbf{s}$, and SAP~\cite{lin2023autoposter} only predict graphic semantic features $\mathbf{f}$.

\subsubsection{Evaluation Metrics}
Besides \textbf{iRecall} proposed in Sec.~\ref{subsubsec:metrics3d} to measure the controllability of layout synthesis, we adopt several graphic and content-aware metrics following previous works~\cite{hsu2023posterlayout, xu2023unsupervised, zhou2022composition}.
Graphic metrics measure the degree of visual harmony among elements (texts, underlay and logos), including Validity (\textbf{Val}) measuring the validity rate of all predicted elements, Overlay (\textbf{Ove}) representing the average Intersection over Union (IoU) for each pair of elements except underlays, Non-Alignment (\textbf{N-Ali}) measuring the extent of spatial non-alignment, and Underlay-Effectiveness (\textbf{Und$_l$} and \textbf{Und$_s$}) indicating the quality of generated underlay.
For the content-aware metric, we use Occlusion (\textbf{Occ}) to evaluate the overall visual impact of the poster by measuring the degree of occlusion of the background subject by elements.
For semantic feature quality assessment, we follow \citet{lin2023autoposter} to calculate the accuracy of predicted dominant colors, stroke colors, and gradient colors in \textit{Lab} space using Mean Squared Error (\textbf{MSE}) and Mean Absolute Error (\textbf{MAE}).

\subsection{Instruction-drive 2D Poster Synthesis}
Table~\ref{tab:postersyn} shows quantitive evaluations on 2D layout generation.
We run all models with 5 different random seeds and report the average score.
\textsc{InstructLayout} considerably outperforms other existing models by a large margin among all metrics. 
Especially, our two-stage diffusion model, which skillfully combines discrete and continuous features, remarkably beats both the single-stage diffusion models, the continuous DDPM~\citep{ho2020denoising} and the discrete  LayoutDM~\citep{Inoue_2023_CVPR}, in terms of controllability and visual harmony.
It highlights the effectiveness of the proposed discrete graph prior and the capacity of modeling fine spatial relationships facilitated by the continuous diffusion process in the second stage.
As presented in Table~\ref{tab:featuresyn}, \textsc{InstructLayout} is also capable of leveraging both image and positional information to mimic the behavior of human designers better and generate graphic semantic features of higher usability.
Qualitative results of generated posters with textual content synthesized by the tagline generator are provided in Appendix~\ref{apx:results_2d}.

\section{Ablation Study}

\begin{figure}
    \centering
    \includegraphics[width=0.5\textwidth]{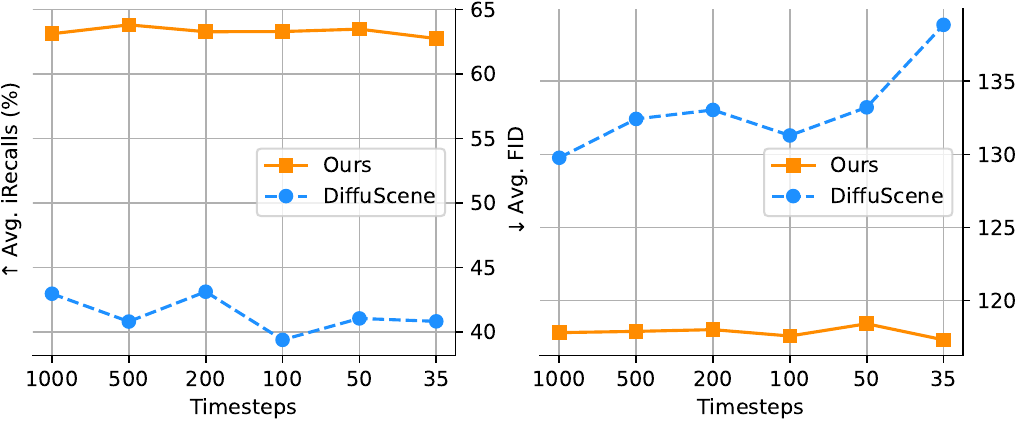}
    \caption{Ablation study on timesteps for diffusion models on 3D scene layout synthesis task.}
    \label{fig:timesteps}
\end{figure}

\begin{table*}[t]
    \centering
    \caption{Ablation study on diffusion timesteps on 2D poster layout synthesis task.}
    \label{tab:timesteps}
    \renewcommand\arraystretch{1.3}
    \resizebox{0.8\textwidth}{!}{
        \begin{tabular}{l|c|ccccc|c}
            \toprule[1.2pt]
            2D Timestep Ablation&
            $\uparrow$ iRecall$_{\%}$ & $\uparrow$ Val$_{\%}$ & $\downarrow$ Ove$_{\times100}$ & $\downarrow$ N-Ali$_{\times100}$ & $\uparrow$ Und$_{l\%}$ & $\uparrow$ Und$_{s\%}$ & $\downarrow$ Occ$_{\%}$ \\
            \midrule[1.2pt]

      LayoutDM ($T=110$)& 65.841 &  93.003 &  2.350 &  0.222 &  86.872 &  82.183 &  7.236   \\
      LayoutDM ($T=60$)& 64.293 &  92.467 &  2.532 &  0.279 &  86.213 &  81.902 &  7.293   \\
      LayoutDM ($T=35$)& 61.825 &  90.145 &  2.813 &  0.468 &  81.911 &  75.312 &  7.470  \\

      \hline

      \cellcolor{mygray}\textbf{Ours} ($T=100+10$)& \cellcolor{mygray}78.268 &  \cellcolor{mygray}99.995 &  \cellcolor{mygray}0.992 &  \cellcolor{mygray}0.193 &  \cellcolor{mygray}96.067 &  \cellcolor{mygray}89.285 &  \cellcolor{mygray}6.587   \\ 
      \cellcolor{mygray}\textbf{Ours} ($T=50+10$)& \cellcolor{mygray}78.833 &  \cellcolor{mygray}99.995 &  \cellcolor{mygray}1.151 &  \cellcolor{mygray}0.188 &  \cellcolor{mygray}95.537 &  \cellcolor{mygray}88.560 &  \cellcolor{mygray}6.580   \\ 
      \cellcolor{mygray}\textbf{Ours} ($T=25+10$)& \cellcolor{mygray}77.862 &  \cellcolor{mygray}99.986 &  \cellcolor{mygray}1.186 &  \cellcolor{mygray}0.192 &  \cellcolor{mygray}94.277 &  \cellcolor{mygray}86.225 &  \cellcolor{mygray}6.662   \\

        \bottomrule[1.2pt]
        \end{tabular}
    }
\end{table*}

\begin{table*}[t]
    \centering
    \caption{Ablation study on different strategies to learn semantic graph prior $p_{\phi}(\mathcal{G}|\mathbf{y})$. ``Perm. Invar.'' means permutation-invariant graph modeling.}
    \label{tab:ablation_3d}
    \renewcommand\arraystretch{1.2}
    \resizebox{0.7\textwidth}{!}{
    \begin{tabular}{l|>{\columncolor{mygray}}c|ccc|c} 
        \toprule[1.2pt]

        Graph Prior &
        Ours &
        Gaussian & Joint Mask  & Uniform &
        Perm. Invar. \\

        \midrule[1.2pt]

        $\uparrow$ iRecall$_{\%}$ &
        \textbf{73.64}\std{1.37} &
        34.18\std{2.53} & 34.21\std{2.79} & 69.22\std{3.25} &
        70.49\std{2.50} \\
        \hline
        $\downarrow$ FID &
        \textbf{114.78}\std{1.19} &
        128.98\std{0.97} & 130.86\std{2.76} & 139.61\std{1.06} &
        116.53\std{1.35} \\
        $\downarrow$ FID$^\text{CLIP}$ &
        \textbf{6.65}\std{0.18} &
        7.30\std{0.03} & 7.59\std{0.17} & 8.82\std{0.24} &
        6.69\std{0.16} \\
        $\downarrow$ KID$_{\times\text{1e-3}}$ &
        \textbf{0.32}\std{0.03} &
        2.63\std{0.73} & 4.82\std{1.69} & 10.55\std{1.19} &
        0.37\std{0.02} \\
        $\ \ $ SCA$_{\%}$ &
        \textbf{56.02}\std{0.91} &
        57.10\std{3.22} & 60.37\std{3.13} & 76.79\std{3.14} &
        58.64\std{1.33} \\

        \bottomrule[1.2pt]
    \end{tabular}
    }
\end{table*}

\subsection{Diffusion Timesteps}\label{subsec:timesteps}
Although containing two diffusion models, our method could achieve better efficiency by reducing the steps of reverse processes without a noticeable decline in performance.
We compare \textsc{InstructLayout} with DiffuScene~\citep{tang2024diffuscene} and LayoutDM~\citep{Inoue_2023_CVPR} in Figure~\ref{fig:timesteps} and Table~\ref{tab:timesteps} for 3D scene and 2D poster layout synthesis tasks respectively.
We try different timesteps $T$ such as ``100+100'', ``100+10'', and ``25+10'', where the first number represents timesteps for semantic graph prior and the latter is for layout decoder.
We find that $T=100$ and $10$ is sufficient for $p_\phi(\mathcal{G}|\mathbf{y})$ and $p_\theta(\mathcal{S}|\mathcal{G})$ respectively.
This stems from the fact that each stage in \textsc{InstructLayout} tackles an easier denoising task compared to single-stage diffusion models such as DiffuScene~\citep{tang2024diffuscene} and LayoutDM~\citep{chai2023layoutdm}.

\subsection{Learning Semantic Graph Prior}\label{subsec:ablation}

We then explore different strategies to learn semantic graph prior.
All experiments are conducted on the 3D bedroom dataset.
Quantitative results are presented in Table~\ref{tab:ablation_3d}.

\noindent
\textbf{Transition Matrices for Learning Graph Prior.}
We investigate the effects of different transition matrices for learning the proposed semantic graph prior, including:
(1) Embed all categorical variables into their one-hot encodings and diffuse them by Gaussian kernels, which is similar to \citet{niu2020permutation} and \citet{jo2022score};
(2) Jointly masking $\mathbf{F}$ and $\mathbf{E}$ along with nodes $\mathbf{C}$ in a graph, so only the attributes of other objects can be utilized for recovery;
(3) Adopt uniform transition matrices without mask states, similar to \citet{vignac2022digress}.
Evaluations on both controllability and fidelity reveal the advantages of our independent mask strategy.

\noindent
\textbf{Permutation Non-invariance.}
Different to previous studies on graph generation~\citep{niu2020permutation,jo2022score,vignac2022digress}, we depart from the permutation-invariant modeling convention to ease the semantic graph prior learning process.
We strive to preserve exchangeable graph distributions by randomly shuffling object orders during training.
Performance for invariant graph prior is provided in the last column of Table~\ref{tab:ablation_3d}.
Its performance declines due to the unnecessary imposition of invariance in scene synthesis.

\section{Conclusion}\label{sec:conclusion}

By integrating a semantic graph prior and a layout decoder, we propose a novel generative framework, \textsc{InstructLayout}, that significantly improves the controllability and fidelity of 2D and 3D layout synthesis, providing a user-friendly interface through instructions in natural languages.
Benefiting from the design of semantic graph prior, our method can also apply to diverse applications without any fine-tuning.
The controllability and versatility positions \textsc{InstructLayout} as a promising tool.
Meanwhile, given the rapid development of large language models (LLMs), integrating an LLM into our instruction-driven pipeline holds significant promise for further enhancing generation interactivity.
We hope this work can help in practical scenarios, such as streamlining e-commerce poster design and facilitating interior design, and serves as a good starting point for creating high-quality instruction datasets in 2D poster and 3D scene applications.

\ifCLASSOPTIONcaptionsoff
    \newpage
\fi
\bibliographystyle{IEEEtranN}
\bibliography{references}

\begin{thebibliography}{82}
\providecommand{\natexlab}[1]{#1}
\providecommand{\url}[1]{#1}
\csname url@samestyle\endcsname
\providecommand{\newblock}{\relax}
\providecommand{\bibinfo}[2]{#2}
\providecommand{\BIBentrySTDinterwordspacing}{\spaceskip=0pt\relax}
\providecommand{\BIBentryALTinterwordstretchfactor}{4}
\providecommand{\BIBentryALTinterwordspacing}{\spaceskip=\fontdimen2\font plus
\BIBentryALTinterwordstretchfactor\fontdimen3\font minus \fontdimen4\font\relax}
\providecommand{\BIBforeignlanguage}[2]{{%
\expandafter\ifx\csname l@#1\endcsname\relax
\typeout{** WARNING: IEEEtranN.bst: No hyphenation pattern has been}%
\typeout{** loaded for the language `#1'. Using the pattern for}%
\typeout{** the default language instead.}%
\else
\language=\csname l@#1\endcsname
\fi
#2}}
\providecommand{\BIBdecl}{\relax}
\BIBdecl

\bibitem[Qi et~al.(2018)Qi, Zhu, Huang, Jiang, and Zhu]{qi2018human}
S.~Qi, Y.~Zhu, S.~Huang, C.~Jiang, and S.-C. Zhu, ``Human-centric indoor scene synthesis using stochastic grammar,'' in \emph{Proceedings of the IEEE Conference on Computer Vision and Pattern Recognition (CVPR)}, 2018, pp. 5899--5908.

\bibitem[Ritchie et~al.(2019)Ritchie, Wang, and Lin]{ritchie2019fast}
D.~Ritchie, K.~Wang, and Y.-a. Lin, ``Fast and flexible indoor scene synthesis via deep convolutional generative models,'' in \emph{Proceedings of the IEEE/CVF Conference on Computer Vision and Pattern Recognition (CVPR)}, 2019, pp. 6182--6190.

\bibitem[Zhang et~al.(2020)Zhang, Yang, Ma, Luo, Huth, Vouga, and Huang]{zhang2020deep}
Z.~Zhang, Z.~Yang, C.~Ma, L.~Luo, A.~Huth, E.~Vouga, and Q.~Huang, ``Deep generative modeling for scene synthesis via hybrid representations,'' \emph{ACM Transactions on Graphics (TOG)}, vol.~39, no.~2, pp. 1--21, 2020.

\bibitem[Gupta et~al.(2021)Gupta, Lazarow, Achille, Davis, Mahadevan, and Shrivastava]{Gupta_2021_ICCV}
K.~Gupta, J.~Lazarow, A.~Achille, L.~S. Davis, V.~Mahadevan, and A.~Shrivastava, ``Layouttransformer: Layout generation and completion with self-attention,'' in \emph{Proceedings of the IEEE/CVF International Conference on Computer Vision (ICCV)}, 2021, pp. 1004--1014.

\bibitem[Jyothi et~al.(2019)Jyothi, Durand, He, Sigal, and Mori]{Jyothi_2019_ICCV}
A.~A. Jyothi, T.~Durand, J.~He, L.~Sigal, and G.~Mori, ``Layoutvae: Stochastic scene layout generation from a label set,'' in \emph{Proceedings of the IEEE/CVF International Conference on Computer Vision (ICCV)}, 2019.

\bibitem[Yang et~al.(2021)Yang, Guo, Zhou, and Tong]{yang2021indoor}
M.-J. Yang, Y.-X. Guo, B.~Zhou, and X.~Tong, ``Indoor scene generation from a collection of semantic-segmented depth images,'' in \emph{Proceedings of the IEEE/CVF International Conference on Computer Vision (ICCV)}, 2021, pp. 15\,203--15\,212.

\bibitem[H\"ollein et~al.(2023)H\"ollein, Cao, Owens, Johnson, and Nie{\ss}ner]{hollein2023text2room}
L.~H\"ollein, A.~Cao, A.~Owens, J.~Johnson, and M.~Nie{\ss}ner, ``Text2room: Extracting textured 3d meshes from 2d text-to-image models,'' in \emph{Proceedings of the IEEE/CVF International Conference on Computer Vision (ICCV)}, 2023, pp. 7909--7920.

\bibitem[Hsu et~al.(2023)Hsu, He, Peng, Kong, and Zhang]{hsu2023posterlayout}
H.~Y. Hsu, X.~He, Y.~Peng, H.~Kong, and Q.~Zhang, ``Posterlayout: A new benchmark and approach for content-aware visual-textual presentation layout,'' in \emph{Proceedings of the IEEE/CVF Conference on Computer Vision and Pattern Recognition (CVPR)}, 2023, pp. 6018--6026.

\bibitem[Chai et~al.(2023)Chai, Zhuang, and Yan]{chai2023layoutdm}
S.~Chai, L.~Zhuang, and F.~Yan, ``Layoutdm: Transformer-based diffusion model for layout generation,'' in \emph{Proceedings of the IEEE/CVF Conference on Computer Vision and Pattern Recognition (CVPR)}, 2023, pp. 18\,349--18\,358.

\bibitem[Lin et~al.(2023)Lin, Zhou, Ma, Gao, Fei, Chen, Yu, and Ge]{lin2023autoposter}
J.~Lin, M.~Zhou, Y.~Ma, Y.~Gao, C.~Fei, Y.~Chen, Z.~Yu, and T.~Ge, ``Autoposter: A highly automatic and content-aware design system for advertising poster generation,'' in \emph{Proceedings of the 31st ACM International Conference on Multimedia (ACM MM)}, 2023, pp. 1250--1260.

\bibitem[Song et~al.(2023)Song, Cao, Xu, Kang, Tang, Yuan, and Zhao]{song2023roomdreamer}
L.~Song, L.~Cao, H.~Xu, K.~Kang, F.~Tang, J.~Yuan, and Y.~Zhao, ``Roomdreamer: Text-driven 3d indoor scene synthesis with coherent geometry and texture,'' \emph{arXiv preprint arXiv:2305.11337}, 2023.

\bibitem[Feng et~al.(2023)Feng, Zhu, Fu, Jampani, Akula, He, Basu, Wang, and Wang]{feng2023layoutgpt}
W.~Feng, W.~Zhu, T.-J. Fu, V.~Jampani, A.~R. Akula, X.~He, S.~Basu, X.~E. Wang, and W.~Y. Wang, ``Layout{GPT}: Compositional visual planning and generation with large language models,'' in \emph{Advances in Neural Information Processing Systems (NeurIPS)}, 2023.

\bibitem[Patil et~al.(2023)Patil, Patil, Li, Fisher, Savva, and Zhang]{patil2023advances}
A.~G. Patil, S.~G. Patil, M.~Li, M.~Fisher, M.~Savva, and H.~Zhang, ``Advances in data-driven analysis and synthesis of 3d indoor scenes,'' \emph{Computer Graphics Forum}, 2023.

\bibitem[Lin and Mu(2024)]{lin2024instructscene}
C.~Lin and Y.~Mu, ``Instructscene: Instruction-driven 3d indoor scene synthesis with semantic graph prior,'' \emph{arXiv preprint arXiv:2402.04717}, 2024.

\bibitem[Wang et~al.(2021)Wang, Yeshwanth, and Nie{\ss}ner]{wang2021sceneformer}
X.~Wang, C.~Yeshwanth, and M.~Nie{\ss}ner, ``Sceneformer: Indoor scene generation with transformers,'' in \emph{International Conference on 3D Vision (3DV)}, 2021, pp. 106--115.

\bibitem[Paschalidou et~al.(2021)Paschalidou, Kar, Shugrina, Kreis, Geiger, and Fidler]{paschalidou2021atiss}
D.~Paschalidou, A.~Kar, M.~Shugrina, K.~Kreis, A.~Geiger, and S.~Fidler, ``Atiss: Autoregressive transformers for indoor scene synthesis,'' \emph{Advances in Neural Information Processing Systems (NeurIPS)}, vol.~34, pp. 12\,013--12\,026, 2021.

\bibitem[Cao et~al.(2022)Cao, Ma, Zhou, Liu, Xie, Ge, and Jiang]{cao2022geometry}
Y.~Cao, Y.~Ma, M.~Zhou, C.~Liu, H.~Xie, T.~Ge, and Y.~Jiang, ``Geometry aligned variational transformer for image-conditioned layout generation,'' in \emph{Proceedings of the 30th ACM International Conference on Multimedia (ACM MM)}, 2022, pp. 1561--1571.

\bibitem[Inoue et~al.(2023)Inoue, Kikuchi, Simo-Serra, Otani, and Yamaguchi]{Inoue_2023_CVPR}
N.~Inoue, K.~Kikuchi, E.~Simo-Serra, M.~Otani, and K.~Yamaguchi, ``Layoutdm: Discrete diffusion model for controllable layout generation,'' in \emph{Proceedings of the IEEE/CVF Conference on Computer Vision and Pattern Recognition (CVPR)}, 2023, pp. 10\,167--10\,176.

\bibitem[Liu et~al.(2023{\natexlab{a}})Liu, Xiong, Jones, Nie, Gupta, and O{\u{g}}uz]{liu2023clip}
J.~Liu, W.~Xiong, I.~Jones, Y.~Nie, A.~Gupta, and B.~O{\u{g}}uz, ``Clip-layout: Style-consistent indoor scene synthesis with semantic furniture embedding,'' \emph{arXiv preprint arXiv:2303.03565}, 2023.

\bibitem[Tang et~al.(2024)Tang, Nie, Markhasin, Dai, Thies, and Nie{\ss}ner]{tang2024diffuscene}
J.~Tang, Y.~Nie, L.~Markhasin, A.~Dai, J.~Thies, and M.~Nie{\ss}ner, ``Diffuscene: Scene graph denoising diffusion probabilistic model for generative indoor scene synthesis,'' in \emph{Proceedings of the IEEE/CVF Conference on Computer Vision and Pattern Recognition (CVPR)}, 2024.

\bibitem[Luo et~al.(2020)Luo, Zhang, Wu, and Tenenbaum]{luo2020end}
A.~Luo, Z.~Zhang, J.~Wu, and J.~B. Tenenbaum, ``End-to-end optimization of scene layout,'' in \emph{Proceedings of the IEEE/CVF Conference on Computer Vision and Pattern Recognition (CVPR)}, 2020, pp. 3754--3763.

\bibitem[Dhamo et~al.(2021)Dhamo, Manhardt, Navab, and Tombari]{dhamo2021graph}
H.~Dhamo, F.~Manhardt, N.~Navab, and F.~Tombari, ``Graph-to-3d: End-to-end generation and manipulation of 3d scenes using scene graphs,'' in \emph{Proceedings of the IEEE/CVF International Conference on Computer Vision (ICCV)}, 2021, pp. 16\,352--16\,361.

\bibitem[Zhai et~al.(2023)Zhai, {\"O}rnek, Wu, Di, Tombari, Navab, and Busam]{zhai2023commonscenes}
G.~Zhai, E.~P. {\"O}rnek, S.-C. Wu, Y.~Di, F.~Tombari, N.~Navab, and B.~Busam, ``Commonscenes: Generating commonsense 3d indoor scenes with scene graphs,'' \emph{Advances in Neural Information Processing Systems (NeurIPS)}, 2023.

\bibitem[Wang et~al.(2019)Wang, Lin, Weissmann, Savva, Chang, and Ritchie]{wang2019planit}
K.~Wang, Y.-A. Lin, B.~Weissmann, M.~Savva, A.~X. Chang, and D.~Ritchie, ``Planit: Planning and instantiating indoor scenes with relation graph and spatial prior networks,'' \emph{ACM Transactions on Graphics (TOG)}, vol.~38, no.~4, pp. 1--15, 2019.

\bibitem[Liu et~al.(2023{\natexlab{b}})Liu, Shi, Kuang, Zhu, Li, Han, Cai, Porikli, and Su]{liu2023openshape}
M.~Liu, R.~Shi, K.~Kuang, Y.~Zhu, X.~Li, S.~Han, H.~Cai, F.~Porikli, and H.~Su, ``Openshape: Scaling up 3d shape representation towards open-world understanding,'' \emph{Advances in Neural Information Processing Systems (NeurIPS)}, 2023.

\bibitem[Li et~al.(2022)Li, Li, Xiong, and Hoi]{li2022blip}
J.~Li, D.~Li, C.~Xiong, and S.~Hoi, ``Blip: Bootstrapping language-image pre-training for unified vision-language understanding and generation,'' in \emph{International Conference on Machine Learning (ICML)}, 2022, pp. 12\,888--12\,900.

\bibitem[Ouyang et~al.(2022)Ouyang, Wu, Jiang, Almeida, Wainwright, Mishkin, Zhang, Agarwal, Slama, Ray, et~al.]{ouyang2022training}
L.~Ouyang, J.~Wu, X.~Jiang, D.~Almeida, C.~Wainwright, P.~Mishkin, C.~Zhang, S.~Agarwal, K.~Slama, A.~Ray \emph{et~al.}, ``Training language models to follow instructions with human feedback,'' \emph{Advances in Neural Information Processing Systems (NeurIPS)}, vol.~35, pp. 27\,730--27\,744, 2022.

\bibitem[OpenAI(2023)]{openai2023gpt4}
OpenAI, ``Gpt-4 technical report,'' \emph{arXiv preprint arXiv:2303.08774}, 2023.

\bibitem[Li et~al.(2019)Li, Patil, Xu, Chaudhuri, Khan, Shamir, Tu, Chen, Cohen-Or, and Zhang]{li2019grains}
M.~Li, A.~G. Patil, K.~Xu, S.~Chaudhuri, O.~Khan, A.~Shamir, C.~Tu, B.~Chen, D.~Cohen-Or, and H.~Zhang, ``Grains: Generative recursive autoencoders for indoor scenes,'' \emph{ACM Transactions on Graphics (TOG)}, vol.~38, no.~2, pp. 1--16, 2019.

\bibitem[Gao et~al.(2023)Gao, Sun, Mo, Lai, Guibas, and Yang]{gao2023scenehgn}
L.~Gao, J.-M. Sun, K.~Mo, Y.-K. Lai, L.~J. Guibas, and J.~Yang, ``Scenehgn: Hierarchical graph networks for 3d indoor scene generation with fine-grained geometry,'' \emph{IEEE Transactions on Pattern Analysis and Machine Intelligence (T-PAMI)}, 2023.

\bibitem[Purkait et~al.(2020)Purkait, Zach, and Reid]{purkait2020sg}
P.~Purkait, C.~Zach, and I.~Reid, ``Sg-vae: Scene grammar variational autoencoder to generate new indoor scenes,'' in \emph{European Conference on Computer Vision (ECCV)}, 2020, pp. 155--171.

\bibitem[Zhou et~al.(2019{\natexlab{a}})Zhou, While, and Kalogerakis]{zhou2019scenegraphnet}
Y.~Zhou, Z.~While, and E.~Kalogerakis, ``Scenegraphnet: Neural message passing for 3d indoor scene augmentation,'' in \emph{Proceedings of the IEEE/CVF International Conference on Computer Vision (ICCV)}, 2019, pp. 7384--7392.

\bibitem[Para et~al.(2021)Para, Guerrero, Kelly, Guibas, and Wonka]{para2021generative}
W.~Para, P.~Guerrero, T.~Kelly, L.~J. Guibas, and P.~Wonka, ``Generative layout modeling using constraint graphs,'' in \emph{Proceedings of the IEEE/CVF International Conference on Computer Vision (ICCV)}, 2021, pp. 6690--6700.

\bibitem[Li et~al.(2018)Li, Vinyals, Dyer, Pascanu, and Battaglia]{li2018learning}
Y.~Li, O.~Vinyals, C.~Dyer, R.~Pascanu, and P.~Battaglia, ``Learning deep generative models of graphs,'' \emph{arXiv preprint arXiv:1803.03324}, 2018.

\bibitem[Kingma and Welling(2014)]{kingma2013auto}
D.~P. Kingma and M.~Welling, ``Auto-encoding variational bayes,'' in \emph{International Conference on Learning Representations (ICLR)}, 2014.

\bibitem[Sohn et~al.(2015)Sohn, Lee, and Yan]{sohn2015learning}
K.~Sohn, H.~Lee, and X.~Yan, ``Learning structured output representation using deep conditional generative models,'' \emph{Advances in Neural Information Processing Systems (NeurIPS)}, vol.~28, 2015.

\bibitem[Johnson et~al.(2018)Johnson, Gupta, and Fei-Fei]{johnson2018image}
J.~Johnson, A.~Gupta, and L.~Fei-Fei, ``Image generation from scene graphs,'' in \emph{Proceedings of the IEEE Conference on Computer Vision and Pattern Recognition (CVPR)}, 2018, pp. 1219--1228.

\bibitem[Vaswani et~al.(2017)Vaswani, Shazeer, Parmar, Uszkoreit, Jones, Gomez, Kaiser, and Polosukhin]{vaswani2017attention}
A.~Vaswani, N.~Shazeer, N.~Parmar, J.~Uszkoreit, L.~Jones, A.~N. Gomez, {\L}.~Kaiser, and I.~Polosukhin, ``Attention is all you need,'' \emph{Advances in Neural Information Processing Systems (NeurIPS)}, vol.~30, 2017.

\bibitem[Kikuchi et~al.(2021)Kikuchi, Simo-Serra, Otani, and Yamaguchi]{Kikuchi_2021}
K.~Kikuchi, E.~Simo-Serra, M.~Otani, and K.~Yamaguchi, ``Constrained graphic layout generation via latent optimization,'' in \emph{Proceedings of the 29th ACM International Conference on Multimedia (ACM MM)}, 2021.

\bibitem[Zhou et~al.(2022)Zhou, Xu, Ma, Ge, Jiang, and Xu]{zhou2022composition}
M.~Zhou, C.~Xu, Y.~Ma, T.~Ge, Y.~Jiang, and W.~Xu, ``Composition-aware graphic layout gan for visual-textual presentation designs,'' \emph{arXiv preprint arXiv:2205.00303}, 2022.

\bibitem[Xu et~al.(2023)Xu, Zhou, Ge, Jiang, and Xu]{xu2023unsupervised}
C.~Xu, M.~Zhou, T.~Ge, Y.~Jiang, and W.~Xu, ``Unsupervised domain adaption with pixel-level discriminator for image-aware layout generation,'' in \emph{Proceedings of the IEEE/CVF Conference on Computer Vision and Pattern Recognition (CVPR)}, 2023, pp. 10\,114--10\,123.

\bibitem[Gao et~al.(2022)Gao, Hou, Zhang, Ge, Jiang, and Wang]{gao2022caponimage}
Y.~Gao, X.~Hou, Y.~Zhang, T.~Ge, Y.~Jiang, and P.~Wang, ``Caponimage: Context-driven dense-captioning on image,'' in \emph{Proceedings of the 2022 Conference on Empirical Methods in Natural Language Processing (EMNLP)}, 2022, pp. 3449--3465.

\bibitem[You et~al.(2018)You, Ying, Ren, Hamilton, and Leskovec]{you2018graphrnn}
J.~You, R.~Ying, X.~Ren, W.~Hamilton, and J.~Leskovec, ``Graphrnn: Generating realistic graphs with deep auto-regressive models,'' in \emph{International Conference on Machine Learning (ICML)}, 2018, pp. 5708--5717.

\bibitem[Garg et~al.(2021)Garg, Dhamo, Farshad, Musatian, Navab, and Tombari]{garg2021unconditional}
S.~Garg, H.~Dhamo, A.~Farshad, S.~Musatian, N.~Navab, and F.~Tombari, ``Unconditional scene graph generation,'' in \emph{Proceedings of the IEEE/CVF International Conference on Computer Vision (ICCV)}, 2021, pp. 16\,362--16\,371.

\bibitem[Simonovsky and Komodakis(2018)]{simonovsky2018graphvae}
M.~Simonovsky and N.~Komodakis, ``Graphvae: Towards generation of small graphs using variational autoencoders,'' in \emph{International Conference on Artificial Neural Networks (ICANN)}, 2018, pp. 412--422.

\bibitem[Verma et~al.(2022)Verma, De, Agrawal, Vinay, and Chakrabarti]{verma2022varscene}
T.~Verma, A.~De, Y.~Agrawal, V.~Vinay, and S.~Chakrabarti, ``Varscene: A deep generative model for realistic scene graph synthesis,'' in \emph{International Conference on Machine Learning (ICML)}, 2022, pp. 22\,168--22\,183.

\bibitem[De~Cao and Kipf(2018)]{de2018molgan}
N.~De~Cao and T.~Kipf, ``{MolGAN: An implicit generative model for small molecular graphs},'' \emph{ICML 2018 workshop on Theoretical Foundations and Applications of Deep Generative Models}, 2018.

\bibitem[Martinkus et~al.(2022)Martinkus, Loukas, Perraudin, and Wattenhofer]{martinkus2022spectre}
K.~Martinkus, A.~Loukas, N.~Perraudin, and R.~Wattenhofer, ``Spectre: Spectral conditioning helps to overcome the expressivity limits of one-shot graph generators,'' in \emph{International Conference on Machine Learning (ICML)}, 2022, pp. 15\,159--15\,179.

\bibitem[Niu et~al.(2020)Niu, Song, Song, Zhao, Grover, and Ermon]{niu2020permutation}
C.~Niu, Y.~Song, J.~Song, S.~Zhao, A.~Grover, and S.~Ermon, ``Permutation invariant graph generation via score-based generative modeling,'' in \emph{International Conference on Artificial Intelligence and Statistics (AISTATS)}, 2020, pp. 4474--4484.

\bibitem[Jo et~al.(2022)Jo, Lee, and Hwang]{jo2022score}
J.~Jo, S.~Lee, and S.~J. Hwang, ``Score-based generative modeling of graphs via the system of stochastic differential equations,'' in \emph{International Conference on Machine Learning (ICML)}, 2022, pp. 10\,362--10\,383.

\bibitem[Vignac et~al.(2023)Vignac, Krawczuk, Siraudin, Wang, Cevher, and Frossard]{vignac2022digress}
C.~Vignac, I.~Krawczuk, A.~Siraudin, B.~Wang, V.~Cevher, and P.~Frossard, ``Digress: Discrete denoising diffusion for graph generation,'' in \emph{International Conference on Learning Representations (ICLR)}, 2023.

\bibitem[Kong et~al.(2023)Kong, Cui, Sun, Zhuang, Prakash, and Zhang]{kong2023autoregressive}
L.~Kong, J.~Cui, H.~Sun, Y.~Zhuang, B.~A. Prakash, and C.~Zhang, ``Autoregressive diffusion model for graph generation,'' in \emph{International Conference on Machine Learning (ICML)}, vol. 202, 2023, pp. 17\,391--17\,408.

\bibitem[Longland et~al.(2022)Longland, Liebowitz, Moore, and Kanhere]{longland2022text}
M.~Longland, D.~Liebowitz, K.~Moore, and S.~S. Kanhere, ``Text-conditioned graph generation using discrete graph variational autoencoders,'' 2022.

\bibitem[Lo et~al.(2023)Lo, Datar, and Sridhar]{lo2023lic}
R.~Lo, A.~Datar, and A.~Sridhar, ``Lic-gan: Language information conditioned graph generative gan model,'' \emph{arXiv preprint arXiv:2306.01937}, 2023.

\bibitem[Sohl-Dickstein et~al.(2015)Sohl-Dickstein, Weiss, Maheswaranathan, and Ganguli]{sohl2015deep}
J.~Sohl-Dickstein, E.~Weiss, N.~Maheswaranathan, and S.~Ganguli, ``Deep unsupervised learning using nonequilibrium thermodynamics,'' in \emph{International Conference on Machine Learning (ICML)}, 2015, pp. 2256--2265.

\bibitem[Ho et~al.(2020)Ho, Jain, and Abbeel]{ho2020denoising}
J.~Ho, A.~Jain, and P.~Abbeel, ``Denoising diffusion probabilistic models,'' \emph{Advances in Neural Information Processing Systems (NeurIPS)}, vol.~33, pp. 6840--6851, 2020.

\bibitem[Song et~al.(2020)Song, Sohl-Dickstein, Kingma, Kumar, Ermon, and Poole]{song2020score}
Y.~Song, J.~Sohl-Dickstein, D.~P. Kingma, A.~Kumar, S.~Ermon, and B.~Poole, ``Score-based generative modeling through stochastic differential equations,'' in \emph{International Conference on Learning Representations (ICLR)}, 2020.

\bibitem[Hoogeboom et~al.(2021)Hoogeboom, Nielsen, Jaini, Forr{\'e}, and Welling]{hoogeboom2021argmax}
E.~Hoogeboom, D.~Nielsen, P.~Jaini, P.~Forr{\'e}, and M.~Welling, ``Argmax flows and multinomial diffusion: Learning categorical distributions,'' \emph{Advances in Neural Information Processing Systems (NeurIPS)}, vol.~34, pp. 12\,454--12\,465, 2021.

\bibitem[Austin et~al.(2021)Austin, Johnson, Ho, Tarlow, and Van Den~Berg]{austin2021structured}
J.~Austin, D.~D. Johnson, J.~Ho, D.~Tarlow, and R.~Van Den~Berg, ``Structured denoising diffusion models in discrete state-spaces,'' \emph{Advances in Neural Information Processing Systems (NeurIPS)}, vol.~34, pp. 17\,981--17\,993, 2021.

\bibitem[Zhang et~al.(2016)Zhang, Isola, and Efros]{zhang2016colorful}
R.~Zhang, P.~Isola, and A.~A. Efros, ``Colorful image colorization,'' in \emph{European Conference on Computer Vision (ECCV)}, 2016, pp. 649--666.

\bibitem[Jang et~al.(2016)Jang, Gu, and Poole]{jang2016categorical}
E.~Jang, S.~Gu, and B.~Poole, ``Categorical reparameterization with gumbel-softmax,'' in \emph{International Conference on Learning Representations (ICLR)}, 2016.

\bibitem[Ramesh et~al.(2021)Ramesh, Pavlov, Goh, Gray, Voss, Radford, Chen, and Sutskever]{ramesh2021zero}
A.~Ramesh, M.~Pavlov, G.~Goh, S.~Gray, C.~Voss, A.~Radford, M.~Chen, and I.~Sutskever, ``Zero-shot text-to-image generation,'' in \emph{International Conference on Machine Learning (ICML)}, 2021, pp. 8821--8831.

\bibitem[Gu et~al.(2022)Gu, Chen, Bao, Wen, Zhang, Chen, Yuan, and Guo]{gu2022vector}
S.~Gu, D.~Chen, J.~Bao, F.~Wen, B.~Zhang, D.~Chen, L.~Yuan, and B.~Guo, ``Vector quantized diffusion model for text-to-image synthesis,'' in \emph{Proceedings of the IEEE/CVF Conference on Computer Vision and Pattern Recognition (CVPR)}, 2022, pp. 10\,696--10\,706.

\bibitem[Zhou et~al.(2019{\natexlab{b}})Zhou, Barnes, Lu, Yang, and Li]{zhou2019continuity}
Y.~Zhou, C.~Barnes, J.~Lu, J.~Yang, and H.~Li, ``On the continuity of rotation representations in neural networks,'' in \emph{Proceedings of the IEEE/CVF Conference on Computer Vision and Pattern Recognition (CVPR)}, 2019, pp. 5745--5753.

\bibitem[Perez et~al.(2018)Perez, Strub, De~Vries, Dumoulin, and Courville]{perez2018film}
E.~Perez, F.~Strub, H.~De~Vries, V.~Dumoulin, and A.~Courville, ``Film: Visual reasoning with a general conditioning layer,'' in \emph{Proceedings of the AAAI conference on Artificial Intelligence (AAAI)}, vol.~32, 2018.

\bibitem[Dwivedi and Bresson(2021)]{dwivedi2021generalization}
V.~P. Dwivedi and X.~Bresson, ``A generalization of transformer networks to graphs,'' in \emph{AAAI Workshop on Deep Learning on Graphs: Methods and Applications}, 2021.

\bibitem[Ba et~al.(2016)Ba, Kiros, and Hinton]{ba2016layer}
J.~L. Ba, J.~R. Kiros, and G.~E. Hinton, ``Layer normalization,'' \emph{arXiv preprint arXiv:1607.06450}, 2016.

\bibitem[Dhariwal and Nichol(2021)]{dhariwal2021diffusion}
P.~Dhariwal and A.~Nichol, ``Diffusion models beat gans on image synthesis,'' \emph{Advances in Neural Information Processing Systems (NeurIPS)}, vol.~34, pp. 8780--8794, 2021.

\bibitem[Radford et~al.(2021)Radford, Kim, Hallacy, Ramesh, Goh, Agarwal, Sastry, Askell, Mishkin, Clark, et~al.]{radford2021learning}
A.~Radford, J.~W. Kim, C.~Hallacy, A.~Ramesh, G.~Goh, S.~Agarwal, G.~Sastry, A.~Askell, P.~Mishkin, J.~Clark \emph{et~al.}, ``Learning transferable visual models from natural language supervision,'' in \emph{International Conference on Machine Learning (ICML)}, 2021, pp. 8748--8763.

\bibitem[Lei et~al.(2023)Lei, Deng, Shen, Guibas, and Daniilidis]{lei2023nap}
J.~Lei, C.~Deng, B.~Shen, L.~Guibas, and K.~Daniilidis, ``Nap: Neural 3d articulation prior,'' in \emph{Advances in Neural Information Processing Systems (NeurIPS)}, 2023.

\bibitem[Fu et~al.(2021)Fu, Cai, Gao, Zhang, Wang, Li, Zeng, Sun, Jia, Zhao, et~al.]{fu20213d}
H.~Fu, B.~Cai, L.~Gao, L.-X. Zhang, J.~Wang, C.~Li, Q.~Zeng, C.~Sun, R.~Jia, B.~Zhao \emph{et~al.}, ``3d-front: 3d furnished rooms with layouts and semantics,'' in \emph{Proceedings of the IEEE/CVF International Conference on Computer Vision (ICCV)}, 2021, pp. 10\,933--10\,942.

\bibitem[Heusel et~al.(2017)Heusel, Ramsauer, Unterthiner, Nessler, and Hochreiter]{heusel2017gans}
M.~Heusel, H.~Ramsauer, T.~Unterthiner, B.~Nessler, and S.~Hochreiter, ``Gans trained by a two time-scale update rule converge to a local nash equilibrium,'' \emph{Advances in Neural Information Processing Systems (NeurIPS)}, vol.~30, 2017.

\bibitem[Kynk{\"a}{\"a}nniemi et~al.(2022)Kynk{\"a}{\"a}nniemi, Karras, Aittala, Aila, and Lehtinen]{kynkaanniemi2022role}
T.~Kynk{\"a}{\"a}nniemi, T.~Karras, M.~Aittala, T.~Aila, and J.~Lehtinen, ``The role of imagenet classes in fr{\'e}chet inception distance,'' in \emph{International Conference on Learning Representations (ICLR)}, 2022.

\bibitem[Bi{\'n}kowski et~al.(2018)Bi{\'n}kowski, Sutherland, Arbel, and Gretton]{binkowski2018demystifying}
M.~Bi{\'n}kowski, D.~J. Sutherland, M.~Arbel, and A.~Gretton, ``Demystifying mmd gans,'' in \emph{International Conference on Learning Representations (ICLR)}, 2018.

\bibitem[Community(2018)]{blender}
\BIBentryALTinterwordspacing
B.~O. Community, \emph{Blender - a 3D modelling and rendering package}, Blender Foundation, Stichting Blender Foundation, Amsterdam, 2018. [Online]. Available: \url{http://www.blender.org}
\BIBentrySTDinterwordspacing

\bibitem[Meng et~al.(2021)Meng, He, Song, Song, Wu, Zhu, and Ermon]{meng2021sdedit}
C.~Meng, Y.~He, Y.~Song, J.~Song, J.~Wu, J.-Y. Zhu, and S.~Ermon, ``Sdedit: Guided image synthesis and editing with stochastic differential equations,'' in \emph{International Conference on Learning Representations (ICLR)}, 2021.

\bibitem[Nichol et~al.(2022)Nichol, Dhariwal, Ramesh, Shyam, Mishkin, Mcgrew, Sutskever, and Chen]{nichol2022glide}
A.~Q. Nichol, P.~Dhariwal, A.~Ramesh, P.~Shyam, P.~Mishkin, B.~Mcgrew, I.~Sutskever, and M.~Chen, ``Glide: Towards photorealistic image generation and editing with text-guided diffusion models,'' in \emph{International Conference on Machine Learning (ICML)}, 2022, pp. 16\,784--16\,804.

\bibitem[Suvorov et~al.(2022)Suvorov, Logacheva, Mashikhin, Remizova, Ashukha, Silvestrov, Kong, Goka, Park, and Lempitsky]{suvorov2022resolution}
R.~Suvorov, E.~Logacheva, A.~Mashikhin, A.~Remizova, A.~Ashukha, A.~Silvestrov, N.~Kong, H.~Goka, K.~Park, and V.~Lempitsky, ``Resolution-robust large mask inpainting with fourier convolutions,'' in \emph{Proceedings of the IEEE/CVF Winter Conference on Applications of Computer Vision (WACV)}, 2022, pp. 2149--2159.

\bibitem[Sun et~al.(2025)Sun, Liu, Gu, Lim, Bhat, Tombari, Li, Haber, and Wu]{sun2025layoutvlm}
F.-Y. Sun, W.~Liu, S.~Gu, D.~Lim, G.~Bhat, F.~Tombari, M.~Li, N.~Haber, and J.~Wu, ``Layoutvlm: Differentiable optimization of 3d layout via vision-language models,'' in \emph{Proceedings of the Computer Vision and Pattern Recognition Conference (CVPR)}, 2025.

\bibitem[Pang et~al.(2025)Pang, Lin, Jian, He, and Torr]{pang2025paper2poster}
W.~Pang, K.~Q. Lin, X.~Jian, X.~He, and P.~Torr, ``Paper2poster: Towards multimodal poster automation from scientific papers,'' \emph{arXiv preprint arXiv:2505.21497}, 2025.

\bibitem[Lin et~al.(2025)Lin, Lin, Pan, Yan, Feng, Mu, and Fragkiadaki]{lin2025partcrafter}
Y.~Lin, C.~Lin, P.~Pan, H.~Yan, Y.~Feng, Y.~Mu, and K.~Fragkiadaki, ``Partcrafter: Structured 3d mesh generation via compositional latent diffusion transformers,'' \emph{arXiv preprint arXiv:2506.05573}, 2025.

\bibitem[Chen et~al.(2025)Chen, Lai, Gao, Ye, Chen, Shi, Shao, Lin, Fei, Xing, et~al.]{chen2025postercraft}
S.~Chen, J.~Lai, J.~Gao, T.~Ye, H.~Chen, H.~Shi, S.~Shao, Y.~Lin, S.~Fei, Z.~Xing \emph{et~al.}, ``Postercraft: Rethinking high-quality aesthetic poster generation in a unified framework,'' \emph{arXiv preprint arXiv:2506.10741}, 2025.

\end{thebibliography}

\vspace{-1cm}

\begin{IEEEbiography}[{\includegraphics[width=1in,height=1.25in,clip,keepaspectratio]{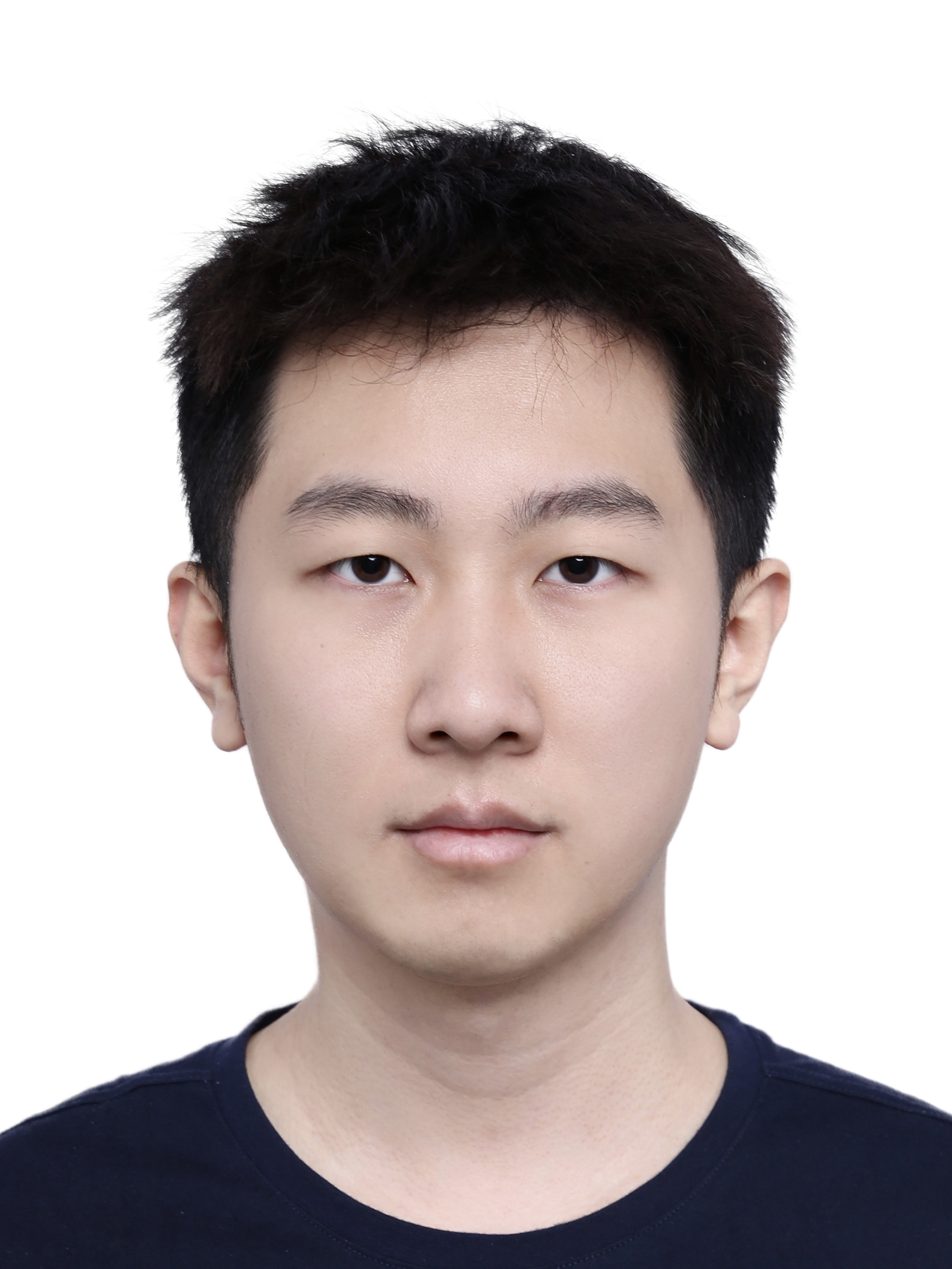}}]{Chenguo Lin}
is a Ph.D. student of computer science and artificial intelligence at Wangxuan Institute of Computer Technology, Peking University. His research interests include generative models and multi-modal representation learning, especially in the 3D domain.
\end{IEEEbiography}
\vspace{-1.2cm}

\begin{IEEEbiography}[{\includegraphics[width=1in,height=1.25in,clip,keepaspectratio]{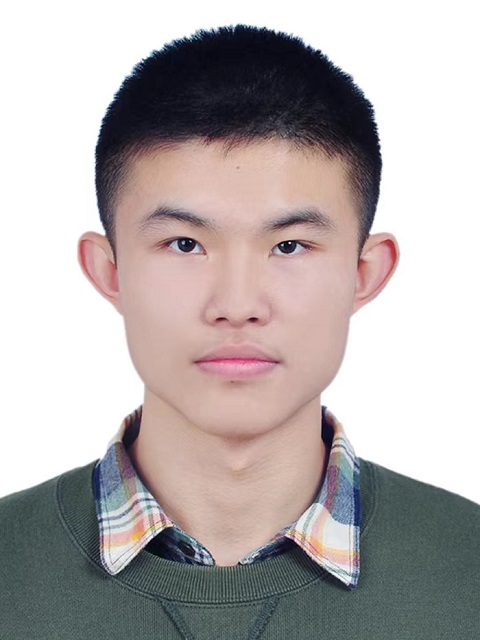}}]{Yuchen Lin}
is an undergraduate student from the School of Electronics Engineering and Computer Science, Peking University. His research interests lie in computer vision, especially 3D scene understanding and generation.
\end{IEEEbiography}
\vspace{-1.2cm}

\begin{IEEEbiography}[{\includegraphics[width=1in,height=1.25in,clip,keepaspectratio]{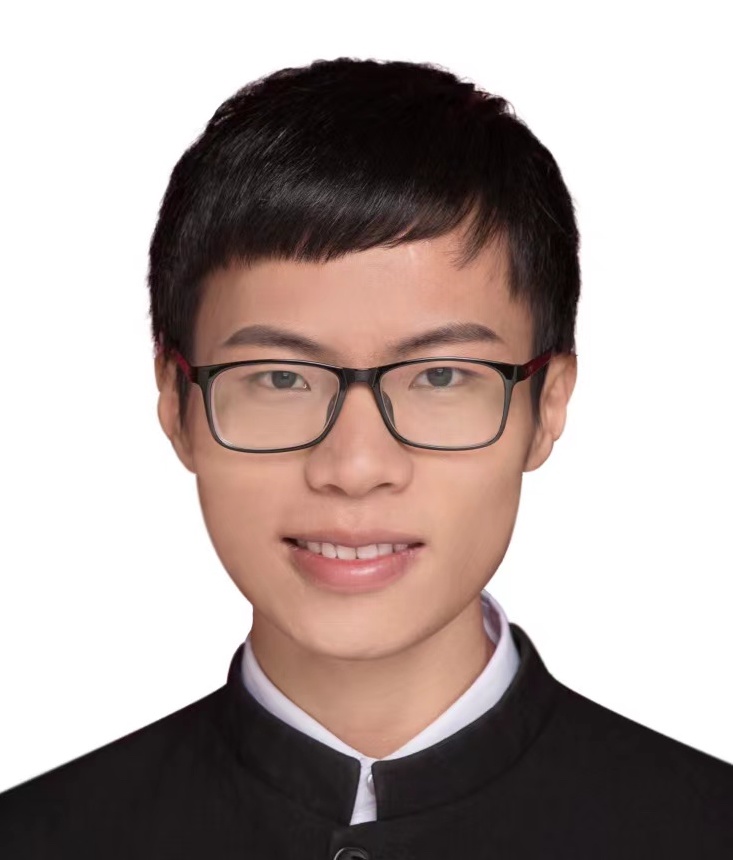}}]{Panwang Pan}
is a senior algorithm engineer of PICO Artificial Intelligence Group within ByteDance Ltd. His research interests lie in computer vision, especially 3D reconstruction, understanding and manipulation.
\end{IEEEbiography}
\vspace{-1.2cm}

\begin{IEEEbiography}[{\includegraphics[width=1in,height=1.25in,clip,keepaspectratio]{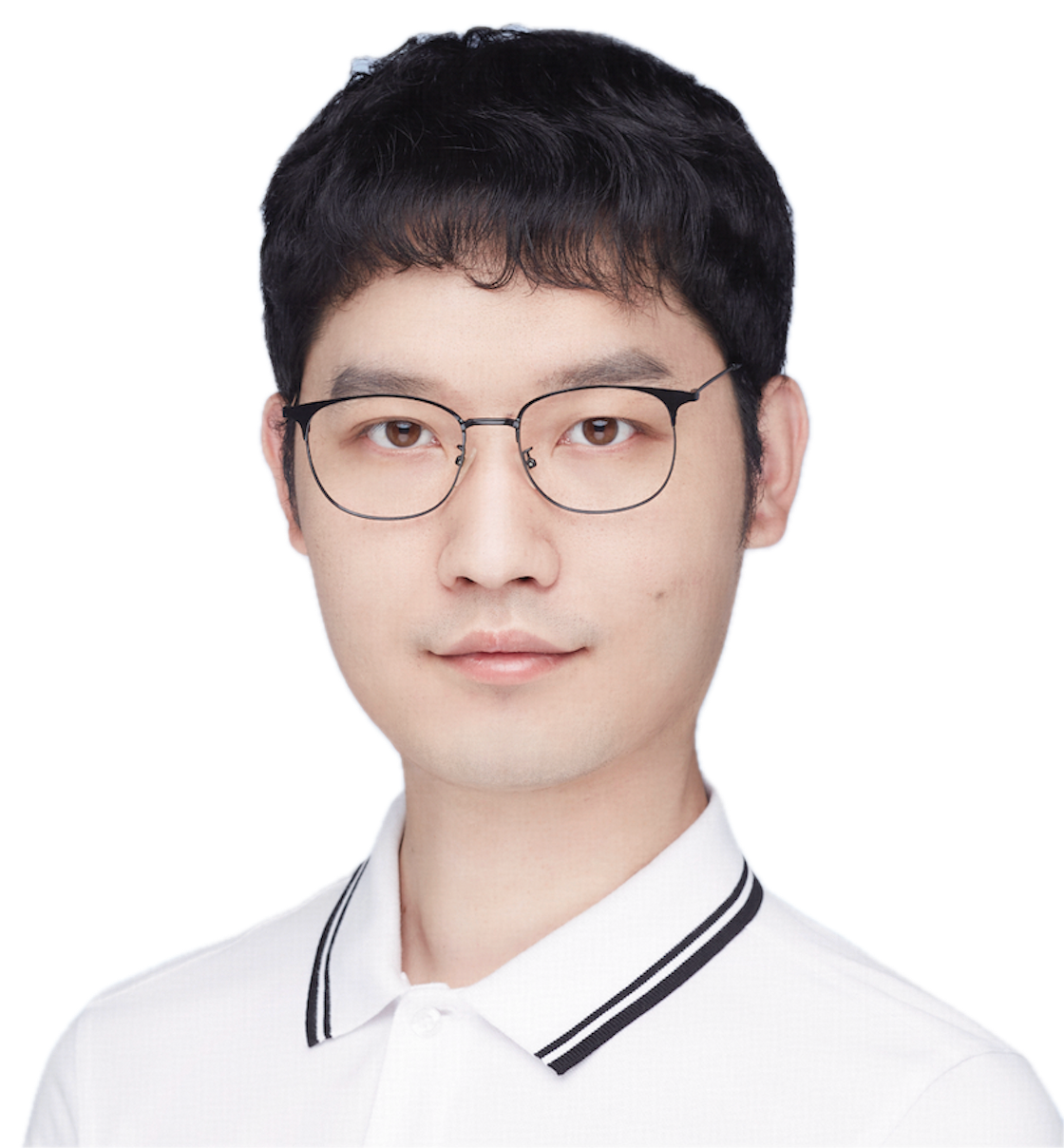}}]{Xuanyan Zhang}
is a senior algorithm engineer in ByteDance’s PICO AI R\&D team
. He is committed to research and development in the direction of computer vision, mainly in the fields of neural architecture design and 3D generation. \end{IEEEbiography}
\vspace{-1.2cm}

\begin{IEEEbiography}[{\includegraphics[width=1in,height=1.25in,clip,keepaspectratio]{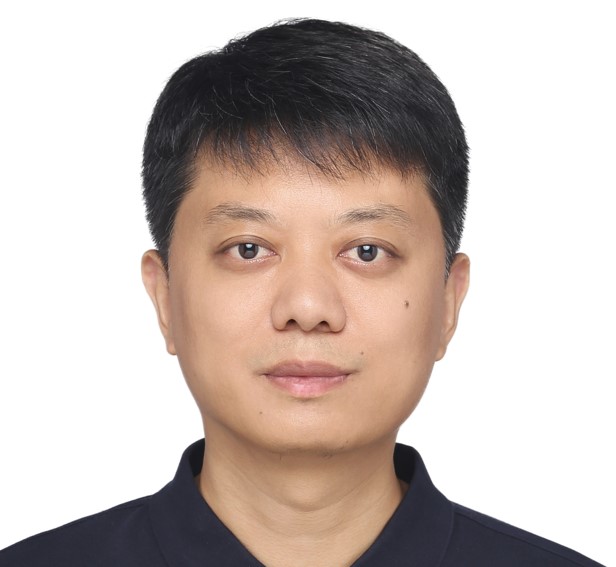}}]{Yadong Mu}
is a tenured associate professor and independent PI leading the Machine Intelligence Lab at Wangxuan Institute of Computer Technology, Peking University. Before joining Peking University, he worked at National University of Singapore, the DVMM lab of Columbia University, Huawei Noah's Ark Lab in Hong Kong, and the Multimedia Department of AT\&T Labs, New Jersey, U.S.A. He obtained both the B.S. and Ph.D. degrees from Peking University.
\end{IEEEbiography}

\clearpage
\appendices
\section{Dataset Details}\label{apx:dataset}

\subsection{3D Indoor Scene-Instruction Paired Dataset}\label{apx:dataset_3d}
Following previous works~\citep{paschalidou2021atiss,liu2023clip,tang2024diffuscene}, we use three types of indoor rooms in 3D-FRONT~\citep{fu20213d} and preprocess the dataset by filtering some problematic samples, resulting in 4041 bedrooms, 813 living rooms and 900 dining rooms.
The number of objects $N_i$ in the valid bedrooms is between 3 and 12 with 21 object categories, i.e., $K_c=21$. While for living and dining rooms, $N_i$ varies from 3 to 21 and $K_c=24$.
We use the same data split for training and evaluation as ATISS~\citep{paschalidou2021atiss}.

The original 3D-FRONT dataset does not contain any descriptions of room layout or object appearance details.
To advance research in the field of text-conditional indoor scene generation, we carefully curate a high-quality dataset with paired scenes and instructions for interior design through a multi-step process:
\begin{enumerate}
    \item \textbf{Spatial Relation Extraction}: View-dependent spatial relations are initially extracted from the 3D-FRONT dataset using predefined rules listed in Appendix~\ref{apx:reldef}.
    \item \textbf{Object Captioning}: We further enhance the dataset by providing captions to objects using BLIP~\citep{li2022blip}, a powerful model pretrained for vision-language understanding, given furniture 2D thumbnail images from the original dataset~\citep{fu20213d}.
    \item \textbf{Caption Refinement}: As generated captions may not always be accurate, we filter them with corresponding ground-truth categories using ChatGPT~\citep{ouyang2022training,openai2023gpt4}, a large language model fine-tuned for instruction-based tasks. This results in precise and expressive descriptions of each object in the scene. The prompt and hyperparameters for ChatGPT to filter captions are provided in Appendix~\ref{apx:prompt}.
    \item \textbf{Instruction Generation}: The final instructions for scene synthesis are derived from $1\sim2$ randomly selected ``(subject, relation, object)" triplets obtained during the first extraction process. Verbs and conjunctions within sentences are also randomly picked to maintain diversity and fluency.
\end{enumerate}

\subsection{2D E-commerce Poster-Instruction Paired Dataset}\label{apx:dataset_2d}
Several efforts have been made to create public datasets for 2D poster design, as shown in Table~\ref{tab:2d_dataset}.
CGL-GAN~\citep{zhou2022composition} and PosterLayout~\citep{hsu2023posterlayout} provide large-scale poster datasets accompanied by bounding box annotations. 
CapOnImage2M~\citep{gao2022caponimage} gathers product descriptions and annotated taglines for elements in posters, offering richer information for poster generation. 
AutoPoster~\citep{lin2023autoposter} innovatively introduces a dataset featuring tagline and element attribute annotations. 
However, to the best of our knowledge, no existing dataset provides comprehensive information for instruction-driven poster synthesis. 

\begin{figure*}[htbp]
    \centering
    \begin{minipage}[c]{0.05\textwidth}
            \hspace{0.4cm}
            \rotatebox{90}{Original}
    \end{minipage}
    \hfill
    \begin{subfigure}[c]{0.153\textwidth}
        \centering
        \includegraphics[width=\textwidth]{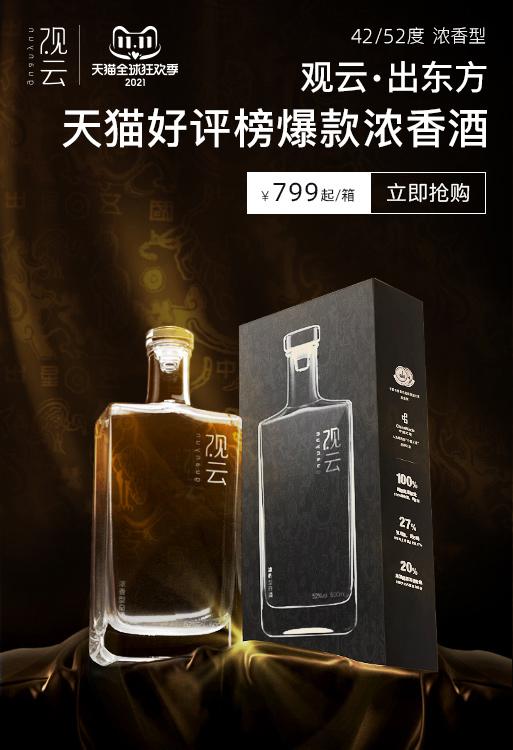}
    \end{subfigure}
    \hfill
    \begin{subfigure}[c]{0.153\textwidth}
        \centering
        \includegraphics[width=\textwidth]{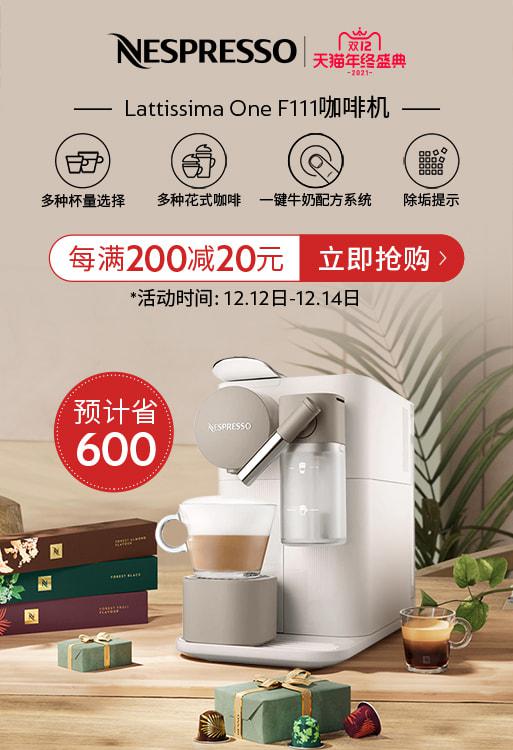}
    \end{subfigure}
    \hfill
    \begin{subfigure}[c]{0.153\textwidth}
        \centering
        \includegraphics[width=\textwidth]{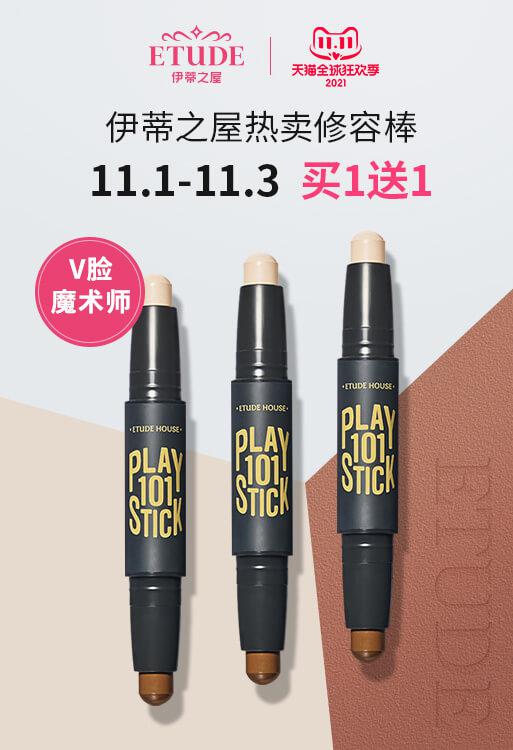}
    \end{subfigure}
    \hfill
    \begin{subfigure}[c]{0.153\textwidth}
        \centering
        \includegraphics[width=\textwidth]{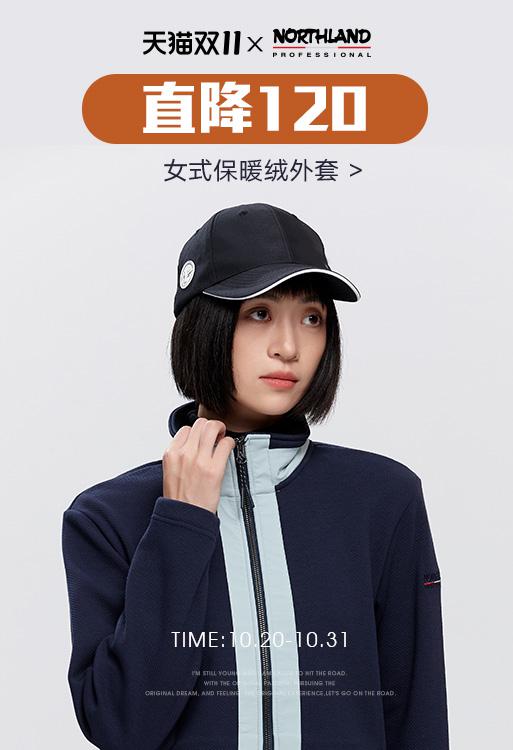}
    \end{subfigure}
    \hfill
    \begin{subfigure}[c]{0.153\textwidth}
        \centering
        \includegraphics[width=\textwidth]{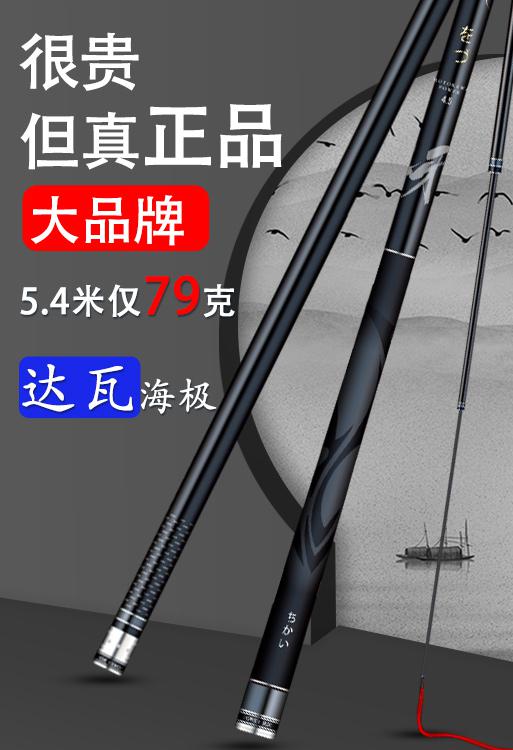}
    \end{subfigure}
    \hfill
    \begin{subfigure}[c]{0.153\textwidth}
        \centering
        \includegraphics[width=\textwidth]{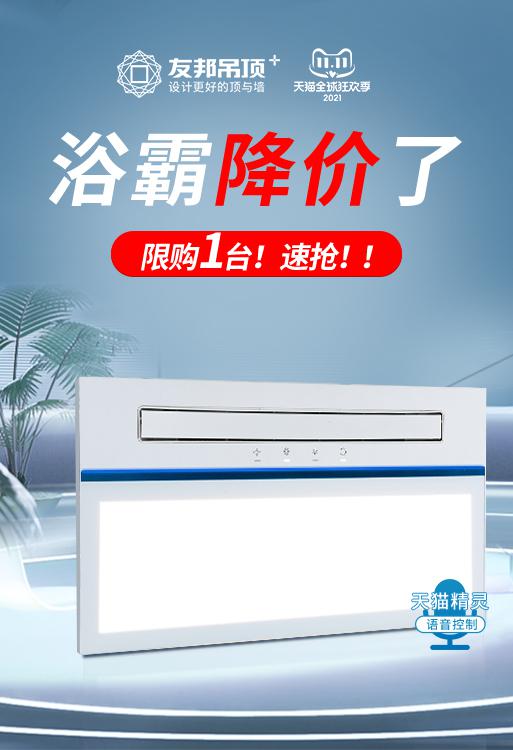}
    \end{subfigure}

    \begin{minipage}[c]{0.05\textwidth}
            \hspace{0.4cm}
            \rotatebox{90}{LaMa~\citep{suvorov2022resolution}}
    \end{minipage}
    \hfill
    \begin{subfigure}[c]{0.153\textwidth}
        \centering
        \includegraphics[width=\textwidth]{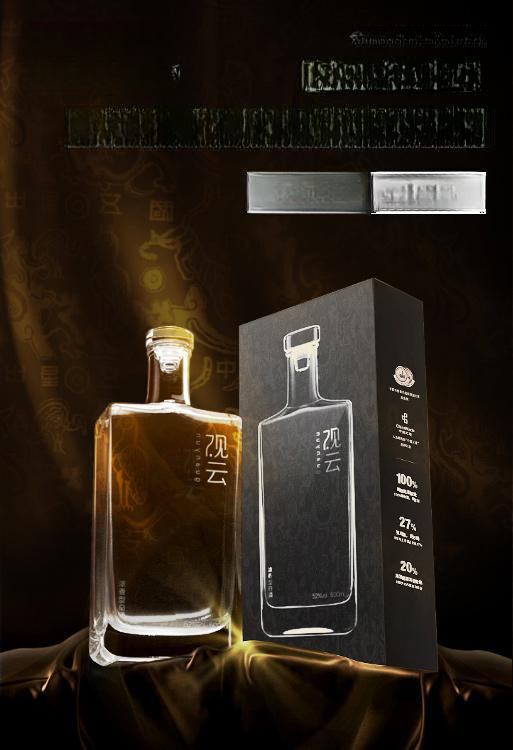}
    \end{subfigure}
    \hfill
    \begin{subfigure}[c]{0.153\textwidth}
        \centering
        \includegraphics[width=\textwidth]{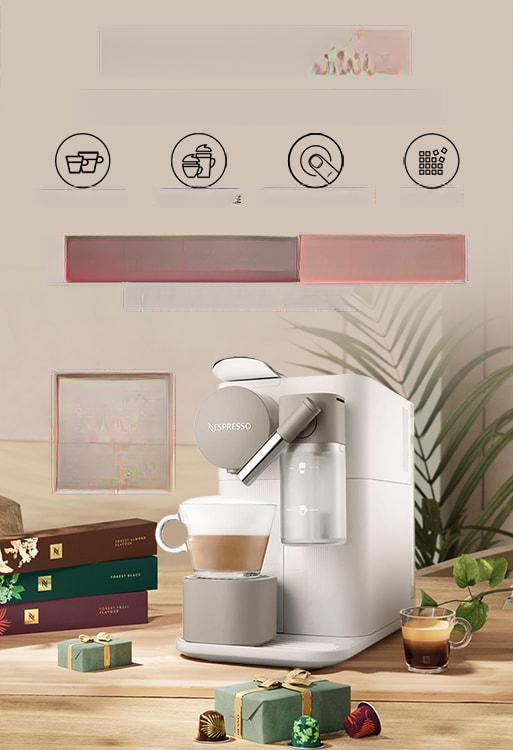}
    \end{subfigure}
    \hfill
    \begin{subfigure}[c]{0.153\textwidth}
        \centering
        \includegraphics[width=\textwidth]{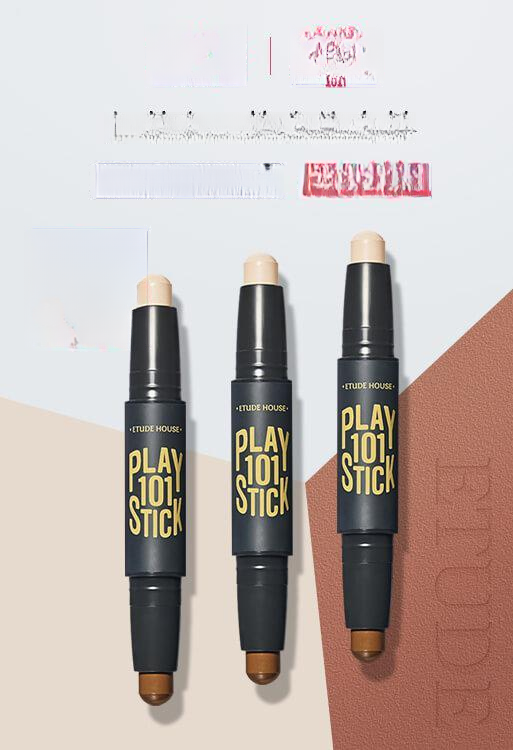}
    \end{subfigure}
    \hfill
    \begin{subfigure}[c]{0.153\textwidth}
        \centering
        \includegraphics[width=\textwidth]{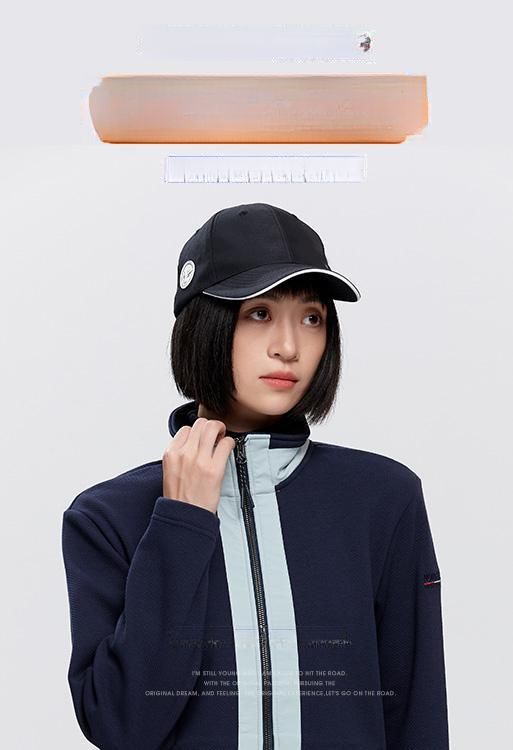}
    \end{subfigure}
    \hfill
    \begin{subfigure}[c]{0.153\textwidth}
        \centering
        \includegraphics[width=\textwidth]{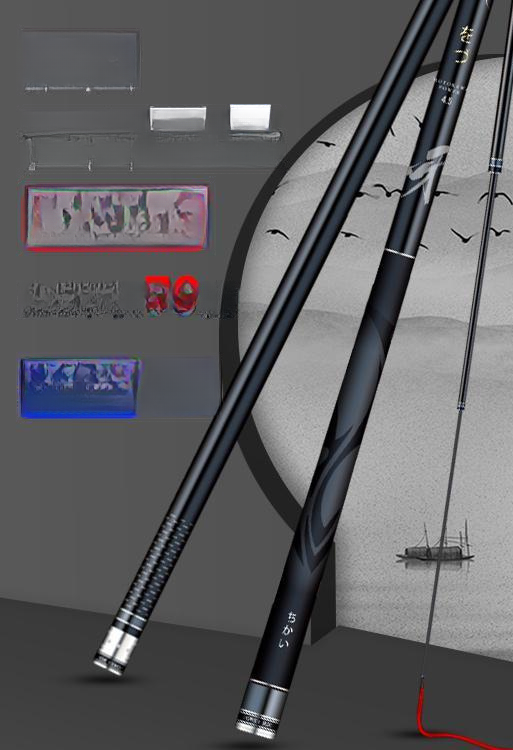}
    \end{subfigure}
    \hfill
    \begin{subfigure}[c]{0.153\textwidth}
        \centering
        \includegraphics[width=\textwidth]{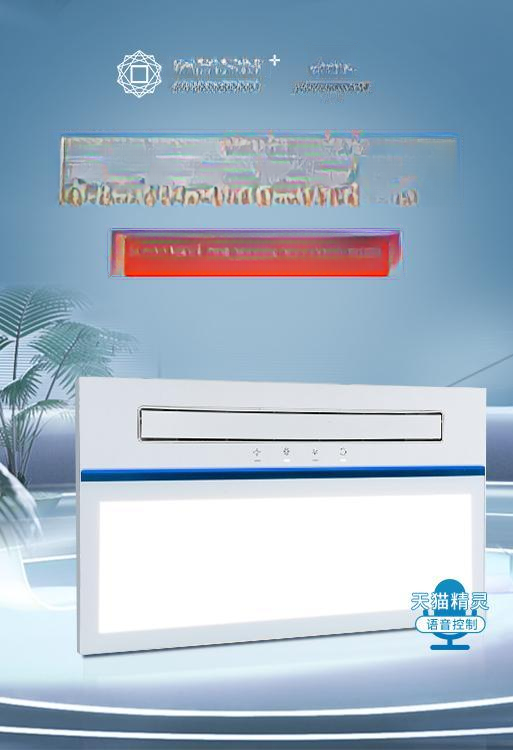}
    \end{subfigure}

    \begin{minipage}[c]{0.05\textwidth}
            \hspace{0.4cm}
            \rotatebox{90}{Ours}
    \end{minipage}
    \hfill
    \begin{subfigure}[c]{0.153\textwidth}
        \centering
        \includegraphics[width=\textwidth]{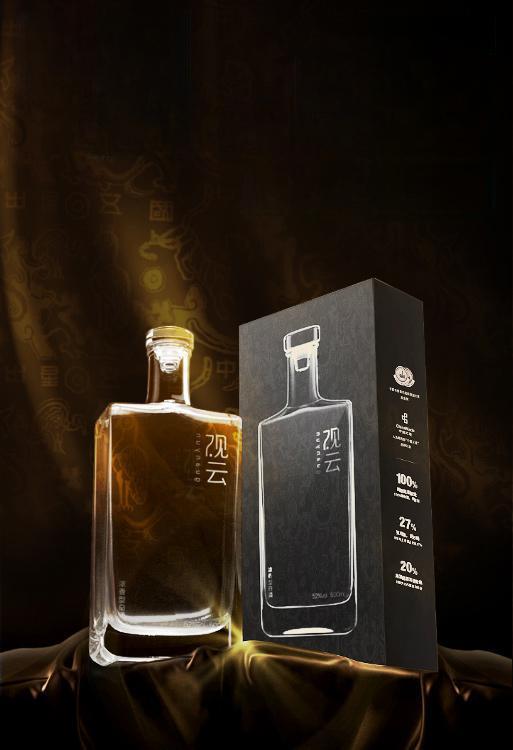}
    \end{subfigure}
    \hfill
    \begin{subfigure}[c]{0.153\textwidth}
        \centering
        \includegraphics[width=\textwidth]{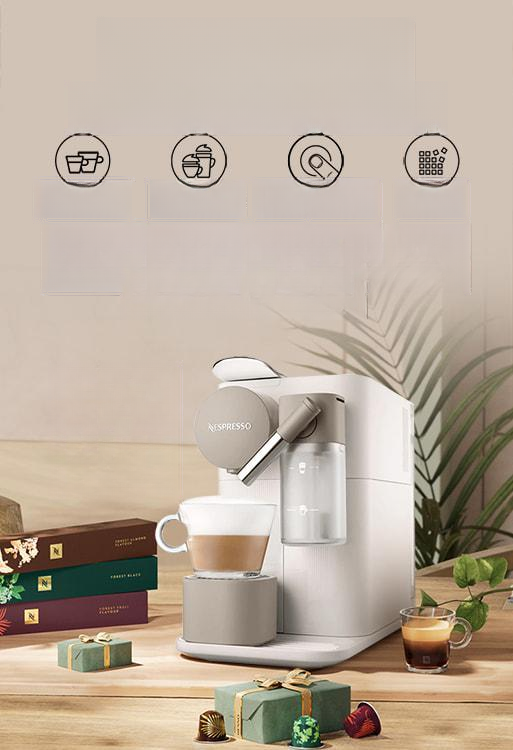}
    \end{subfigure}
    \hfill
    \begin{subfigure}[c]{0.153\textwidth}
        \centering
        \includegraphics[width=\textwidth]{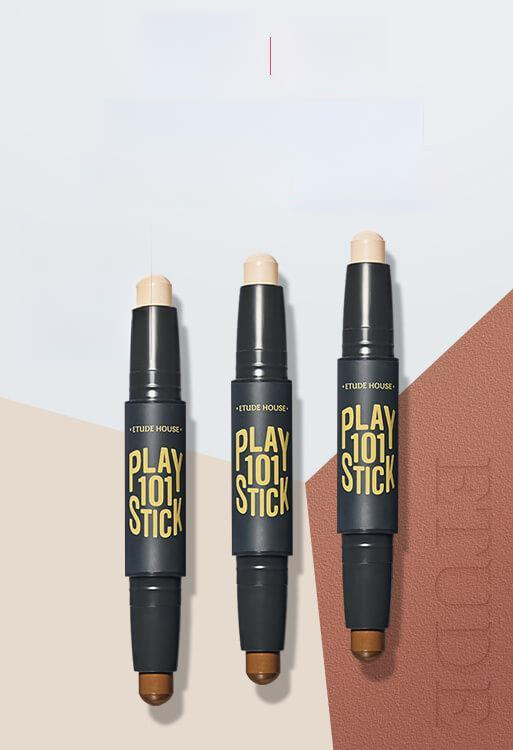}
    \end{subfigure}
    \hfill
    \begin{subfigure}[c]{0.153\textwidth}
        \centering
        \includegraphics[width=\textwidth]{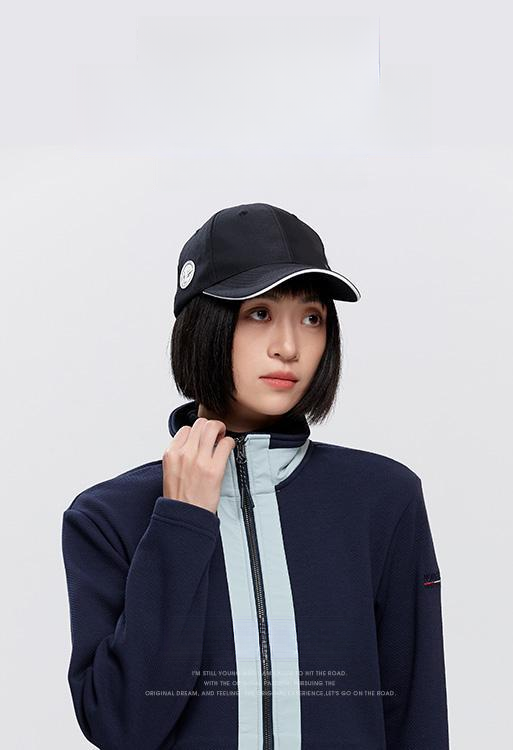}
    \end{subfigure}
    \hfill
    \begin{subfigure}[c]{0.153\textwidth}
        \centering
        \includegraphics[width=\textwidth]{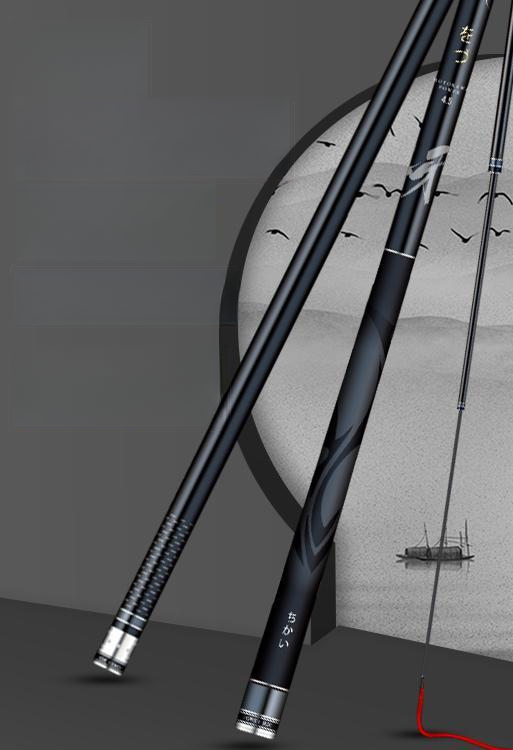}
    \end{subfigure}
    \hfill
    \begin{subfigure}[c]{0.153\textwidth}
        \centering
        \includegraphics[width=\textwidth]{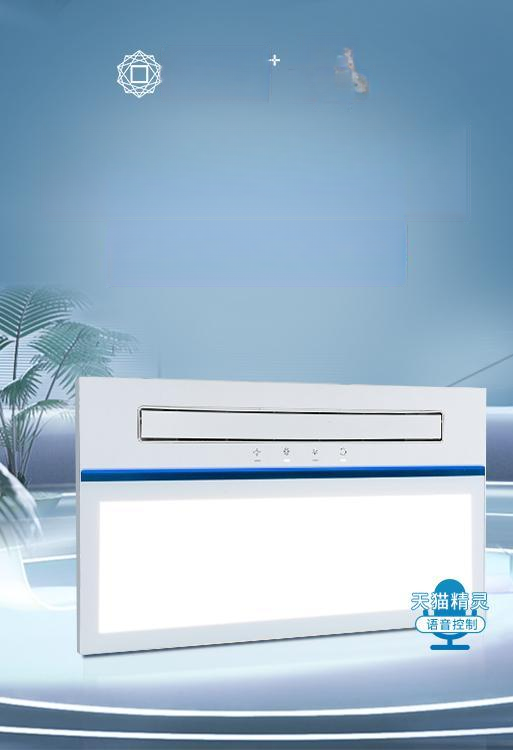}
    \end{subfigure}
    \caption{Comparisons of inpainting results between our strategy and the original LaMa~\citep{suvorov2022resolution}.}
    \label{fig:inpainting}
    \vspace{-0.2cm}
\end{figure*}

\begin{figure*}[t]
    \centering
    \includegraphics[width=\textwidth]{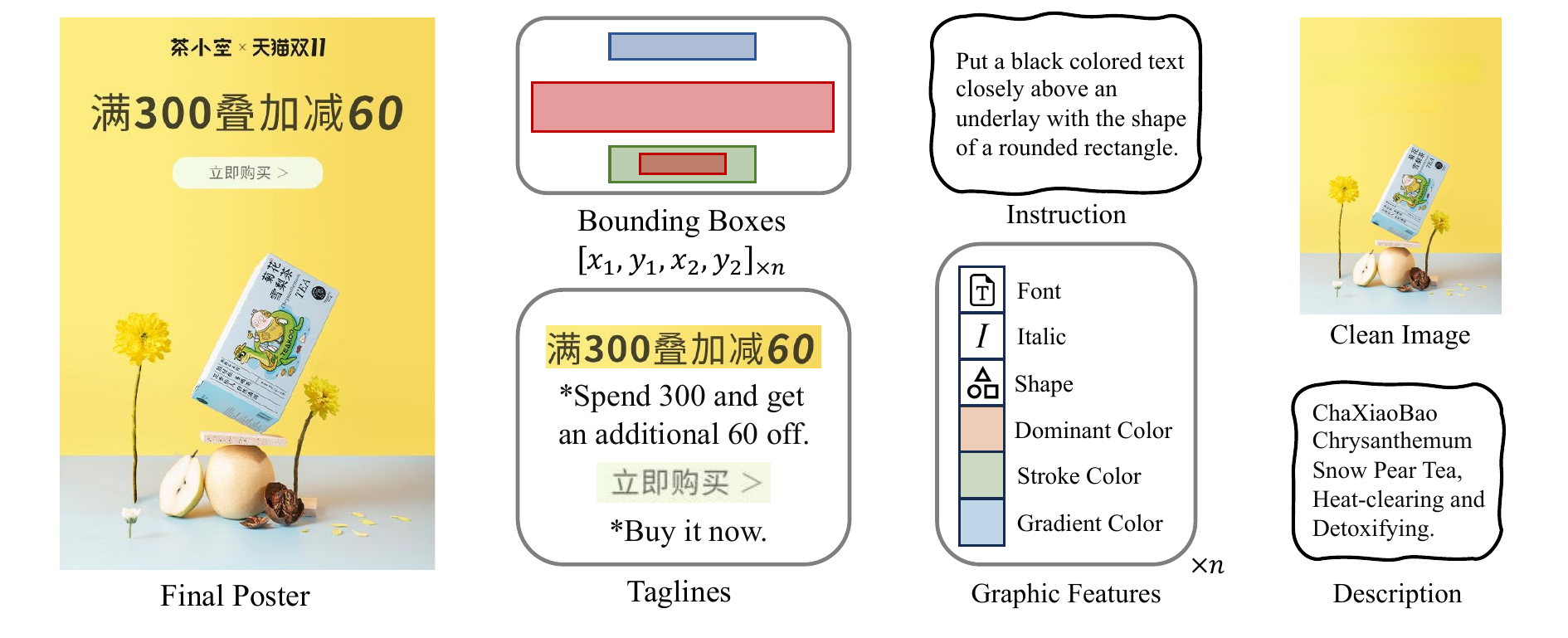}
    \caption{An example from our comprehensive 2D E-commerce poster dataset.}
    \label{fig:2d_example}
\end{figure*}

In this work, we bridge this gap by constructing a high-quality E-commerce poster dataset based on AutoPoster~\citep{lin2023autoposter}:
\begin{enumerate}
    \item \textbf{Instruction Generation}: Firstly, we employ predefined rules as shown in Appendix~\ref{apx:reldef} to extract spatial relations between 2D objects based on the place of their bounding boxes.
    With the annotation of graphic features, we can derive descriptions of objects in terms of colors, shapes, and fonts.
    Since the names of colors in LAB space can be ambiguous, we opt to use the nearest match in \texttt{Web Standard Color} palette as the name of each color.
    Then, we randomly choose several ``(subject, relation, object)'' triplets and pick up verbs and conjunctions to formulate an instruction.
    \item \textbf{Product Descriptions}: We obtained product descriptions from the Internet by web crawler techniques and manually checked and filtered them, resulting in \textbf{73,070} pieces of product descriptions.
    \item \textbf{Clean Product Images}: In accordance with the improved strategy outlined in Sec.~\ref{subsec:dataset_2d} for inpainting final posters, we obtained clean product images of significantly higher quality compared to the previous method.
    The comparisons between previous works~\citep{hsu2023posterlayout,lin2023autoposter} are demonstrated in Figure~\ref{fig:inpainting}.
\end{enumerate}
We provide an example in Figure~\ref{fig:2d_example} to showcase various elements contained in our curated poster dataset.

\subsection{Relation Definations}\label{apx:reldef}
Assume $X$ and $Y$ span the ground or poster plane, and $Z$ is the vertical axis for 3D scenes only.
We use \texttt{Center} to represent the coordinates of a bounding box's center.
\texttt{Height} is the $Z$-axis size of a 3D bounding box.
Relative orientation is computed as $\theta_{so}\coloneqq\text{atan2}(Y_s-Y_o,X_s-X_o)$, where $s$ and $o$ respectively refer to ``subject'' and ``object'' in a relationship.
$d(s,o)$ is the ground distance between $s$ and $o$.
$\text{Inside}(s,o)$ indicates whether the subject center is inside the ground bounding box of the object.
We define 11 relationships in both 3D and 2D space as listed in Table~\ref{tab:rules_3d} and Table~\ref{tab:rules_2d}, respectively.

\subsection{Caption Refinement by ChatGPT}\label{apx:prompt}
The generated 3D object captions from BLIP are refined by ChatGPT (\texttt{gpt-3.5-turbo}).
Our prompt to ChatGPT is provided in Table~\ref{tab:chatgpt}.
We set the hyperparameter \texttt{temperature} and \texttt{top\_p} for text generation to \texttt{0.2} and \texttt{0.1} respectively, encouraging more deterministic and focused outputs.

\vspace{-0.5cm}
\section{Computational Costs}\label{apx:cost}
Training takes about 2 days on one NVIDIA A40 GPU for the 3D bedroom and 2D poster datasets, and 3 days for the dining room and living room datasets, due to the increased complexity of semantic graphs in larger scenes.

For inference, under the default configuration (diffusion steps $T=100+10$), generating a batch of 128 samples takes approximately 12 seconds for both 2D and 3D layout synthesis.
Notably, our method can be significantly accelerated by reducing the number of diffusion steps.
For example, setting $T=25+10$ reduces runtime to just 3 seconds, with minimal performance degradation.
The impact of the diffusion step count is analyzed in Figure~\ref{fig:timesteps} and Table~\ref{tab:timesteps}.
We believe that further acceleration is possible through advanced techniques such as flow matching and consistency distillation.

\section{User Study}\label{apx:user}
\begin{figure}[h]
    \centering
    \includegraphics[width=0.5\textwidth]{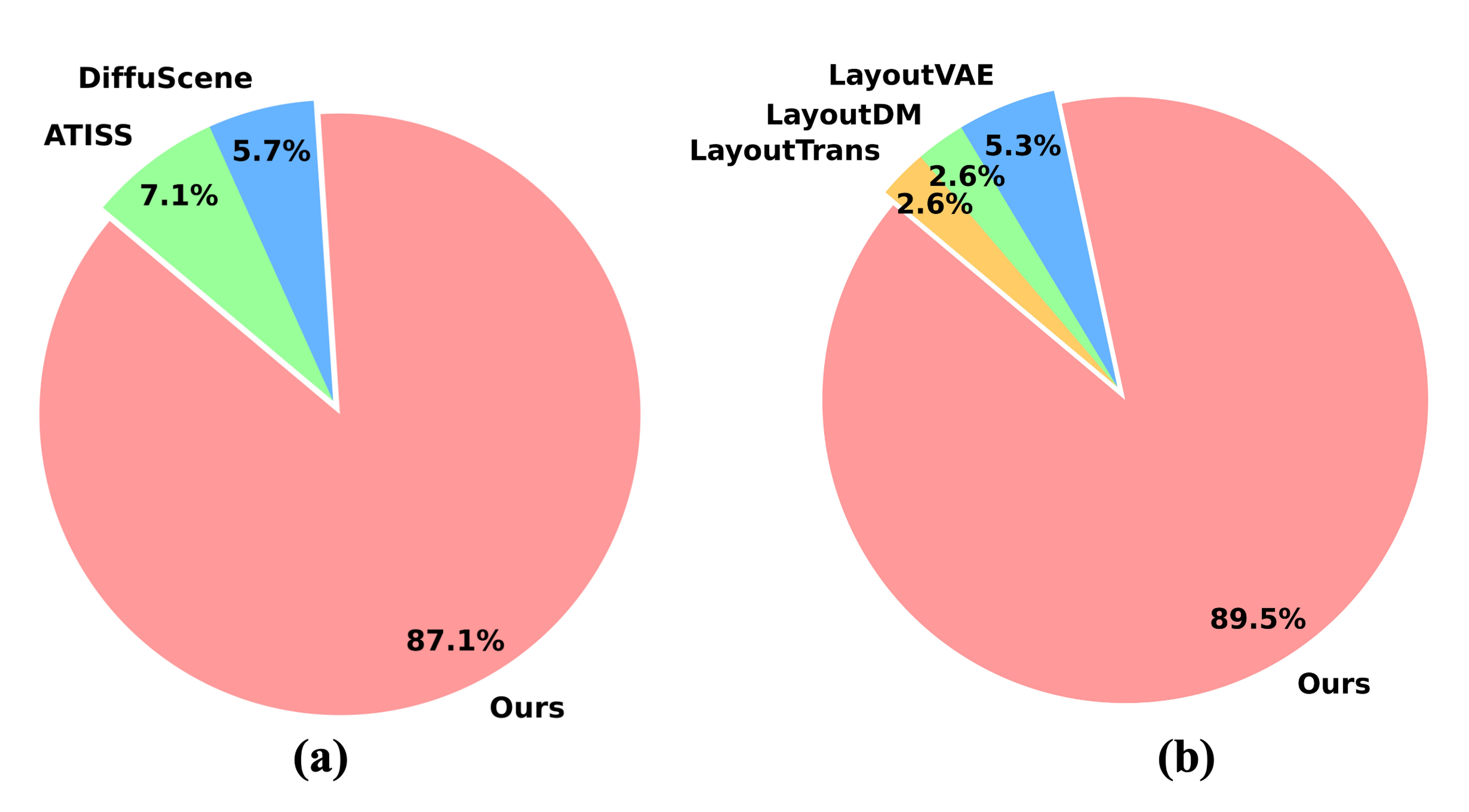}
    \caption{User study on layout synthesis tasks, encompassing (a) 3D scene generation and (b) 2D poster design.}
    \label{fig:user_study}
\end{figure}

For the evaluation of both 3D scene and 2D poster layout synthesis, a total of 360 questionnaires were gathered from 20 participants.
This comprised 200 questionnaires for 3D scene layout synthesis and 160 for 2D poster layout synthesis.
In each instance, participants were tasked with selecting the optimal synthesis result based on its fidelity to the given instructions.
A method's User Choice (UC) score is defined as the percentage of times its output was deemed superior by the participants, and the UC scores for all compared methods sum to 100\% consequently.
As shown in Figure~\ref{fig:user_study}, the empirical results demonstrate that \textsc{InstructLayout} achieved markedly higher preference rates over baseline methods, with user preference scores of 87.1\% for 3D scene synthesis and 89.5\% for 2D poster layout synthesis tasks, respectively.

\section{Qualitative Results}\label{apx:results}

\subsection{Instruction-driven 3D Scene Synthesis}\label{apx:results_3d}
\begin{figure*}[htbp]
    \centering
    \begin{subfigure}{0.24\textwidth}
        \includegraphics[width=\textwidth]{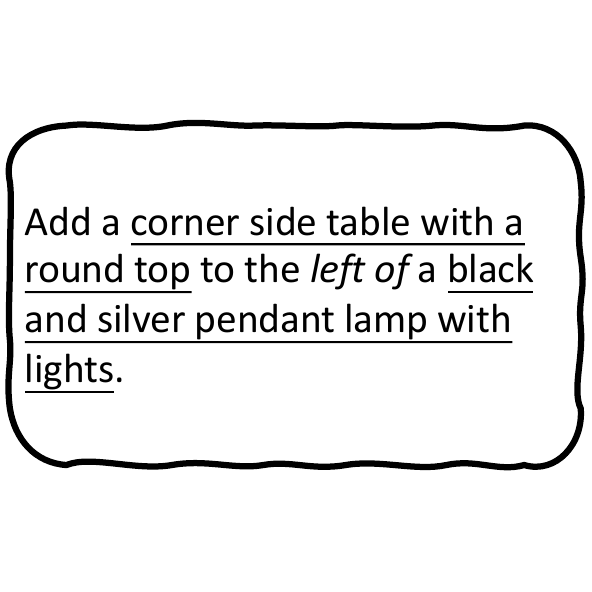}
        \includegraphics[width=\textwidth]{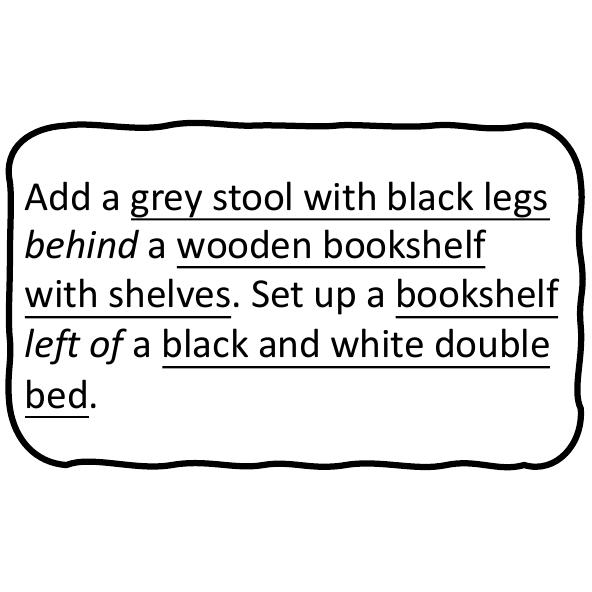}
        \includegraphics[width=\textwidth]{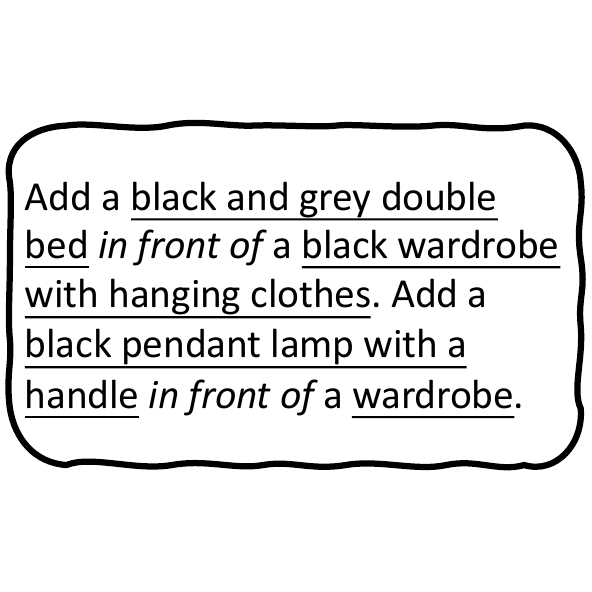}
        \includegraphics[width=\textwidth]{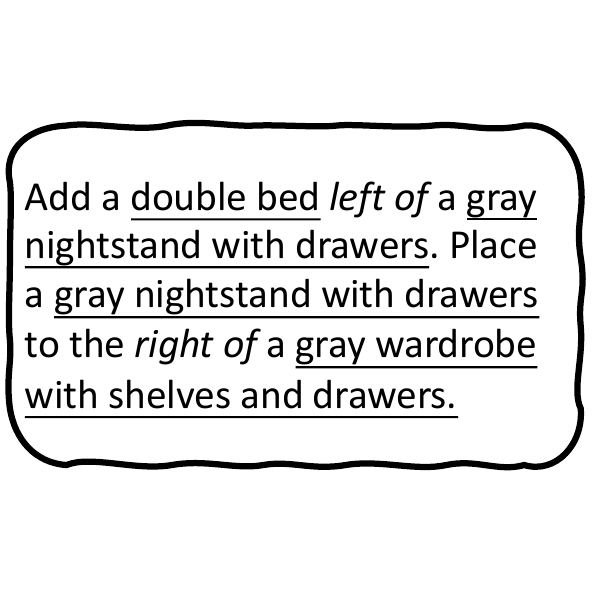}
        \includegraphics[width=\textwidth]{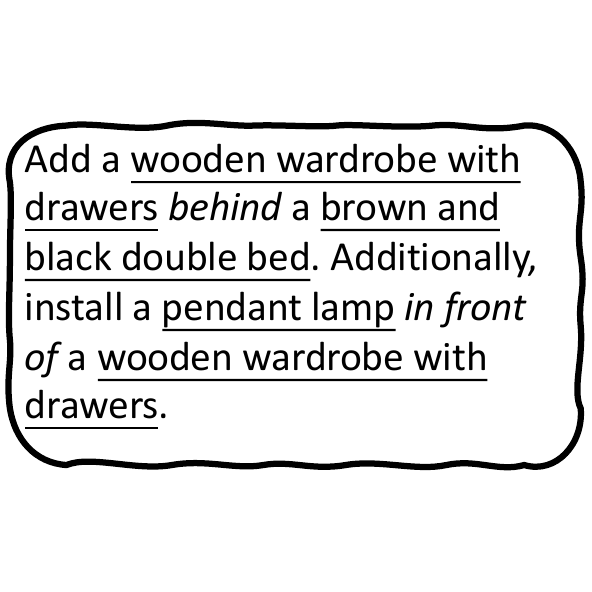}
        \caption{Instructions}
    \end{subfigure}
    \hfill
    \begin{subfigure}{0.24\textwidth}
        \includegraphics[width=\textwidth]{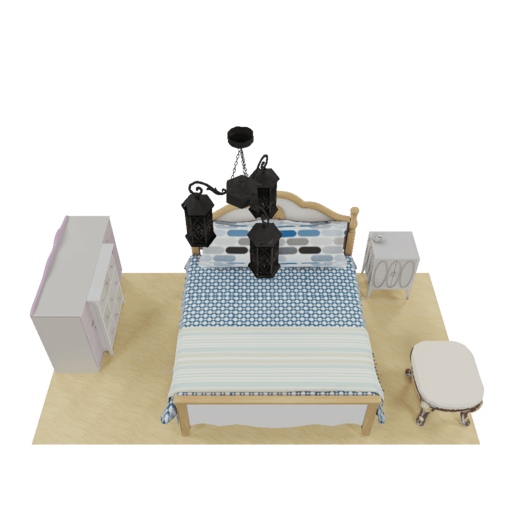}
        \includegraphics[width=\textwidth]{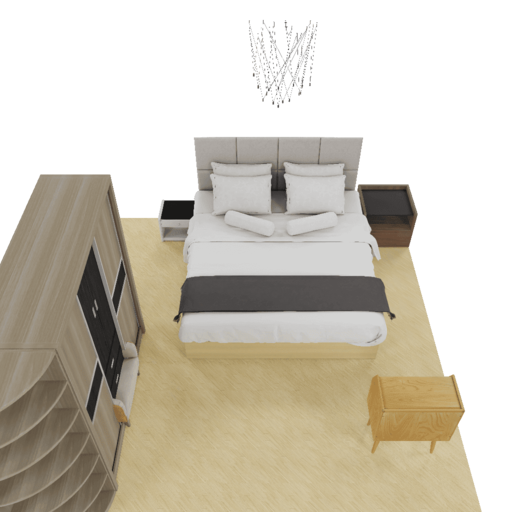}
        \includegraphics[width=\textwidth]{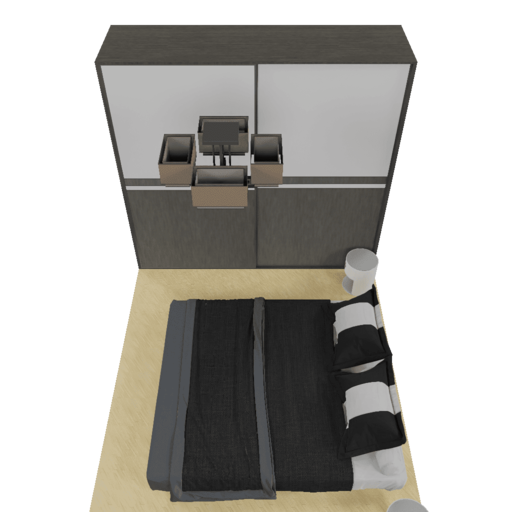}
        \includegraphics[width=\textwidth]{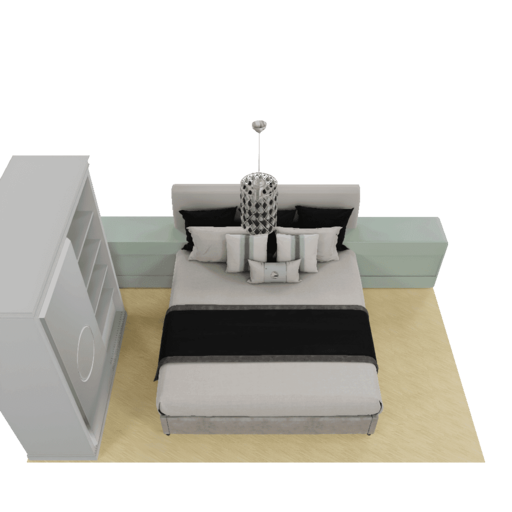}
        \includegraphics[width=\textwidth]{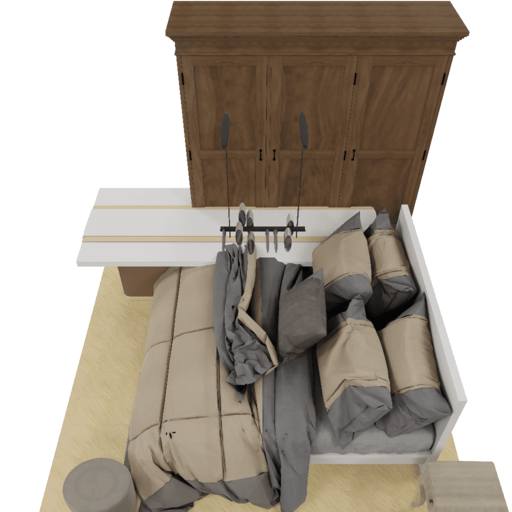}
        \caption{ATISS}
    \end{subfigure}
    \hfill
    \begin{subfigure}{0.24\textwidth}
        \includegraphics[width=\textwidth]{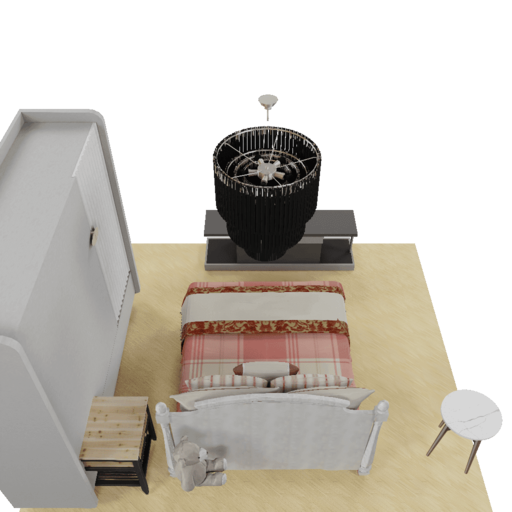}
        \includegraphics[width=\textwidth]{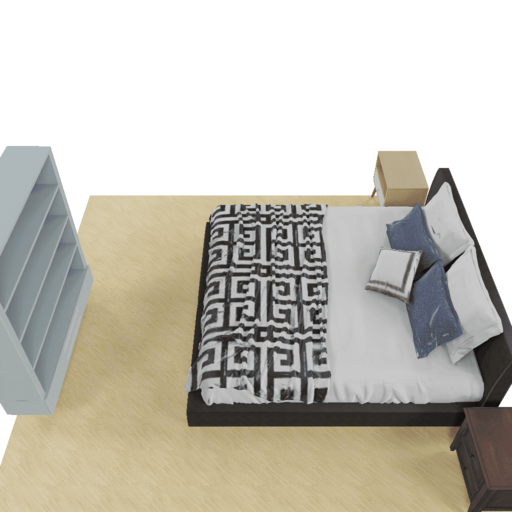}
        \includegraphics[width=\textwidth]{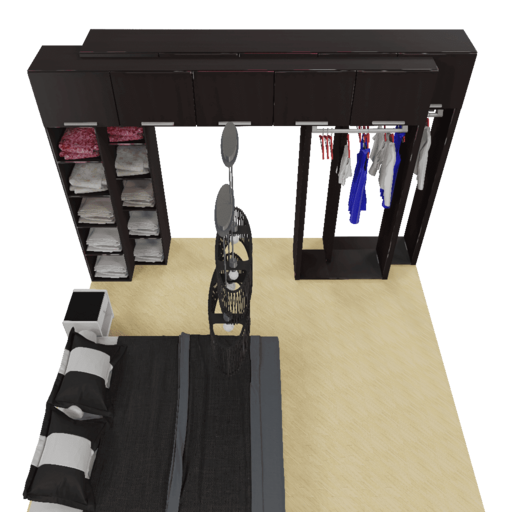}
        \includegraphics[width=\textwidth]{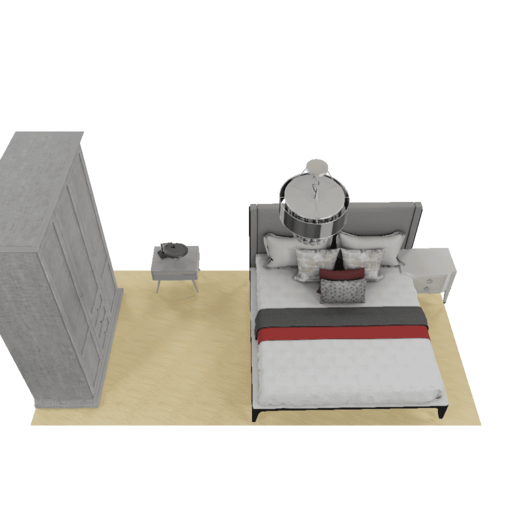}
        \includegraphics[width=\textwidth]{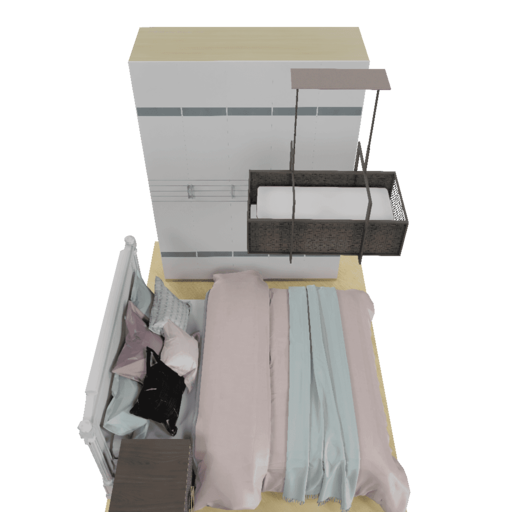}
        \caption{DiffuScene}
    \end{subfigure}
    \hfill
    \begin{subfigure}{0.24\textwidth}
        \includegraphics[width=\textwidth]{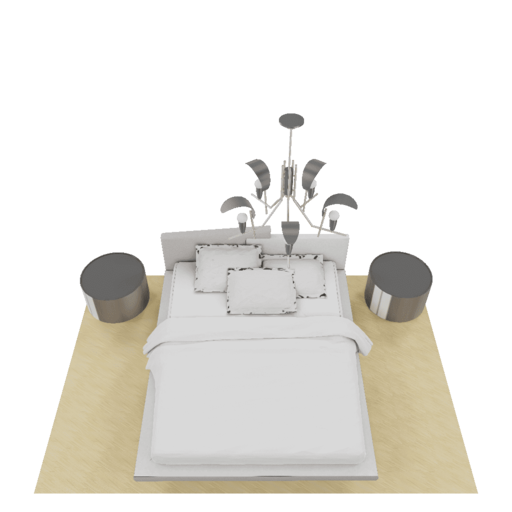}
        \includegraphics[width=\textwidth]{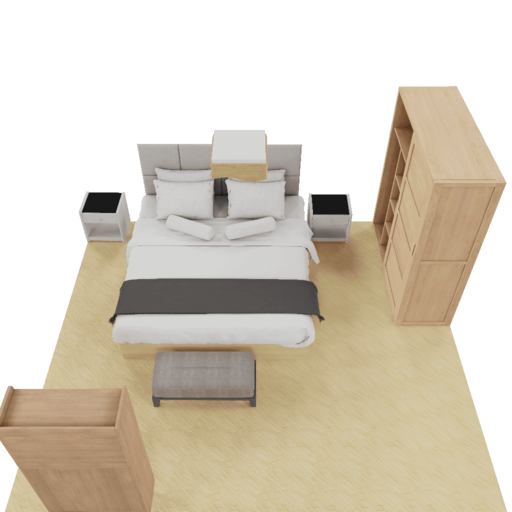}
        \includegraphics[width=\textwidth]{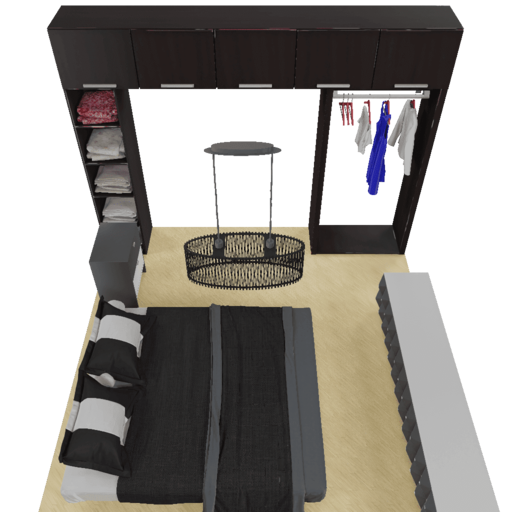}
        \includegraphics[width=\textwidth]{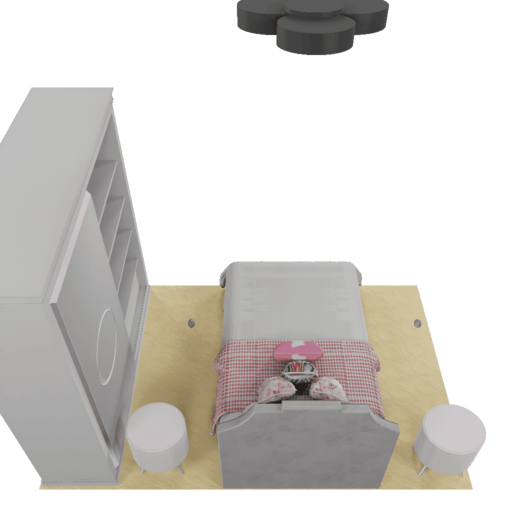}
        \includegraphics[width=\textwidth]{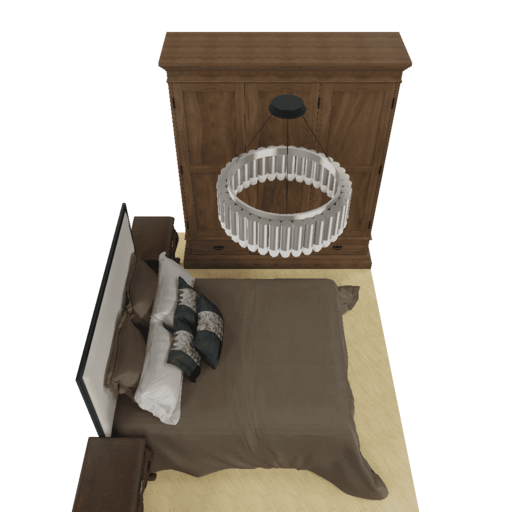}
        \caption{Ours}
    \end{subfigure}
    \caption{Visualizations for instruction-drive synthesized 3D bedrooms by ATISS~\citep{paschalidou2021atiss}, DiffuScene~\citep{tang2024diffuscene} and our method.}
    \label{fig:scenesyn_vis_bedroom}
\end{figure*}

\begin{figure*}[htbp]
    \centering
    \begin{subfigure}{0.24\textwidth}
        \includegraphics[width=\textwidth]{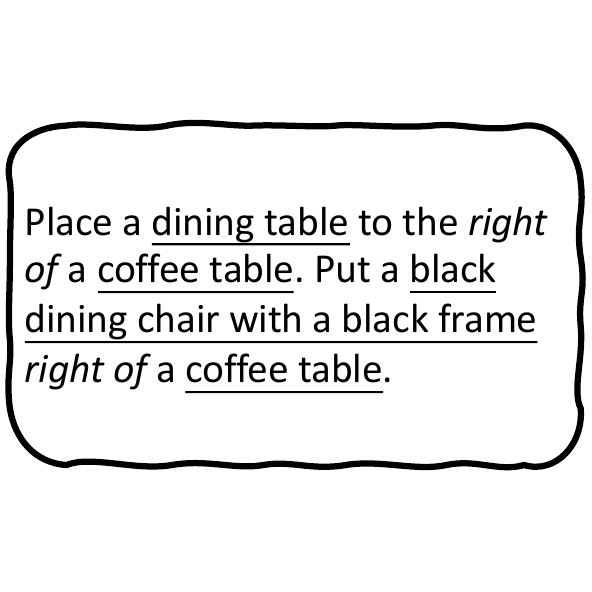}
        \includegraphics[width=\textwidth]{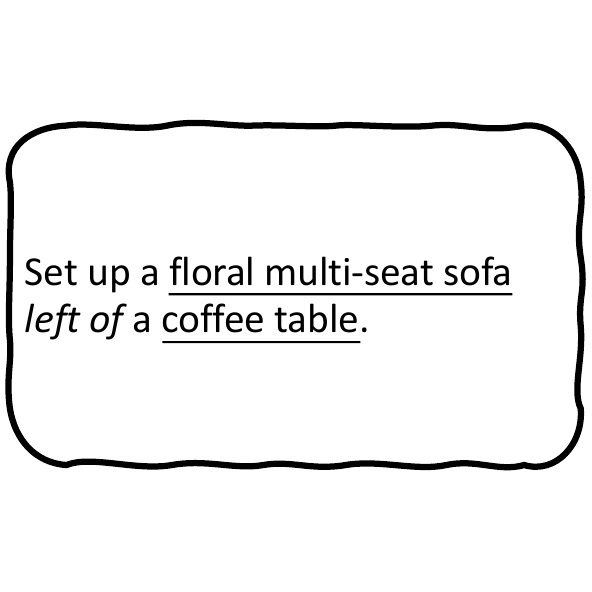}
        \includegraphics[width=\textwidth]{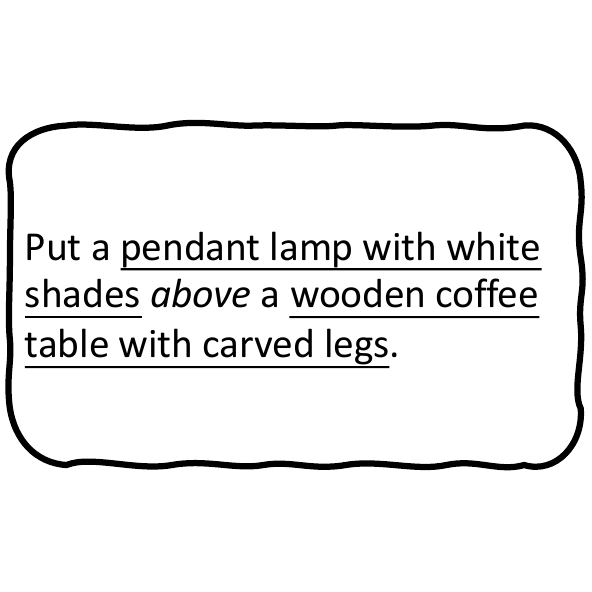}
        \includegraphics[width=\textwidth]{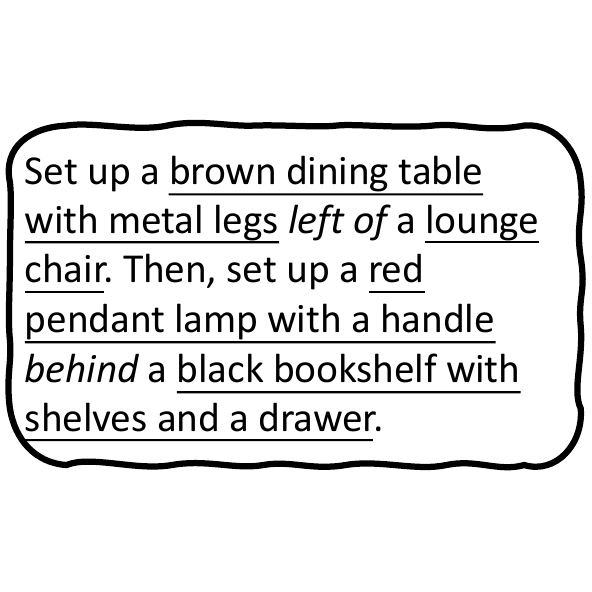}
        \includegraphics[width=\textwidth]{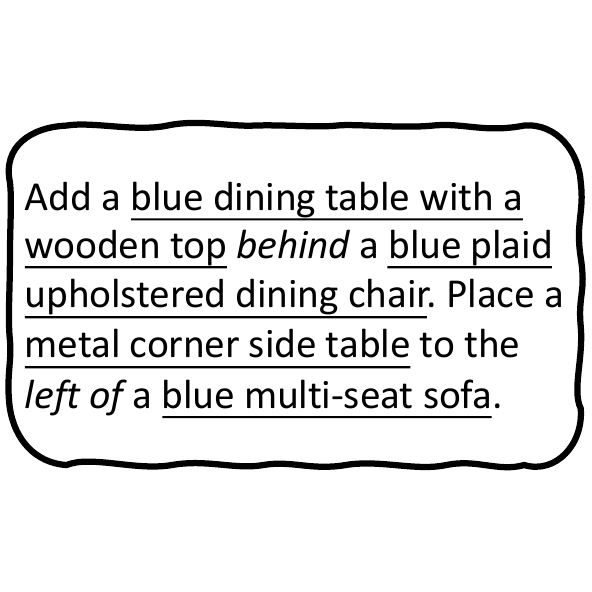}
        \caption{Instructions}
    \end{subfigure}
    \hfill
    \begin{subfigure}{0.24\textwidth}
        \includegraphics[width=\textwidth]{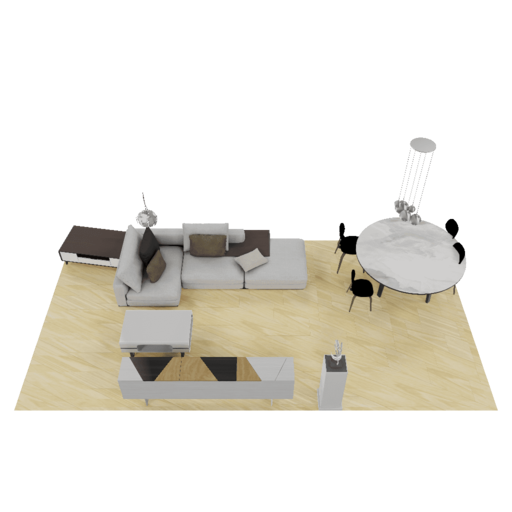}
        \includegraphics[width=\textwidth]{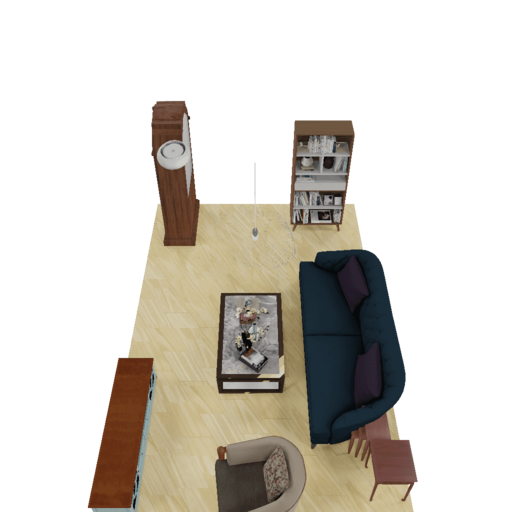}
        \includegraphics[width=\textwidth]{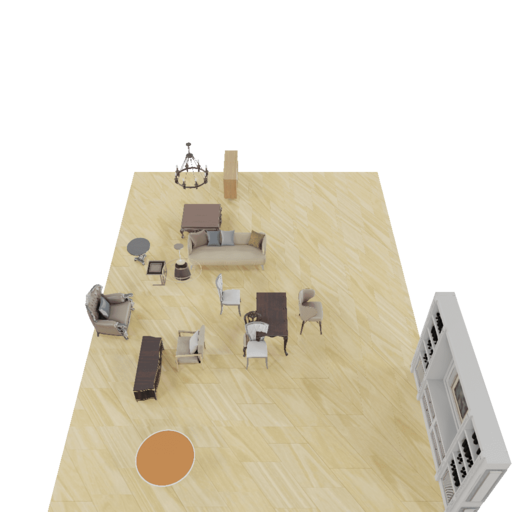}
        \includegraphics[width=\textwidth]{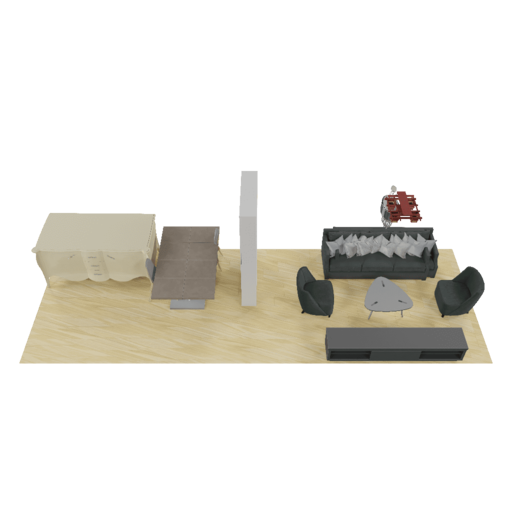}
        \includegraphics[width=\textwidth]{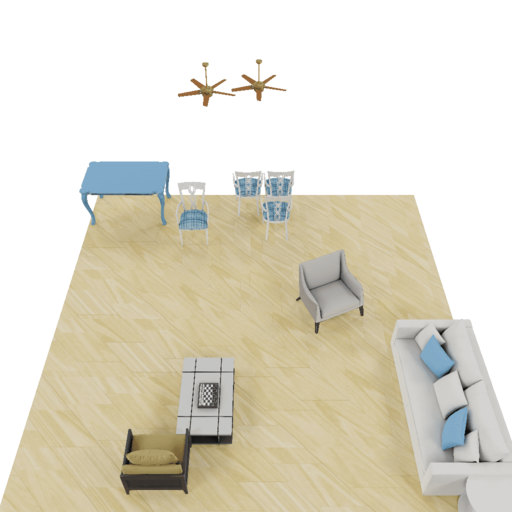}
        \caption{ATISS}
    \end{subfigure}
    \hfill
    \begin{subfigure}{0.24\textwidth}
        \includegraphics[width=\textwidth]{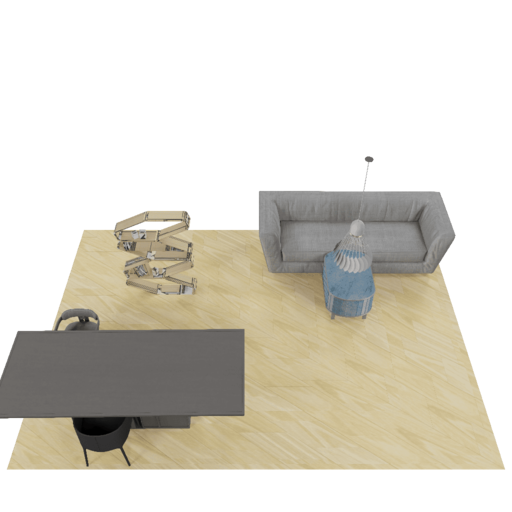}
        \includegraphics[width=\textwidth]{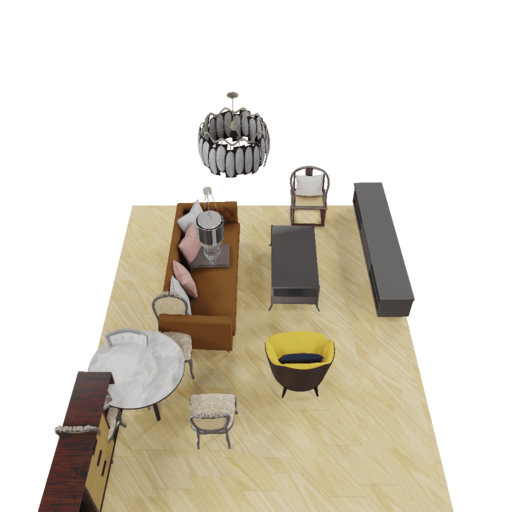}
        \includegraphics[width=\textwidth]{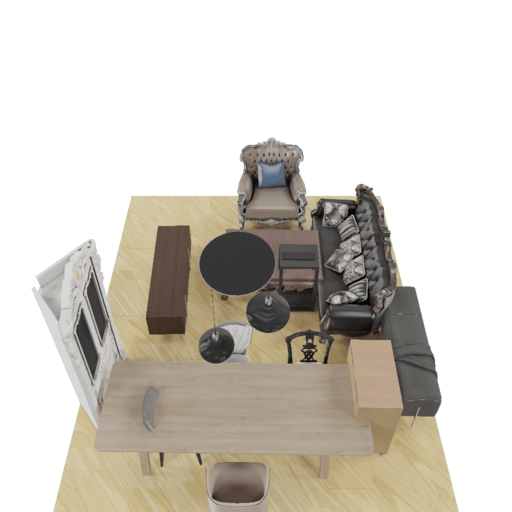}
        \includegraphics[width=\textwidth]{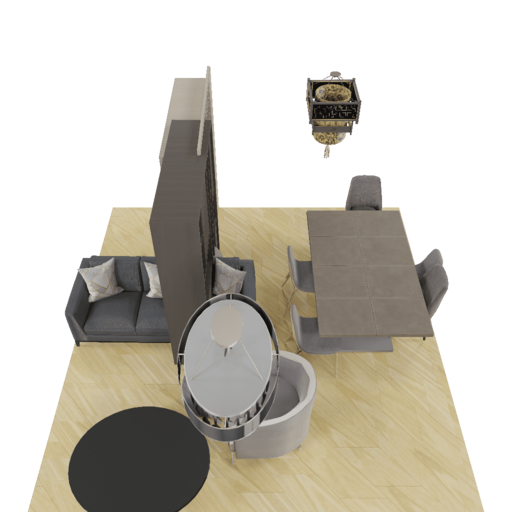}
        \includegraphics[width=\textwidth]{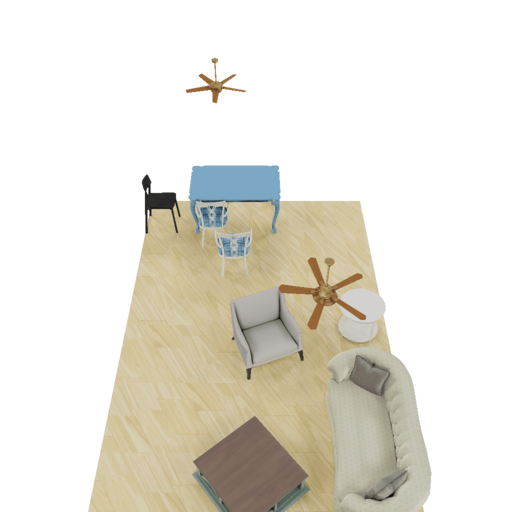}
        \caption{DiffuScene}
    \end{subfigure}
    \hfill
    \begin{subfigure}{0.24\textwidth}
        \includegraphics[width=\textwidth]{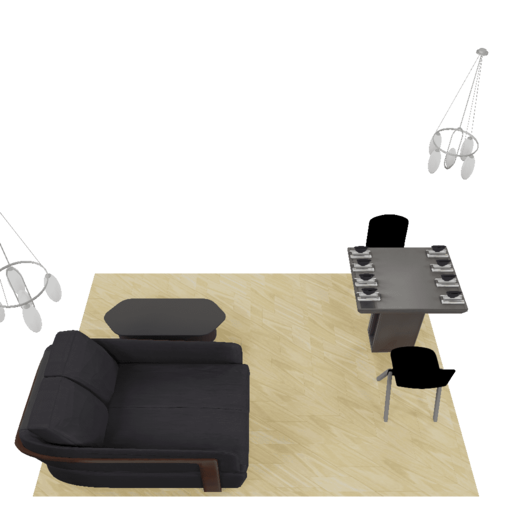}
        \includegraphics[width=\textwidth]{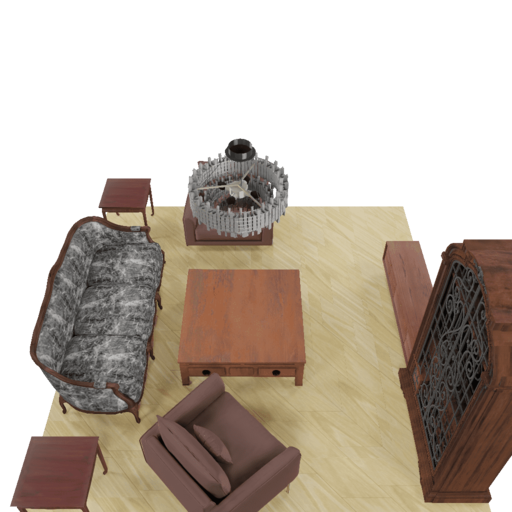}
        \includegraphics[width=\textwidth]{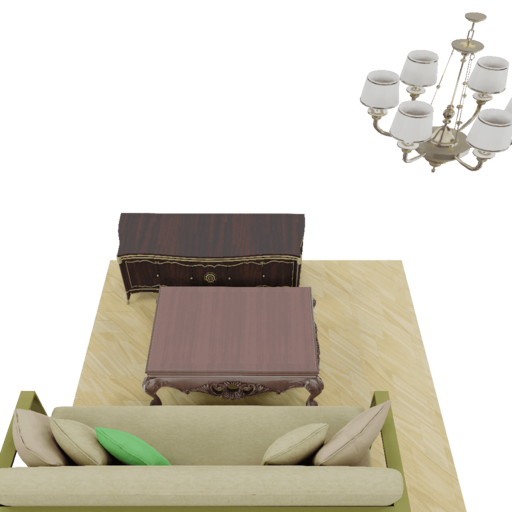}
        \includegraphics[width=\textwidth]{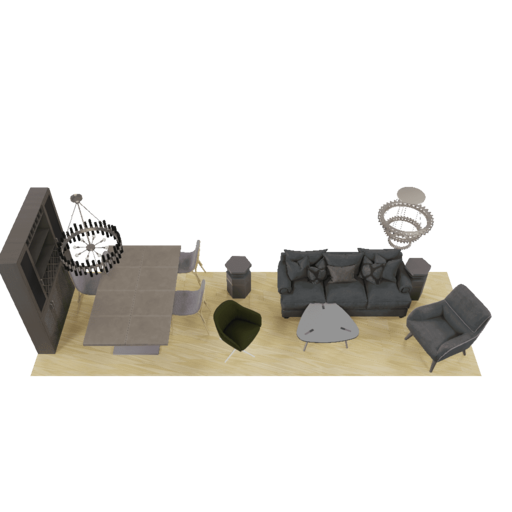}
        \includegraphics[width=\textwidth]{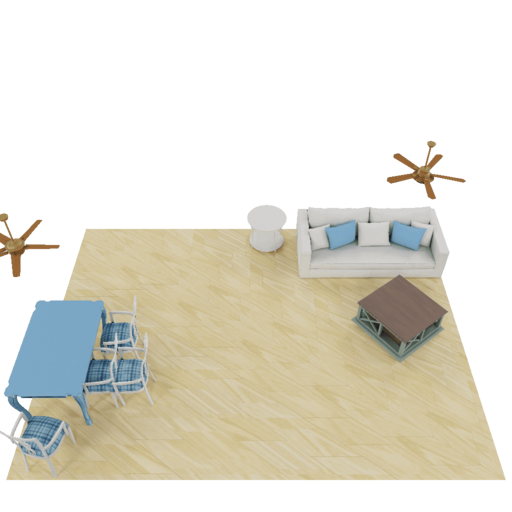}
        \caption{Ours}
    \end{subfigure}
    \caption{Visualizations for instruction-drive synthesized 3D living rooms by ATISS~\citep{paschalidou2021atiss}, DiffuScene~\citep{tang2024diffuscene} and our method.}
    \label{fig:scenesyn_vis_livingroom}
\end{figure*}

\begin{figure*}[htbp]
    \centering
    \begin{subfigure}{0.24\textwidth}
        \includegraphics[width=\textwidth]{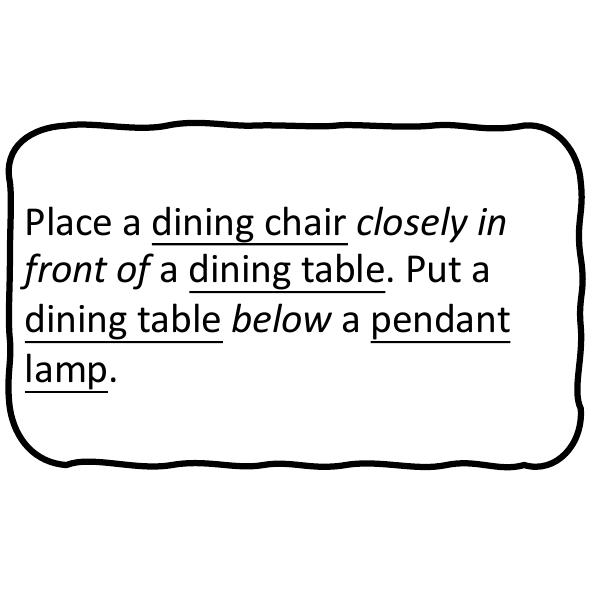}
        \includegraphics[width=\textwidth]{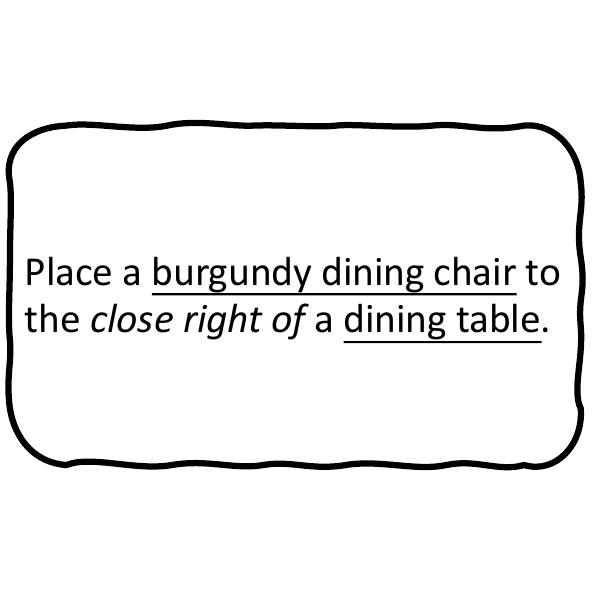}
        \includegraphics[width=\textwidth]{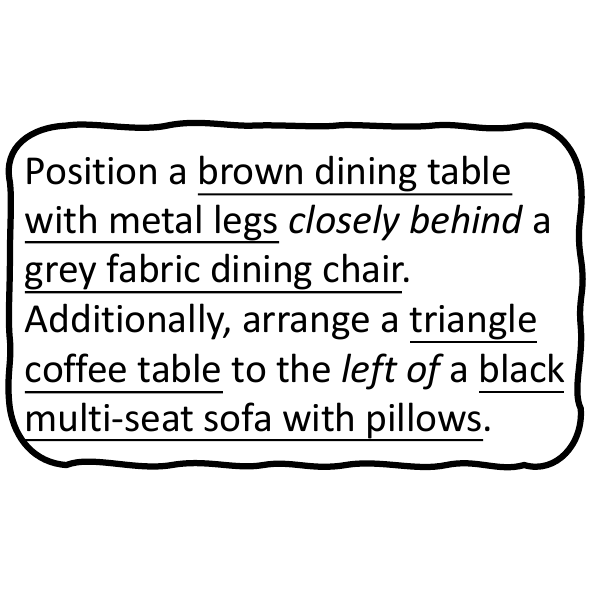}
        \includegraphics[width=\textwidth]{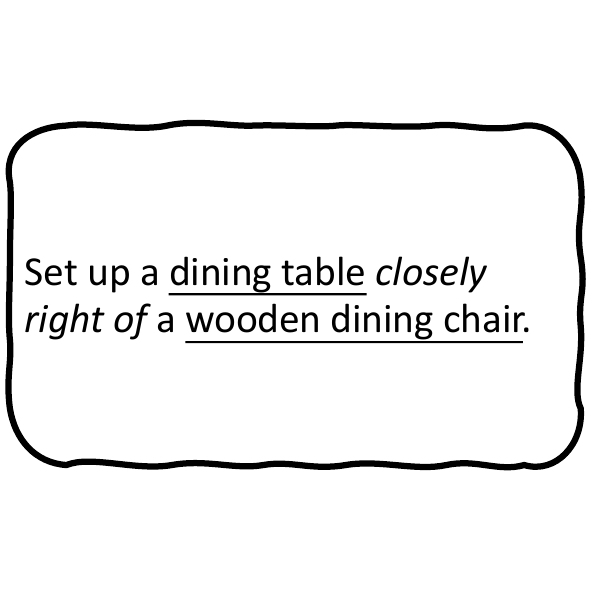}
        \includegraphics[width=\textwidth]{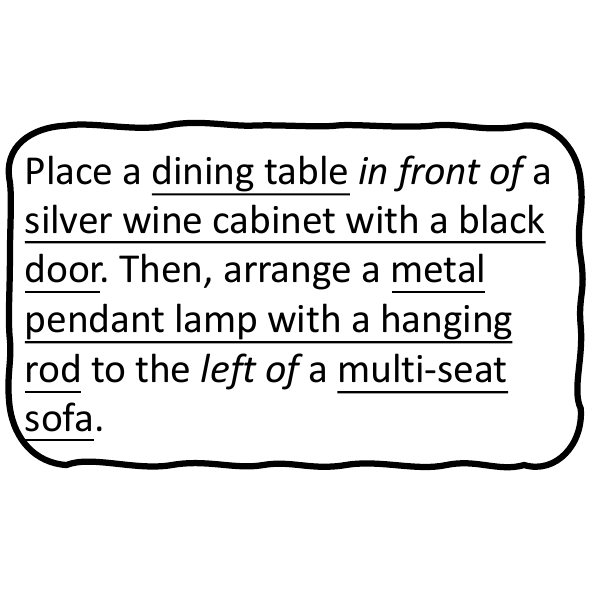}
        \caption{Instructions}
    \end{subfigure}
    \hfill
    \begin{subfigure}{0.24\textwidth}
        \includegraphics[width=\textwidth]{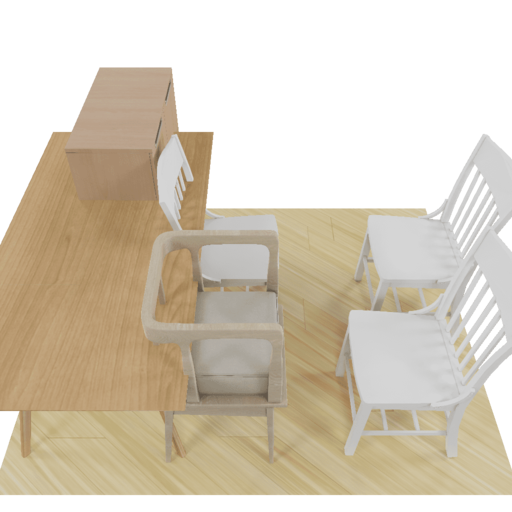}
        \includegraphics[width=\textwidth]{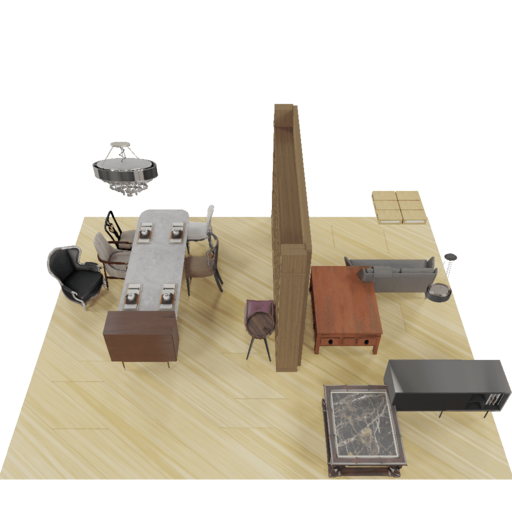}
        \includegraphics[width=\textwidth]{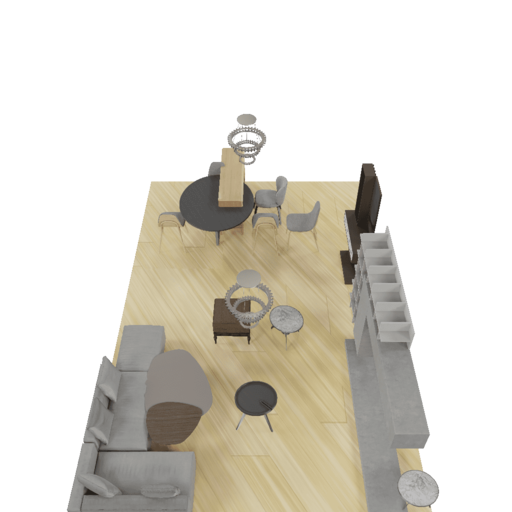}
        \includegraphics[width=\textwidth]{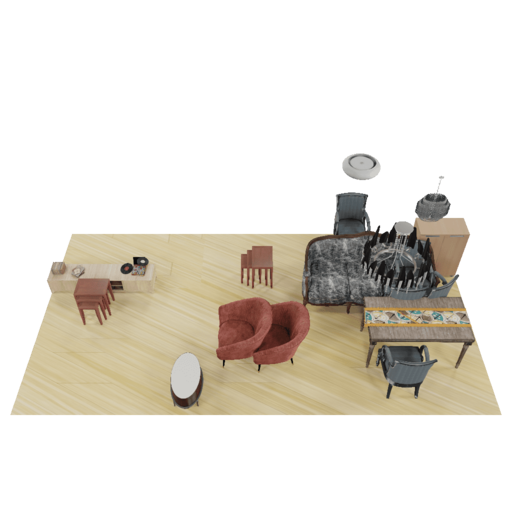}
        \includegraphics[width=\textwidth]{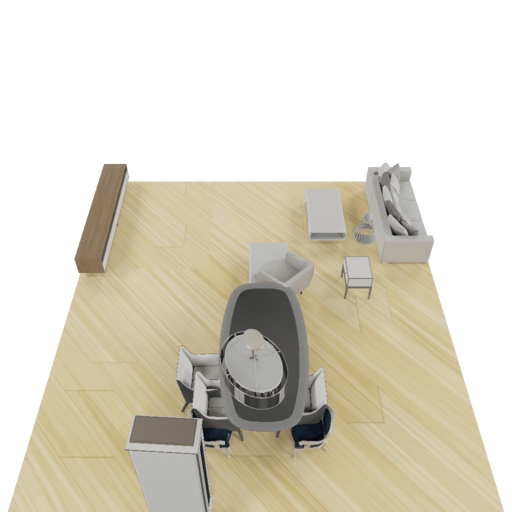}
        \caption{ATISS}
    \end{subfigure}
    \hfill
    \begin{subfigure}{0.24\textwidth}
        \includegraphics[width=\textwidth]{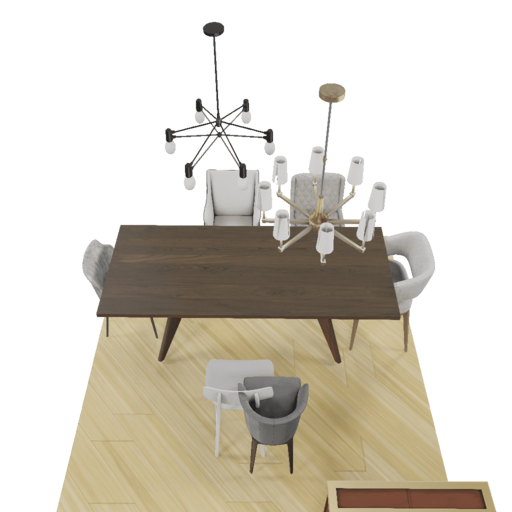}
        \includegraphics[width=\textwidth]{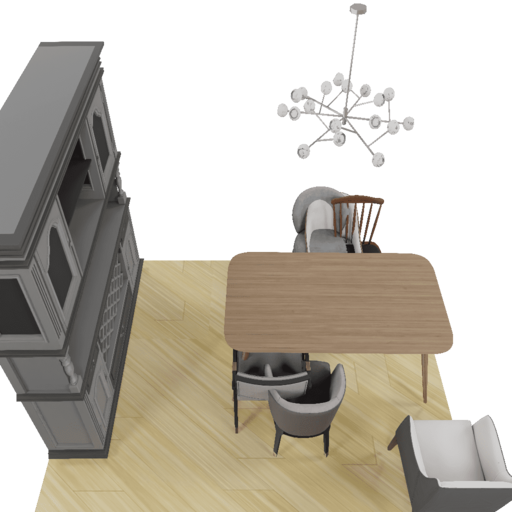}
        \includegraphics[width=\textwidth]{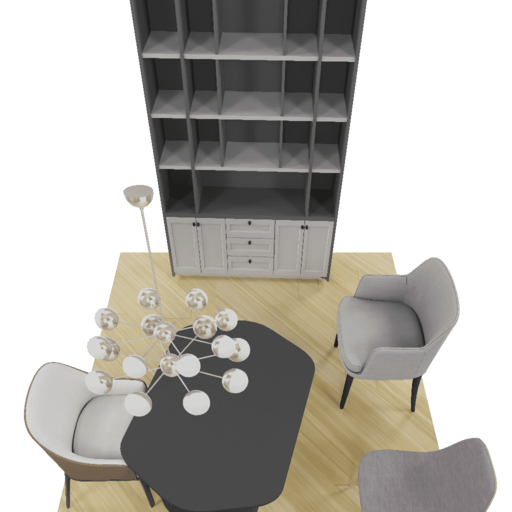}
        \includegraphics[width=\textwidth]{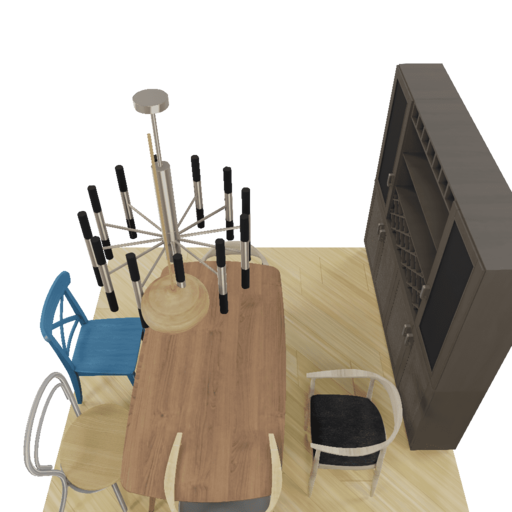}
        \includegraphics[width=\textwidth]{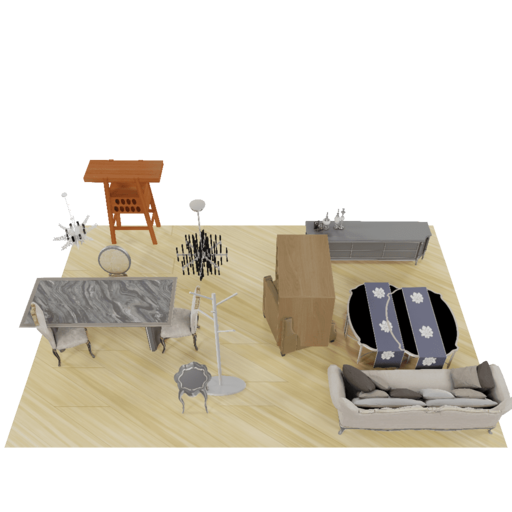}
        \caption{DiffuScene}
    \end{subfigure}
    \hfill
    \begin{subfigure}{0.24\textwidth}
        \includegraphics[width=\textwidth]{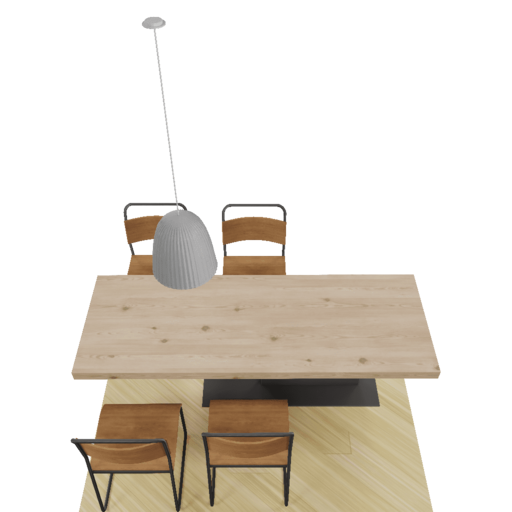}
        \includegraphics[width=\textwidth]{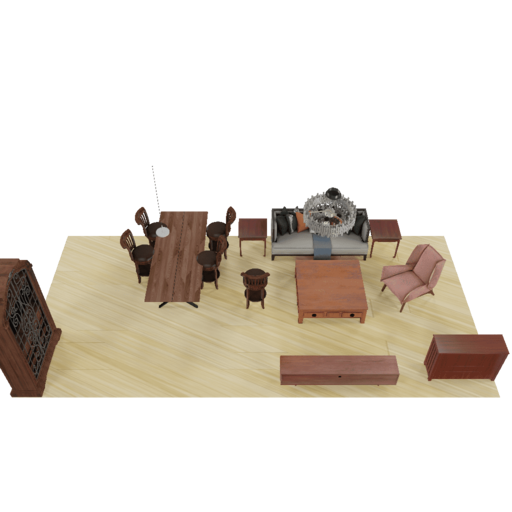}
        \includegraphics[width=\textwidth]{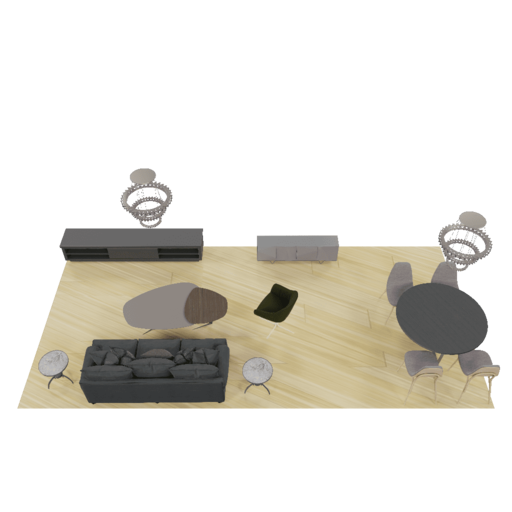}
        \includegraphics[width=\textwidth]{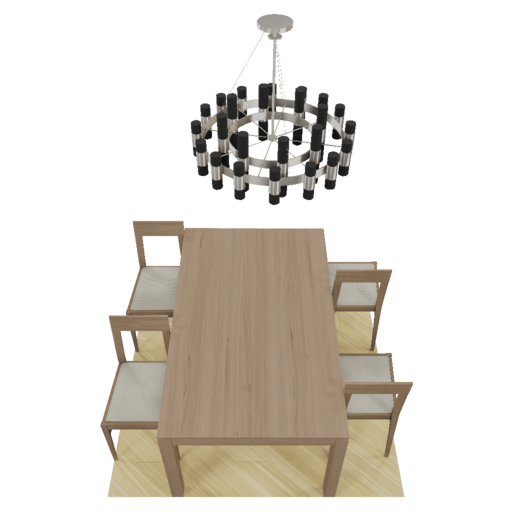}
        \includegraphics[width=\textwidth]{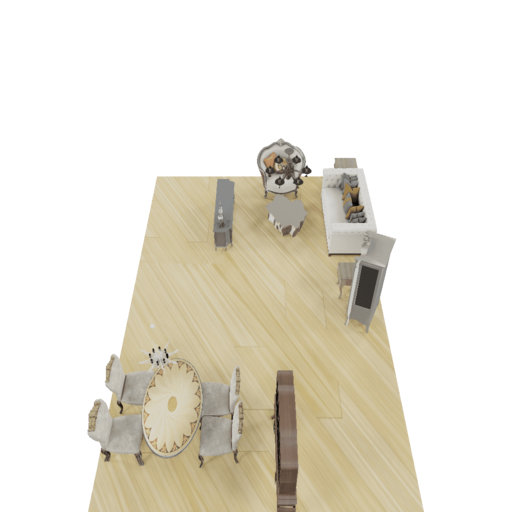}
        \caption{Ours}
    \end{subfigure}
    \caption{Visualizations for instruction-drive synthesized 3D dining rooms by ATISS~\citep{paschalidou2021atiss}, DiffuScene~\citep{tang2024diffuscene} and our method.}
    \label{fig:scenesyn_vis_diningroom}
\end{figure*}

We present visualizations of instruction-driven synthesized 3D scenes in Figure~\ref{fig:scenesyn_vis_bedroom}, \ref{fig:scenesyn_vis_livingroom} and \ref{fig:scenesyn_vis_diningroom}.
Besides the quantitative evaluations shown in Table~\ref{tab:scenesyn}, these qualitative visualizations also show the superiority of our method over previous state-of-the-art approaches in terms of adherence to instructions and generative quality.

\subsection{Zero-shot 3D Scene Applications}\label{apx:app_3d}
\begin{figure*}[htbp]
    \centering
    \begin{subfigure}{0.19\textwidth}
        \includegraphics[width=\textwidth]{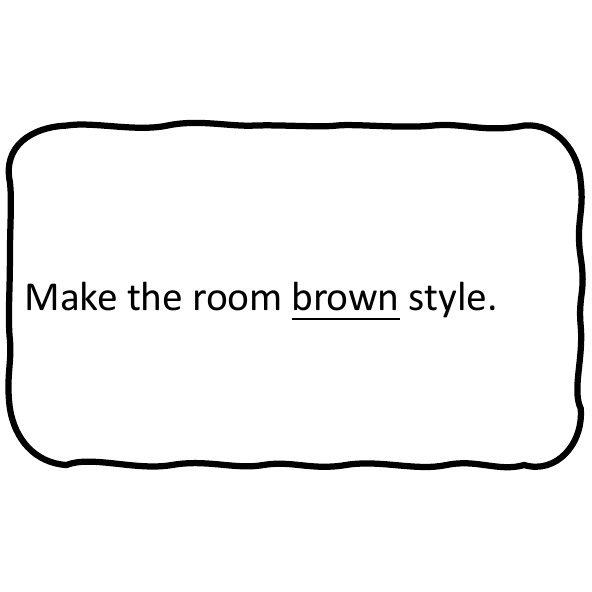}
        \includegraphics[width=\textwidth]{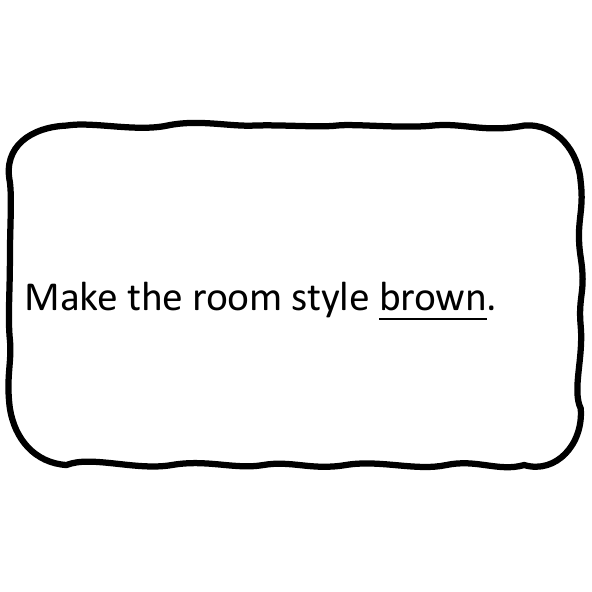}
        \includegraphics[width=\textwidth]{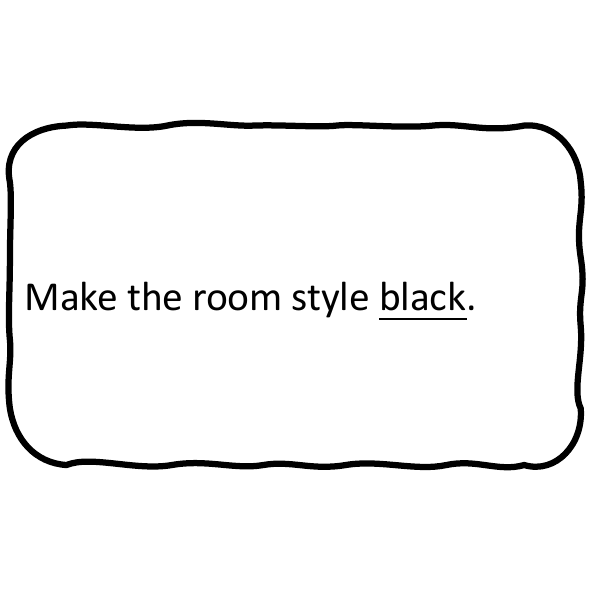}
        \includegraphics[width=\textwidth]{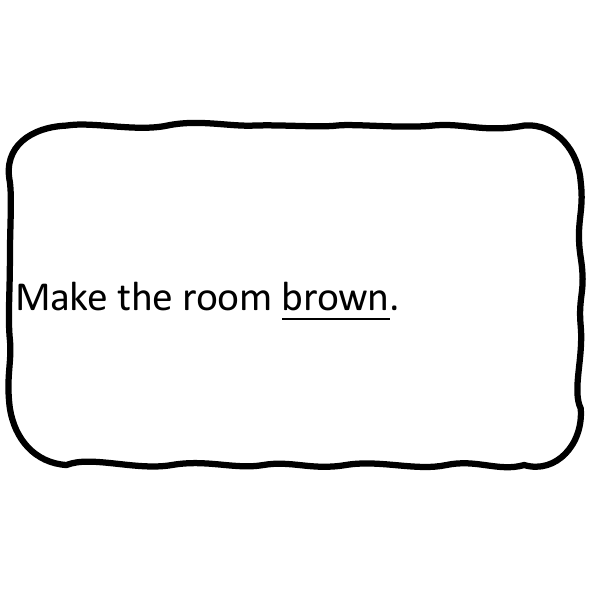}
        \includegraphics[width=\textwidth]{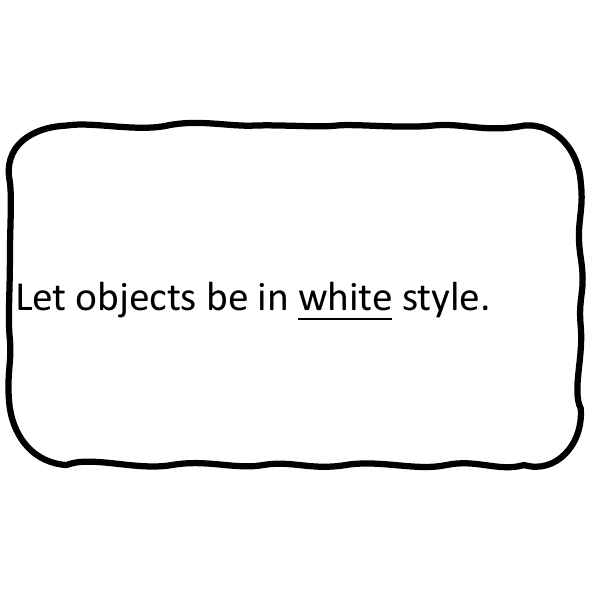}
        \includegraphics[width=\textwidth]{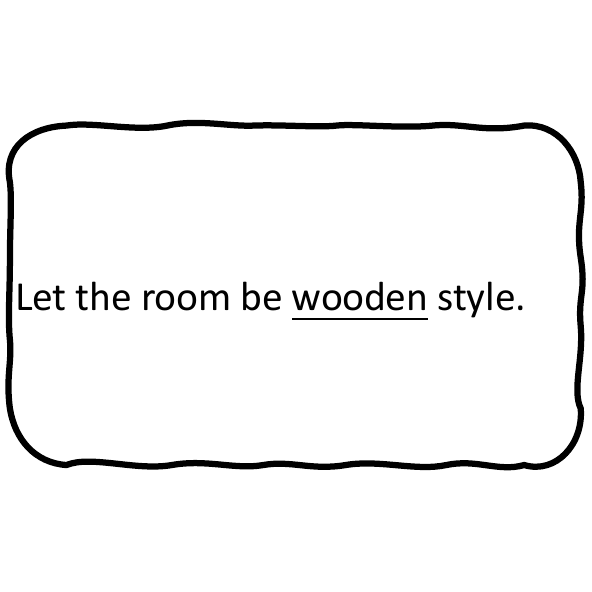}
        \includegraphics[width=\textwidth]{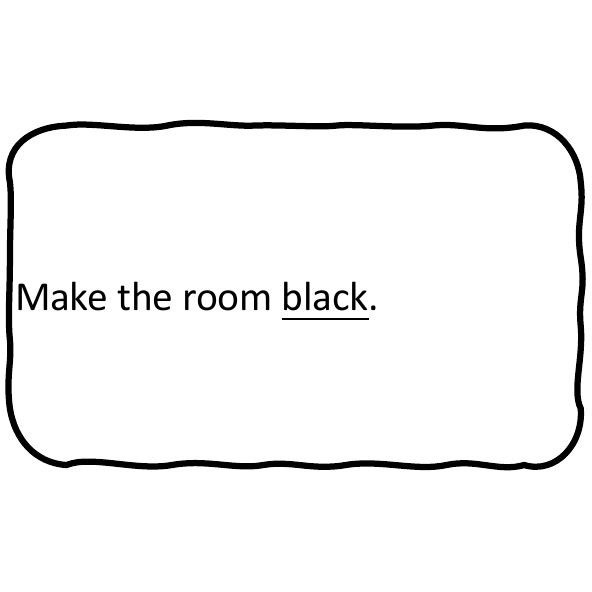}
        \caption{Instructions}
    \end{subfigure}
    \hfill
    \begin{subfigure}{0.19\textwidth}
        \includegraphics[width=\textwidth]{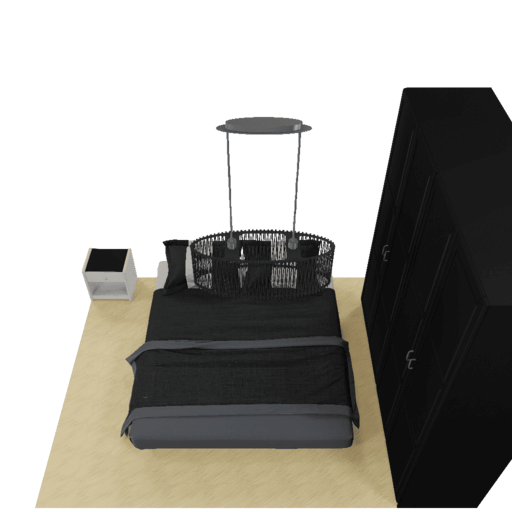}
        \includegraphics[width=\textwidth]{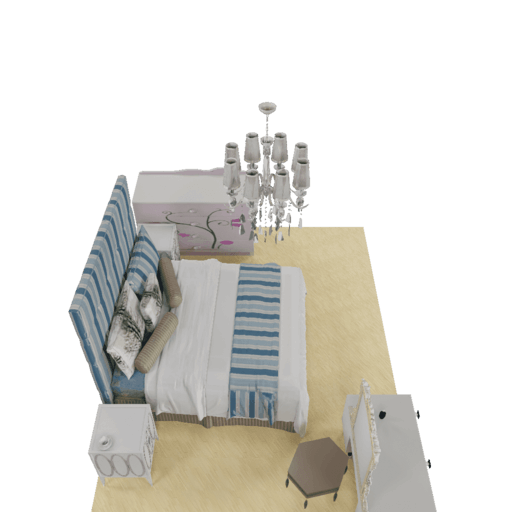}
        \includegraphics[width=\textwidth]{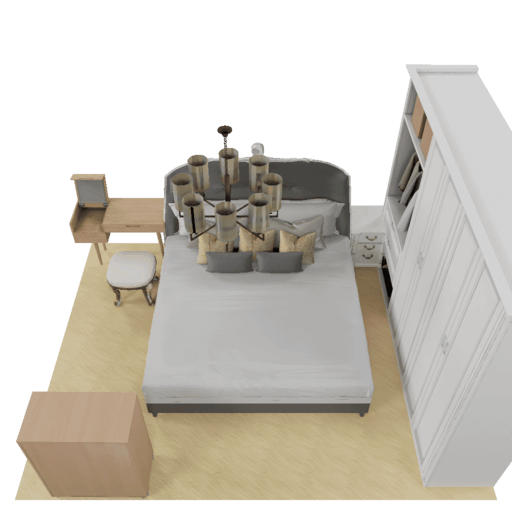}
        \includegraphics[width=\textwidth]{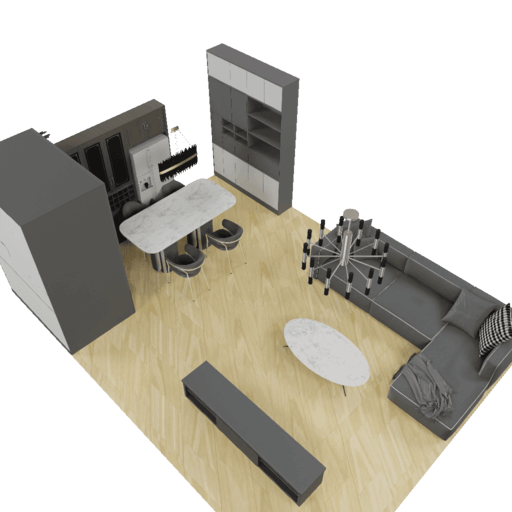}
        \includegraphics[width=\textwidth]{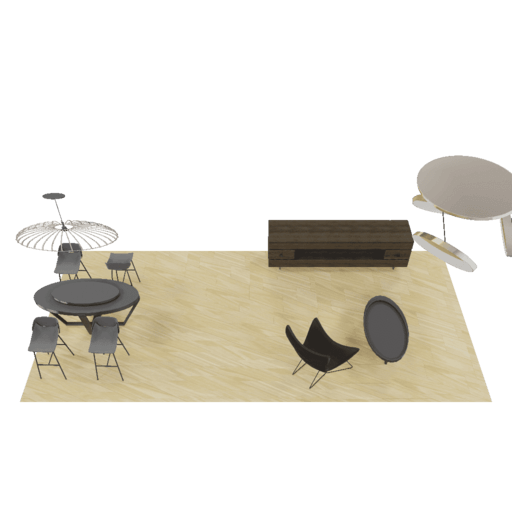}
        \includegraphics[width=\textwidth]{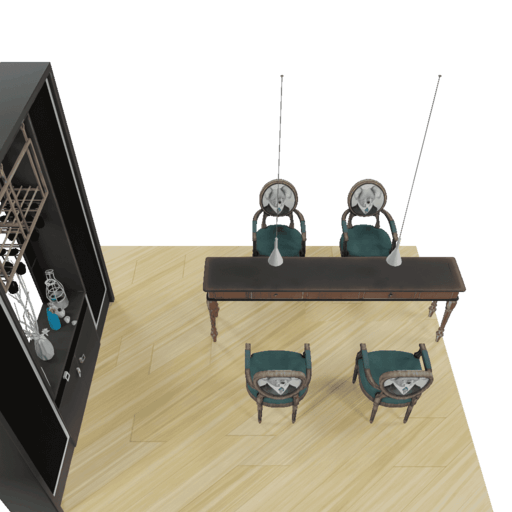}
        \includegraphics[width=\textwidth]{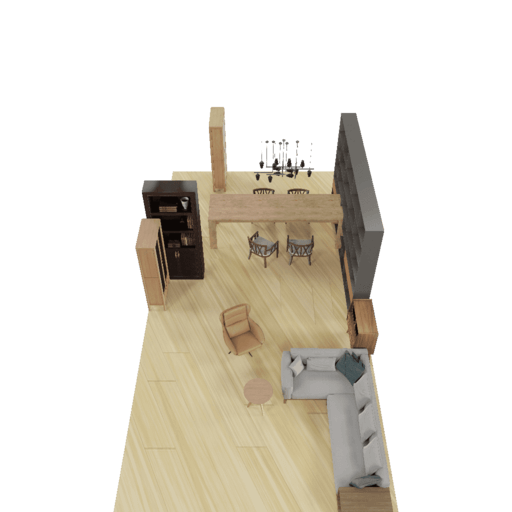}
        \caption{Original Scenes}
    \end{subfigure}
    \hfill
    \begin{subfigure}{0.19\textwidth}
        \includegraphics[width=\textwidth]{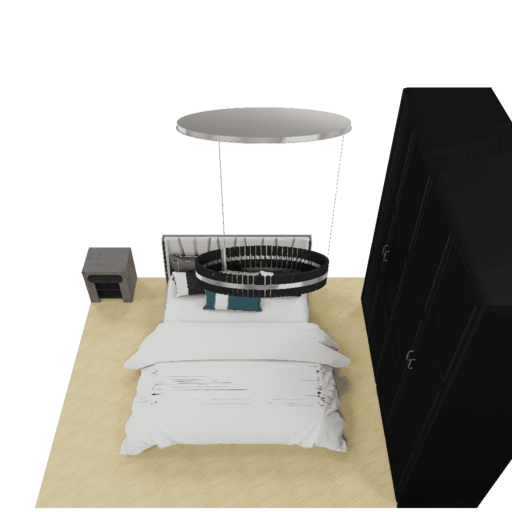}
        \includegraphics[width=\textwidth]{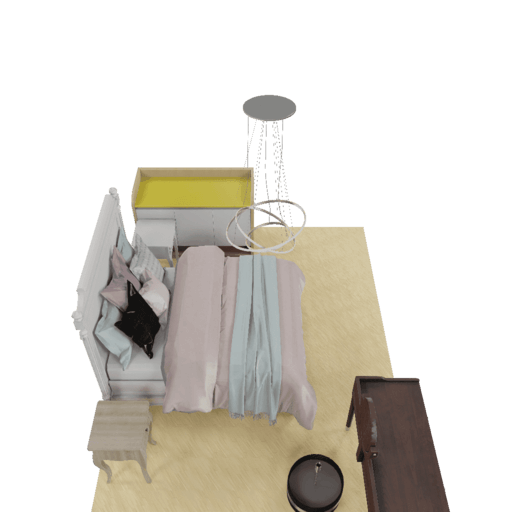}
        \includegraphics[width=\textwidth]{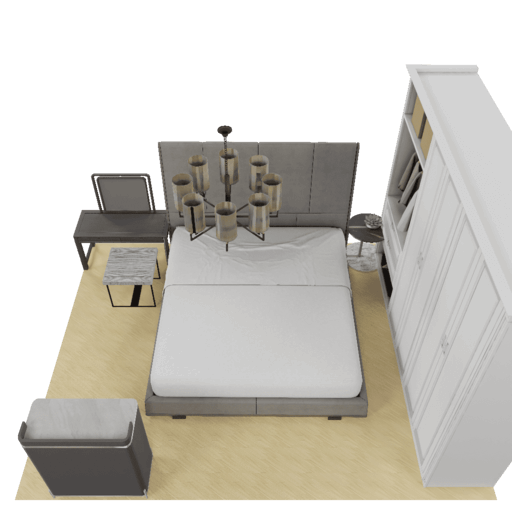}
        \includegraphics[width=\textwidth]{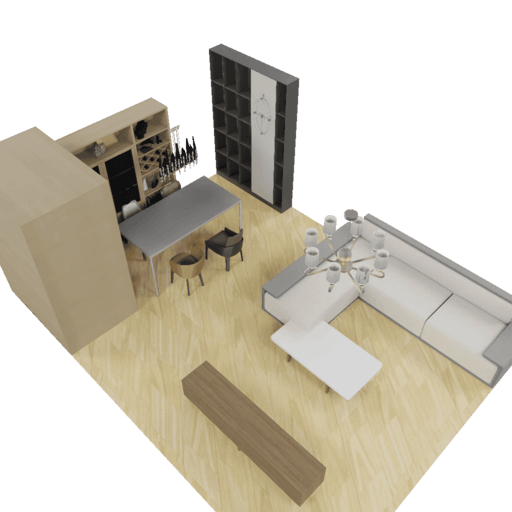}
        \includegraphics[width=\textwidth]{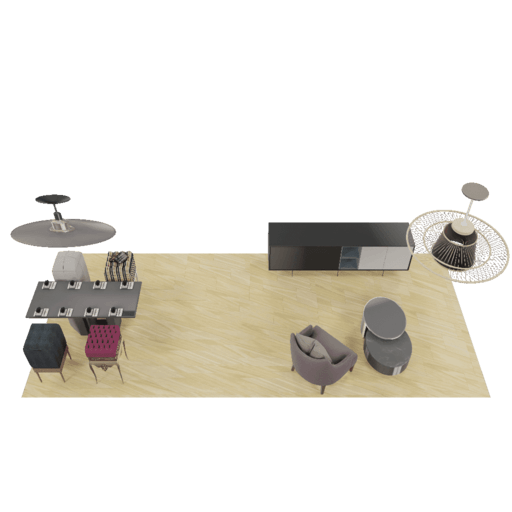}
        \includegraphics[width=\textwidth]{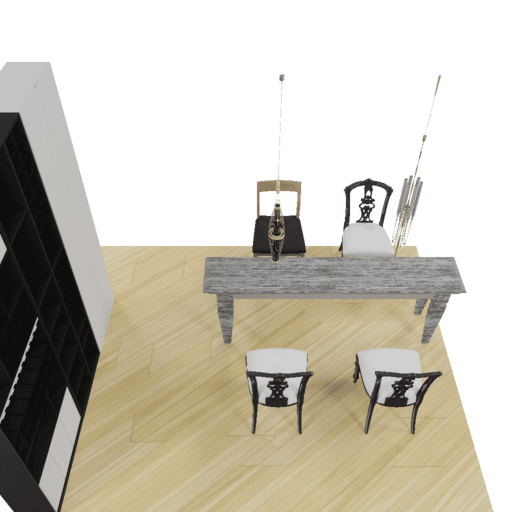}
        \includegraphics[width=\textwidth]{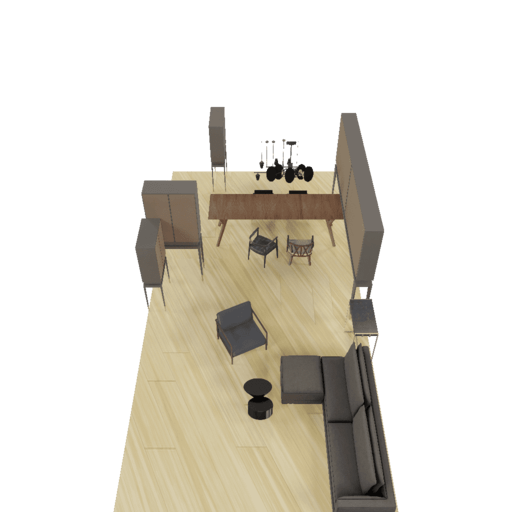}
        \caption{ATISS}
    \end{subfigure}
    \hfill
    \begin{subfigure}{0.19\textwidth}
        \includegraphics[width=\textwidth]{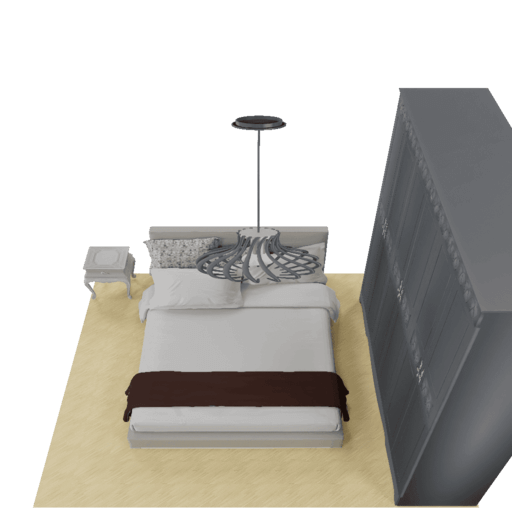}
        \includegraphics[width=\textwidth]{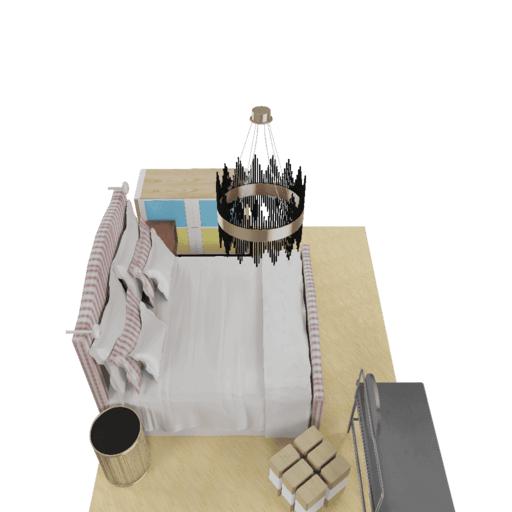}
        \includegraphics[width=\textwidth]{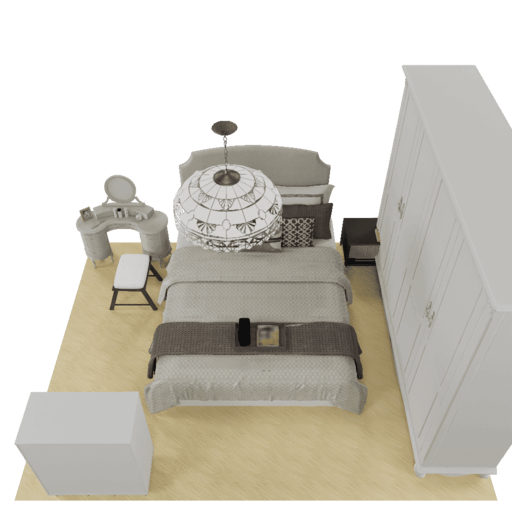}
        \includegraphics[width=\textwidth]{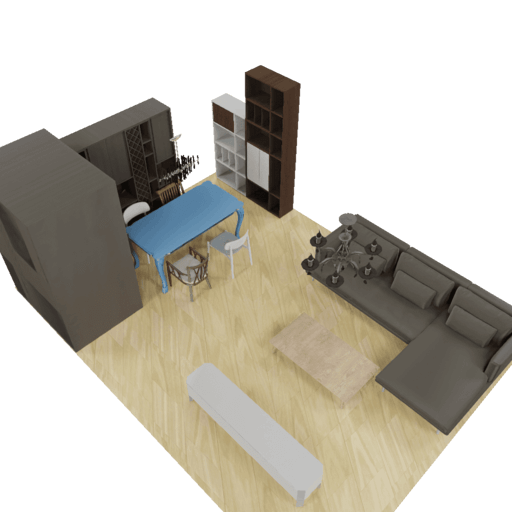}
        \includegraphics[width=\textwidth]{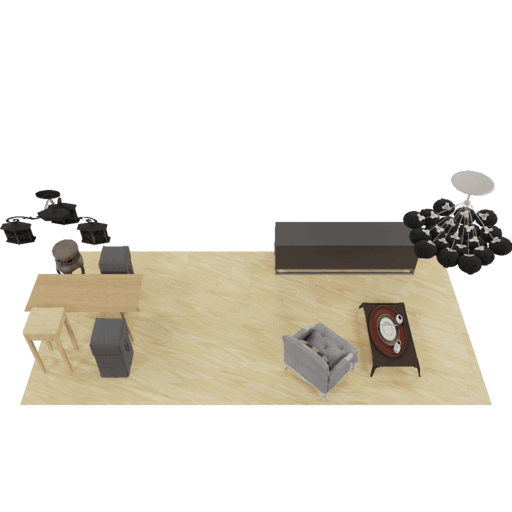}
        \includegraphics[width=\textwidth]{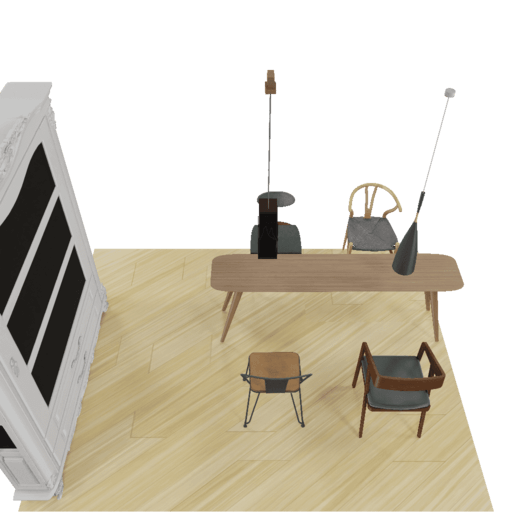}
        \includegraphics[width=\textwidth]{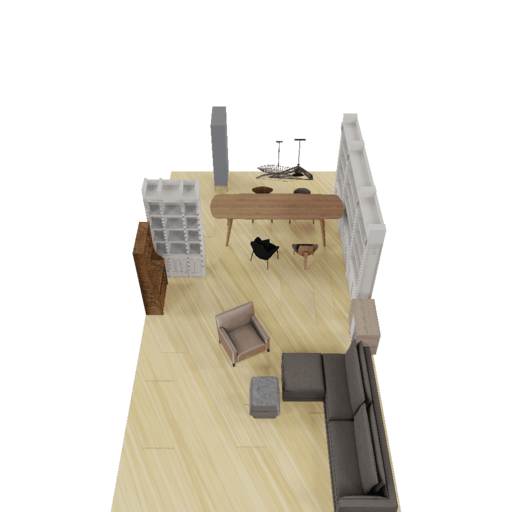}
        \caption{DiffuScene}
    \end{subfigure}
    \hfill
    \begin{subfigure}{0.19\textwidth}
        \includegraphics[width=\textwidth]{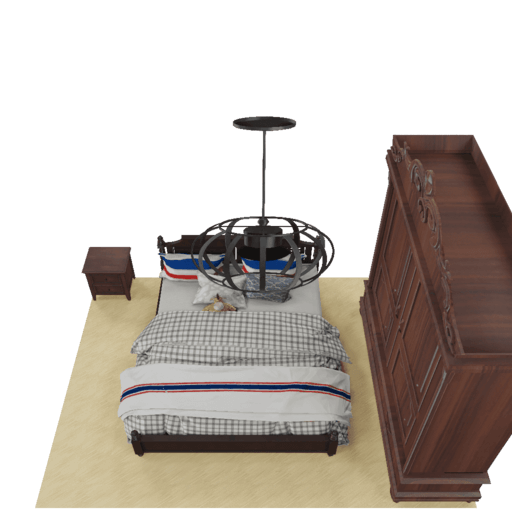}
        \includegraphics[width=\textwidth]{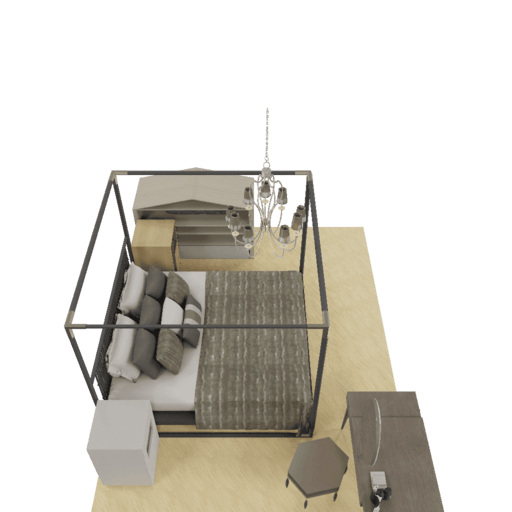}
        \includegraphics[width=\textwidth]{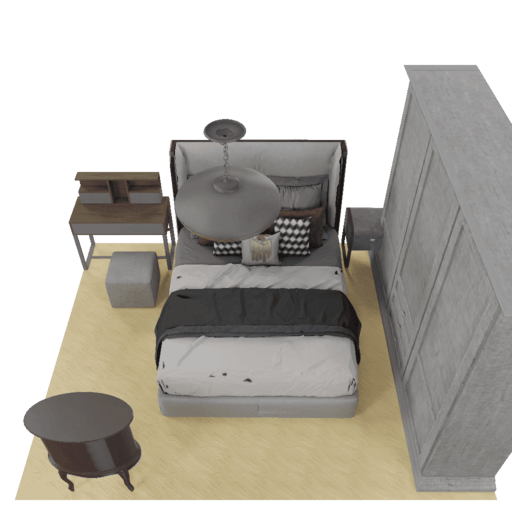}
        \includegraphics[width=\textwidth]{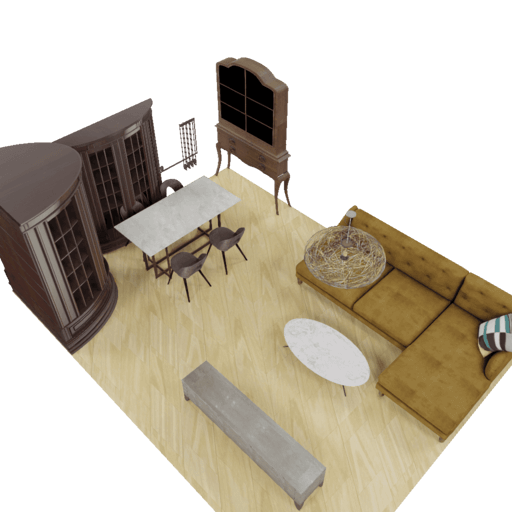}
        \includegraphics[width=\textwidth]{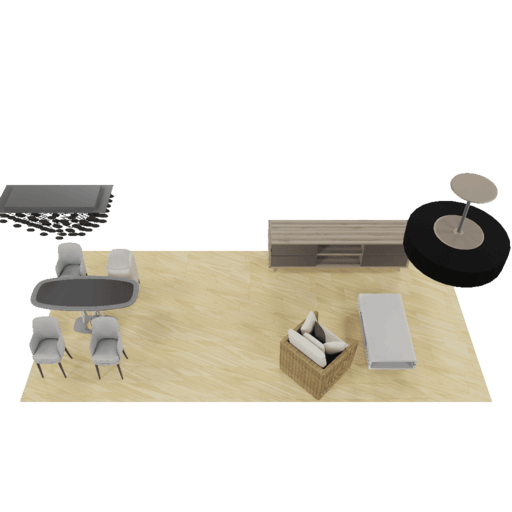}
        \includegraphics[width=\textwidth]{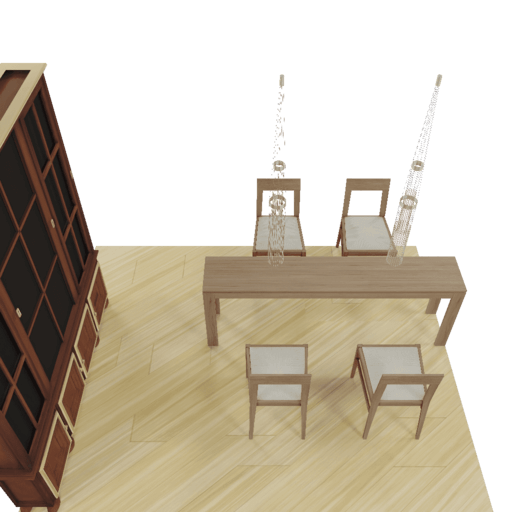}
        \includegraphics[width=\textwidth]{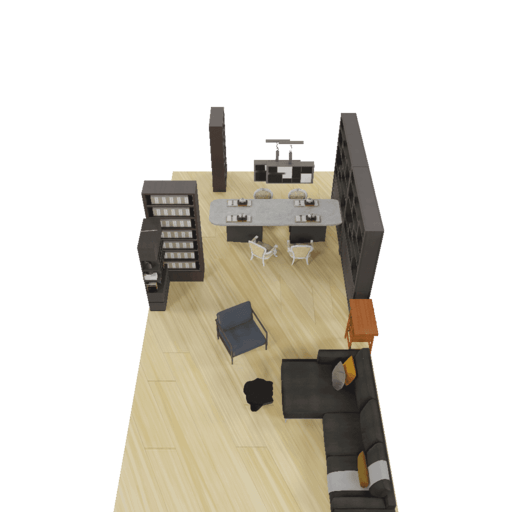}
        \caption{Ours}
    \end{subfigure}
    \caption{Visualizations for instruction-drive 3D scenes stylization by ATISS~\citep{paschalidou2021atiss}, DiffuScene~\citep{tang2024diffuscene} and our method.}
    \label{fig:stylization_vis}
\end{figure*}

\begin{figure*}[htbp]
    \centering
    \begin{subfigure}{0.19\textwidth}    
        \includegraphics[width=\textwidth]{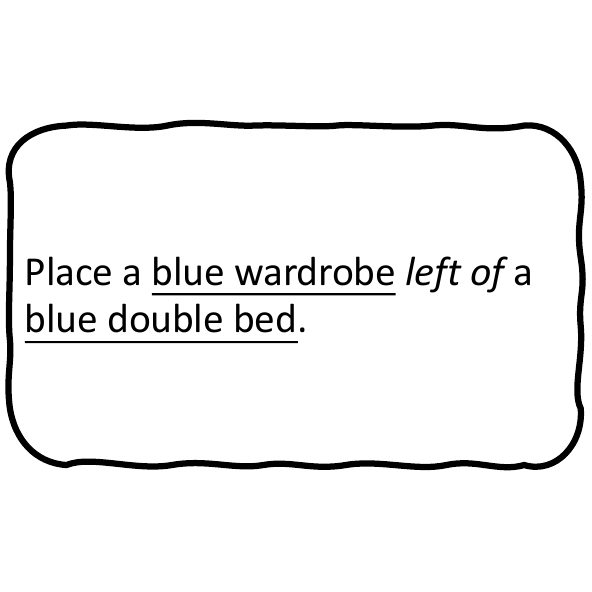}
        \includegraphics[width=\textwidth]{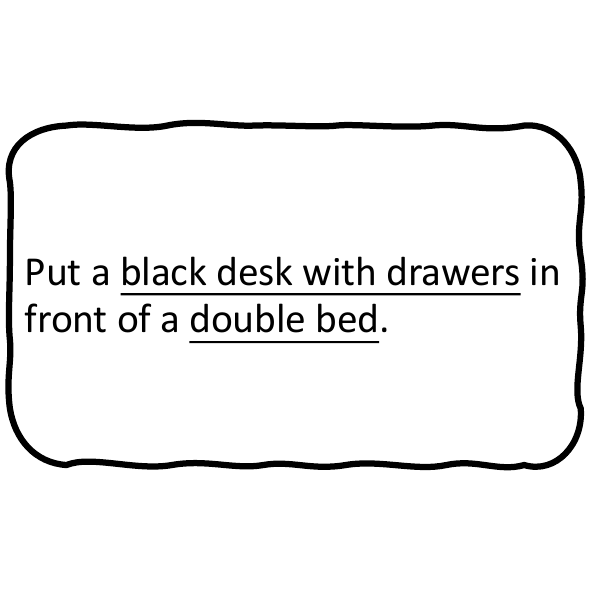}
        \includegraphics[width=\textwidth]{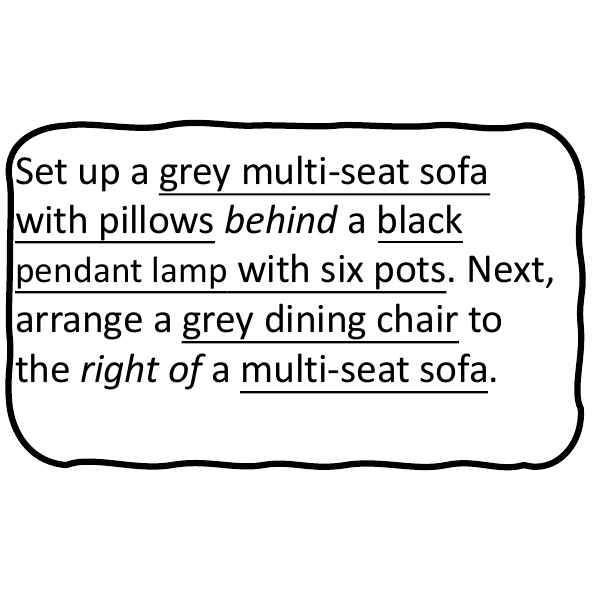}
        \includegraphics[width=\textwidth]{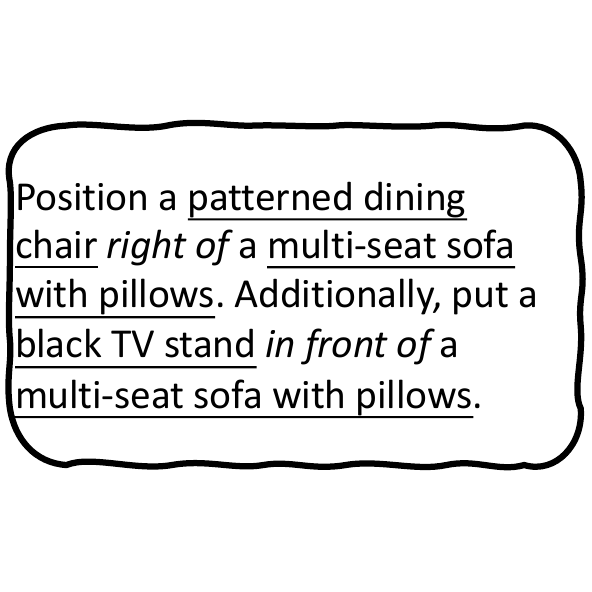}
        \includegraphics[width=\textwidth]{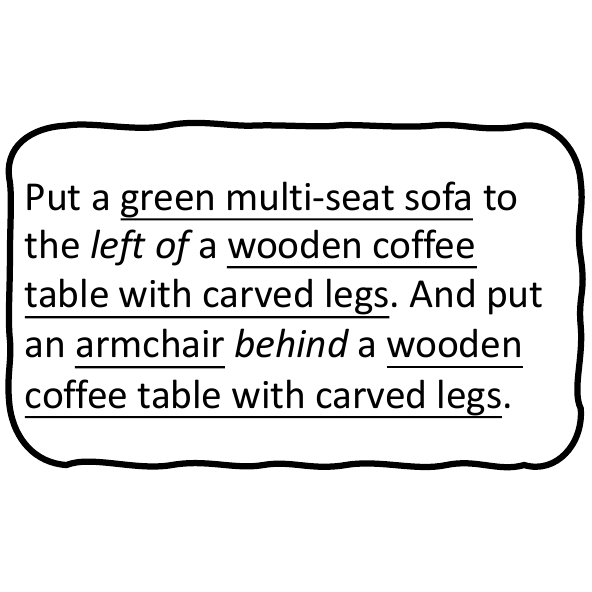}
        \includegraphics[width=\textwidth]{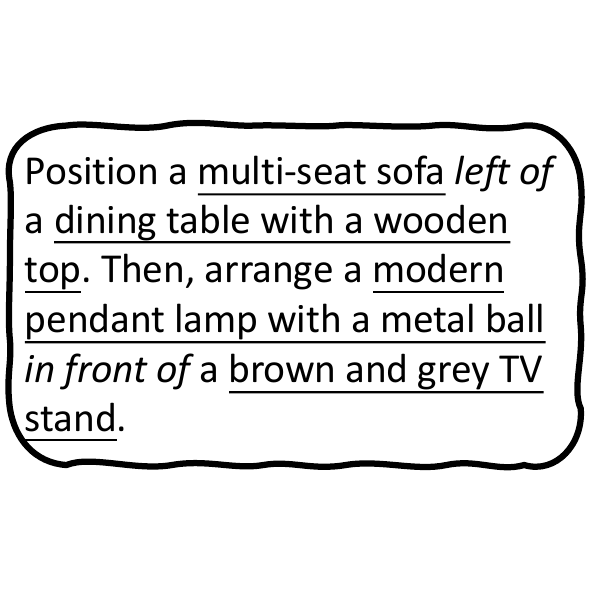}
        \includegraphics[width=\textwidth]{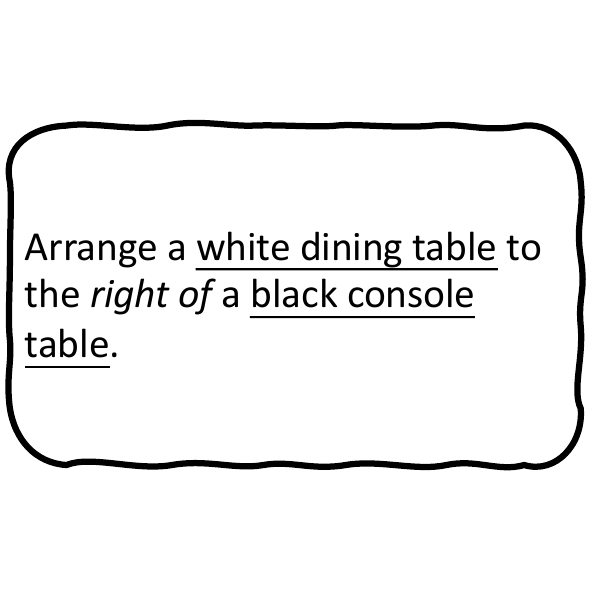}
        \caption{Instructions}
    \end{subfigure}
    \hfill
    \begin{subfigure}{0.19\textwidth}
        \includegraphics[width=\textwidth]{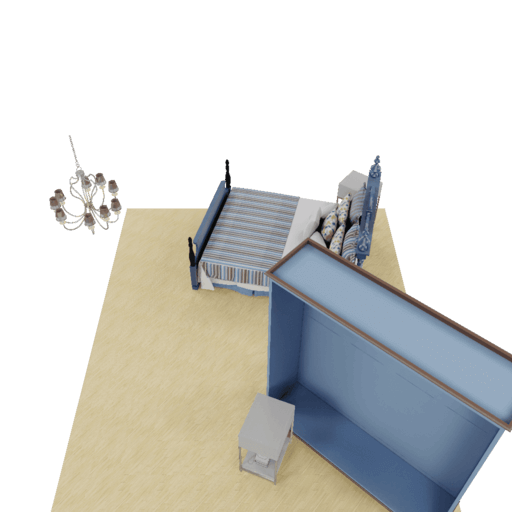}
        \includegraphics[width=\textwidth]{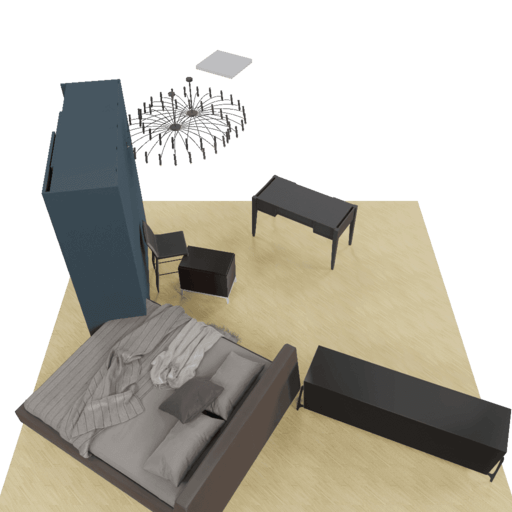}
        \includegraphics[width=\textwidth]{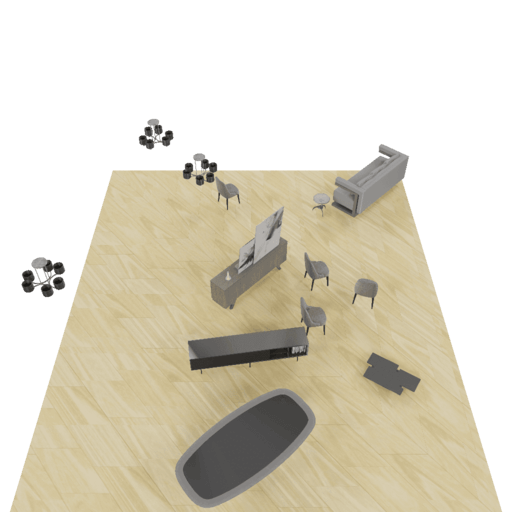}
        \includegraphics[width=\textwidth]{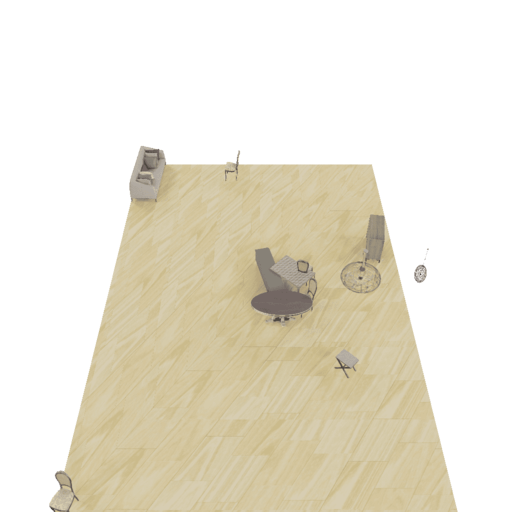}
        \includegraphics[width=\textwidth]{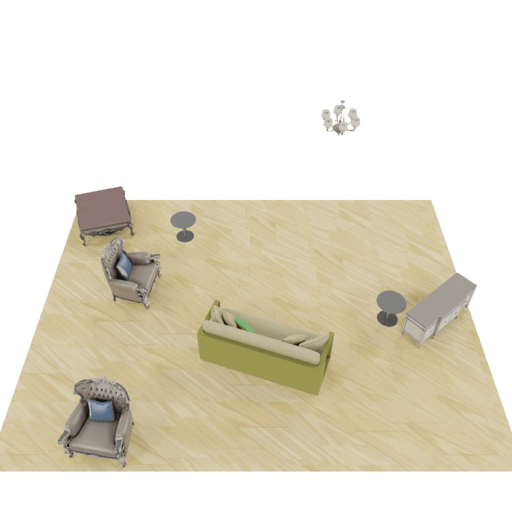}
        \includegraphics[width=\textwidth]{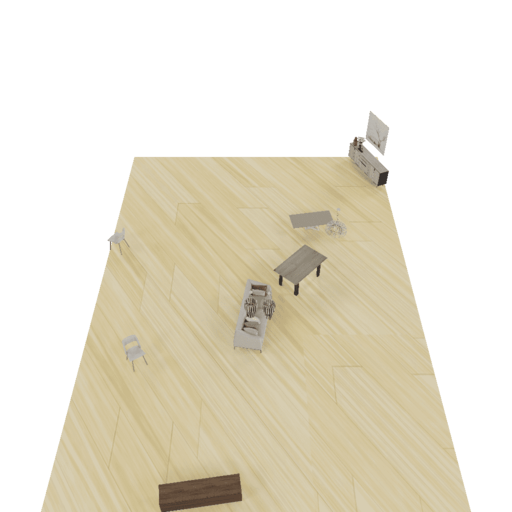}
        \includegraphics[width=\textwidth]{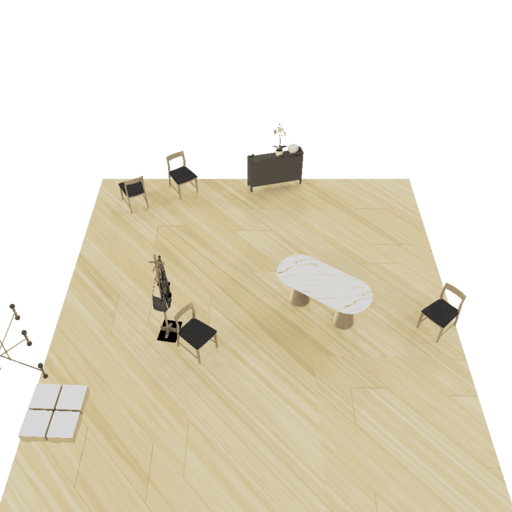}
        \caption{Messy Scenes}
    \end{subfigure}
    \hfill
    \begin{subfigure}{0.19\textwidth}
        \includegraphics[width=\textwidth]{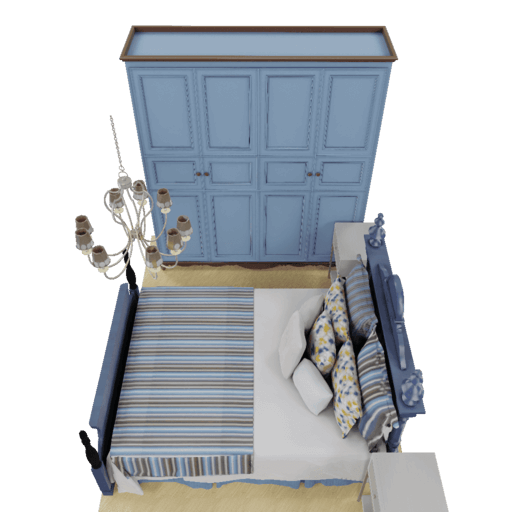}
        \includegraphics[width=\textwidth]{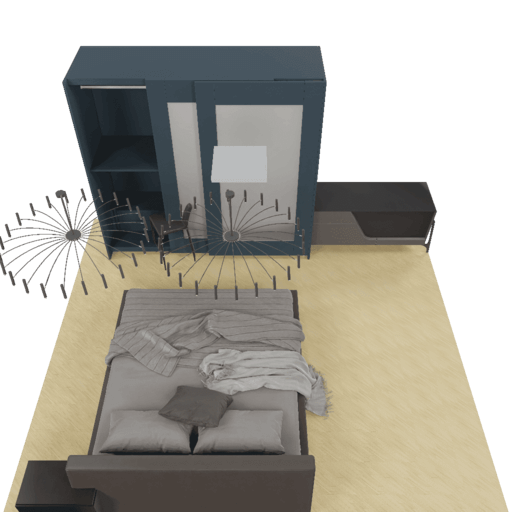}
        \includegraphics[width=\textwidth]{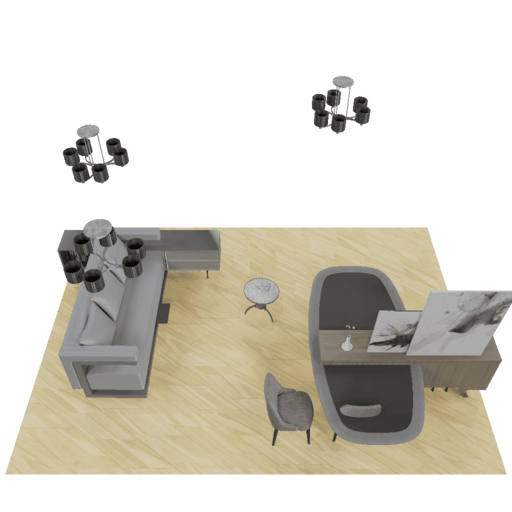}
        \includegraphics[width=\textwidth]{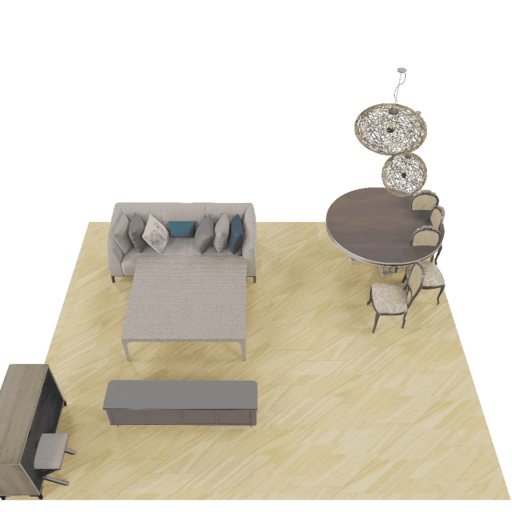}
        \includegraphics[width=\textwidth]{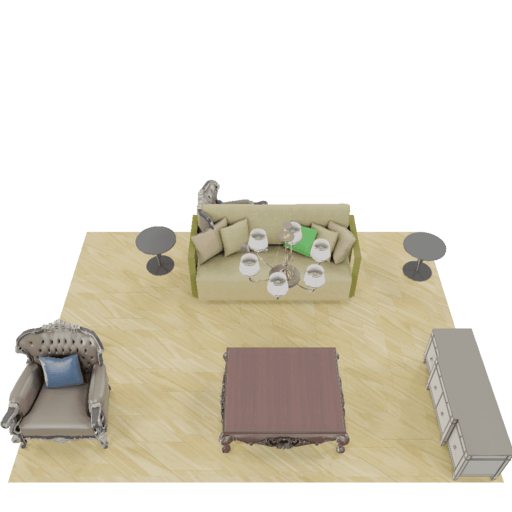}
        \includegraphics[width=\textwidth]{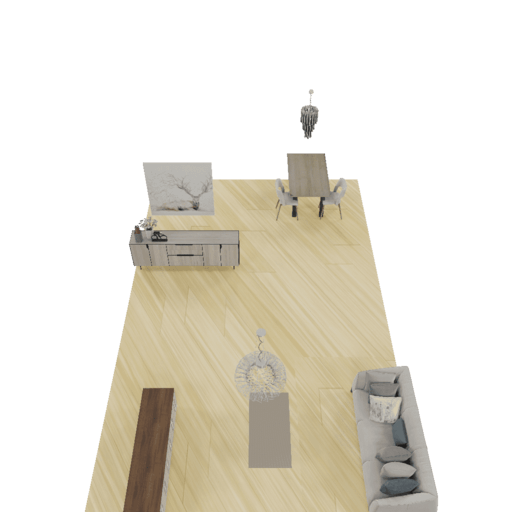}
        \includegraphics[width=\textwidth]{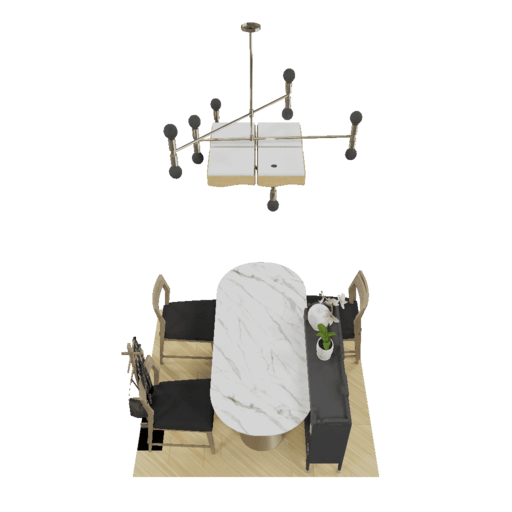}
        \caption{ATISS}
    \end{subfigure}
    \hfill
    \begin{subfigure}{0.19\textwidth}
        \includegraphics[width=\textwidth]{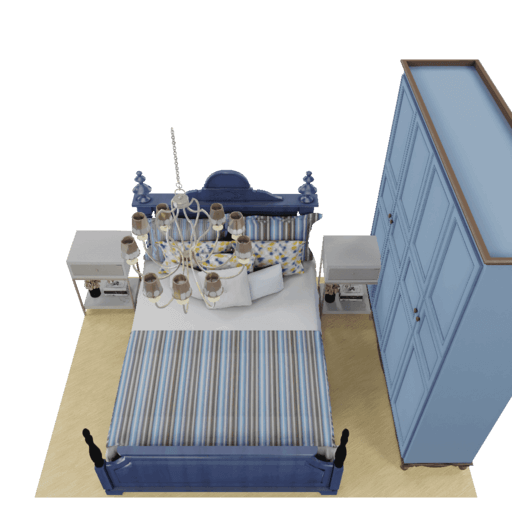}
        \includegraphics[width=\textwidth]{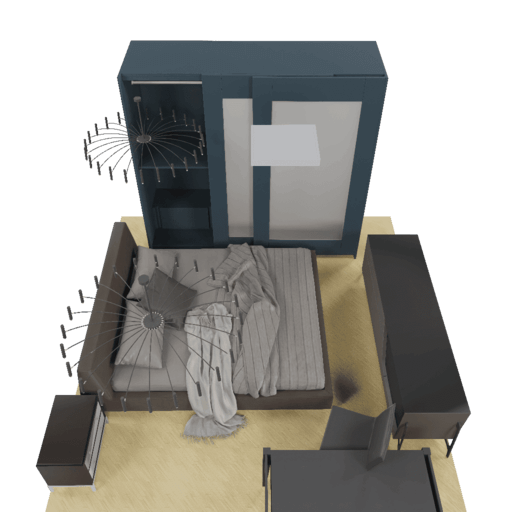}
        \includegraphics[width=\textwidth]{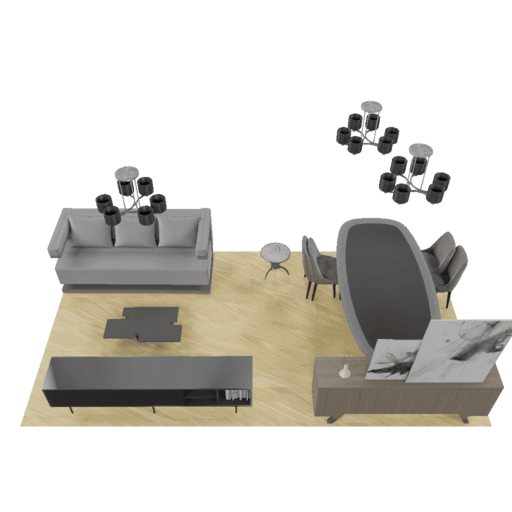}
        \includegraphics[width=\textwidth]{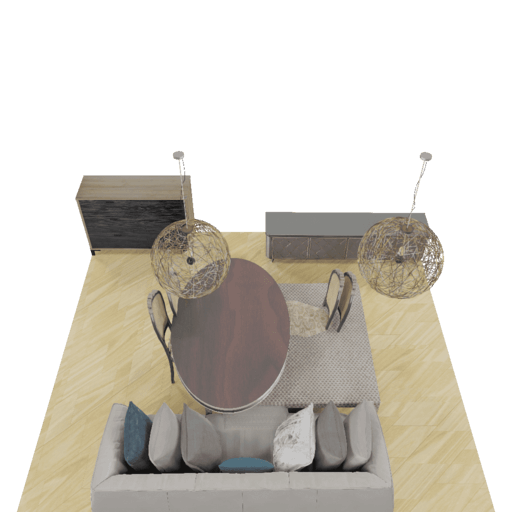}
        \includegraphics[width=\textwidth]{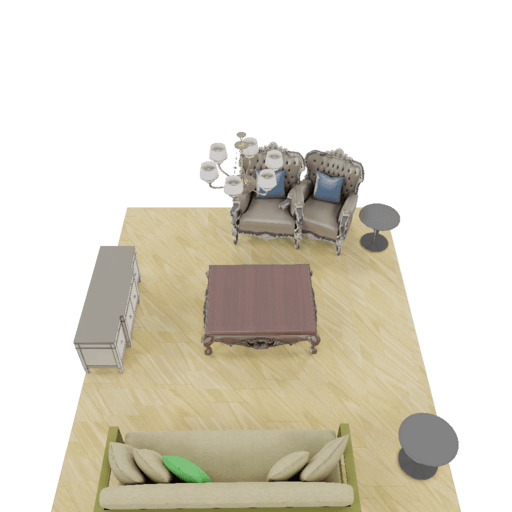}
        \includegraphics[width=\textwidth]{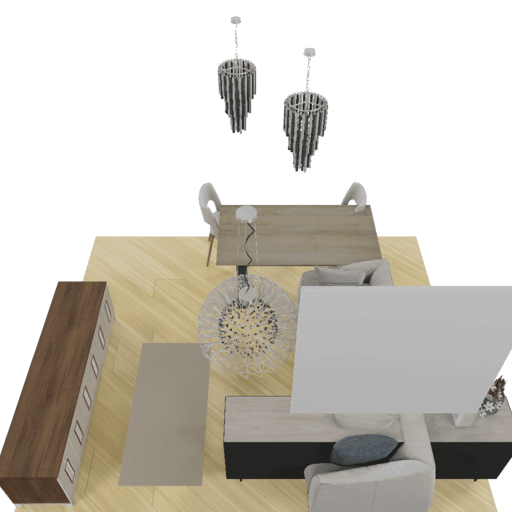}
        \includegraphics[width=\textwidth]{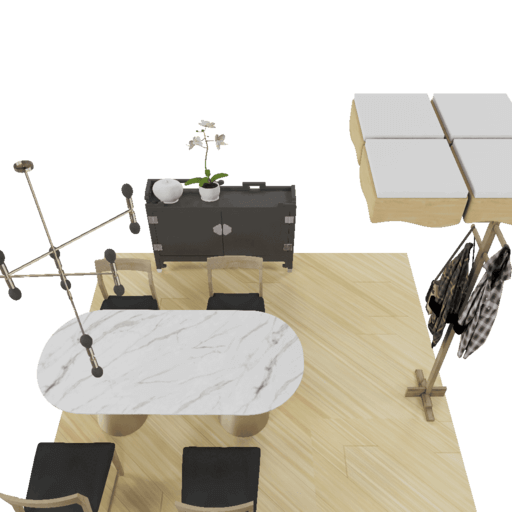}
        \caption{DiffuScene}
    \end{subfigure}
    \hfill
    \begin{subfigure}{0.19\textwidth}
        \includegraphics[width=\textwidth]{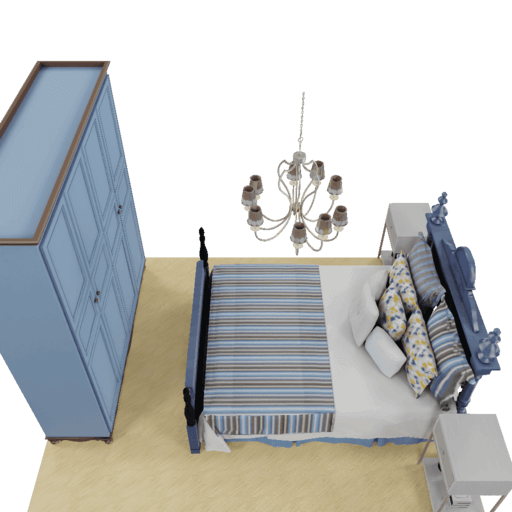}
        \includegraphics[width=\textwidth]{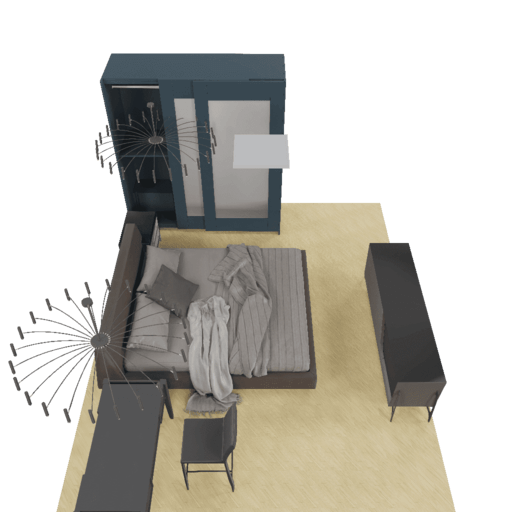}
        \includegraphics[width=\textwidth]{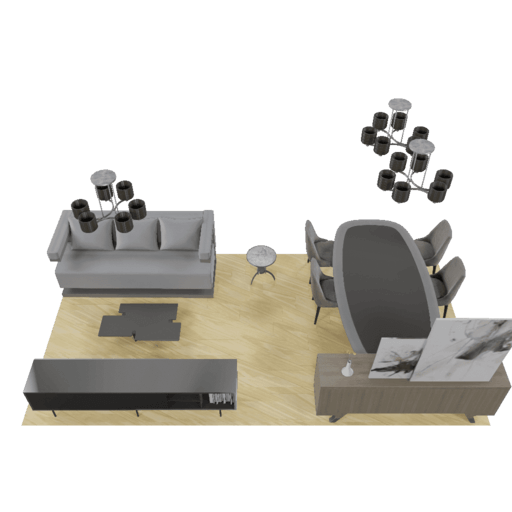}
        \includegraphics[width=\textwidth]{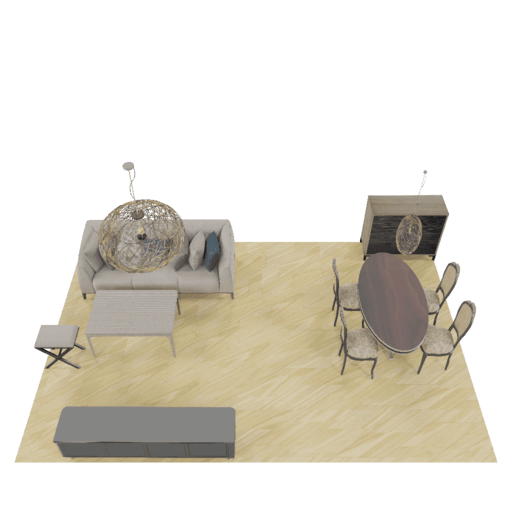}
        \includegraphics[width=\textwidth]{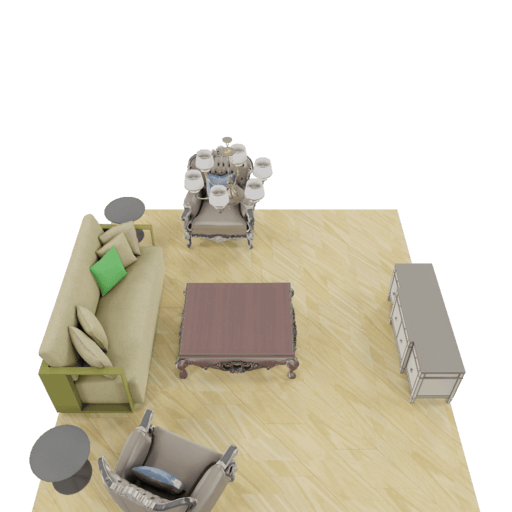}
        \includegraphics[width=\textwidth]{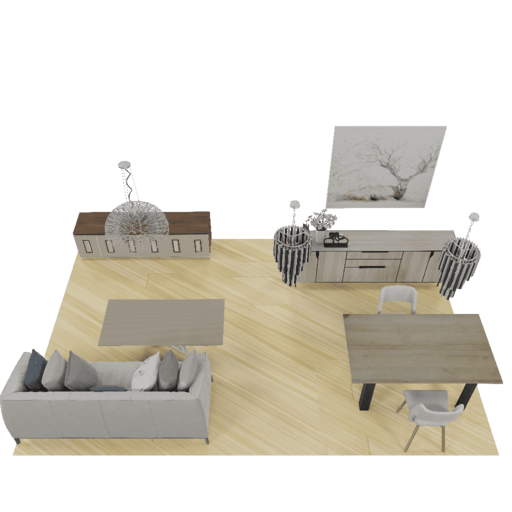}
        \includegraphics[width=\textwidth]{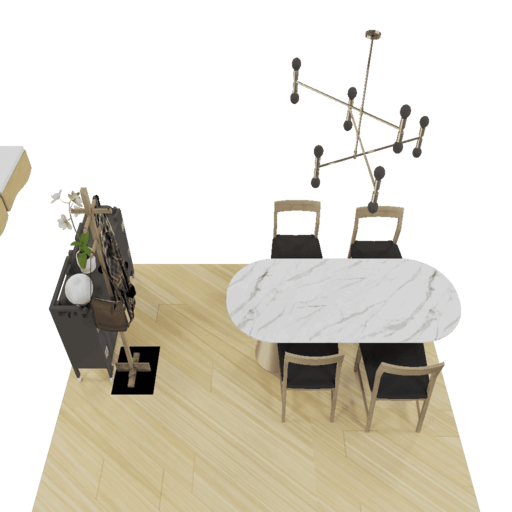}
        \caption{Ours}
    \end{subfigure}
    \caption{Visualizations for instruction-drive 3D scenes re-arrangement by ATISS~\citep{paschalidou2021atiss}, DiffuScene~\citep{tang2024diffuscene} and our method.}
    \label{fig:rearrangement_vis}
\end{figure*}

\begin{figure*}[htbp]
    \centering
    \begin{subfigure}{0.19\textwidth}
        \includegraphics[width=\textwidth]{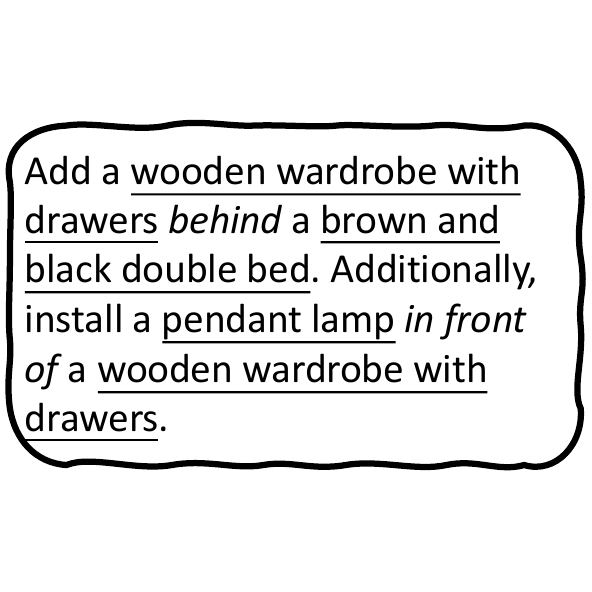}
        \includegraphics[width=\textwidth]{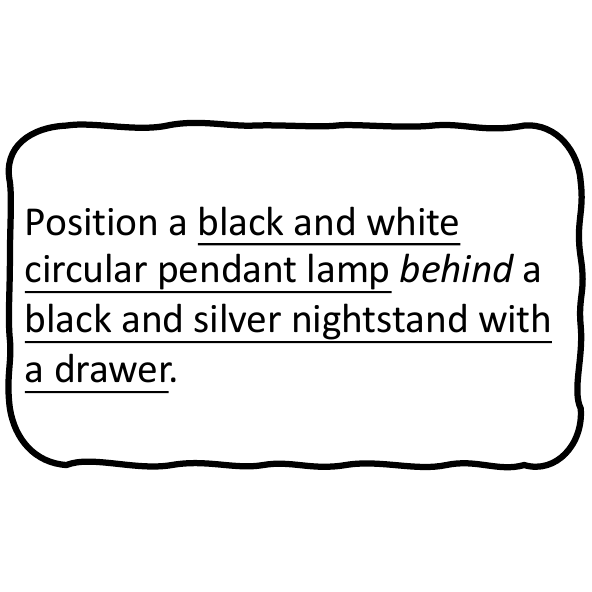}
        \includegraphics[width=\textwidth]{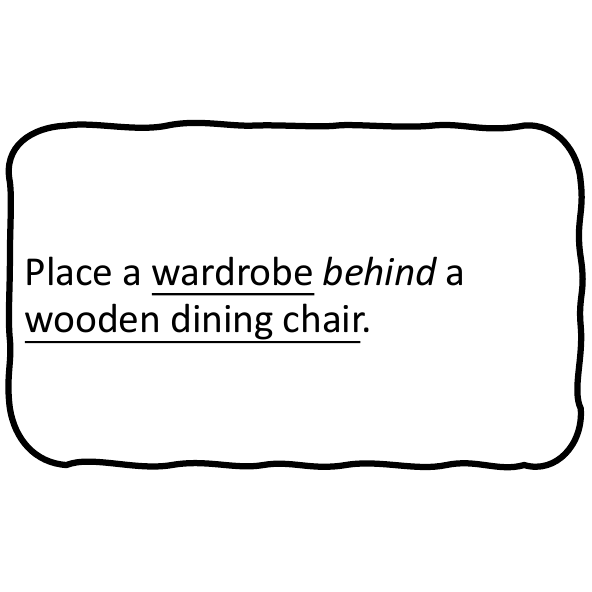}
        \includegraphics[width=\textwidth]{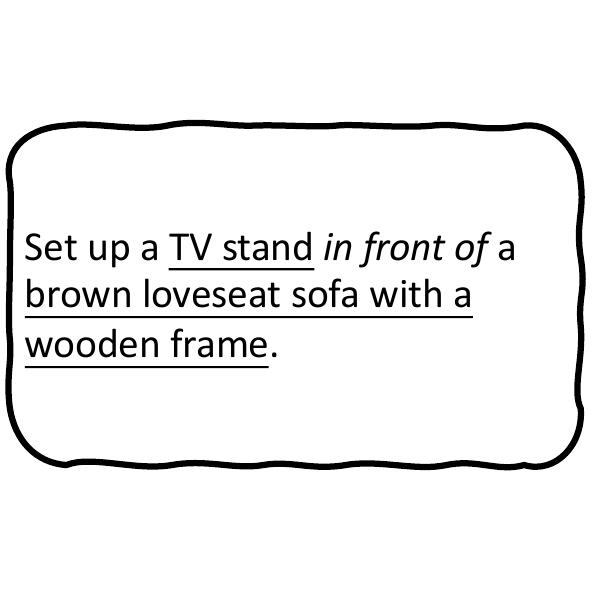}
        \includegraphics[width=\textwidth]{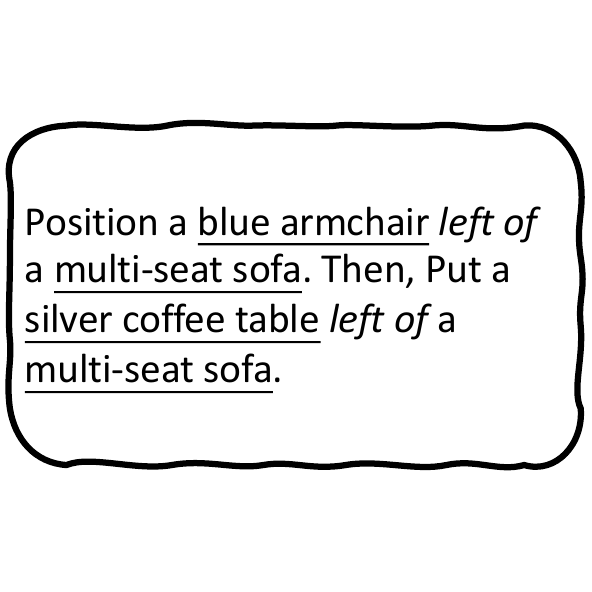}
        \includegraphics[width=\textwidth]{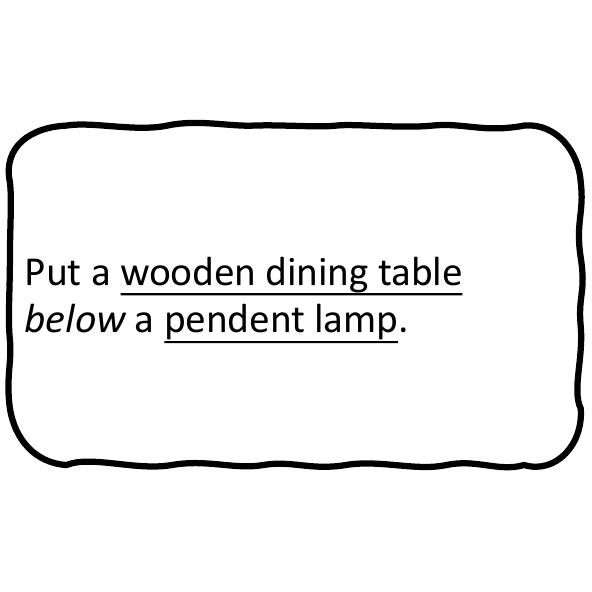}
        \includegraphics[width=\textwidth]{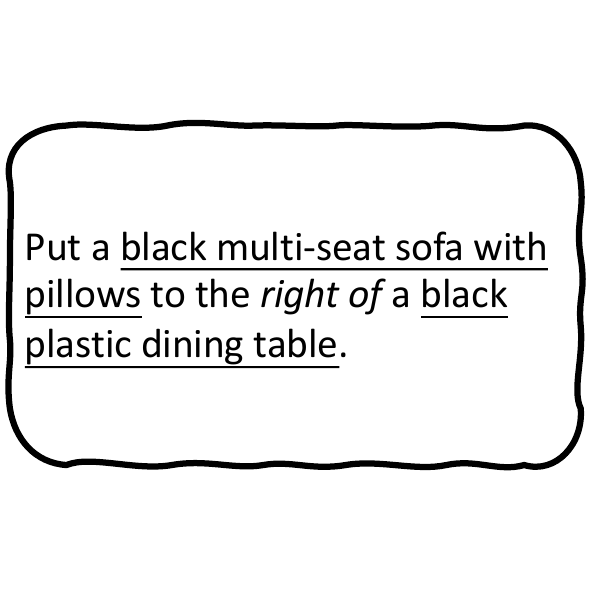}
        \caption{Instructions}
    \end{subfigure}
    \hfill
    \begin{subfigure}{0.19\textwidth}
        \includegraphics[width=\textwidth]{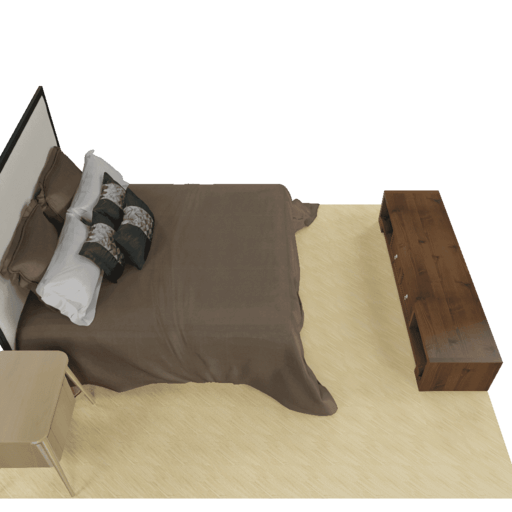}
        \includegraphics[width=\textwidth]{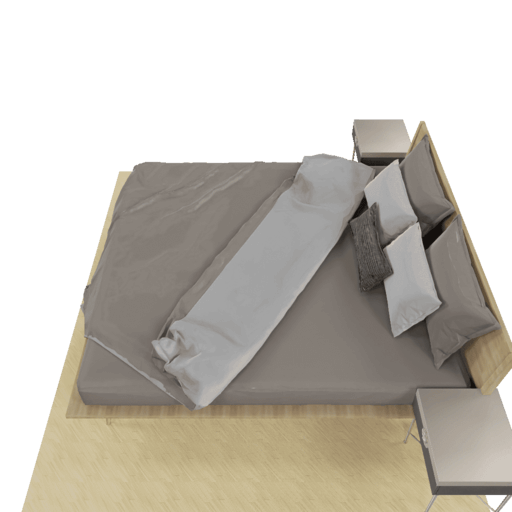}
        \includegraphics[width=\textwidth]{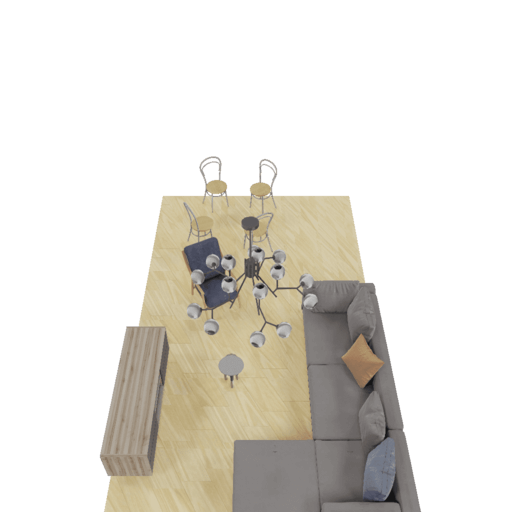}
        \includegraphics[width=\textwidth]{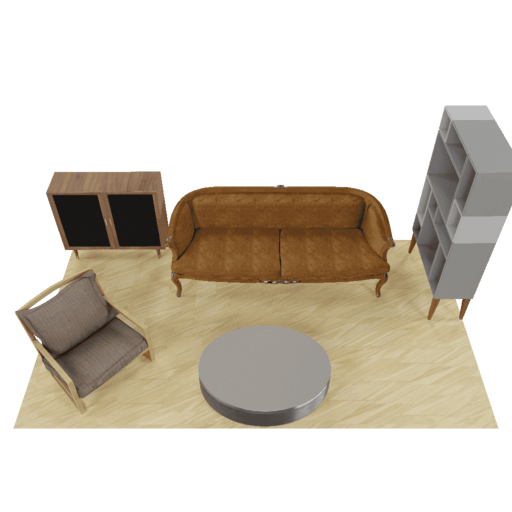}
        \includegraphics[width=\textwidth]{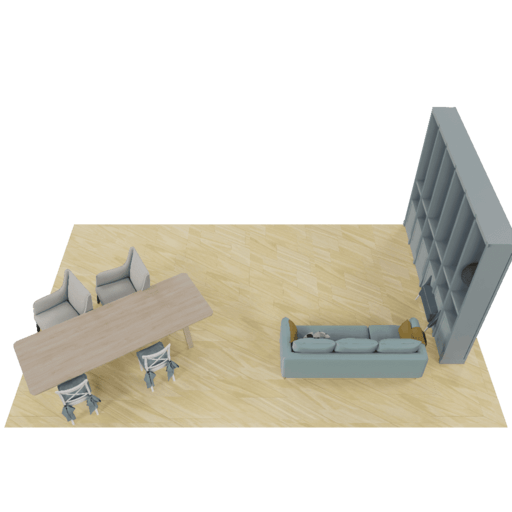}
        \includegraphics[width=\textwidth]{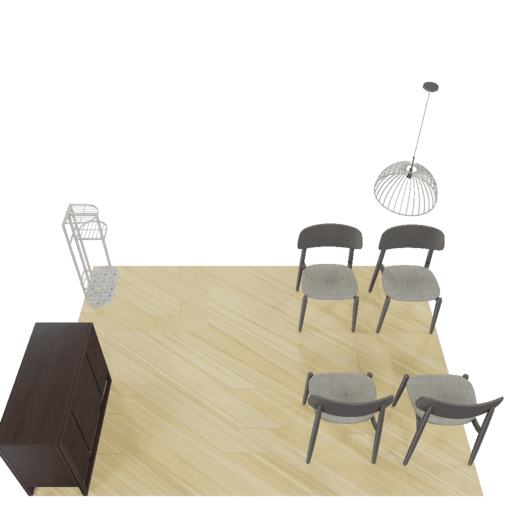}
        \includegraphics[width=\textwidth]{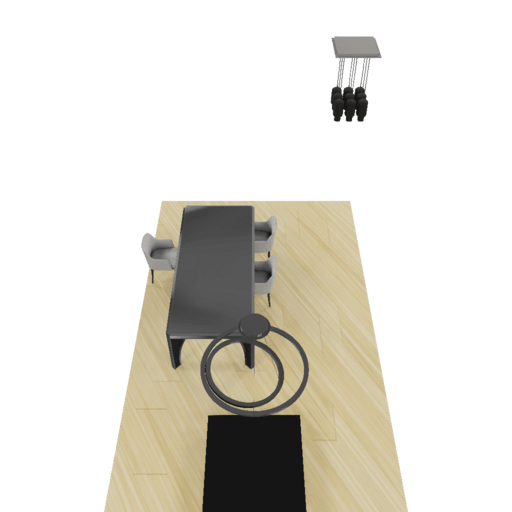}
        \caption{Partial Scenes}
    \end{subfigure}
    \hfill
    \begin{subfigure}{0.19\textwidth}
        \includegraphics[width=\textwidth]{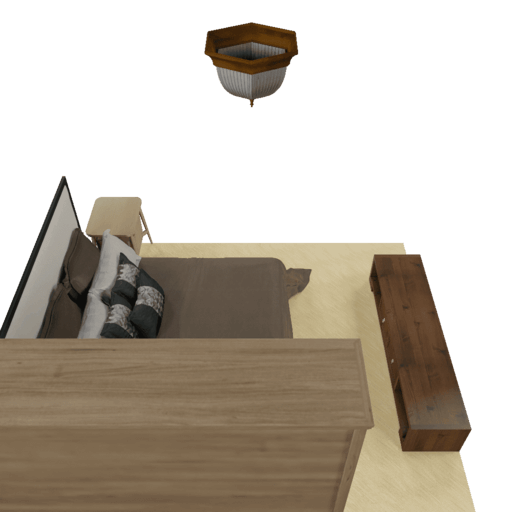}
        \includegraphics[width=\textwidth]{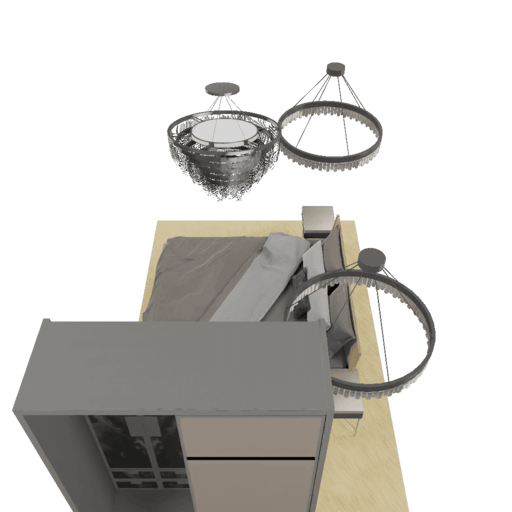}
        \includegraphics[width=\textwidth]{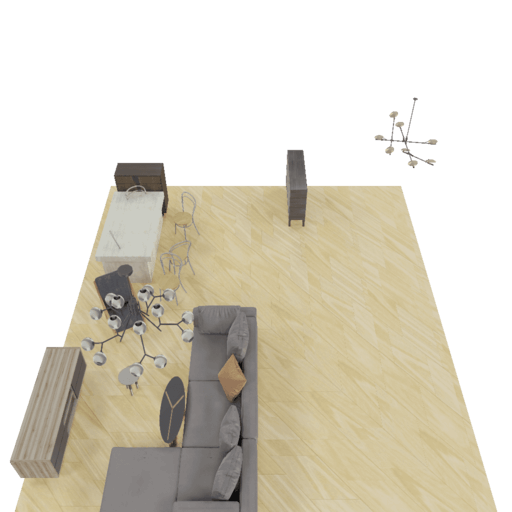}
        \includegraphics[width=\textwidth]{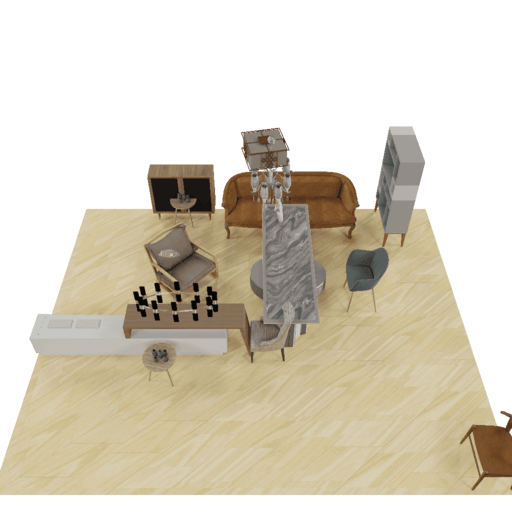}
        \includegraphics[width=\textwidth]{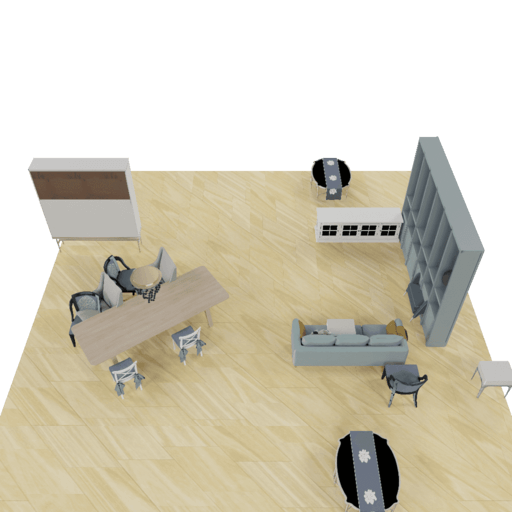}
        \includegraphics[width=\textwidth]{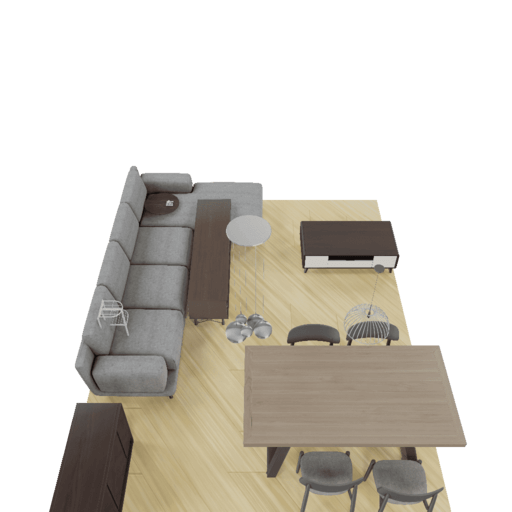}
        \includegraphics[width=\textwidth]{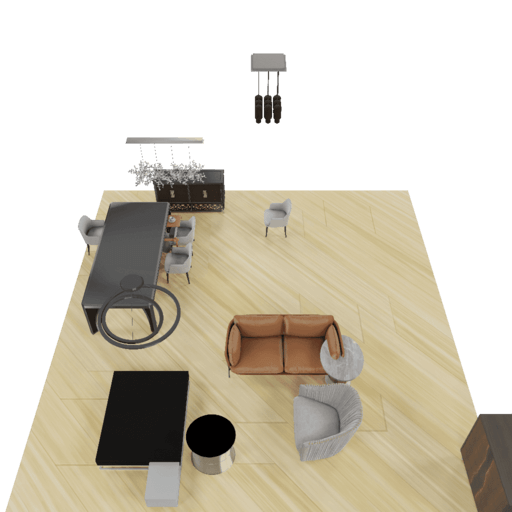}
        \caption{ATISS}
    \end{subfigure}
    \hfill
    \begin{subfigure}{0.19\textwidth}
        \includegraphics[width=\textwidth]{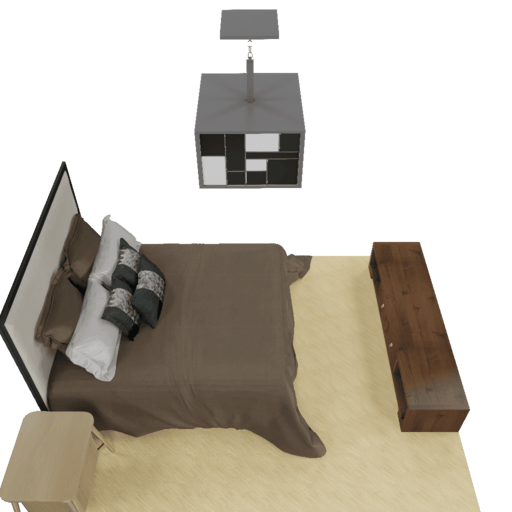}
        \includegraphics[width=\textwidth]{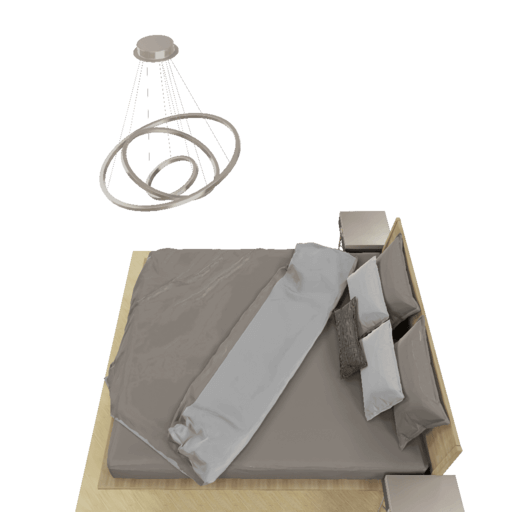}
        \includegraphics[width=\textwidth]{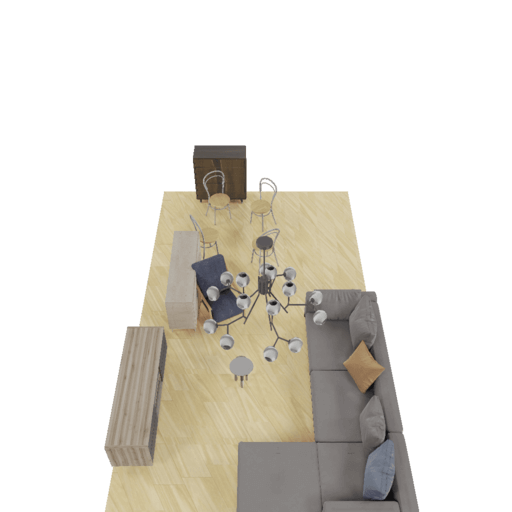}
        \includegraphics[width=\textwidth]{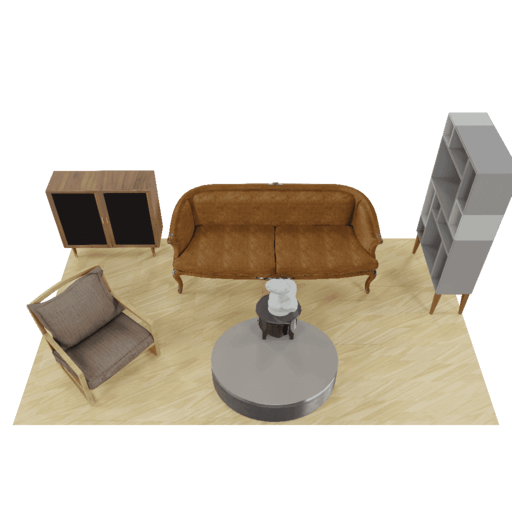}
        \includegraphics[width=\textwidth]{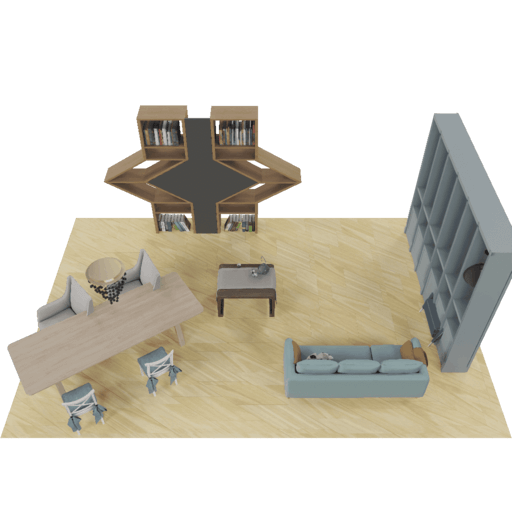}
        \includegraphics[width=\textwidth]{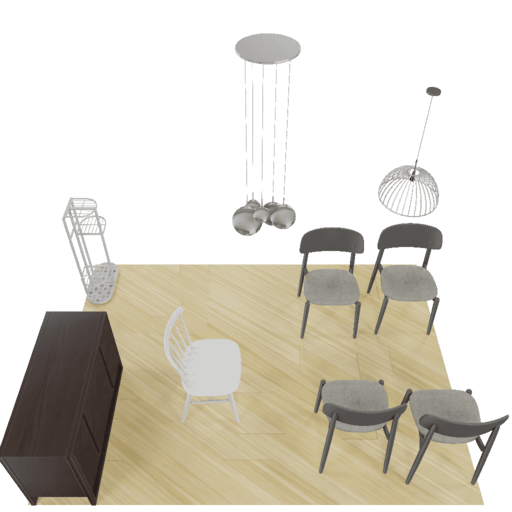}
        \includegraphics[width=\textwidth]{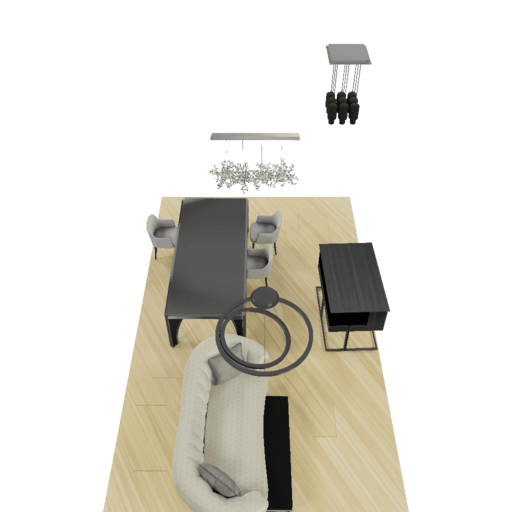}
        \caption{DiffuScene}
    \end{subfigure}
    \hfill
    \begin{subfigure}{0.19\textwidth}
        \includegraphics[width=\textwidth]{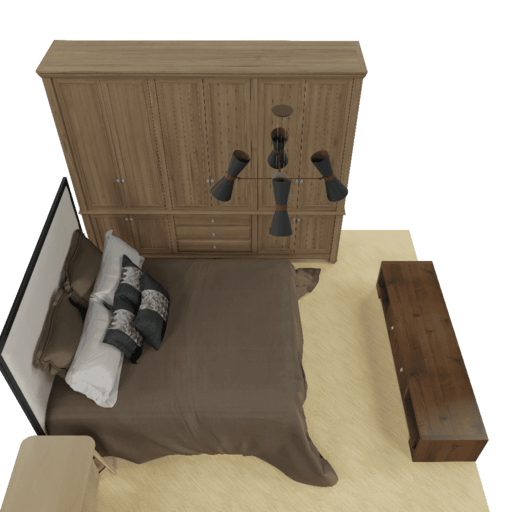}
        \includegraphics[width=\textwidth]{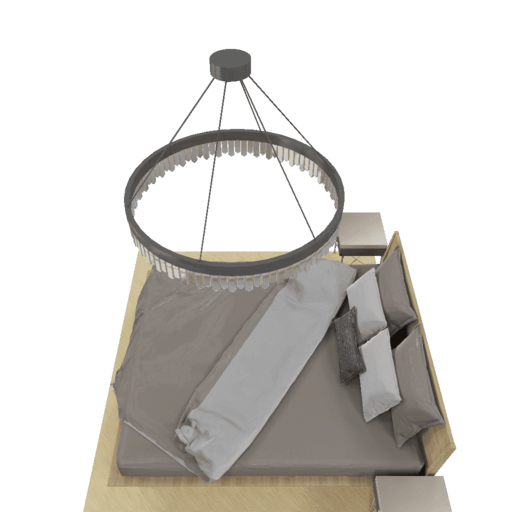}
        \includegraphics[width=\textwidth]{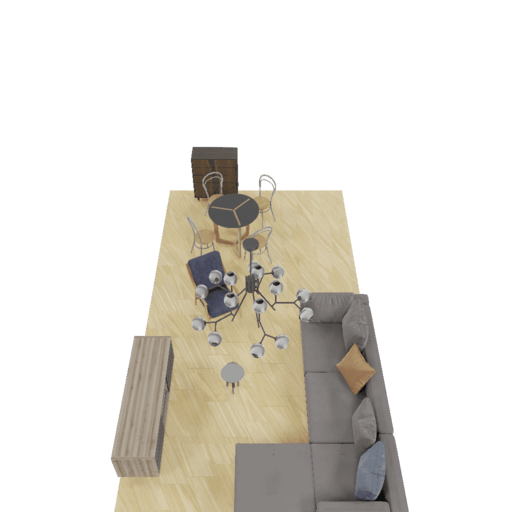}
        \includegraphics[width=\textwidth]{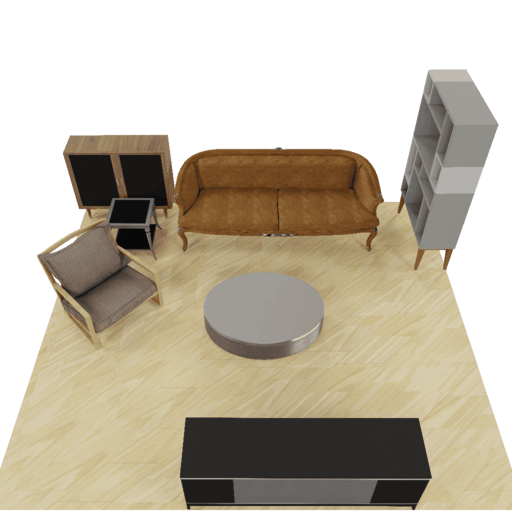}
        \includegraphics[width=\textwidth]{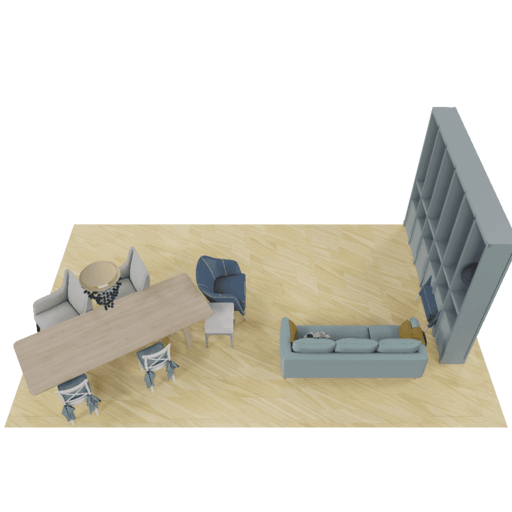}
        \includegraphics[width=\textwidth]{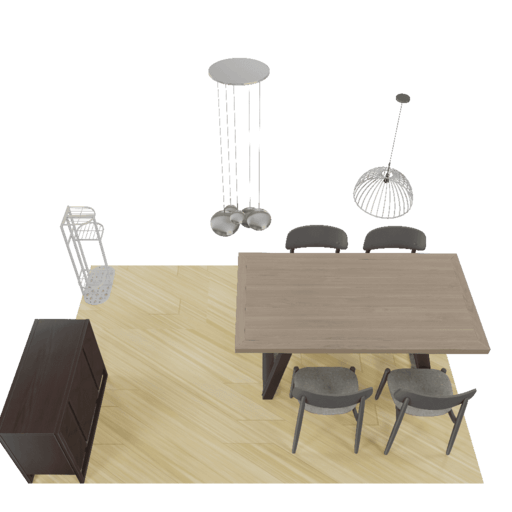}
        \includegraphics[width=\textwidth]{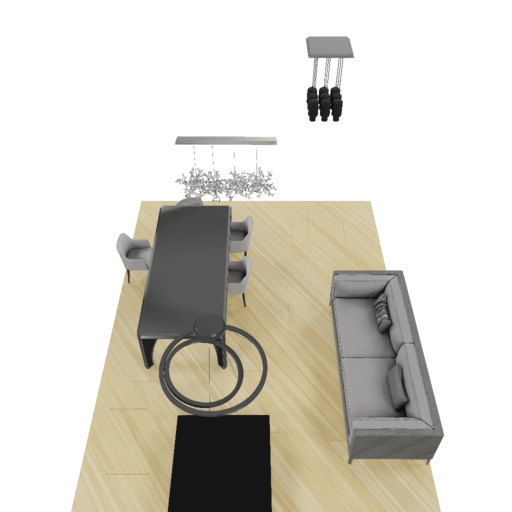}
        \caption{Ours}
    \end{subfigure}
    \caption{Visualizations for instruction-drive 3D scenes completion by ATISS~\citep{paschalidou2021atiss}, DiffuScene~\citep{tang2024diffuscene} and our method.}
    \label{fig:completion_vis}
\end{figure*}

\begin{figure*}[htbp]
    \centering
    \begin{subfigure}{0.25\textwidth}
        \includegraphics[width=\textwidth]{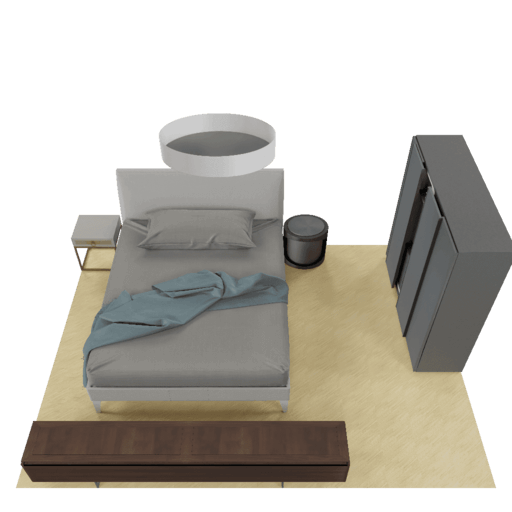}
        \includegraphics[width=\textwidth]{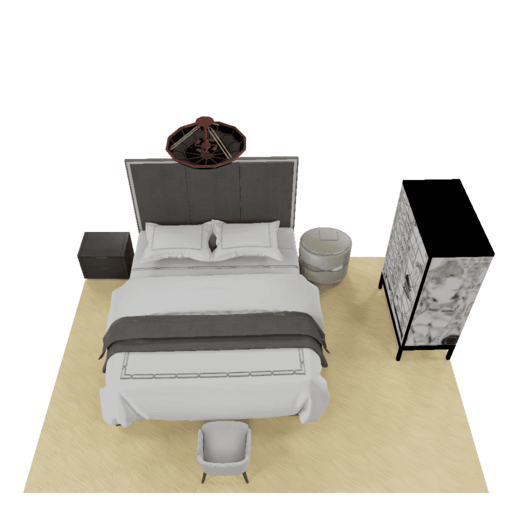}
        \includegraphics[width=\textwidth]{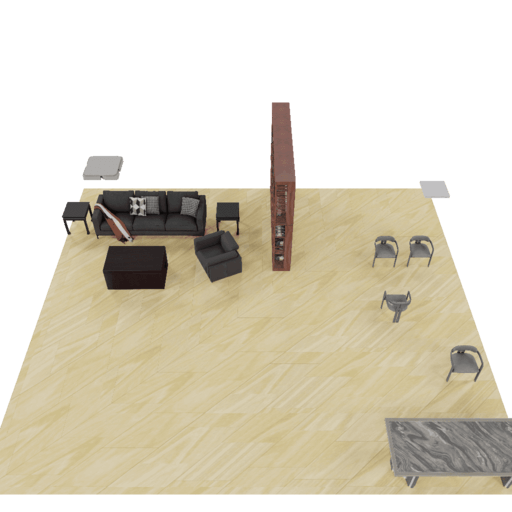}
        \includegraphics[width=\textwidth]{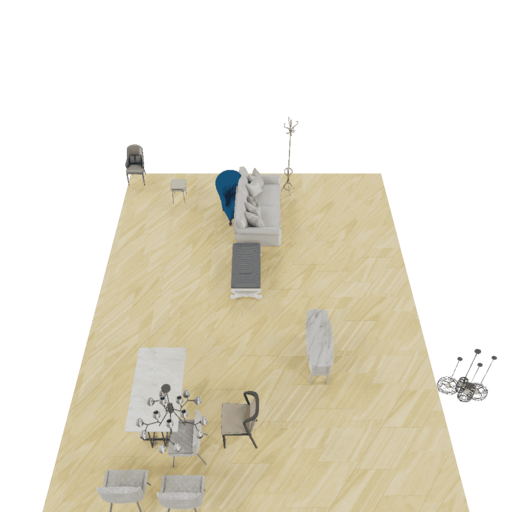}
        \includegraphics[width=\textwidth]{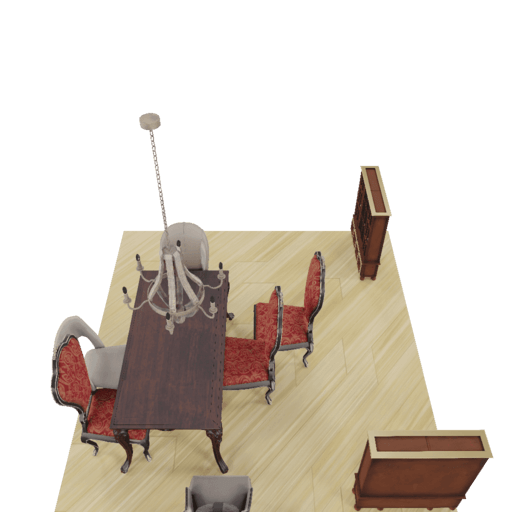}
        \caption{ATISS}
    \end{subfigure}
    \hfill
    \begin{subfigure}{0.25\textwidth}
        \includegraphics[width=\textwidth]{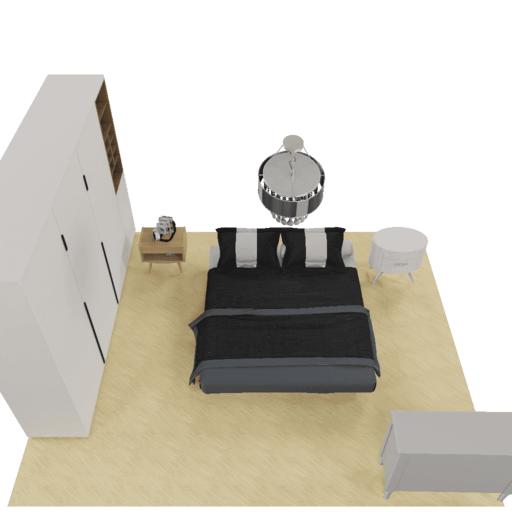}
        \includegraphics[width=\textwidth]{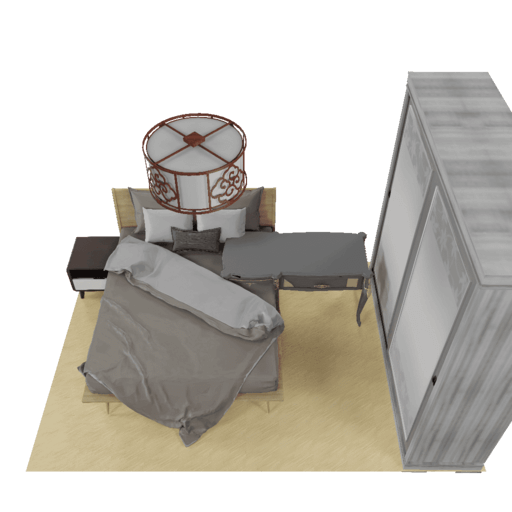}
        \includegraphics[width=\textwidth]{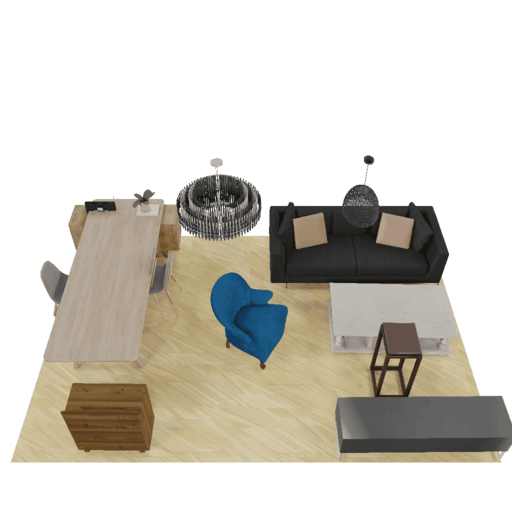}
        \includegraphics[width=\textwidth]{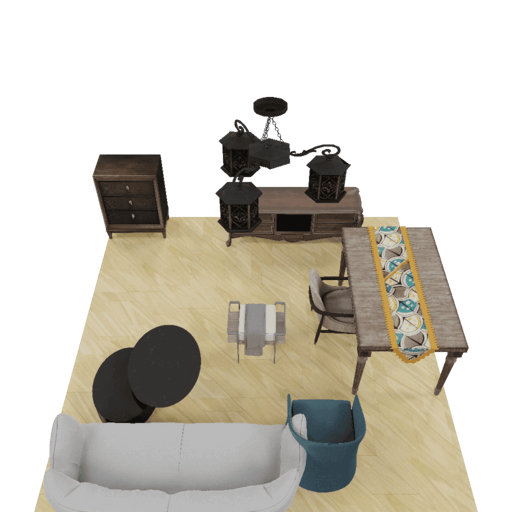}
        \includegraphics[width=\textwidth]{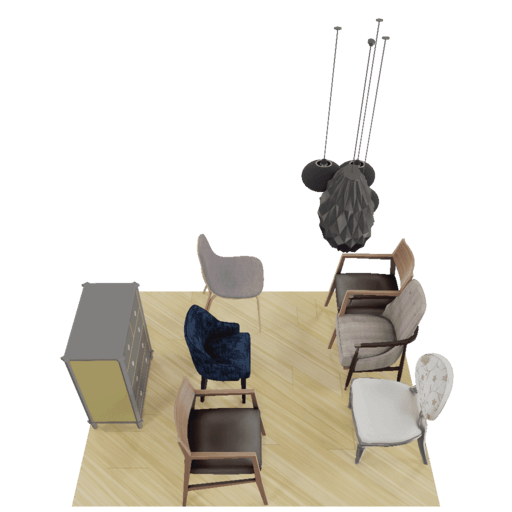}
        \caption{DiffuScene}
    \end{subfigure}
    \hfill
    \begin{subfigure}{0.25\textwidth}
        \includegraphics[width=\textwidth]{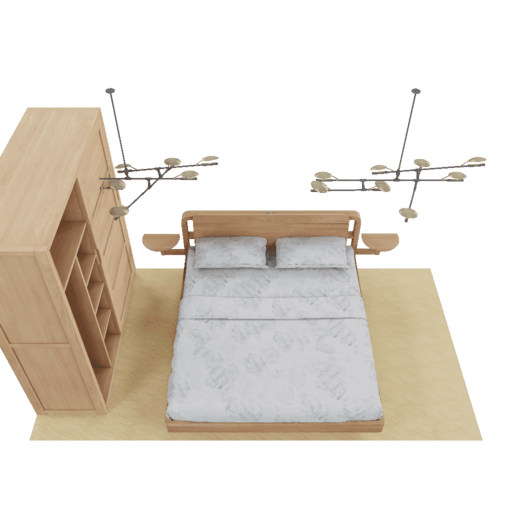}
        \includegraphics[width=\textwidth]{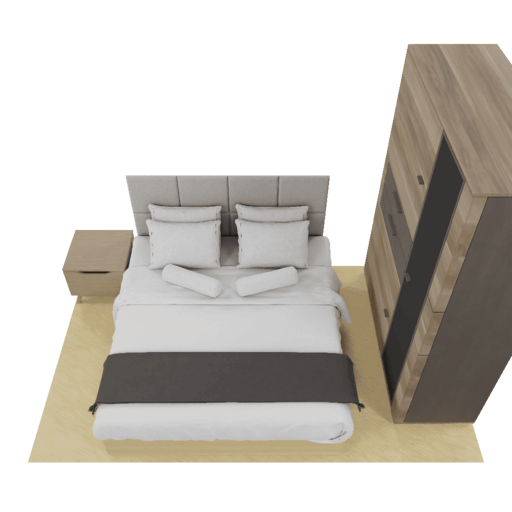}
        \includegraphics[width=\textwidth]{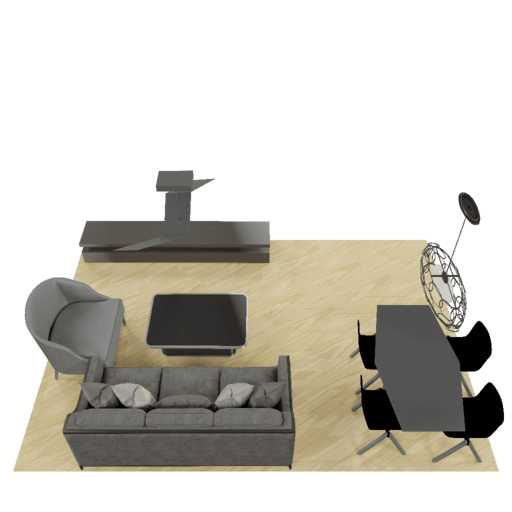}
        \includegraphics[width=\textwidth]{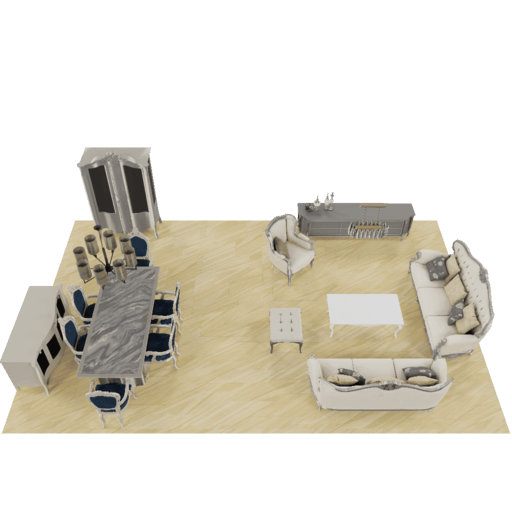}
        \includegraphics[width=\textwidth]{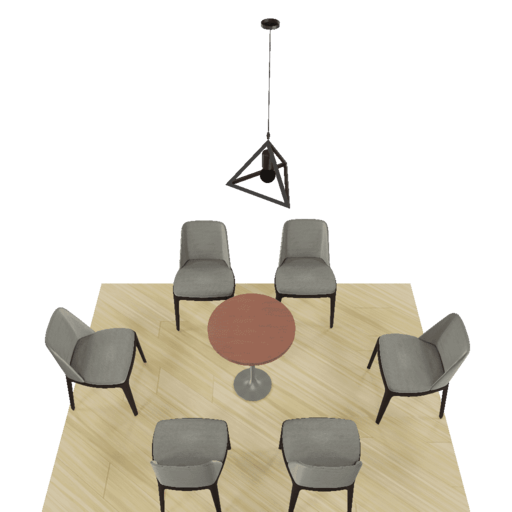}
        \caption{Ours}
    \end{subfigure}
    \caption{Visualizations for unconditional 3D scenes stylization by ATISS~\citep{paschalidou2021atiss}, DiffuScene~\citep{tang2024diffuscene} and our method.}
    \label{fig:unconditional_vis}
\end{figure*}

We present visualizations illustrating various zero-shot instruction-driven applications, including stylization, re-arrangement, completion, and unconditional 3D scene synthesis in Figure~\ref{fig:stylization_vis}, \ref{fig:rearrangement_vis}, \ref{fig:completion_vis} and \ref{fig:unconditional_vis} respectively.
We find that the autoregressive model ATISS~\citep{paschalidou2021atiss} tends to generate redundant objects, resulting in chaotic synthesized scenes.
DiffuScene~\citep{tang2024diffuscene} encounters challenges in accurately modeling object semantic features, often yielding objects that lack coherence in terms of style or pairing, thereby diminishing the aesthetic appeal of the synthesized scenes.
Moreover, both baseline models frequently struggle to follow the provided instructions during conditional generation.

\subsection{Instruction-driven 2D Poster Synthesis}\label{apx:results_2d}
\begin{figure*}[htbp]
    \centering

    \begin{minipage}[t]{0.2\textwidth}
        \begin{minipage}[t]{0.05\textwidth}
            
        \end{minipage}
        \hfill
        \begin{minipage}[t]{0.92\textwidth}
            \centering
            (a) Instruction \& Product Information
        \end{minipage}
    \end{minipage}
    \hfill
    \begin{minipage}[t]{0.15\textwidth}
        \centering
        (b) LayoutTrans
    \end{minipage}
    \hfill
    \begin{minipage}[t]{0.15\textwidth}
        \centering
        (c) LayoutVAE
    \end{minipage}
    \hfill
    \begin{minipage}[t]{0.15\textwidth}
        \centering
        (d) LayoutDM
    \end{minipage}
    \hfill
    \begin{minipage}[t]{0.15\textwidth}
        \centering
        (e) Ours
    \end{minipage}
    \hfill
    \begin{minipage}[t]{0.15\textwidth}
        \centering
        (f) Ours (Rendered)
    \end{minipage}
    
    \begin{minipage}[c]{0.2\textwidth}
        \begin{minipage}[c]{0.05\textwidth}
            \rotatebox{90}{\small Instruction}
        \end{minipage}
        \hfill
        \begin{subfigure}[c]{0.92\textwidth}
            \centering
            \includegraphics[width=\textwidth]{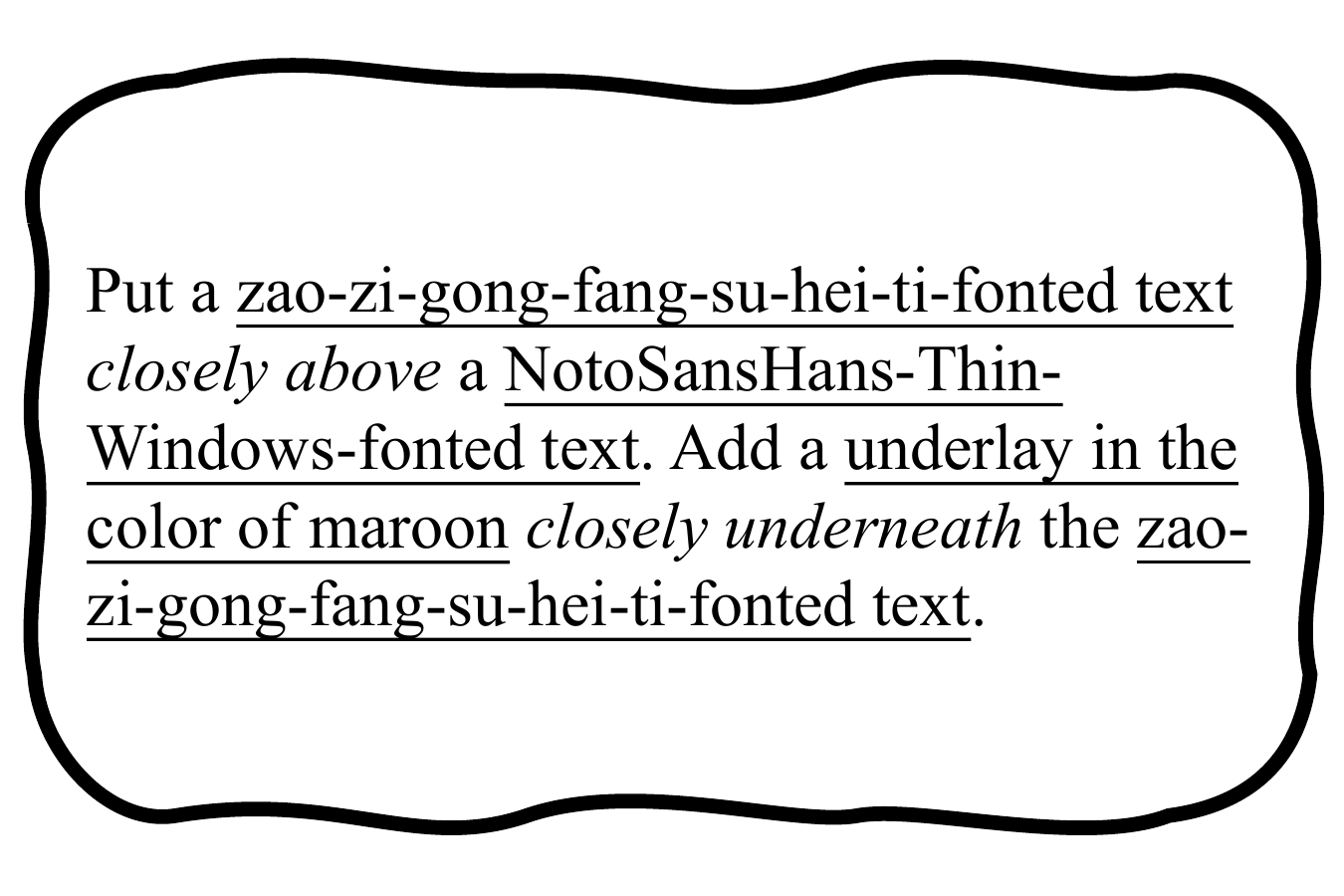}
        \end{subfigure}
        \begin{minipage}[c]{0.05\textwidth}
            \rotatebox{90}{\small Information}
        \end{minipage}
        \hfill
        \begin{subfigure}[c]{0.92\textwidth}
            \centering
            \includegraphics[width=\textwidth]{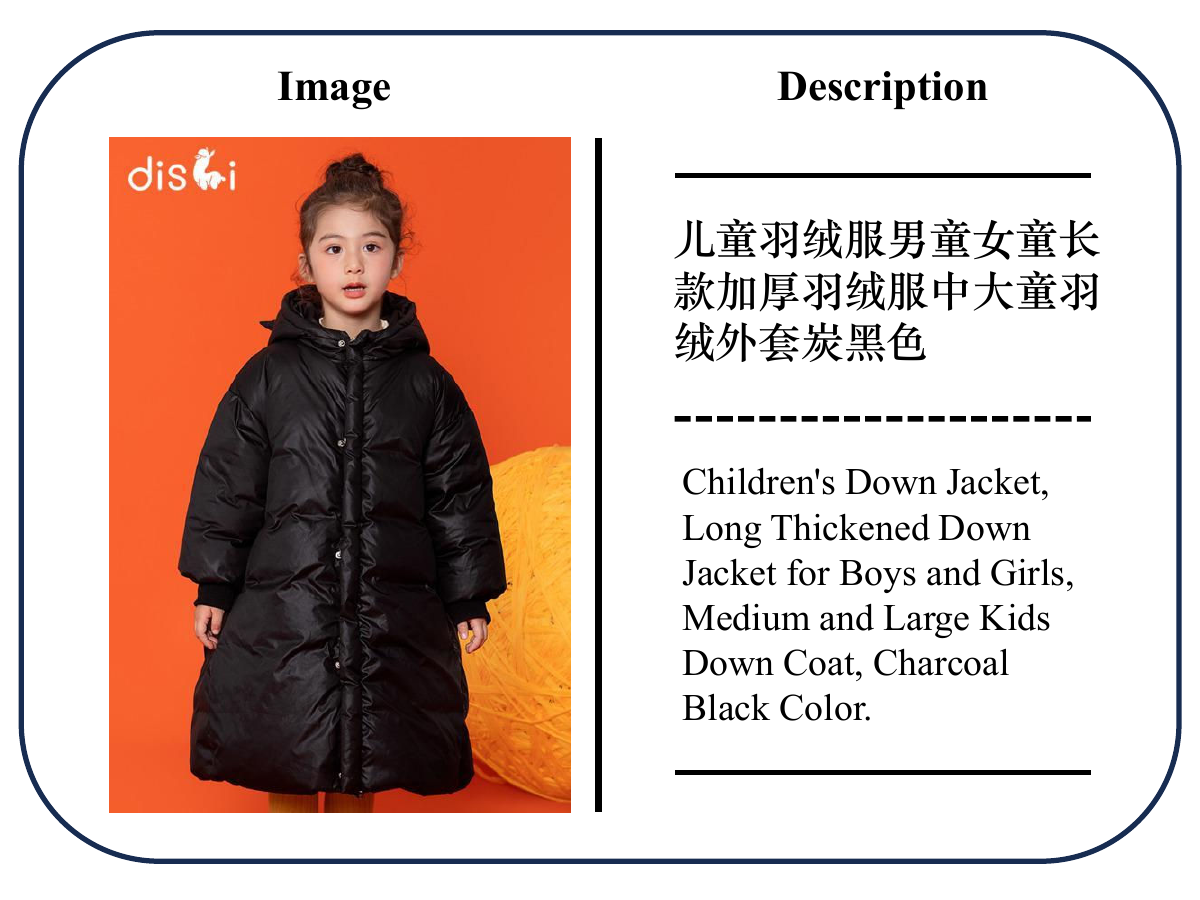}
        \end{subfigure}
    \end{minipage}
    \hfill
    \begin{subfigure}[c]{0.15\textwidth}
        \centering
        \includegraphics[width=\textwidth]{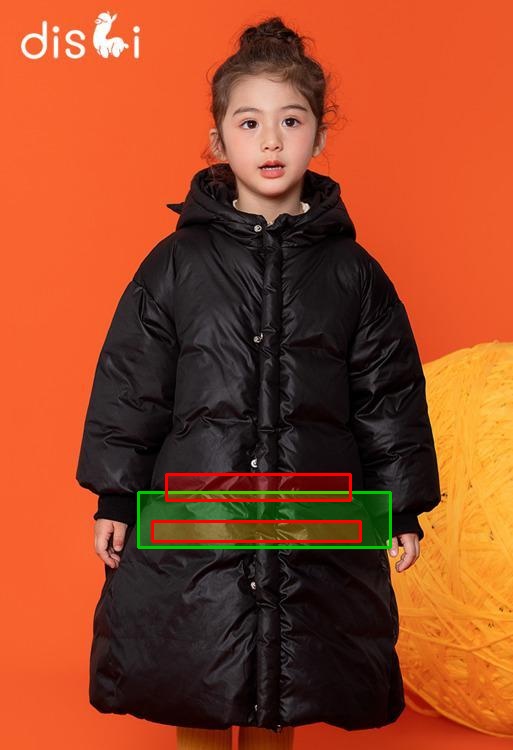}
    \end{subfigure}
    \hfill
    \begin{subfigure}[c]{0.15\textwidth}
        \centering
        \includegraphics[width=\textwidth]{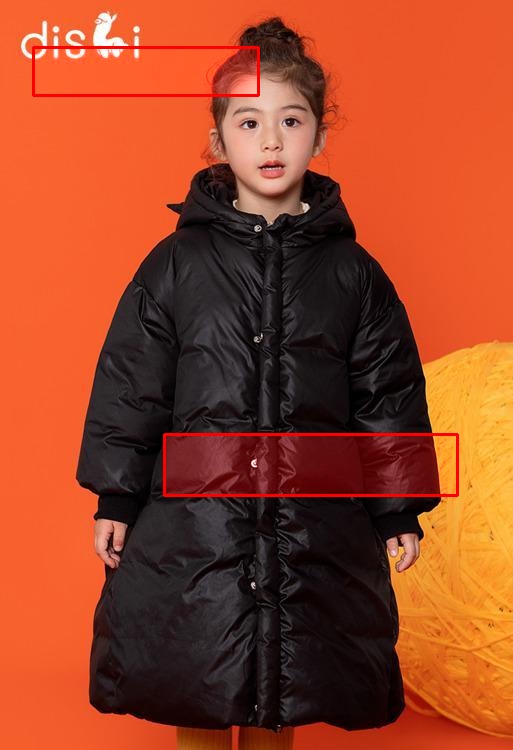}
    \end{subfigure}
    \hfill
    \begin{subfigure}[c]{0.15\textwidth}
        \centering
        \includegraphics[width=\textwidth]{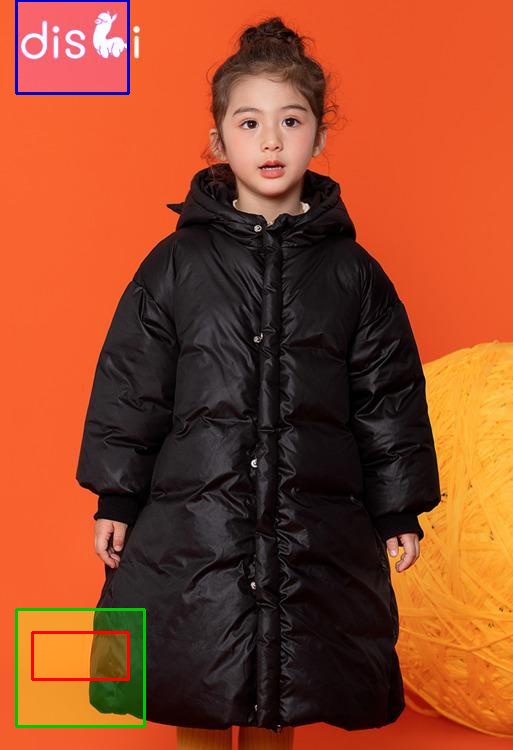}
    \end{subfigure}
    \hfill
    \begin{subfigure}[c]{0.15\textwidth}
        \centering
        \includegraphics[width=\textwidth]{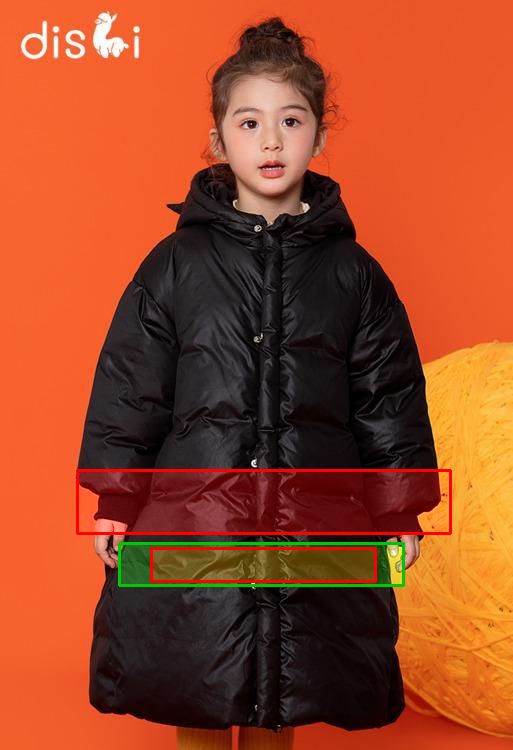}
    \end{subfigure}
    \hfill
    \begin{subfigure}[c]{0.15\textwidth}
        \centering
        \includegraphics[width=\textwidth]{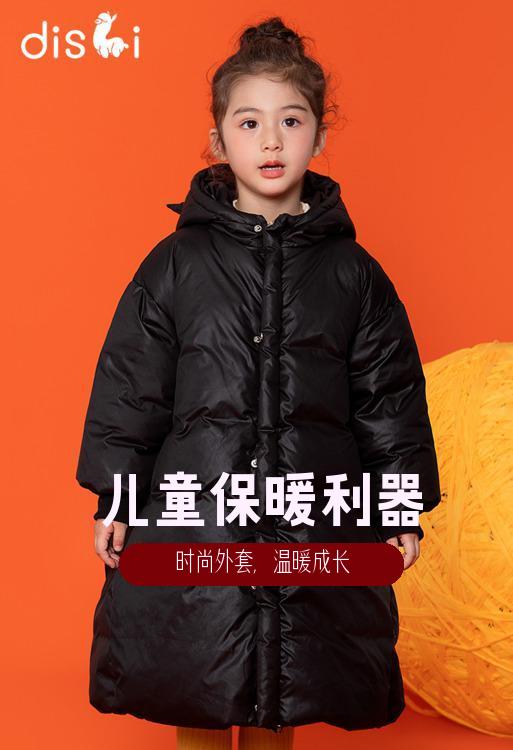}
    \end{subfigure}

    \begin{minipage}[c]{\textwidth}
    \centering
       Synthesized Taglines: [Children's Warmth Essential; Stylish Jacket for Cozy Growth.]
    \end{minipage}
    
    \begin{minipage}[c]{0.2\textwidth}
        \begin{minipage}[c]{0.05\textwidth}
            \rotatebox{90}{\small Instruction}
        \end{minipage}
        \hfill
        \begin{subfigure}[c]{0.92\textwidth}
            \centering
            \includegraphics[width=\textwidth]{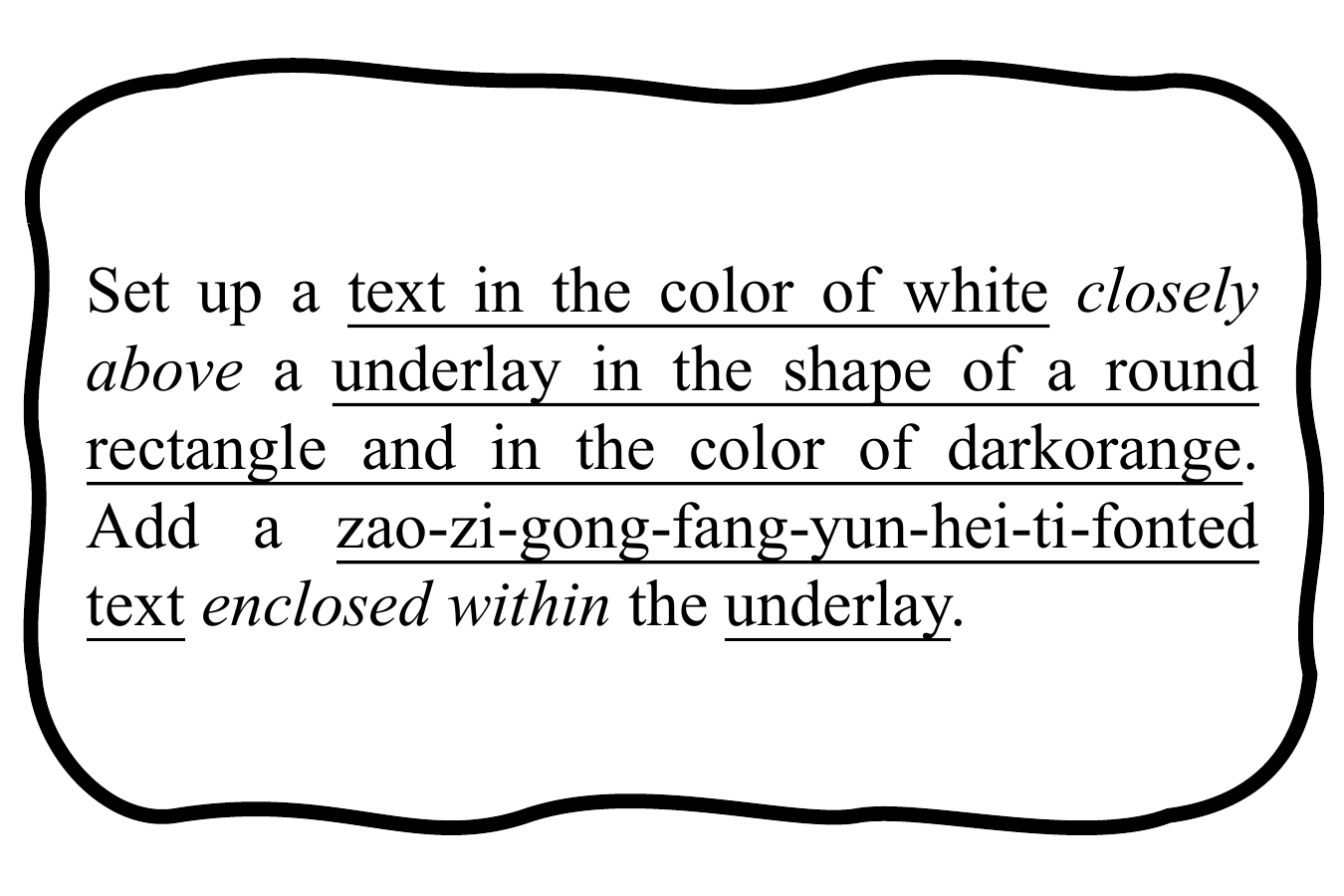}
        \end{subfigure}
        \begin{minipage}[c]{0.05\textwidth}
            \rotatebox{90}{\small Information}
        \end{minipage}
        \hfill
        \begin{subfigure}[c]{0.92\textwidth}
            \centering
            \includegraphics[width=\textwidth]{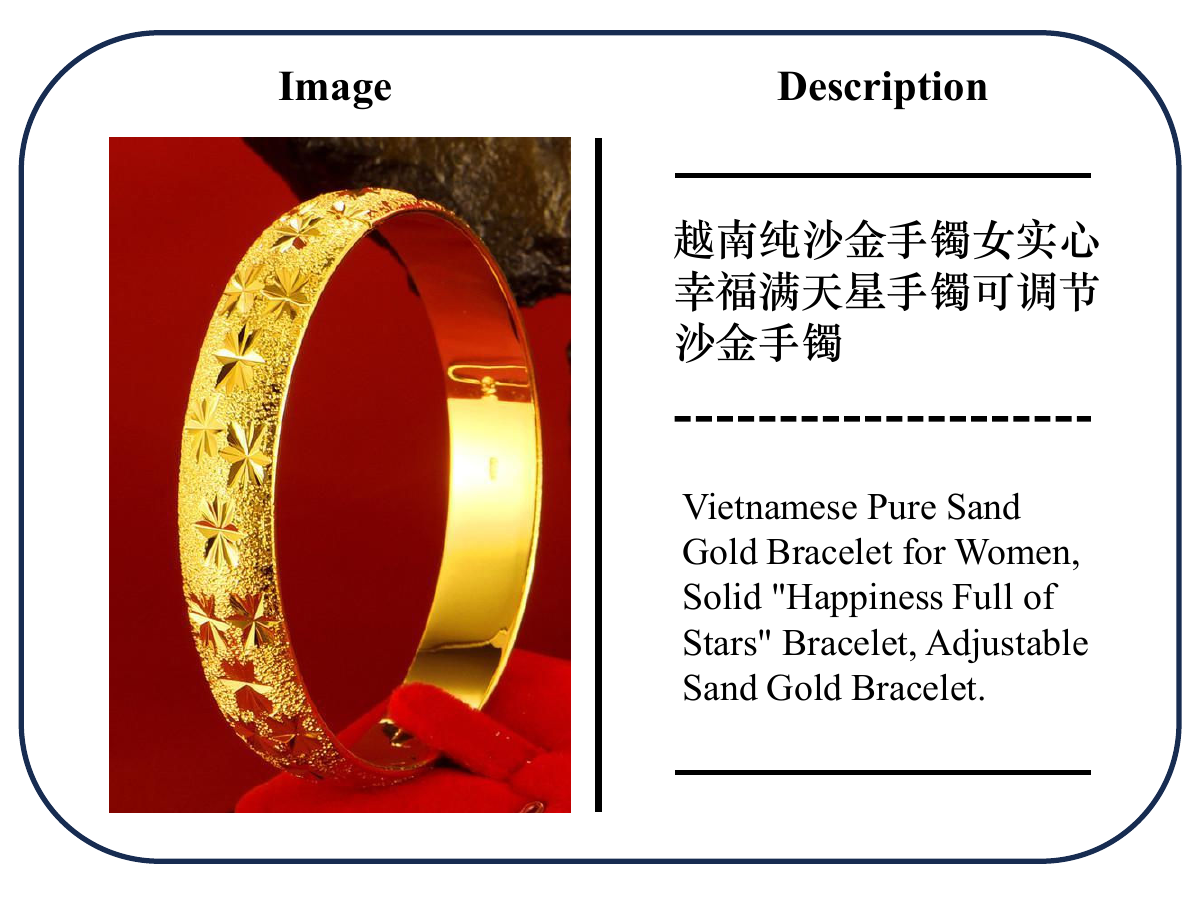}
        \end{subfigure}
    \end{minipage}
    \hfill
    \begin{subfigure}[c]{0.15\textwidth}
        \centering
        \includegraphics[width=\textwidth]{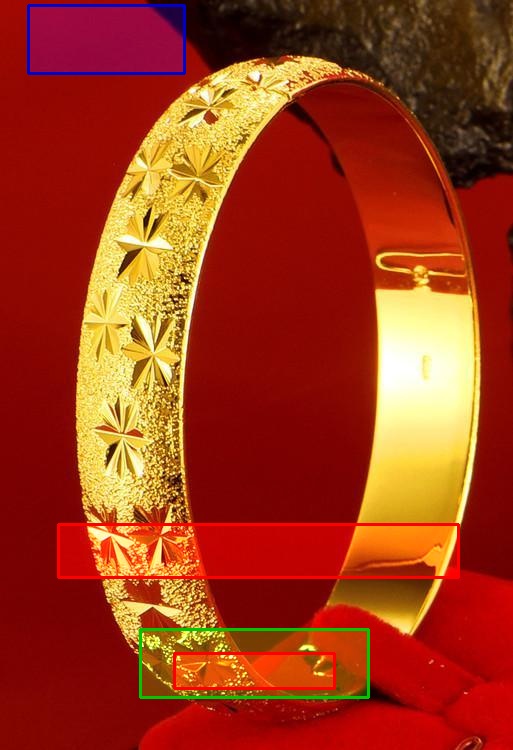}
    \end{subfigure}
    \hfill
    \begin{subfigure}[c]{0.15\textwidth}
        \centering
        \includegraphics[width=\textwidth]{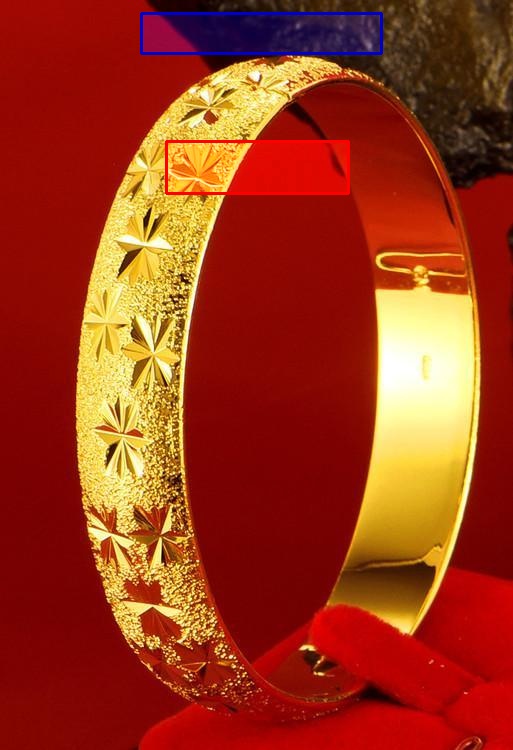}
    \end{subfigure}
    \hfill
    \begin{subfigure}[c]{0.15\textwidth}
        \centering
        \includegraphics[width=\textwidth]{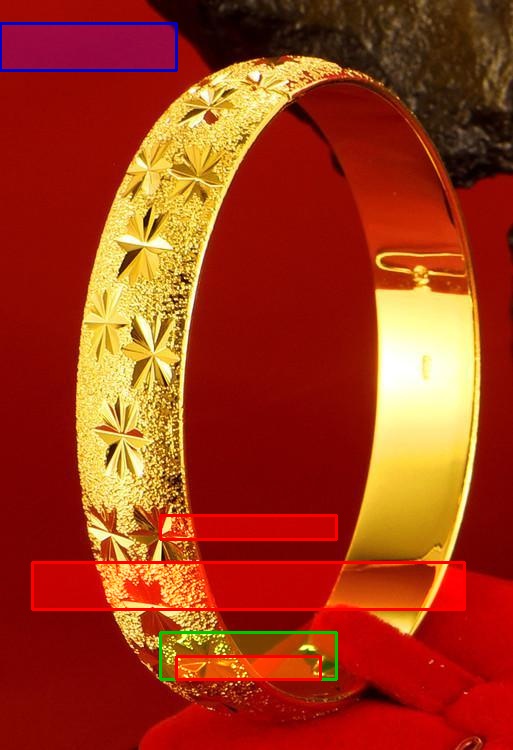}
    \end{subfigure}
    \hfill
    \begin{subfigure}[c]{0.15\textwidth}
        \centering
        \includegraphics[width=\textwidth]{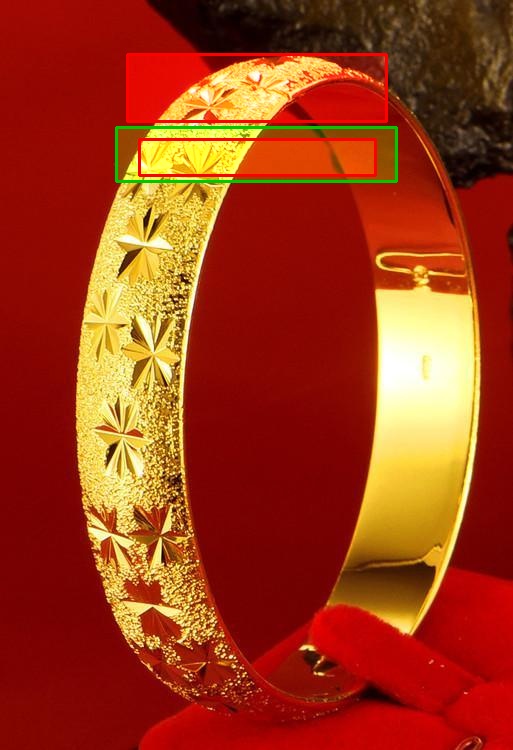}
    \end{subfigure}
    \hfill
    \begin{subfigure}[c]{0.15\textwidth}
        \centering
        \includegraphics[width=\textwidth]{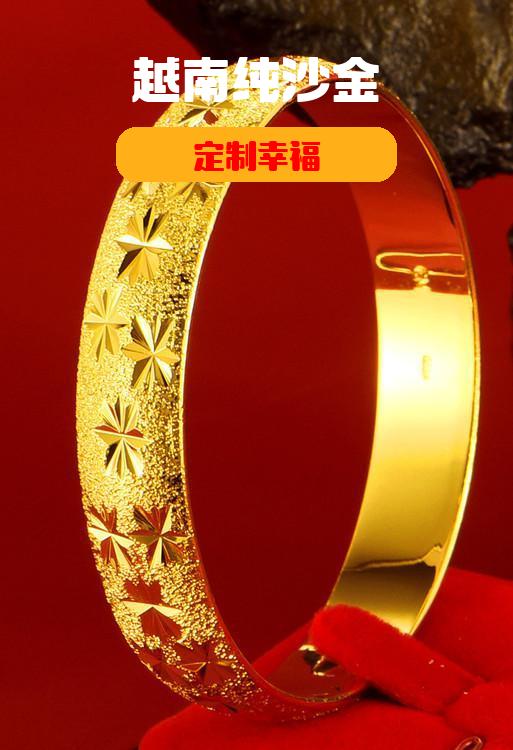}
    \end{subfigure}

    \begin{minipage}[c]{\textwidth}
    \centering
       Synthesized Taglines: [Vietnamese Pure Sand Gold; Customized Happiness.]
    \end{minipage}
    
    \begin{minipage}[c]{0.2\textwidth}
        \begin{minipage}[c]{0.05\textwidth}
            \rotatebox{90}{\small Instruction}
        \end{minipage}
        \hfill
        \begin{subfigure}[c]{0.92\textwidth}
            \centering
            \includegraphics[width=\textwidth]{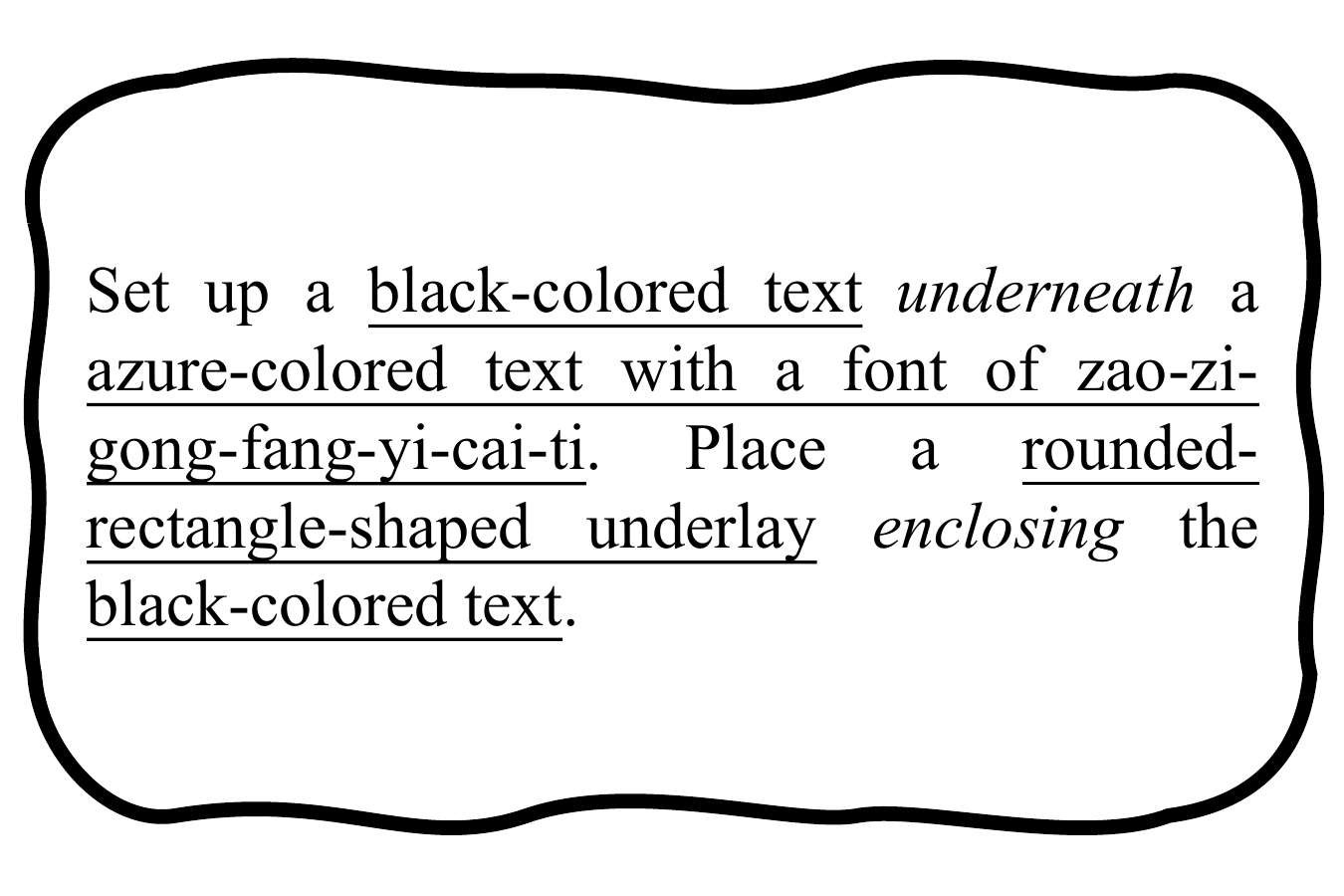}
        \end{subfigure}
        \begin{minipage}[c]{0.05\textwidth}
            \rotatebox{90}{\small Information}
        \end{minipage}
        \hfill
        \begin{subfigure}[c]{0.92\textwidth}
            \centering
            \includegraphics[width=\textwidth]{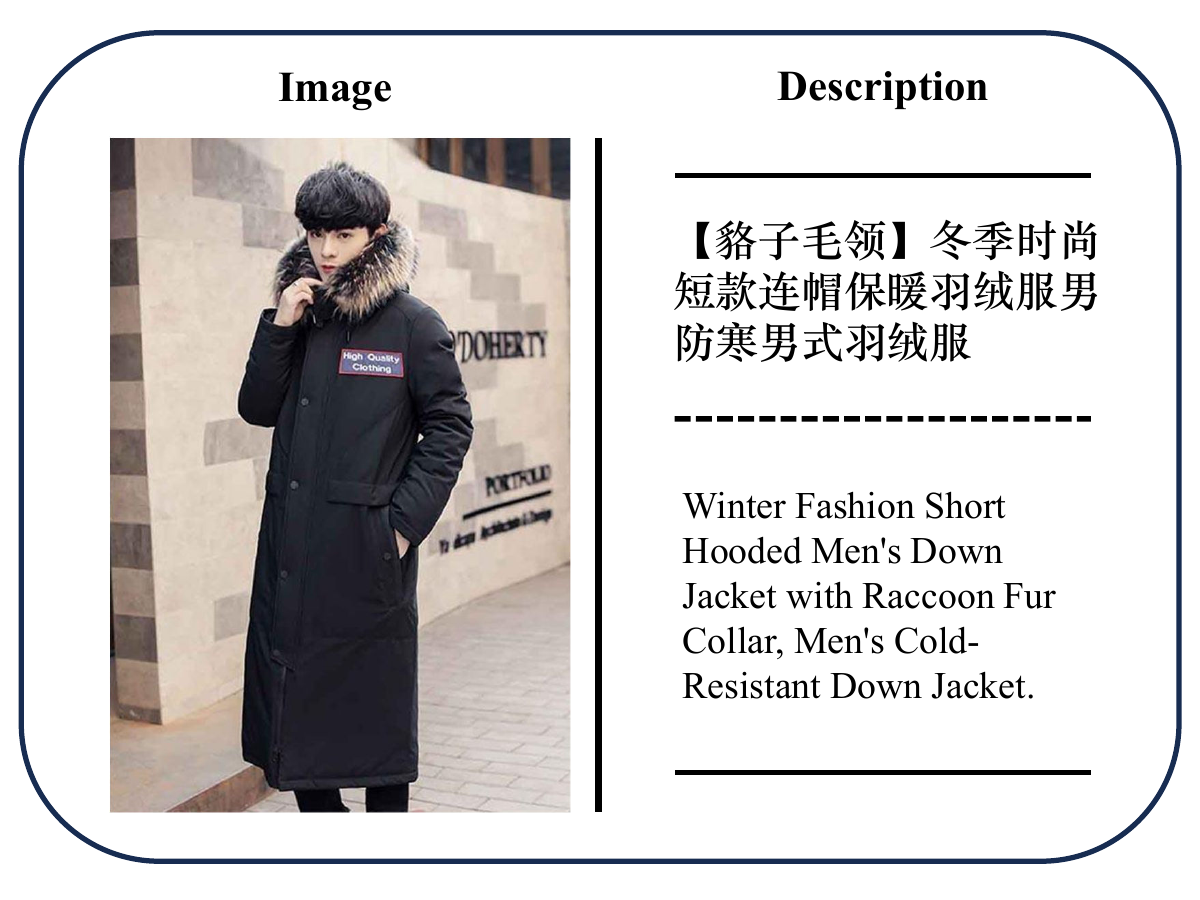}
        \end{subfigure}
    \end{minipage}
    \hfill
    \begin{subfigure}[c]{0.15\textwidth}
        \centering
        \includegraphics[width=\textwidth]{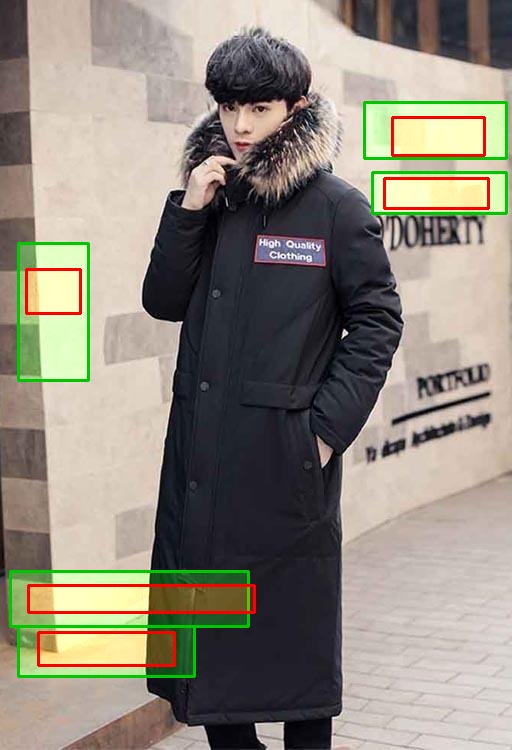}
    \end{subfigure}
    \hfill
    \begin{subfigure}[c]{0.15\textwidth}
        \centering
        \includegraphics[width=\textwidth]{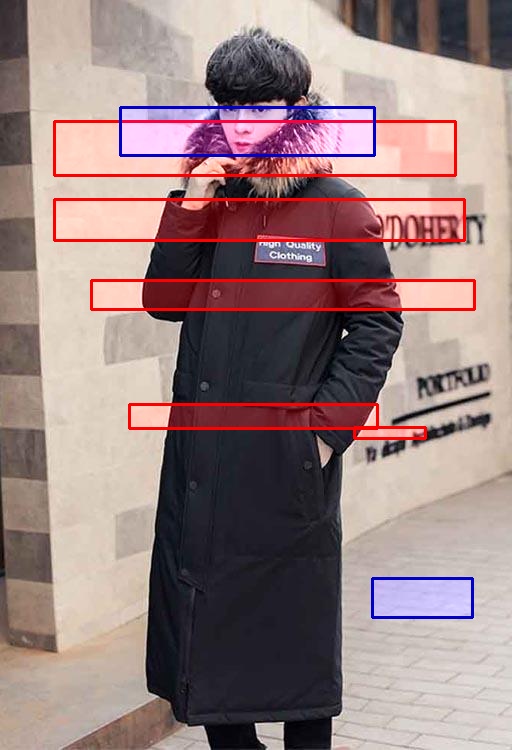}
    \end{subfigure}
    \hfill
    \begin{subfigure}[c]{0.15\textwidth}
        \centering
        \includegraphics[width=\textwidth]{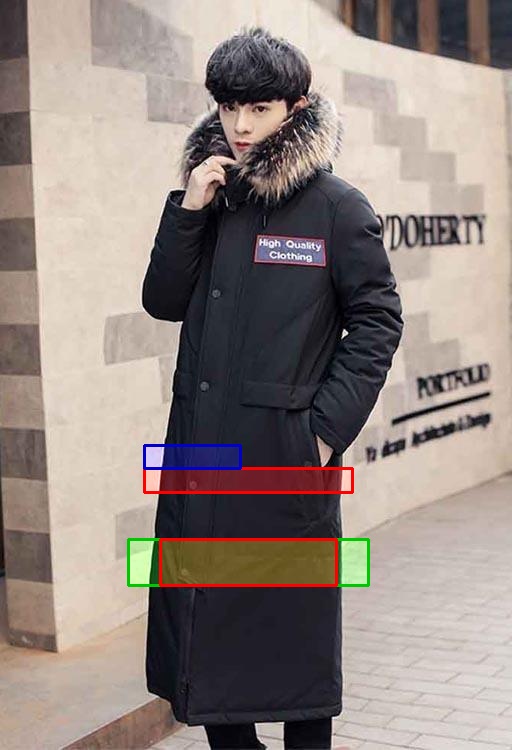}
    \end{subfigure}
    \hfill
    \begin{subfigure}[c]{0.15\textwidth}
        \centering
        \includegraphics[width=\textwidth]{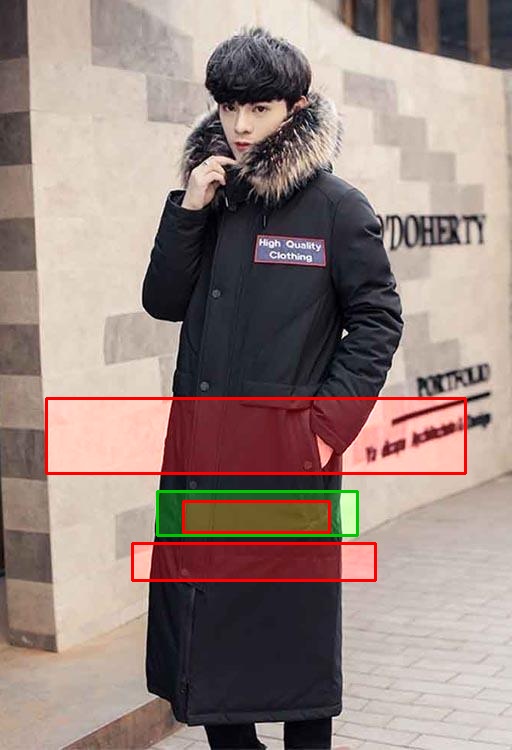}
    \end{subfigure}
    \hfill
    \begin{subfigure}[c]{0.15\textwidth}
        \centering
        \includegraphics[width=\textwidth]{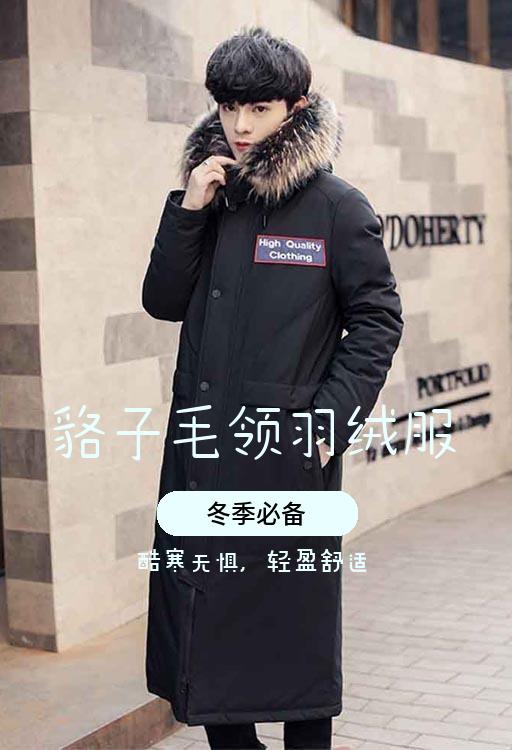}
    \end{subfigure}

    \begin{minipage}[c]{\textwidth}
    \centering
       Synthesized Taglines: [Raccoon Fur Down Jacket; Winter Essential;\\ Brave the Cold with Ease, Lightweight and Comfortable.]
    \end{minipage}
    
    \begin{minipage}[c]{0.2\textwidth}
        \begin{minipage}[c]{0.05\textwidth}
            \rotatebox{90}{\small Instruction}
        \end{minipage}
        \hfill
        \begin{subfigure}[c]{0.92\textwidth}
            \centering
            \includegraphics[width=\textwidth]{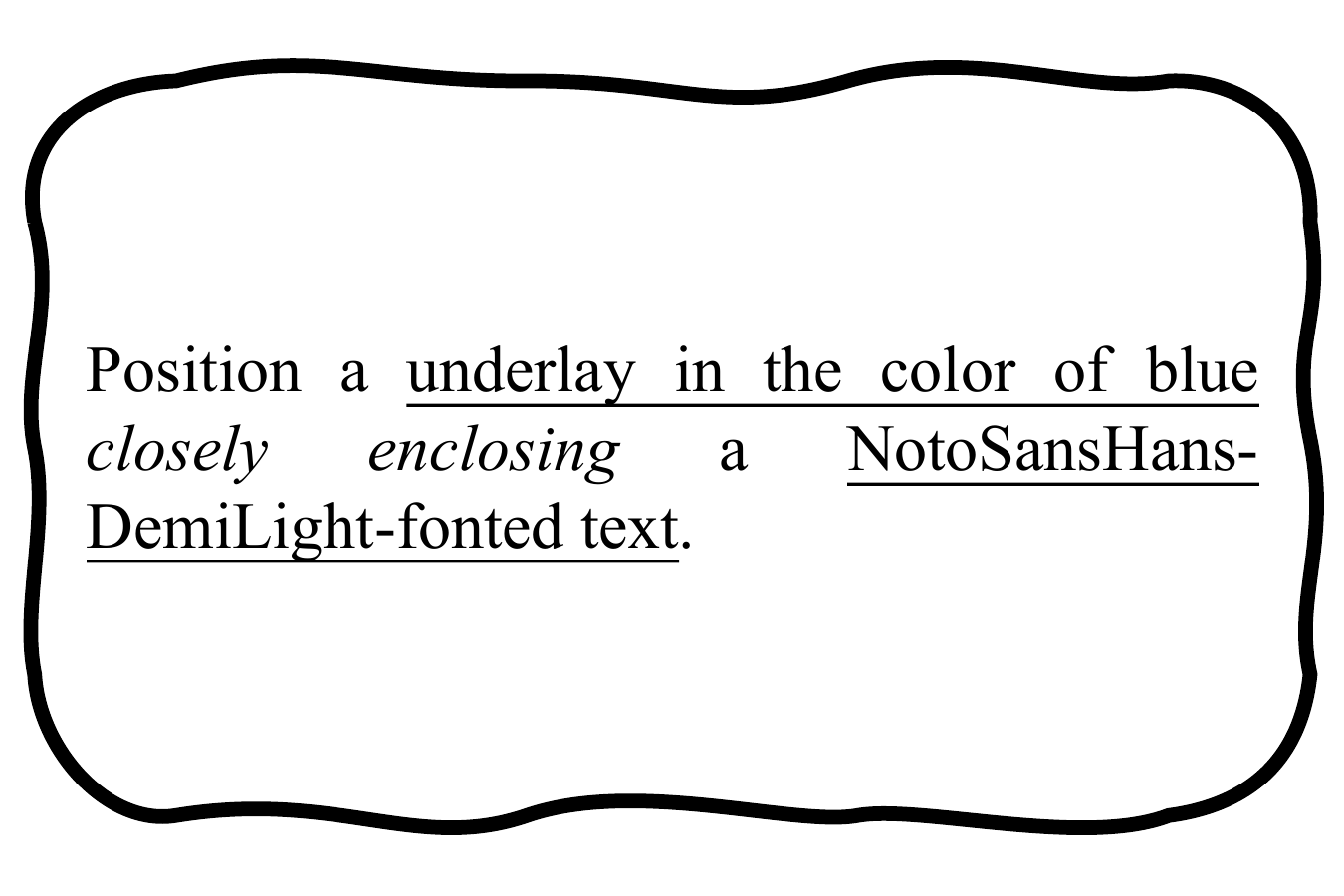}
        \end{subfigure}
        \begin{minipage}[c]{0.05\textwidth}
            \rotatebox{90}{\small Information}
        \end{minipage}
        \hfill
        \begin{subfigure}[c]{0.92\textwidth}
            \centering
            \includegraphics[width=\textwidth]{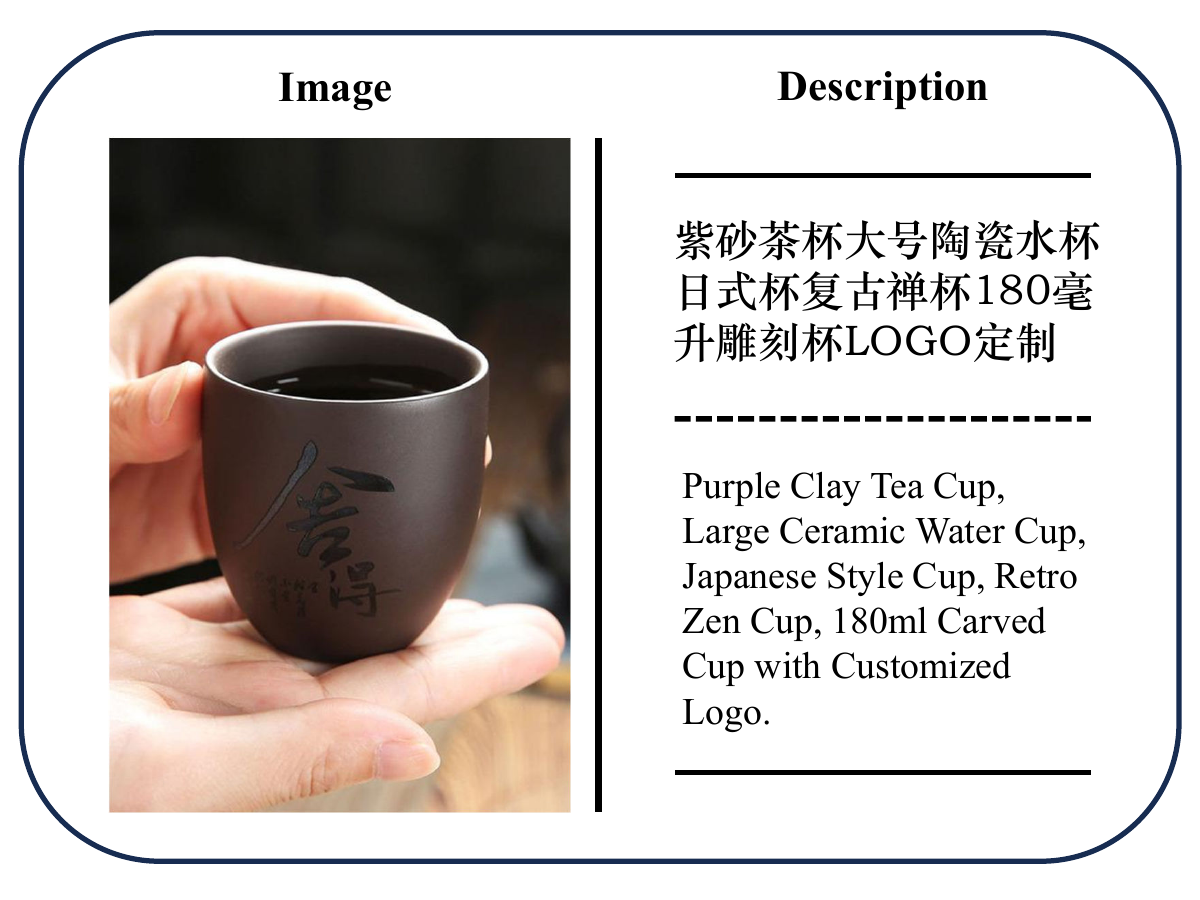}
        \end{subfigure}
    \end{minipage}
    \hfill
    \begin{subfigure}[c]{0.15\textwidth}
        \centering
        \includegraphics[width=\textwidth]{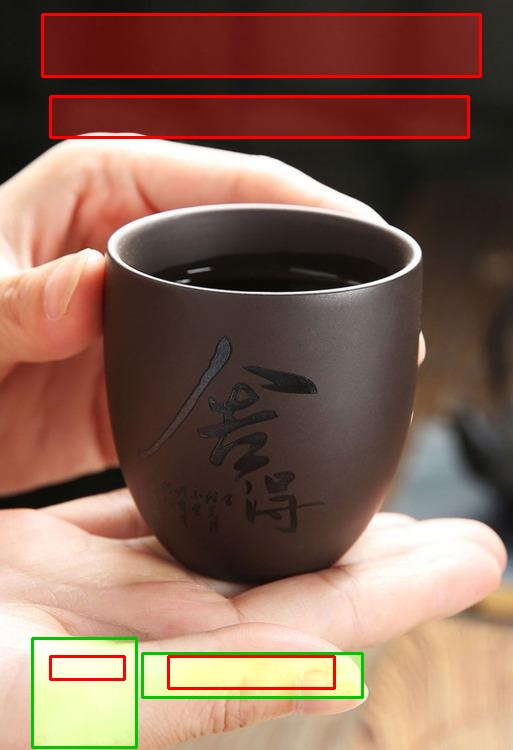}
    \end{subfigure}
    \hfill
    \begin{subfigure}[c]{0.15\textwidth}
        \centering
        \includegraphics[width=\textwidth]{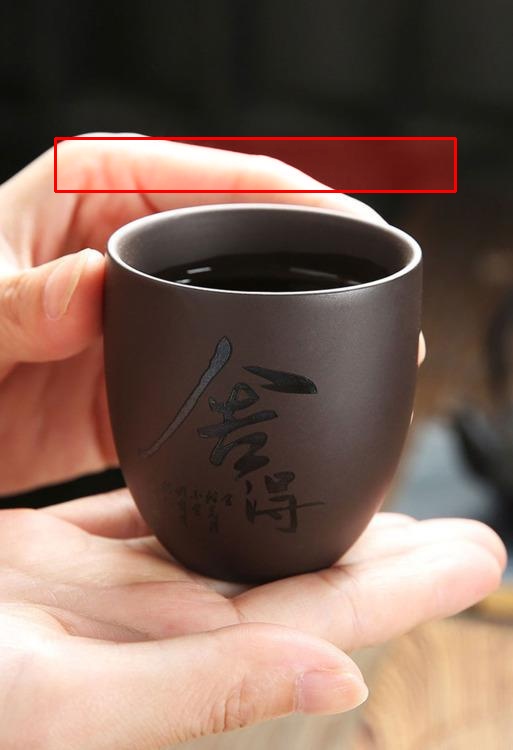}
    \end{subfigure}
    \hfill
    \begin{subfigure}[c]{0.15\textwidth}
        \centering
        \includegraphics[width=\textwidth]{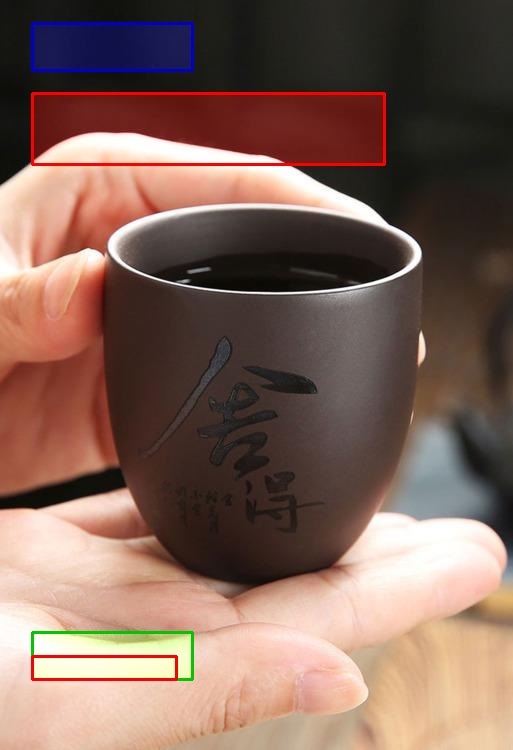}
    \end{subfigure}
    \hfill
    \begin{subfigure}[c]{0.15\textwidth}
        \centering
        \includegraphics[width=\textwidth]{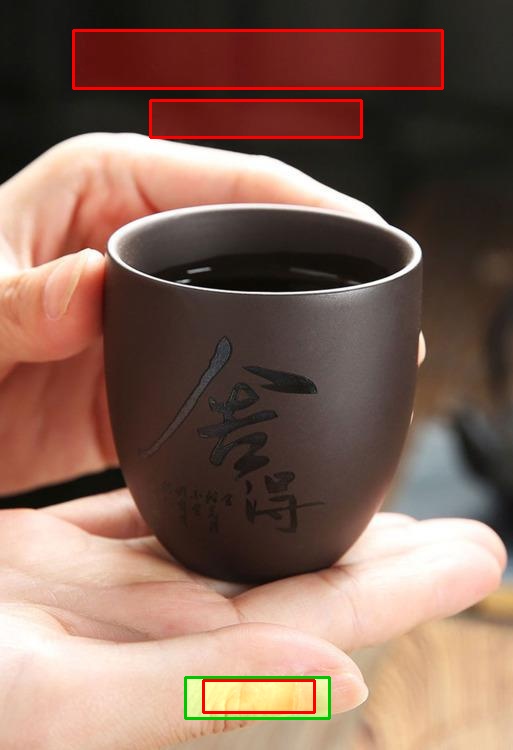}
    \end{subfigure}
    \hfill
    \begin{subfigure}[c]{0.15\textwidth}
        \centering
        \includegraphics[width=\textwidth]{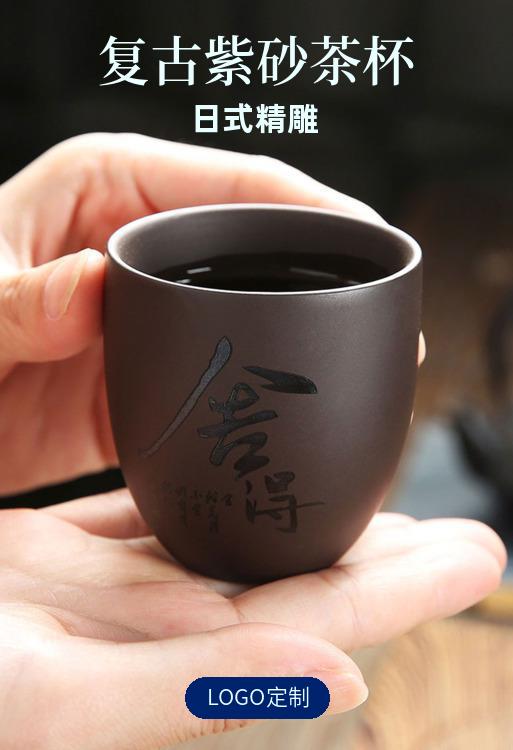}
    \end{subfigure}

    \begin{minipage}[c]{\textwidth}
    \centering
       Synthesized Taglines: [Vintage Purple Clay Tea Cup; Japanese Fine Carving; Custom Logo Design.]
    \end{minipage}
    
    \caption{Visualizations for instruction-drive synthesized 2D posters by LayoutTrans~\citep{Gupta_2021_ICCV}, LayoutVAE~\citep{Jyothi_2019_ICCV}, LayoutDM~\citep{Inoue_2023_CVPR} and our method. Note that posters are rendered according to synthesized graphic and textual features.}
    \label{fig:2d_vis_1}
\end{figure*}

\begin{figure*}[htbp]
    \centering

     \begin{minipage}[t]{0.2\textwidth}
        \begin{minipage}[t]{0.05\textwidth}
            
        \end{minipage}
        \hfill
        \begin{minipage}[t]{0.92\textwidth}
            \centering
            (a) Instruction \& Product Information
        \end{minipage}
    \end{minipage}
    \hfill
    \begin{minipage}[t]{0.15\textwidth}
        \centering
        (b) LayoutTrans
    \end{minipage}
    \hfill
    \begin{minipage}[t]{0.15\textwidth}
        \centering
        (c) LayoutVAE
    \end{minipage}
    \hfill
    \begin{minipage}[t]{0.15\textwidth}
        \centering
        (d) LayoutDM
    \end{minipage}
    \hfill
    \begin{minipage}[t]{0.15\textwidth}
        \centering
        (e) Ours
    \end{minipage}
    \hfill
    \begin{minipage}[t]{0.15\textwidth}
        \centering
        (f) Ours (Rendered)
    \end{minipage}
    
    \begin{minipage}[c]{0.2\textwidth}
        \begin{minipage}[c]{0.05\textwidth}
            \rotatebox{90}{\small Instruction}
        \end{minipage}
        \hfill
        \begin{subfigure}[c]{0.92\textwidth}
            \centering
            \includegraphics[width=\textwidth]{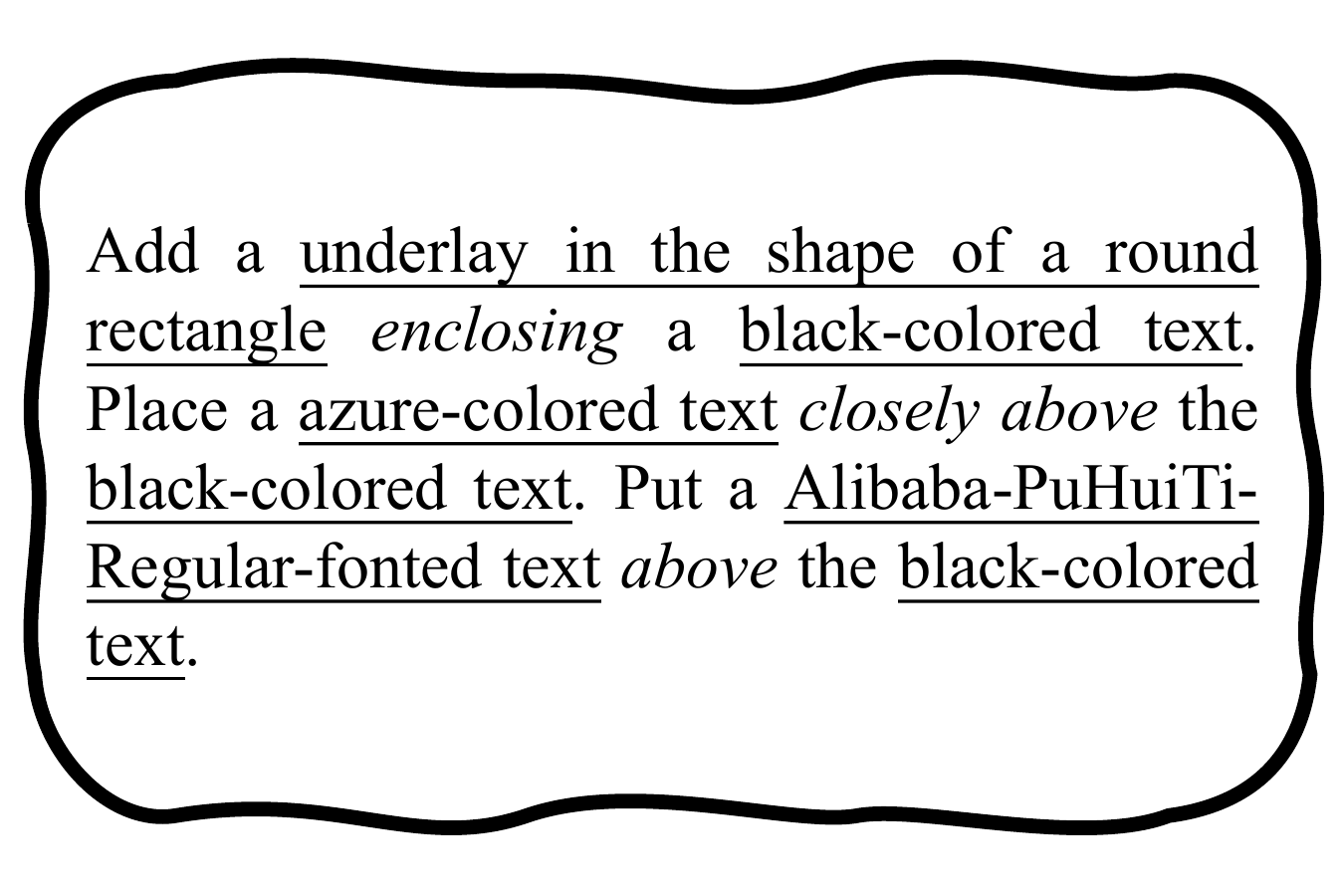}
        \end{subfigure}
        \begin{minipage}[c]{0.05\textwidth}
            \rotatebox{90}{\small Information}
        \end{minipage}
        \hfill
        \begin{subfigure}[c]{0.92\textwidth}
            \centering
            \includegraphics[width=\textwidth]{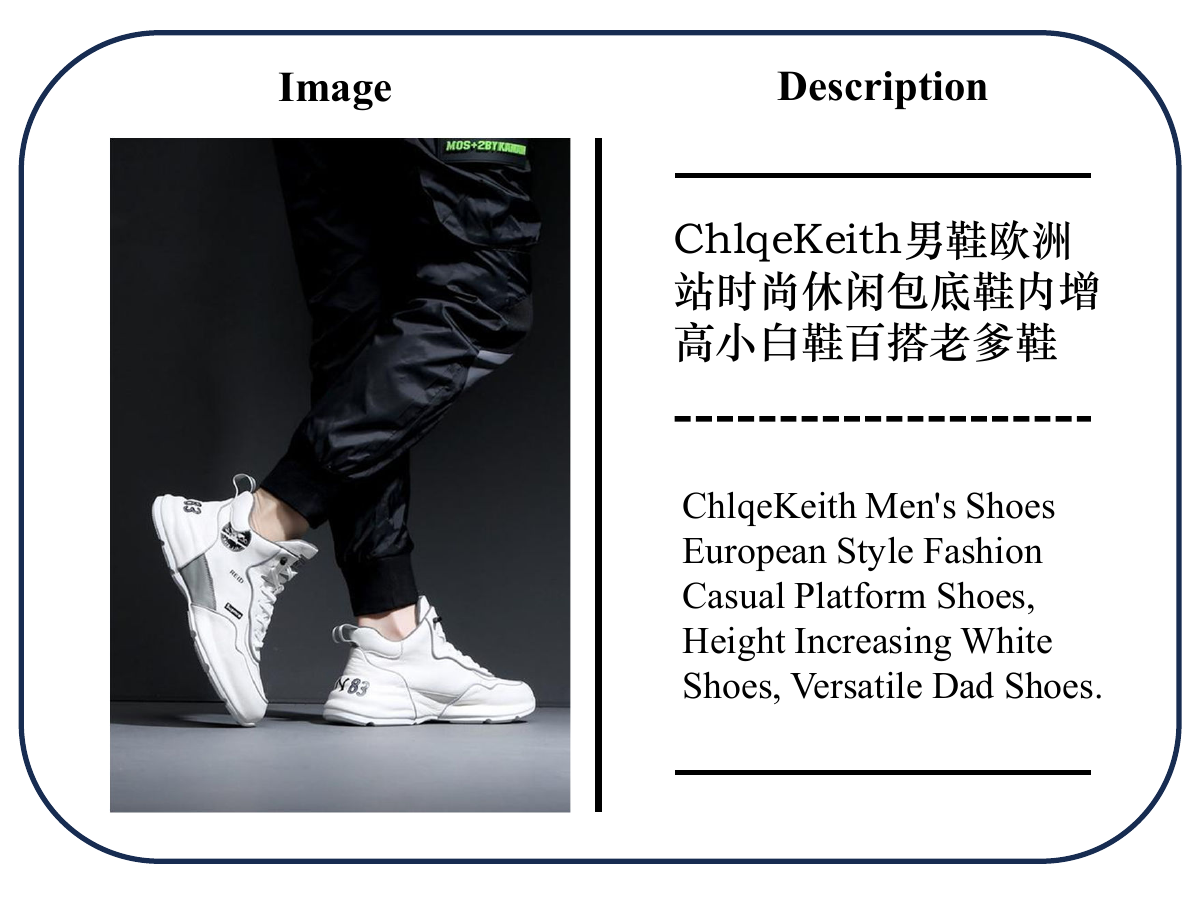}
        \end{subfigure}
    \end{minipage}
    \hfill
    \begin{subfigure}[c]{0.15\textwidth}
        \centering
        \includegraphics[width=\textwidth]{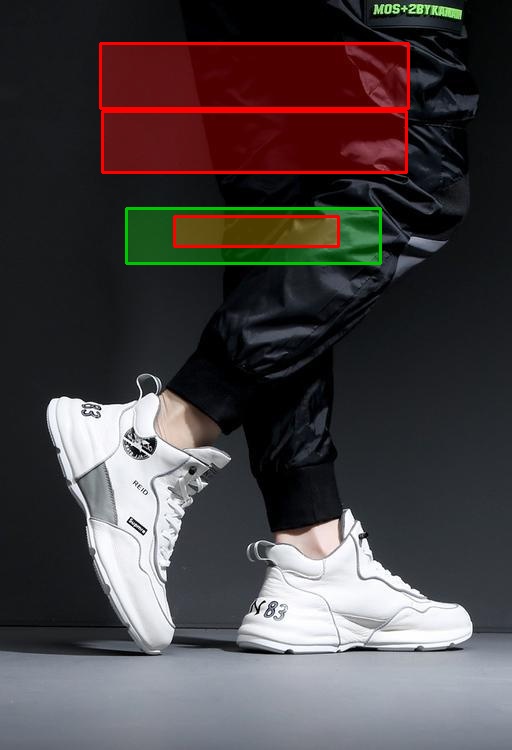}
    \end{subfigure}
    \hfill
    \begin{subfigure}[c]{0.15\textwidth}
        \centering
        \includegraphics[width=\textwidth]{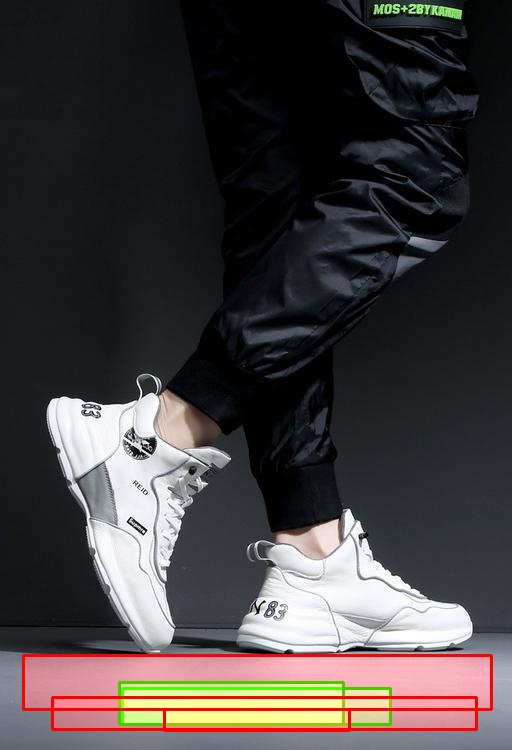}
    \end{subfigure}
    \hfill
    \begin{subfigure}[c]{0.15\textwidth}
        \centering
        \includegraphics[width=\textwidth]{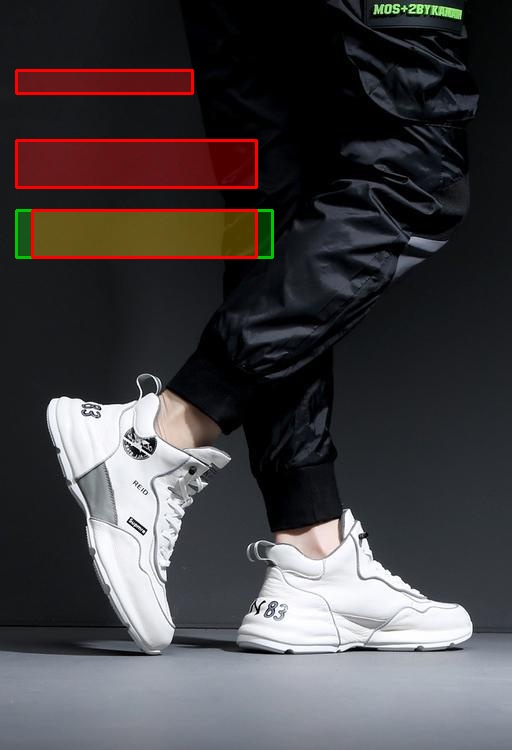}
    \end{subfigure}
    \hfill
    \begin{subfigure}[c]{0.15\textwidth}
        \centering
        \includegraphics[width=\textwidth]{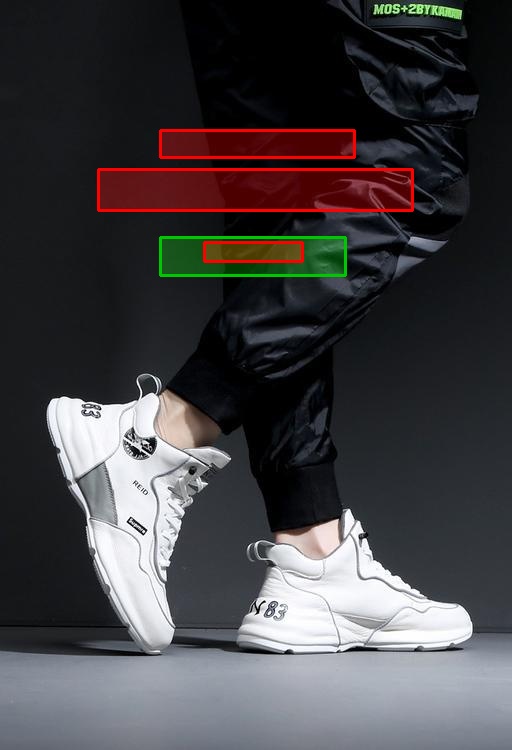}
    \end{subfigure}
    \hfill
    \begin{subfigure}[c]{0.15\textwidth}
        \centering
        \includegraphics[width=\textwidth]{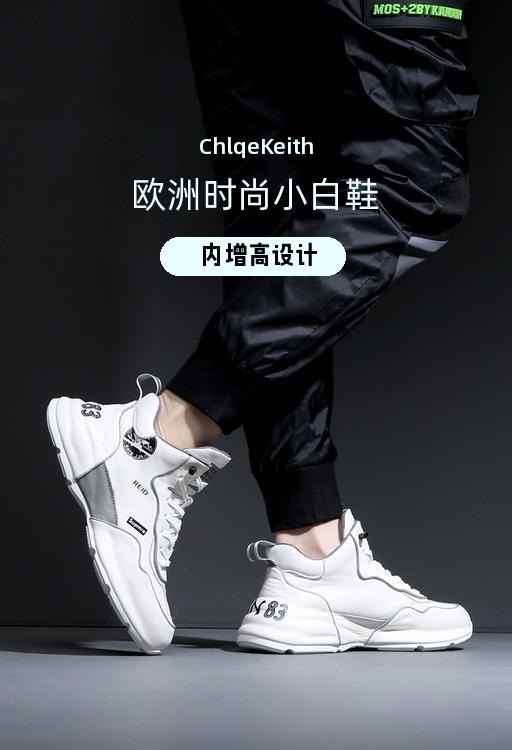}
    \end{subfigure}
    
    \begin{minipage}[c]{\textwidth}
    \centering
       Synthesized Taglines: [European Fashion White Shoes; Internal Height Increase Design.]
    \end{minipage}
    
    \begin{minipage}[c]{0.2\textwidth}
        \begin{minipage}[c]{0.05\textwidth}
            \rotatebox{90}{\small Instruction}
        \end{minipage}
        \hfill
        \begin{subfigure}[c]{0.92\textwidth}
            \centering
            \includegraphics[width=\textwidth]{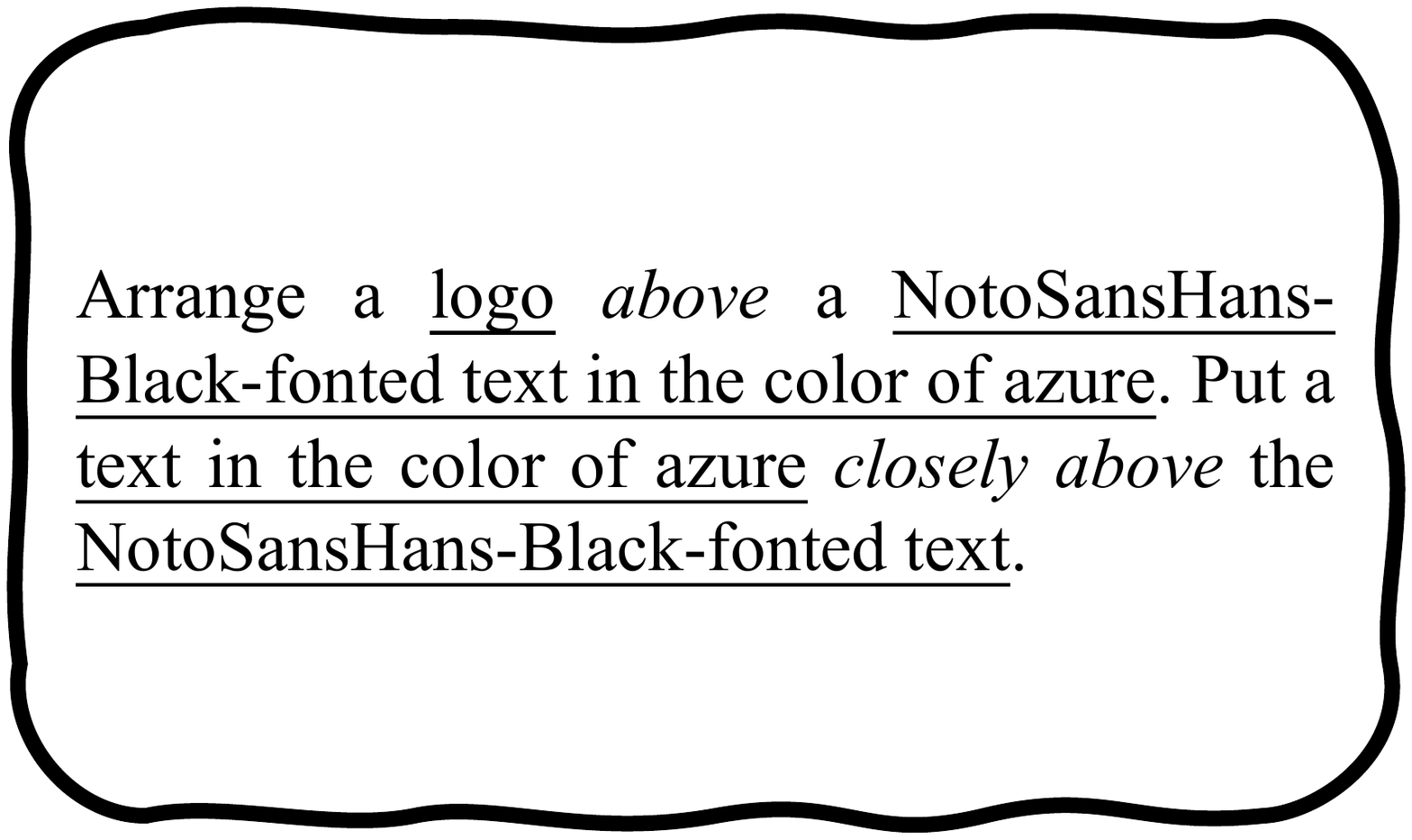}
        \end{subfigure}
        \begin{minipage}[c]{0.05\textwidth}
            \rotatebox{90}{\small Information}
        \end{minipage}
        \hfill
        \begin{subfigure}[c]{0.92\textwidth}
            \centering
            \includegraphics[width=\textwidth]{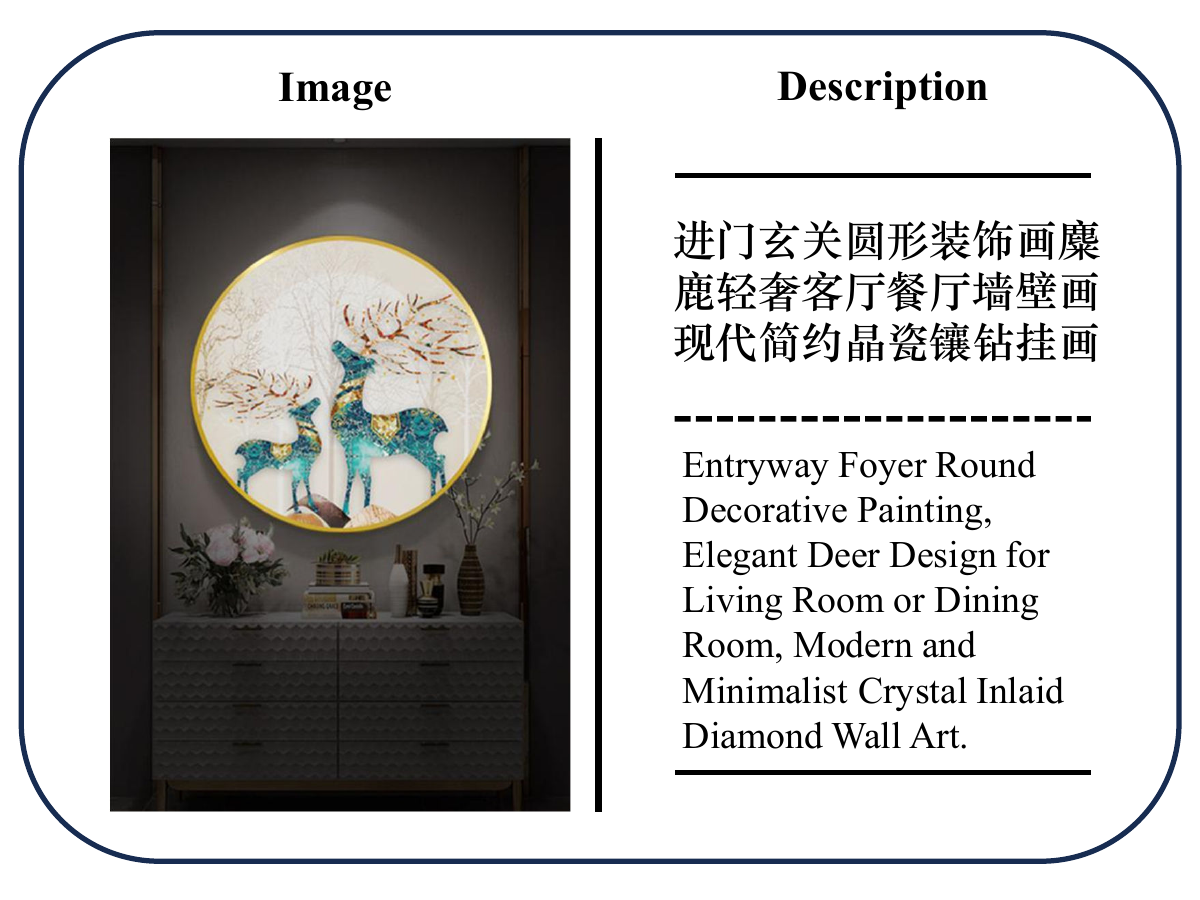}
        \end{subfigure}
    \end{minipage}
    \hfill
    \begin{subfigure}[c]{0.15\textwidth}
        \centering
        \includegraphics[width=\textwidth]{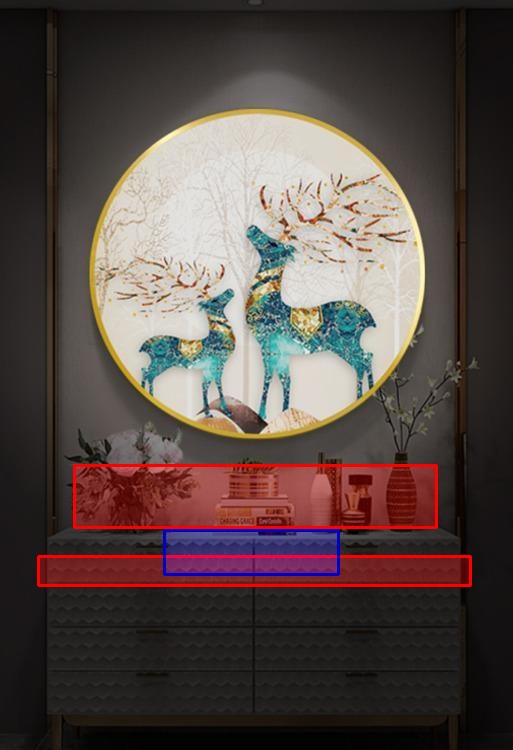}
    \end{subfigure}
    \hfill
    \begin{subfigure}[c]{0.15\textwidth}
        \centering
        \includegraphics[width=\textwidth]{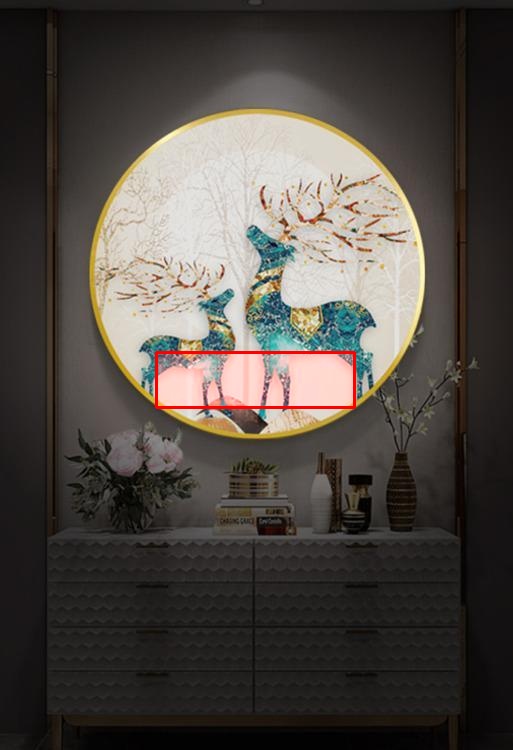}
    \end{subfigure}
    \hfill
    \begin{subfigure}[c]{0.15\textwidth}
        \centering
        \includegraphics[width=\textwidth]{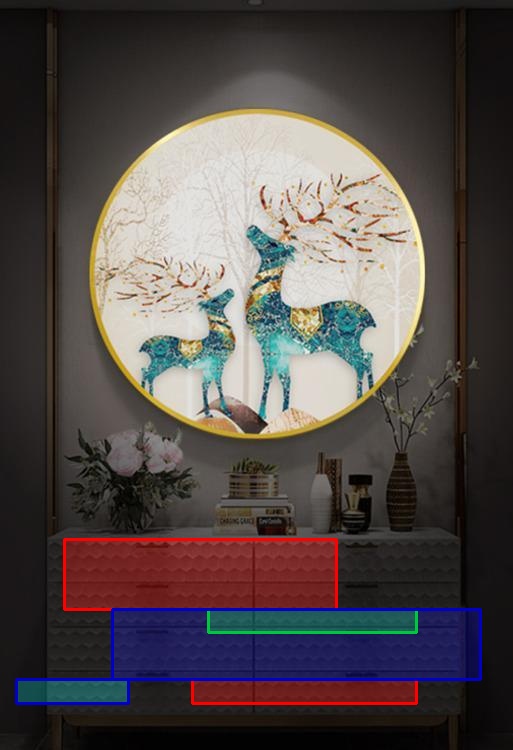}
    \end{subfigure}
    \hfill
    \begin{subfigure}[c]{0.15\textwidth}
        \centering
        \includegraphics[width=\textwidth]{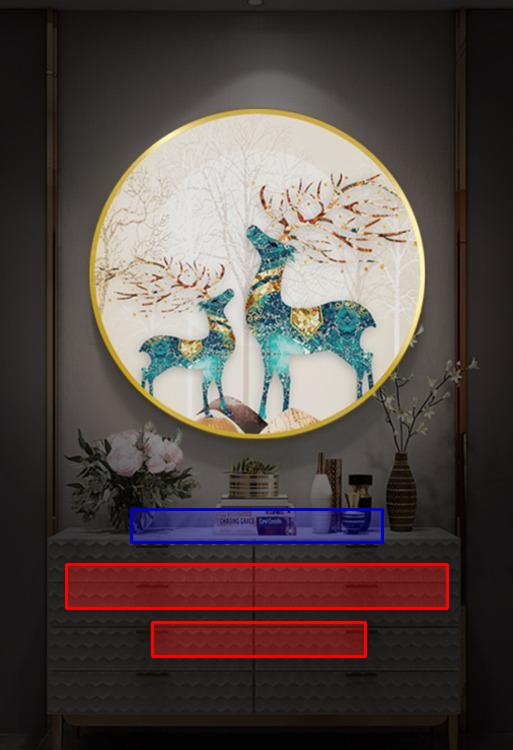}
    \end{subfigure}
    \hfill
    \begin{subfigure}[c]{0.15\textwidth}
        \centering
        \includegraphics[width=\textwidth]{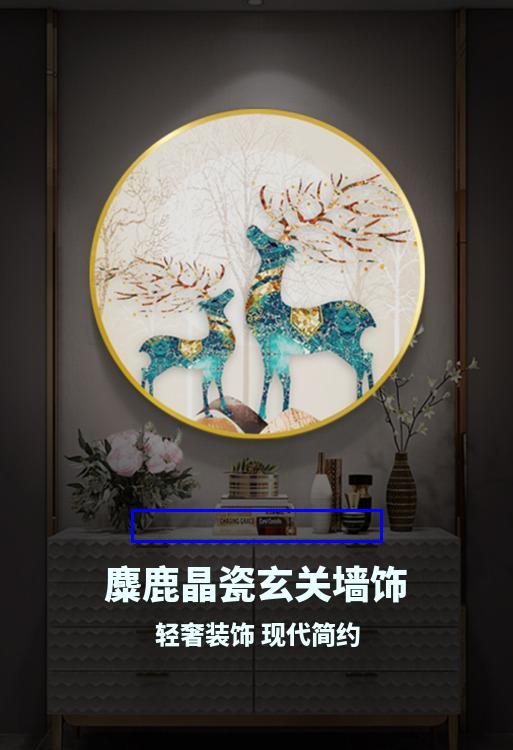}
    \end{subfigure}
    
    \begin{minipage}[c]{\textwidth}
    \centering
       Synthesized Taglines: [Deer Crystal Porcelain Wall Decoration for the Foyer; Light Luxury Modern Minimalist Decor.]
    \end{minipage}
    
    \begin{minipage}[c]{0.2\textwidth}
        \begin{minipage}[c]{0.05\textwidth}
            \rotatebox{90}{\small Instruction}
        \end{minipage}
        \hfill
        \begin{subfigure}[c]{0.92\textwidth}
            \centering
            \includegraphics[width=\textwidth]{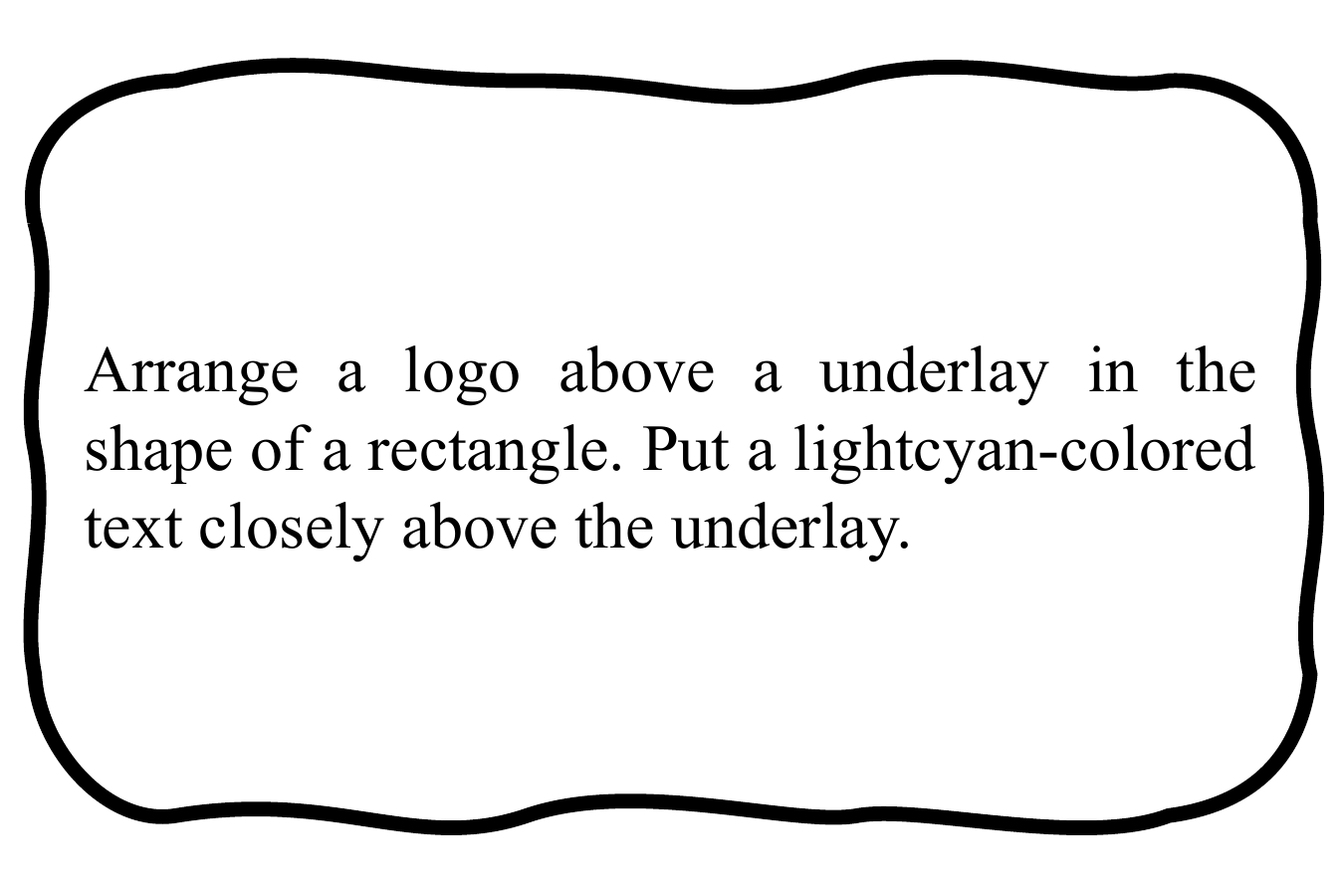}
        \end{subfigure}
        \begin{minipage}[c]{0.05\textwidth}
            \rotatebox{90}{\small Information}
        \end{minipage}
        \hfill
        \begin{subfigure}[c]{0.92\textwidth}
            \centering
            \includegraphics[width=\textwidth]{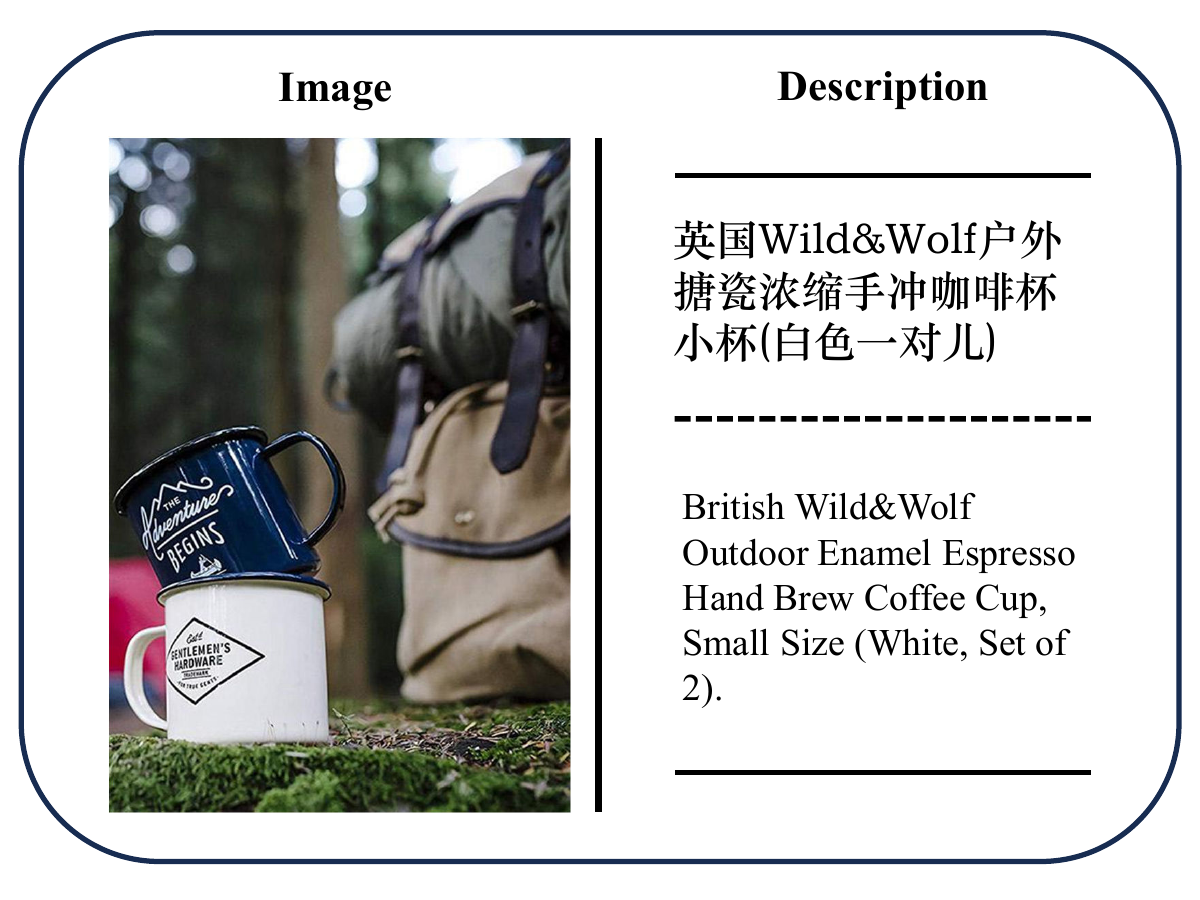}
        \end{subfigure}
    \end{minipage}
    \hfill
    \begin{subfigure}[c]{0.15\textwidth}
        \centering
        \includegraphics[width=\textwidth]{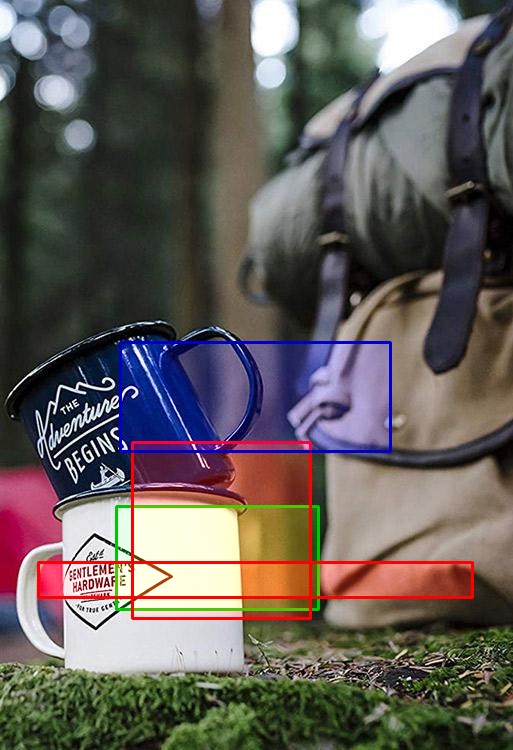}
    \end{subfigure}
    \hfill
    \begin{subfigure}[c]{0.15\textwidth}
        \centering
        \includegraphics[width=\textwidth]{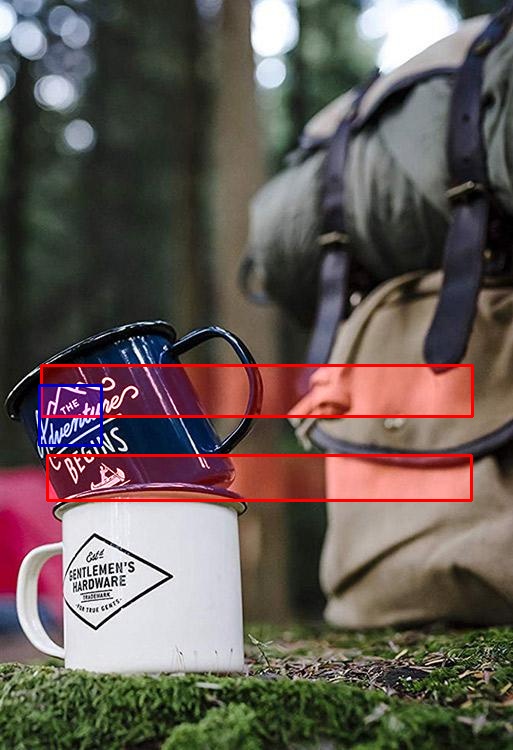}
    \end{subfigure}
    \hfill
    \begin{subfigure}[c]{0.15\textwidth}
        \centering
        \includegraphics[width=\textwidth]{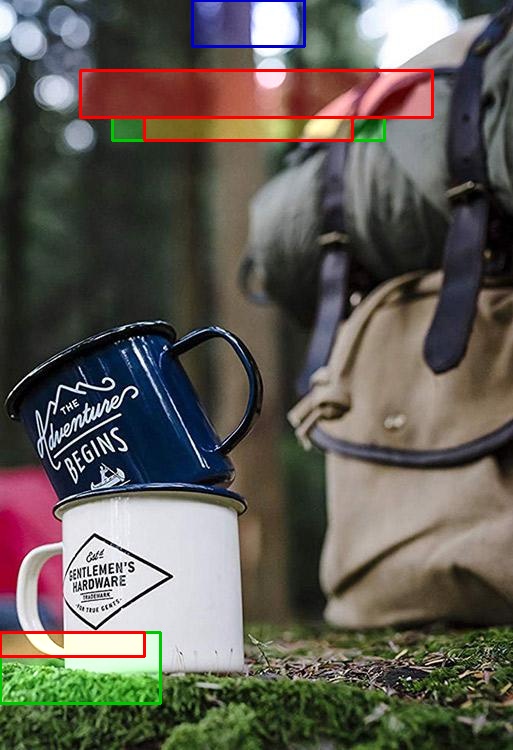}
    \end{subfigure}
    \hfill
    \begin{subfigure}[c]{0.15\textwidth}
        \centering
        \includegraphics[width=\textwidth]{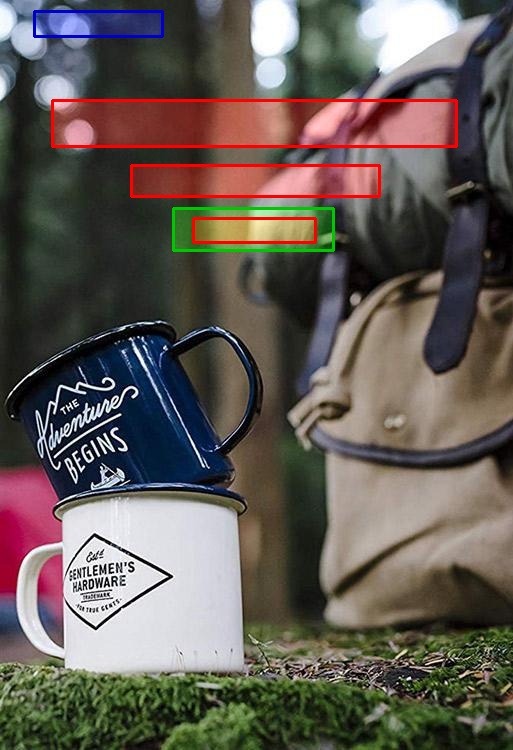}
    \end{subfigure}
    \hfill
    \begin{subfigure}[c]{0.15\textwidth}
        \centering
        \includegraphics[width=\textwidth]{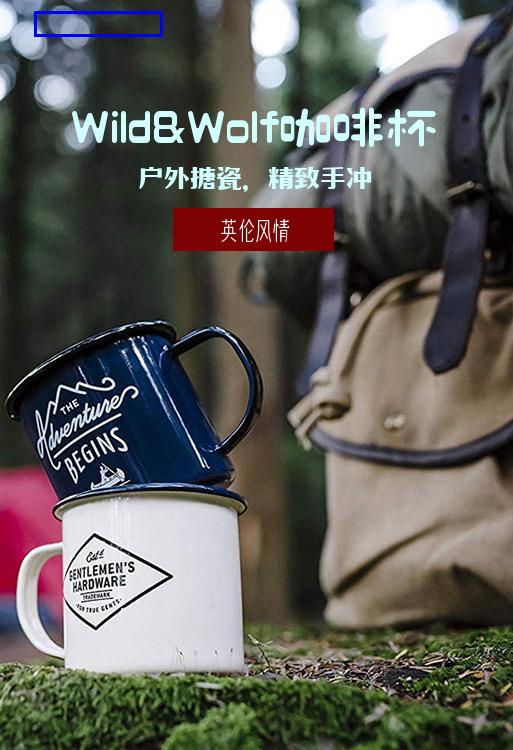}
    \end{subfigure}
    
    \begin{minipage}[c]{\textwidth}
    \centering
       Synthesized Taglines: [Wild\&Wolf Coffee Cup; Delicate Enamel Hand Brewing for Outdoor Use; British Style.]
    \end{minipage}
    
    \begin{minipage}[c]{0.2\textwidth}
        \begin{minipage}[c]{0.05\textwidth}
            \rotatebox{90}{\small Instruction}
        \end{minipage}
        \hfill
        \begin{subfigure}[c]{0.92\textwidth}
            \centering
            \includegraphics[width=\textwidth]{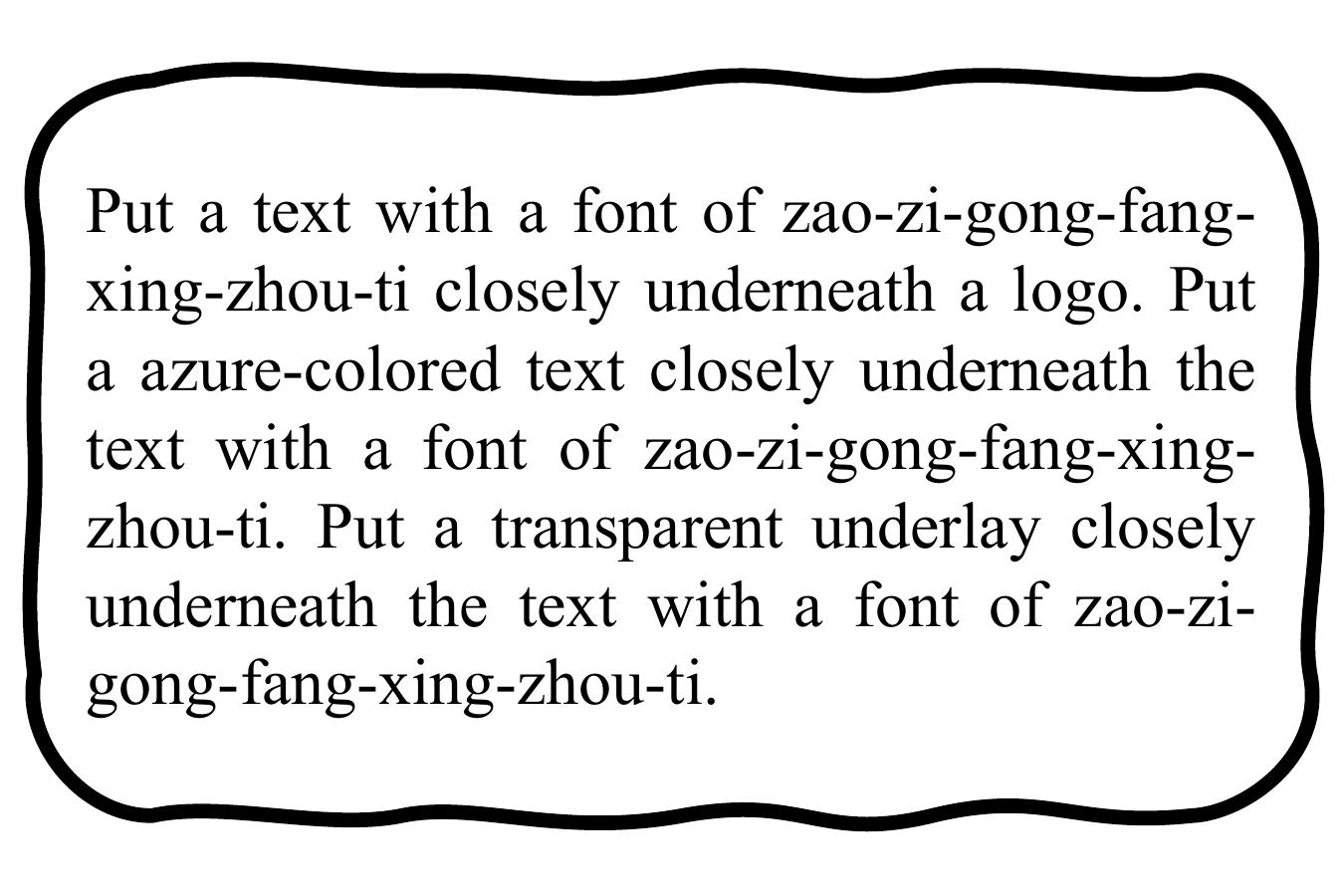}
        \end{subfigure}
        \begin{minipage}[c]{0.05\textwidth}
            \rotatebox{90}{\small Information}
        \end{minipage}
        \hfill
        \begin{subfigure}[c]{0.92\textwidth}
            \centering
            \includegraphics[width=\textwidth]{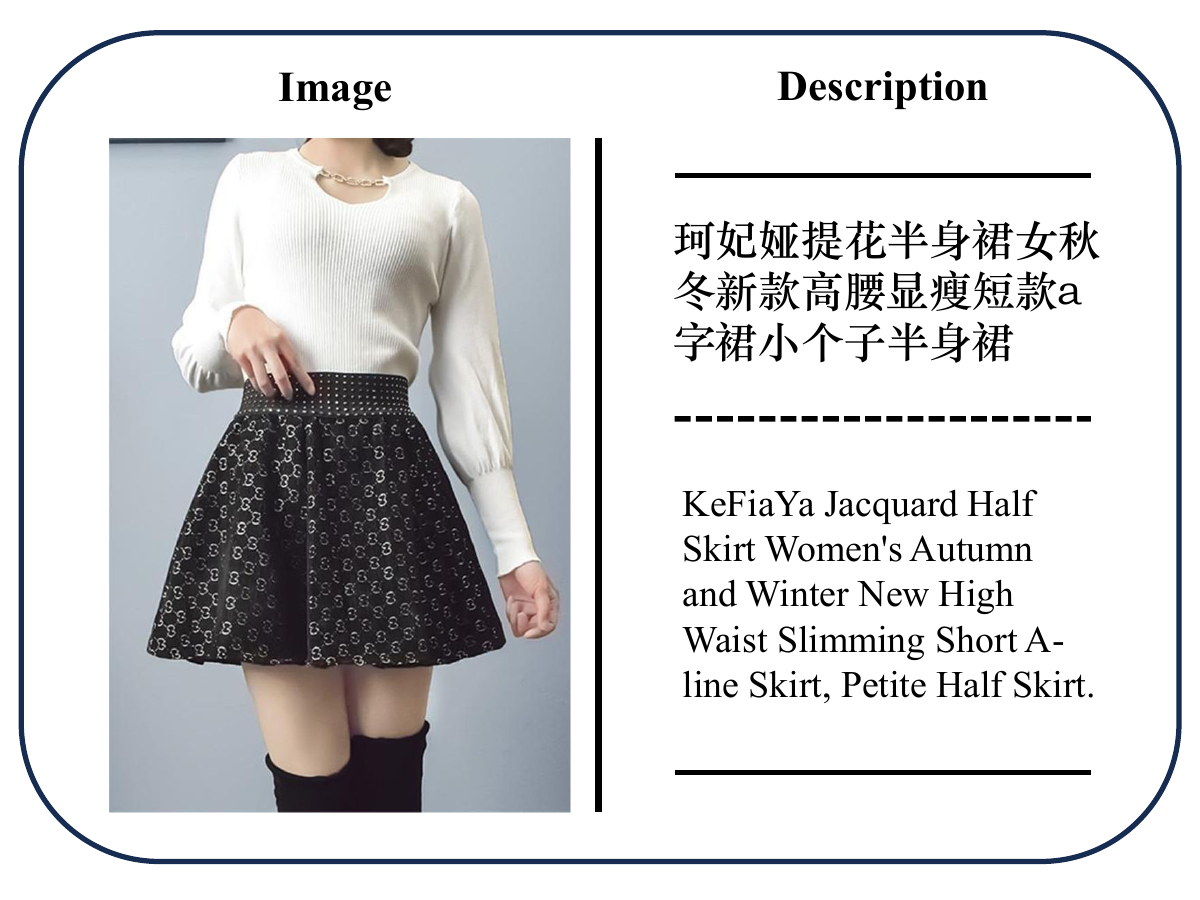}
        \end{subfigure}
    \end{minipage}
    \hfill
    \begin{subfigure}[c]{0.15\textwidth}
        \centering
        \includegraphics[width=\textwidth]{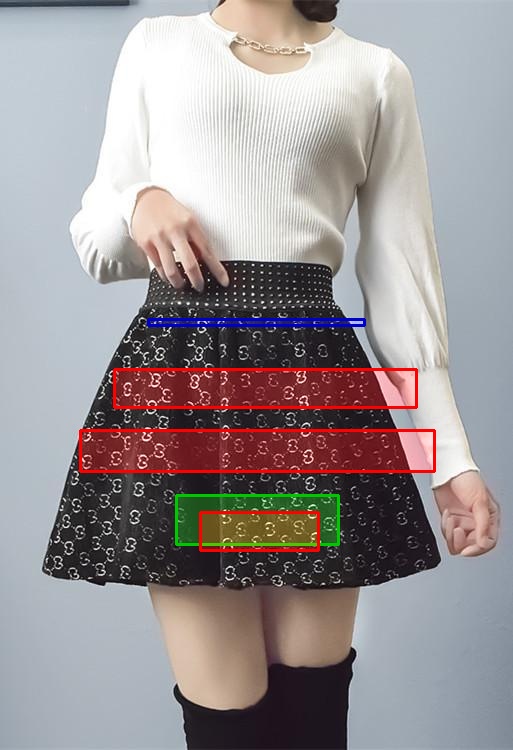}
    \end{subfigure}
    \hfill
    \begin{subfigure}[c]{0.15\textwidth}
        \centering
        \includegraphics[width=\textwidth]{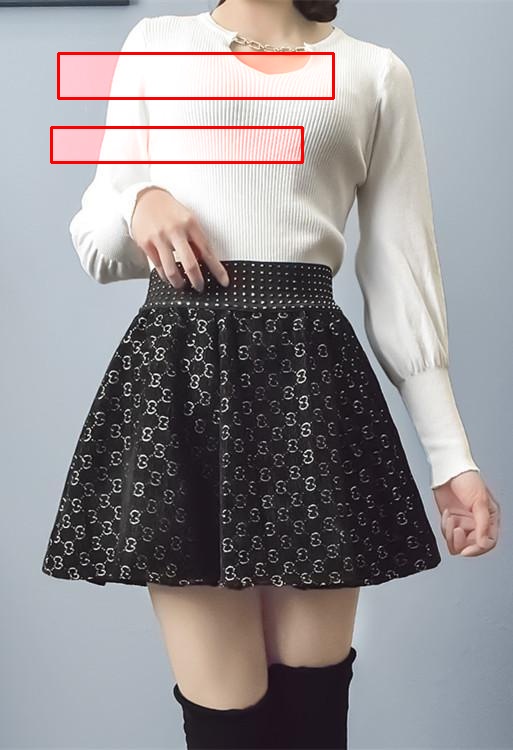}
    \end{subfigure}
    \hfill
    \begin{subfigure}[c]{0.15\textwidth}
        \centering
        \includegraphics[width=\textwidth]{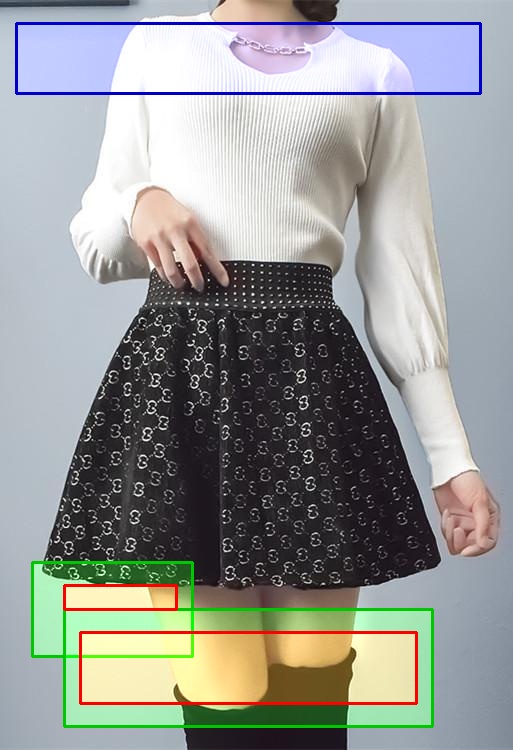}
    \end{subfigure}
    \hfill
    \begin{subfigure}[c]{0.15\textwidth}
        \centering
        \includegraphics[width=\textwidth]{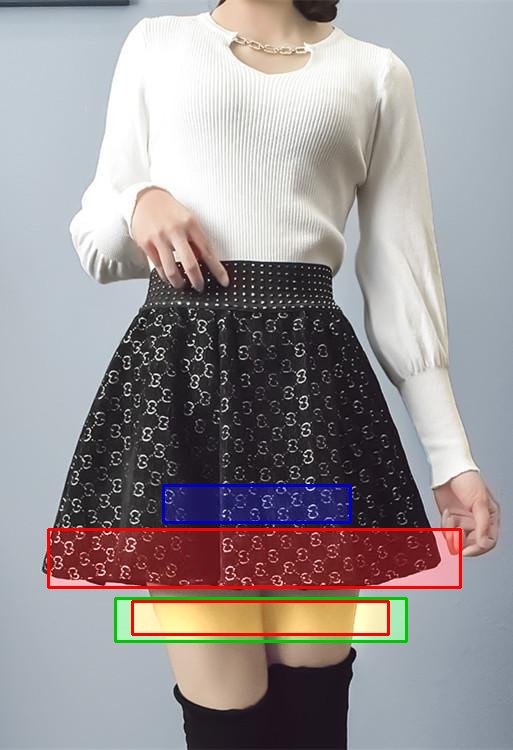}
    \end{subfigure}
    \hfill
    \begin{subfigure}[c]{0.15\textwidth}
        \centering
        \includegraphics[width=\textwidth]{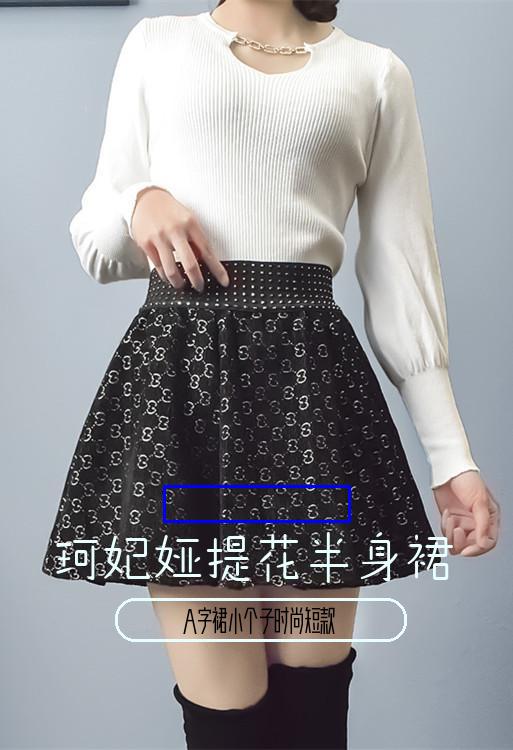}
    \end{subfigure}
    
    \begin{minipage}[c]{\textwidth}
    \centering
       Synthesized Taglines: [KeFiaYa Jacquard Half Skirt; A-line Style Short Skirt for Petite and Fashionable Looks.]
    \end{minipage}
    
    \caption{Visualizations for instruction-drive synthesized 2D posters by LayoutTrans~\citep{Gupta_2021_ICCV}, LayoutVAE~\citep{Jyothi_2019_ICCV}, LayoutDM~\citep{Inoue_2023_CVPR} and our method. Note that posters are rendered according to synthesized graphic and textual features.}
    \label{fig:2d_vis_2}
\end{figure*}

\begin{figure*}[htbp]
    \centering

     \begin{minipage}[t]{0.2\textwidth}
        \begin{minipage}[t]{0.05\textwidth}
            
        \end{minipage}
        \hfill
        \begin{minipage}[t]{0.92\textwidth}
            \centering
            (a) Instruction \& Product Information
        \end{minipage}
    \end{minipage}
    \hfill
    \begin{minipage}[t]{0.15\textwidth}
        \centering
        (b) LayoutTrans
    \end{minipage}
    \hfill
    \begin{minipage}[t]{0.15\textwidth}
        \centering
        (c) LayoutVAE
    \end{minipage}
    \hfill
    \begin{minipage}[t]{0.15\textwidth}
        \centering
        (d) LayoutDM
    \end{minipage}
    \hfill
    \begin{minipage}[t]{0.15\textwidth}
        \centering
        (e) Ours
    \end{minipage}
    \hfill
    \begin{minipage}[t]{0.15\textwidth}
        \centering
        (f) Ours (Rendered)
    \end{minipage}
    
    \begin{minipage}[c]{0.2\textwidth}
        \begin{minipage}[c]{0.05\textwidth}
            \rotatebox{90}{\small Instruction}
        \end{minipage}
        \hfill
        \begin{subfigure}[c]{0.92\textwidth}
            \centering
            \includegraphics[width=\textwidth]{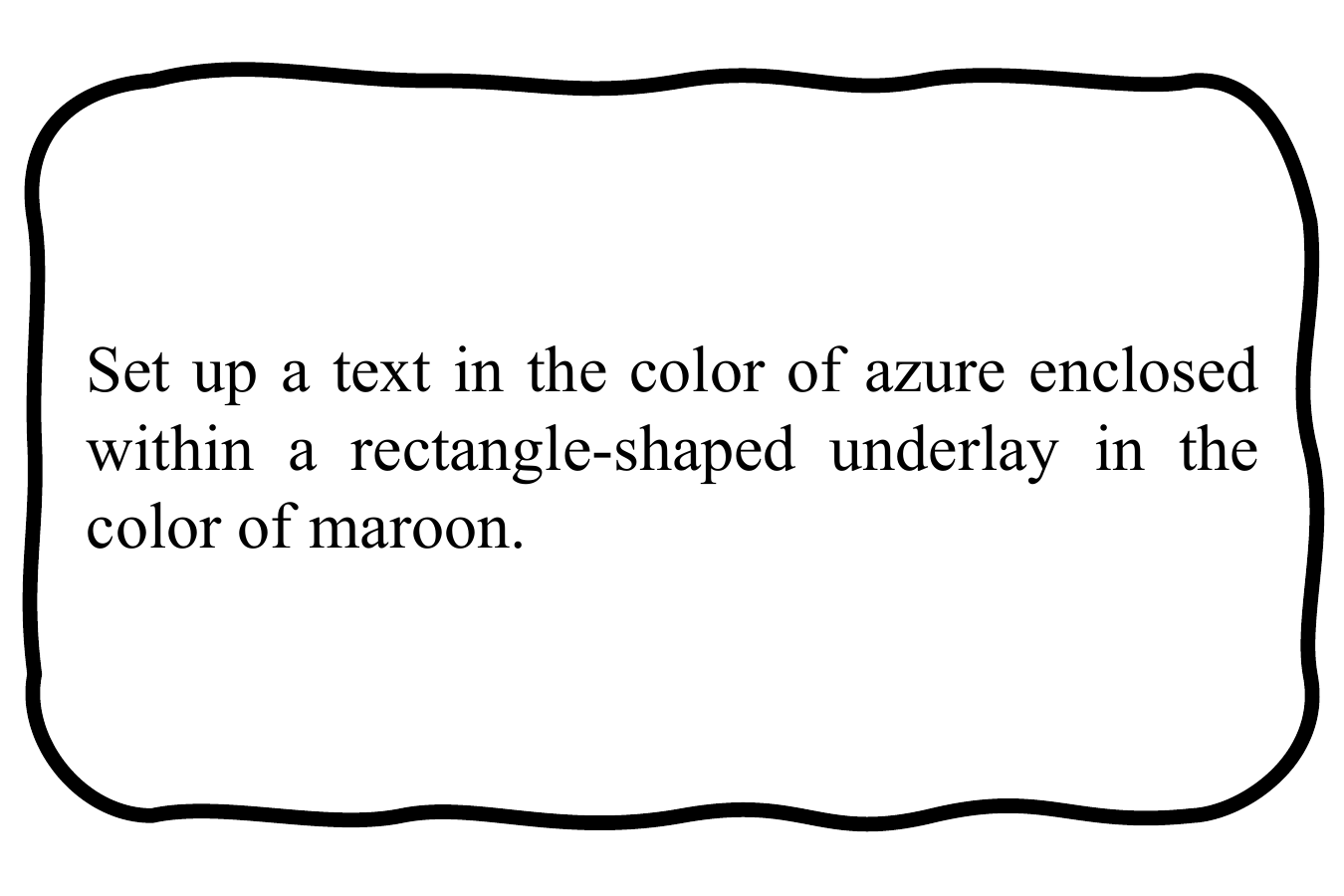}
        \end{subfigure}
        \begin{minipage}[c]{0.05\textwidth}
            \rotatebox{90}{\small Information}
        \end{minipage}
        \hfill
        \begin{subfigure}[c]{0.92\textwidth}
            \centering
            \includegraphics[width=\textwidth]{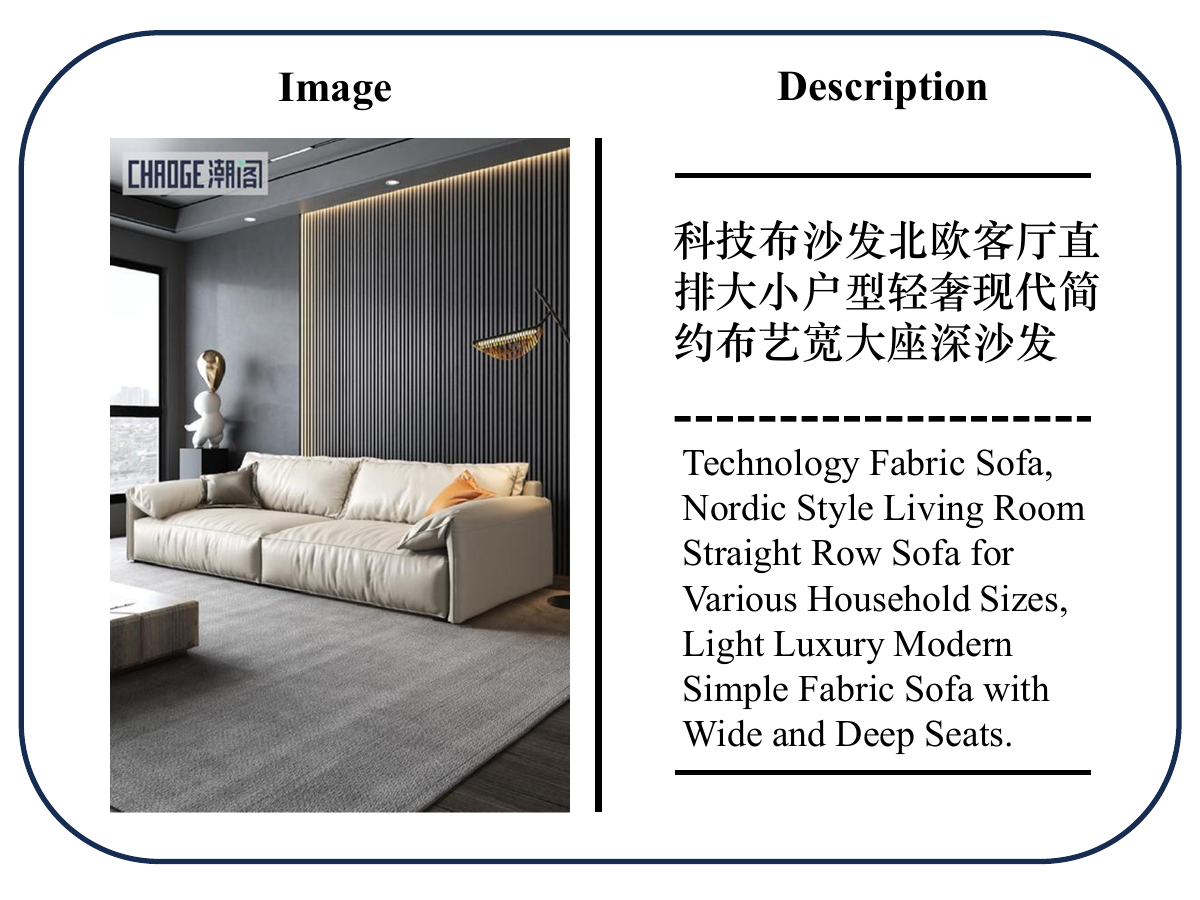}
        \end{subfigure}
    \end{minipage}
    \hfill
    \begin{subfigure}[c]{0.15\textwidth}
        \centering
        \includegraphics[width=\textwidth]{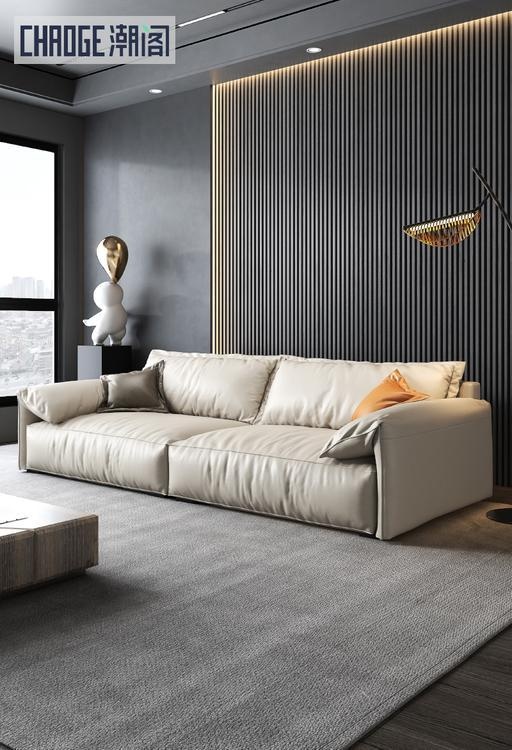}
    \end{subfigure}
    \hfill
    \begin{subfigure}[c]{0.15\textwidth}
        \centering
        \includegraphics[width=\textwidth]{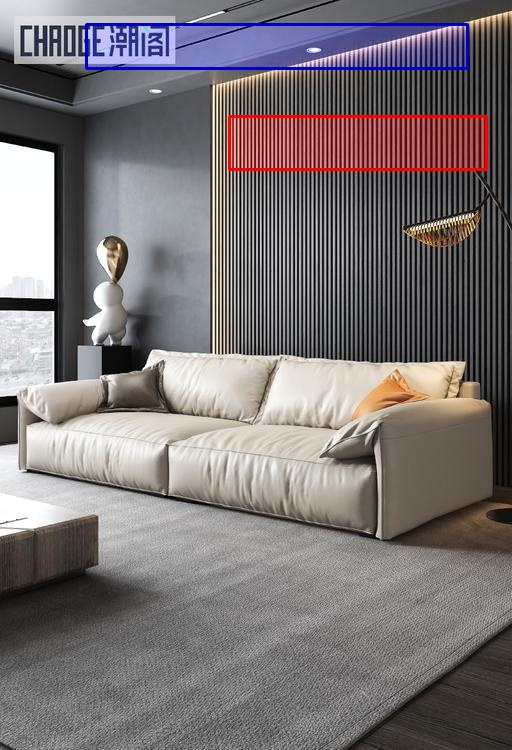}
    \end{subfigure}
    \hfill
    \begin{subfigure}[c]{0.15\textwidth}
        \centering
        \includegraphics[width=\textwidth]{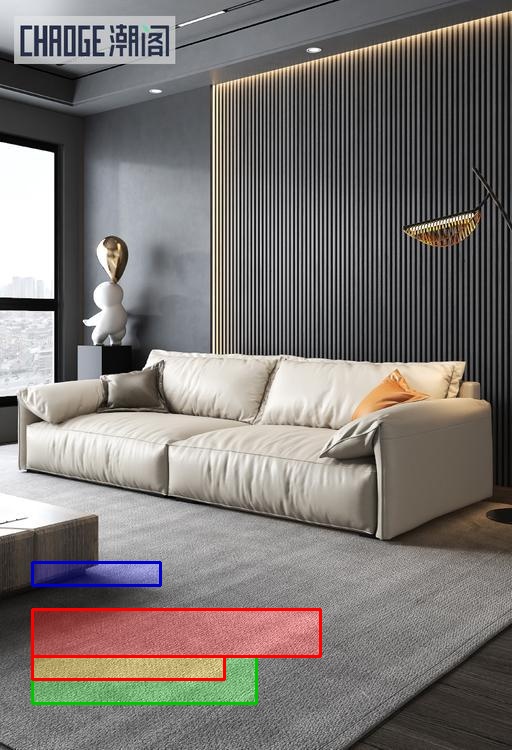}
    \end{subfigure}
    \hfill
    \begin{subfigure}[c]{0.15\textwidth}
        \centering
        \includegraphics[width=\textwidth]{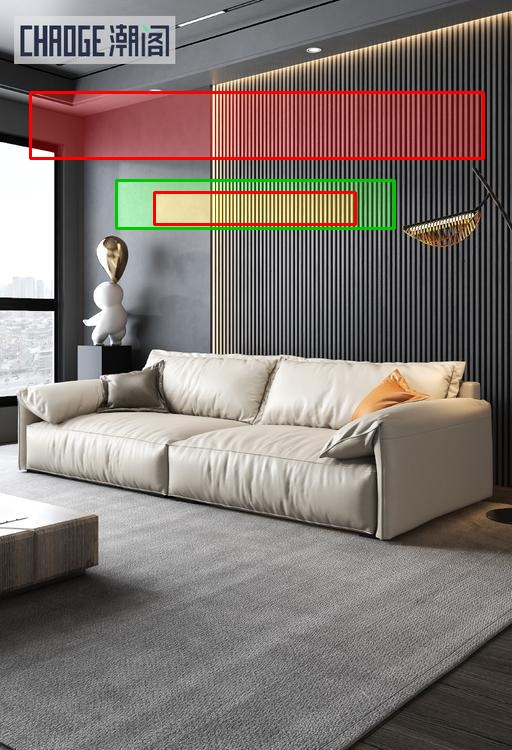}
    \end{subfigure}
    \hfill
    \begin{subfigure}[c]{0.15\textwidth}
        \centering
        \includegraphics[width=\textwidth]{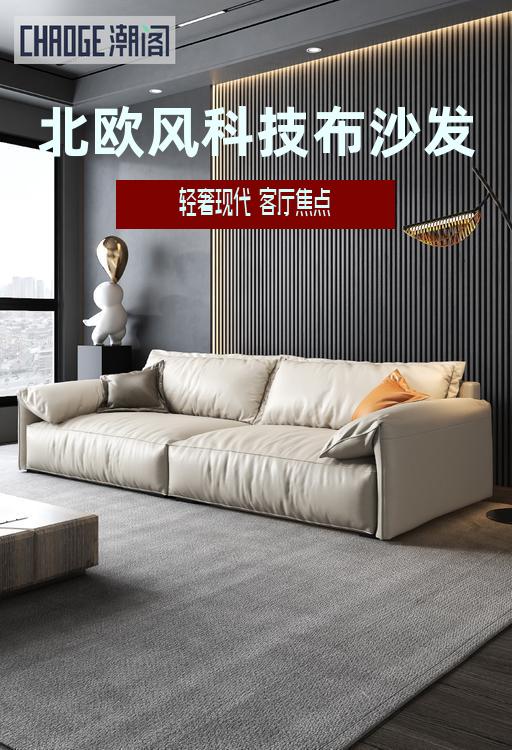}
    \end{subfigure}
    
    \begin{minipage}[c]{\textwidth}
    \centering
       Synthesized Taglines: [Nordic Style Technology Fabric Sofa; Light Luxury Modern Living Room Centerpiece.]
    \end{minipage}
    
    \begin{minipage}[c]{0.2\textwidth}
        \begin{minipage}[c]{0.05\textwidth}
            \rotatebox{90}{\small Instruction}
        \end{minipage}
        \hfill
        \begin{subfigure}[c]{0.92\textwidth}
            \centering
            \includegraphics[width=\textwidth]{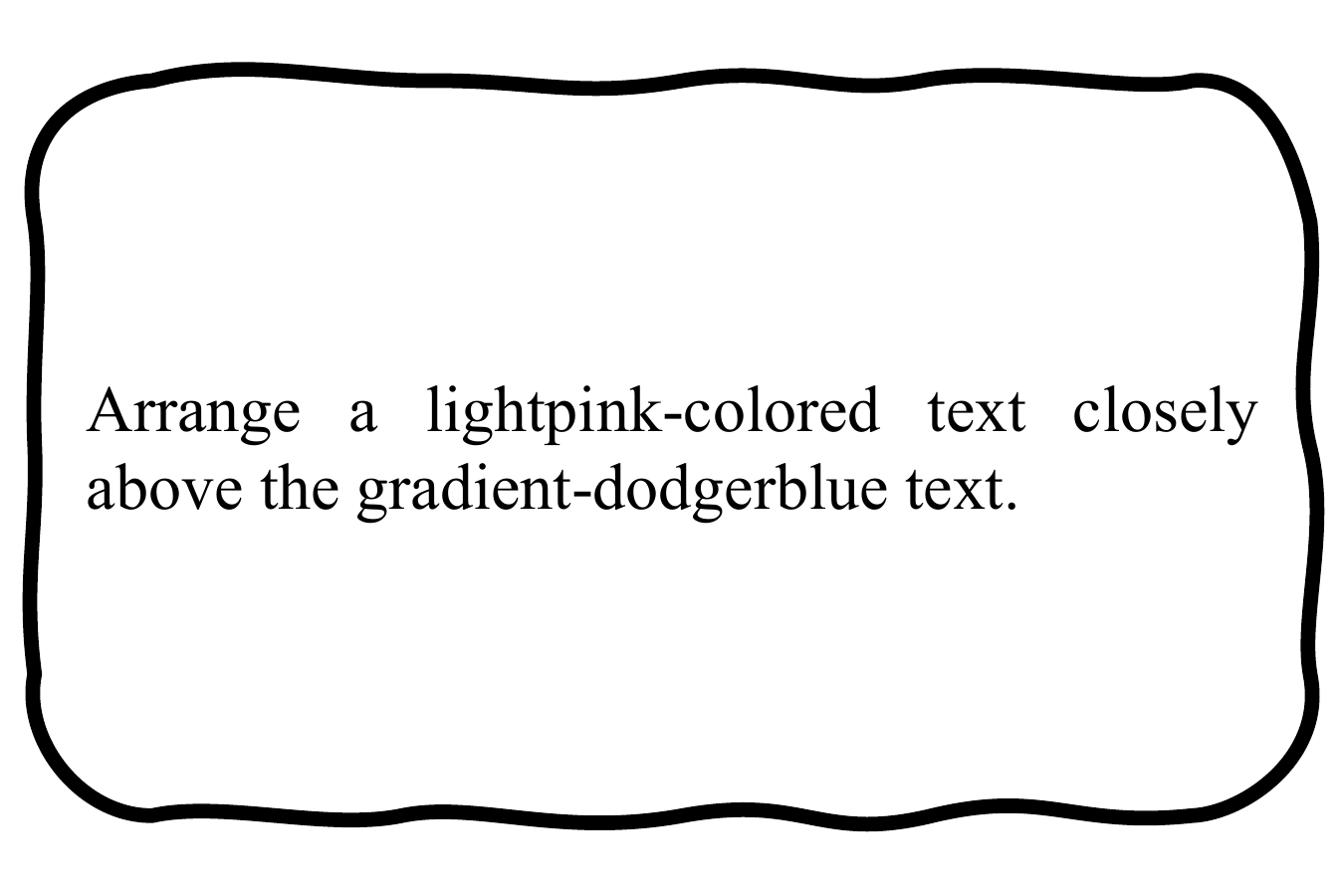}
        \end{subfigure}
        \begin{minipage}[c]{0.05\textwidth}
            \rotatebox{90}{\small Information}
        \end{minipage}
        \hfill
        \begin{subfigure}[c]{0.92\textwidth}
            \centering
            \includegraphics[width=\textwidth]{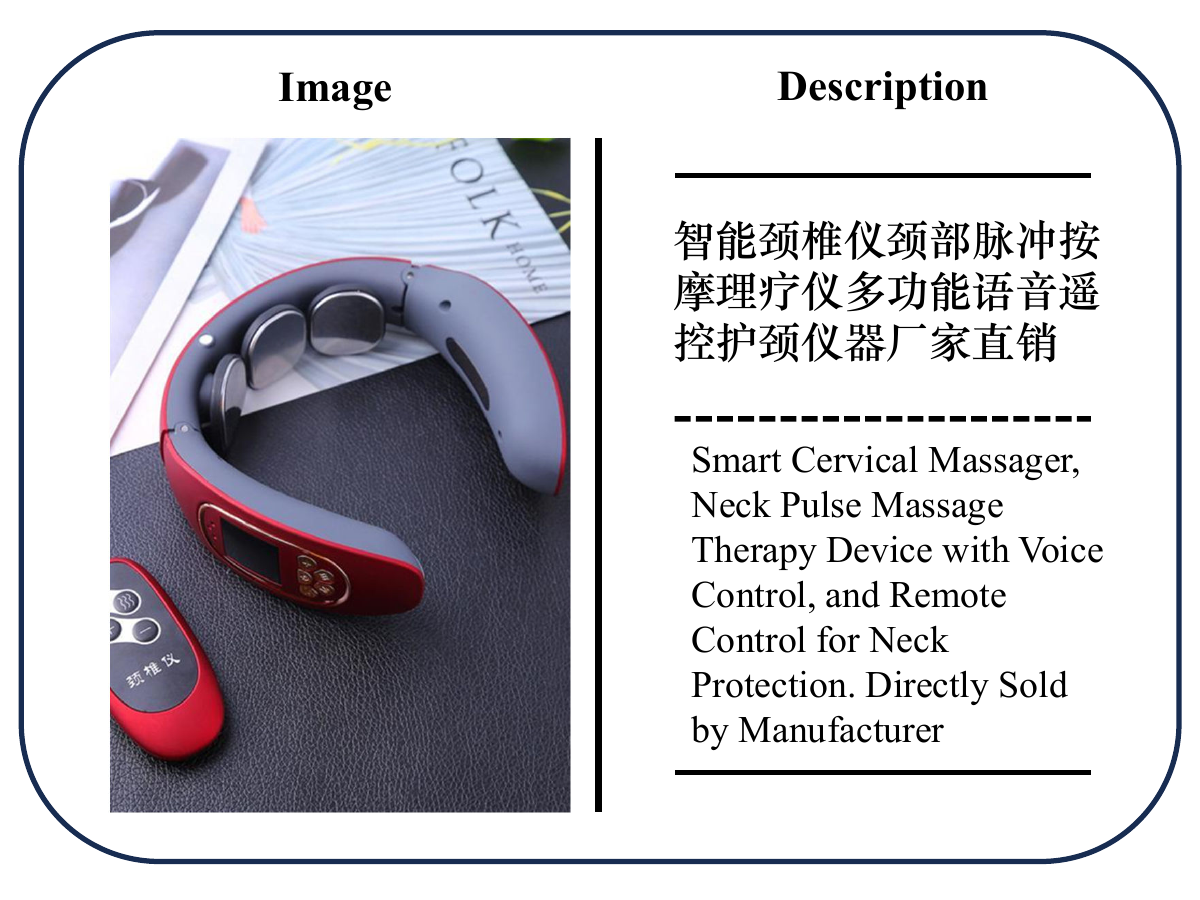}
        \end{subfigure}
    \end{minipage}
    \hfill
    \begin{subfigure}[c]{0.15\textwidth}
        \centering
        \includegraphics[width=\textwidth]{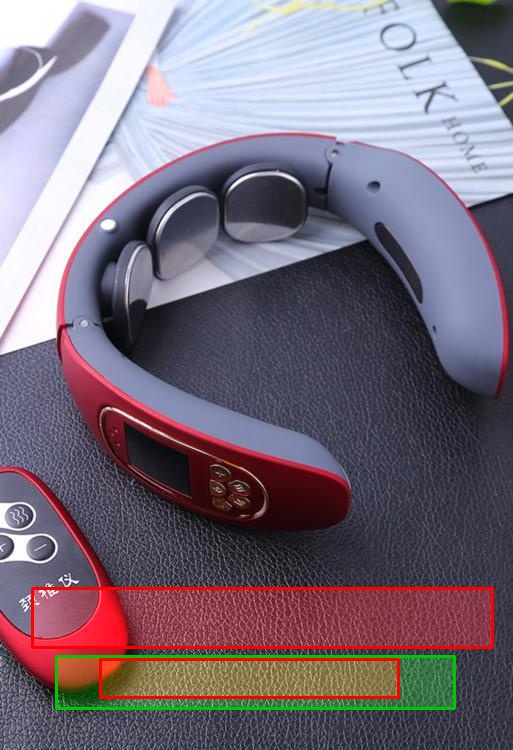}
    \end{subfigure}
    \hfill
    \begin{subfigure}[c]{0.15\textwidth}
        \centering
        \includegraphics[width=\textwidth]{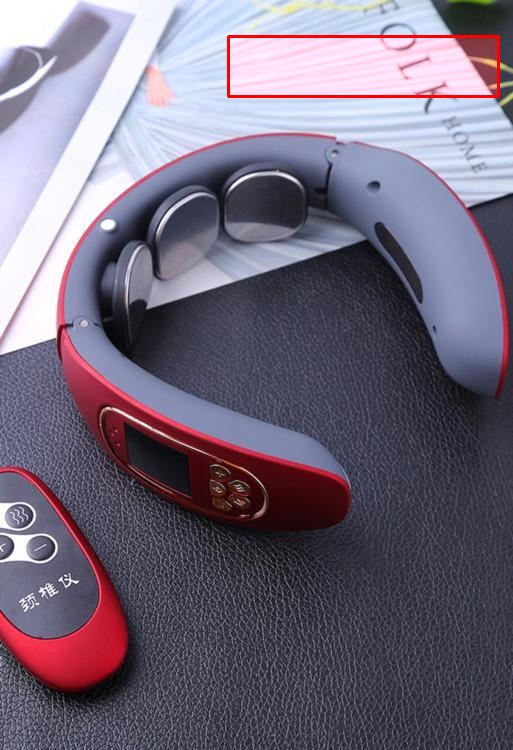}
    \end{subfigure}
    \hfill
    \begin{subfigure}[c]{0.15\textwidth}
        \centering
        \includegraphics[width=\textwidth]{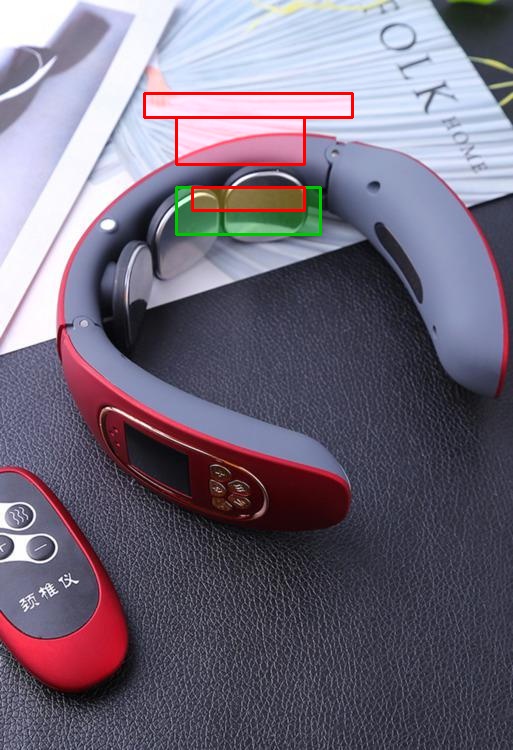}
    \end{subfigure}
    \hfill
    \begin{subfigure}[c]{0.15\textwidth}
        \centering
        \includegraphics[width=\textwidth]{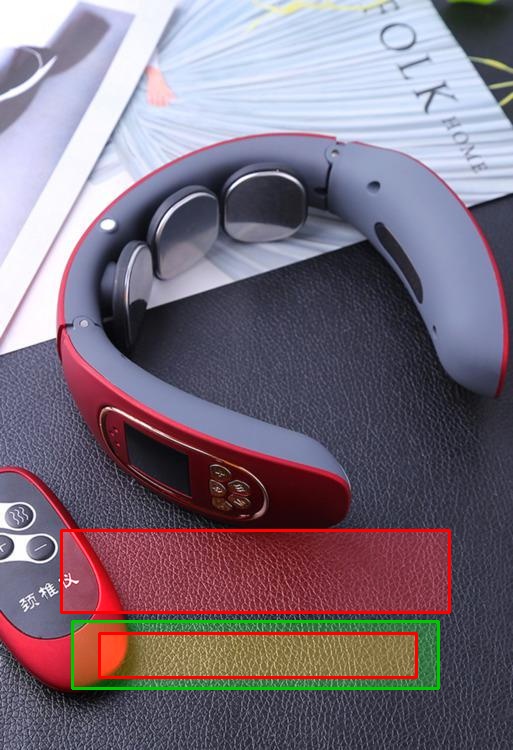}
    \end{subfigure}
    \hfill
    \begin{subfigure}[c]{0.15\textwidth}
        \centering
        \includegraphics[width=\textwidth]{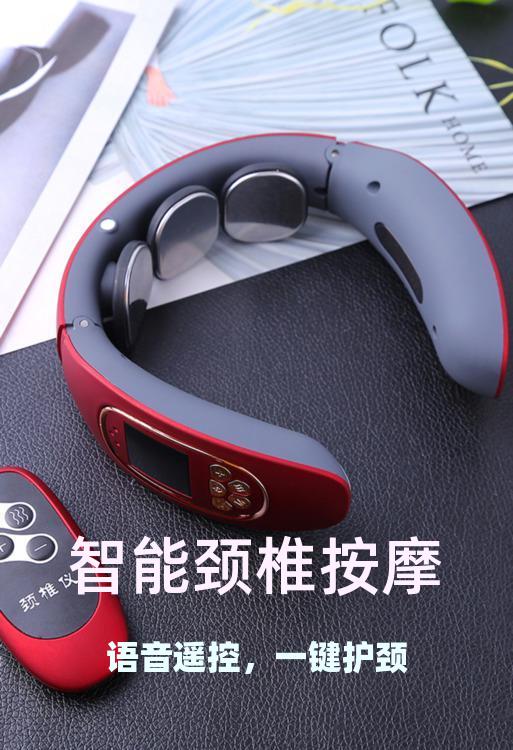}
    \end{subfigure}
    
    \begin{minipage}[c]{\textwidth}
    \centering
       Synthesized Taglines: [Smart Neck Massage; Voice-Controlled One-Button Neck Protection.]
    \end{minipage}
    
    \begin{minipage}[c]{0.2\textwidth}
        \begin{minipage}[c]{0.05\textwidth}
            \rotatebox{90}{\small Instruction}
        \end{minipage}
        \hfill
        \begin{subfigure}[c]{0.92\textwidth}
            \centering
            \includegraphics[width=\textwidth]{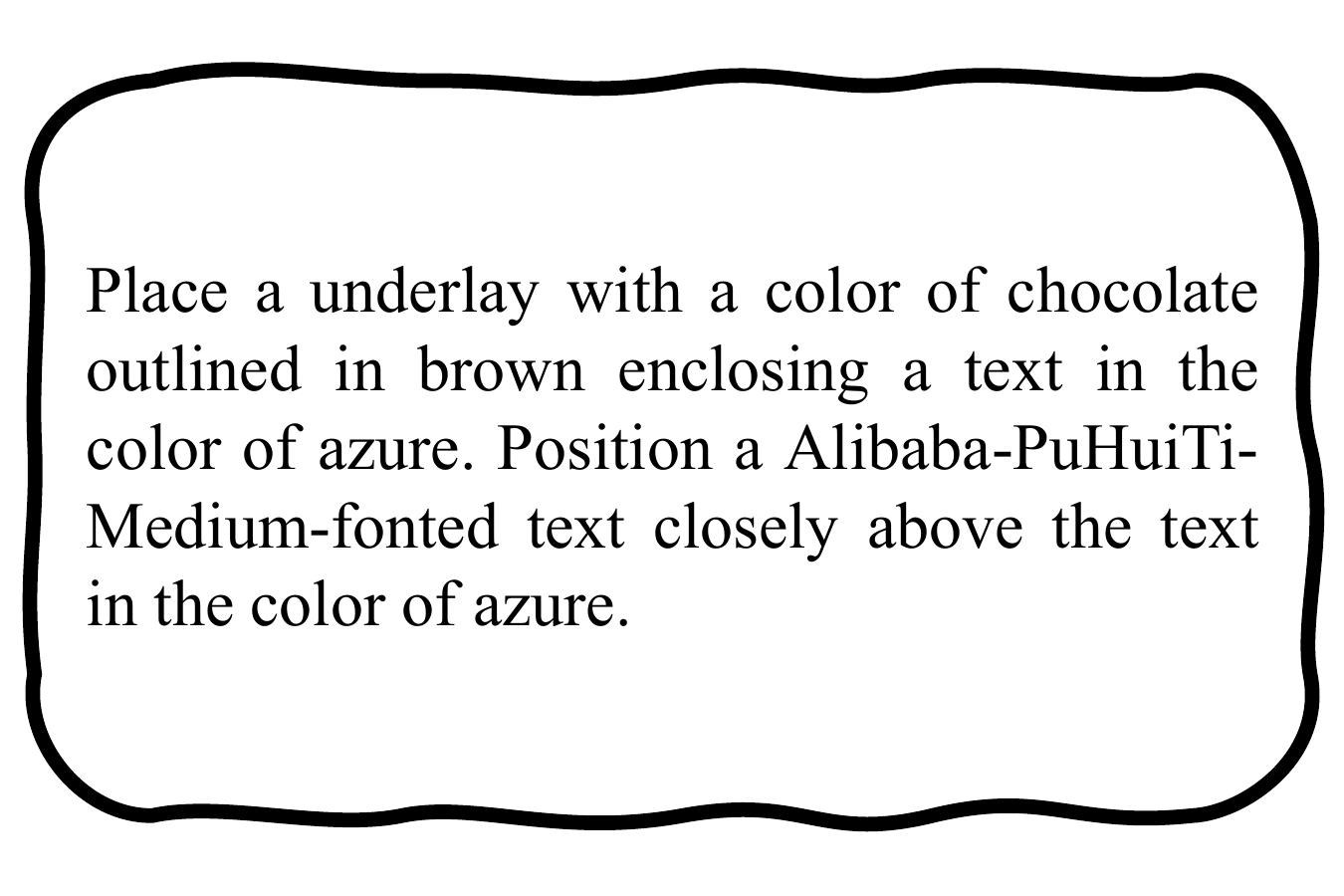}
        \end{subfigure}
        \begin{minipage}[c]{0.05\textwidth}
            \rotatebox{90}{\small Information}
        \end{minipage}
        \hfill
        \begin{subfigure}[c]{0.92\textwidth}
            \centering
            \includegraphics[width=\textwidth]{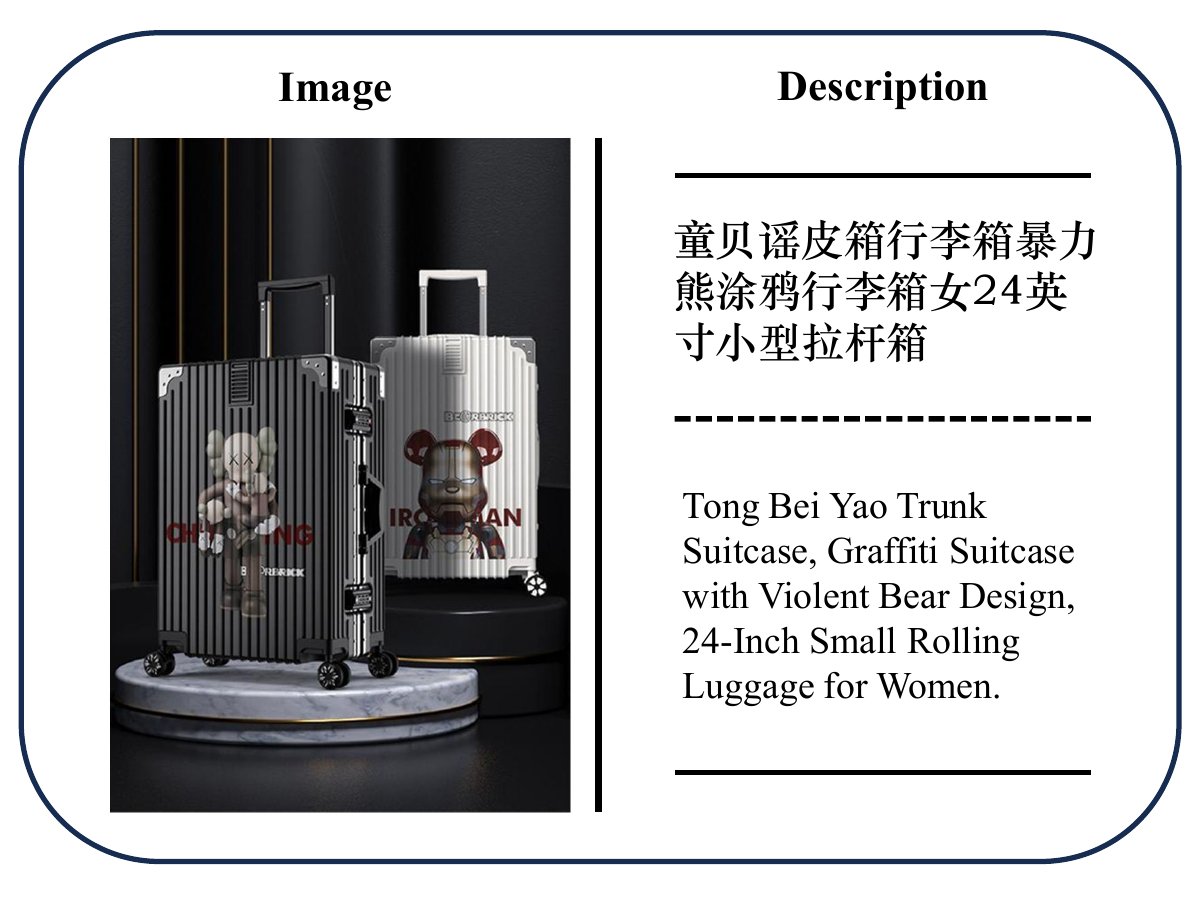}
        \end{subfigure}
    \end{minipage}
    \hfill
    \begin{subfigure}[c]{0.15\textwidth}
        \centering
        \includegraphics[width=\textwidth]{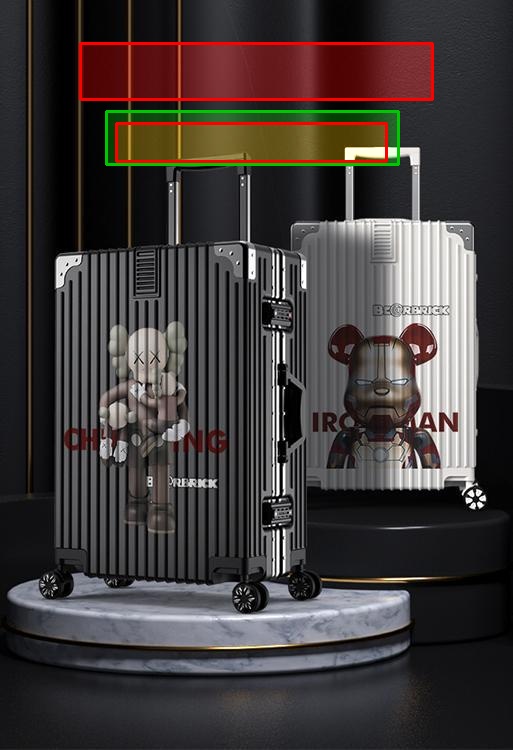}
    \end{subfigure}
    \hfill
    \begin{subfigure}[c]{0.15\textwidth}
        \centering
        \includegraphics[width=\textwidth]{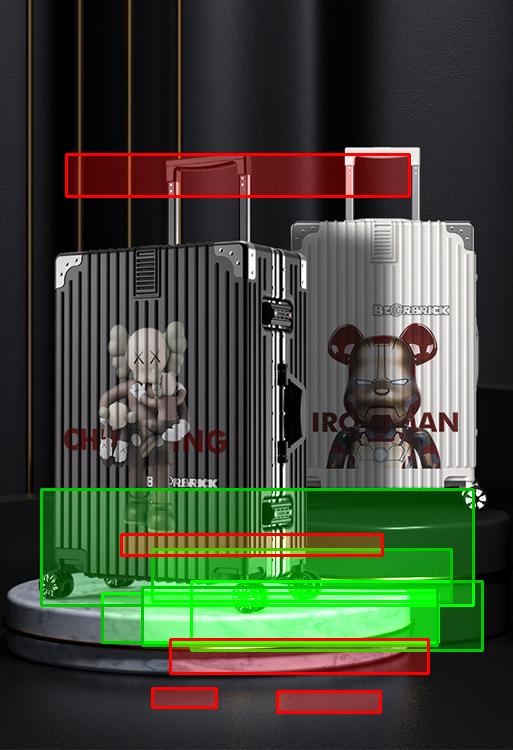}
    \end{subfigure}
    \hfill
    \begin{subfigure}[c]{0.15\textwidth}
        \centering
        \includegraphics[width=\textwidth]{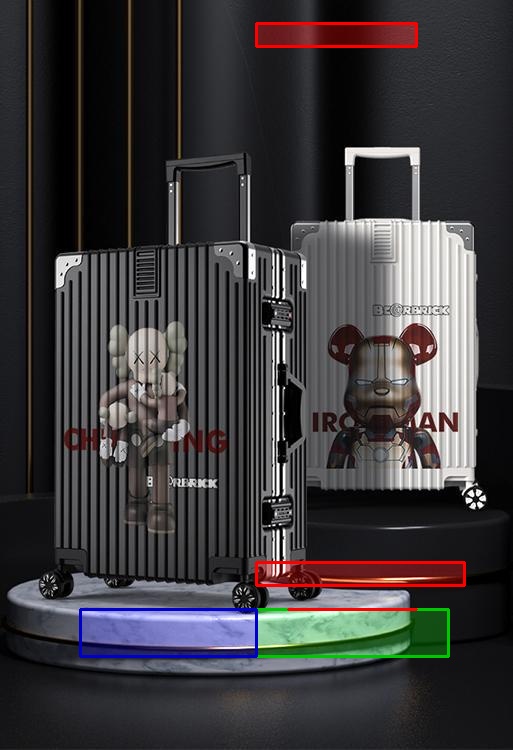}
    \end{subfigure}
    \hfill
    \begin{subfigure}[c]{0.15\textwidth}
        \centering
        \includegraphics[width=\textwidth]{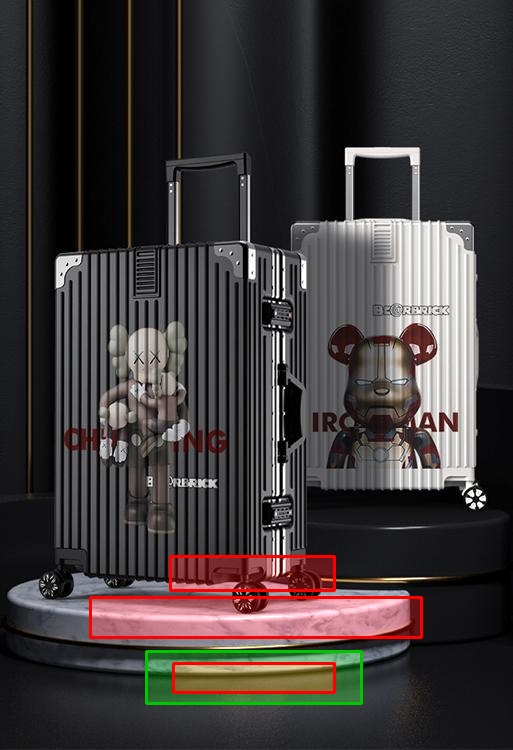}
    \end{subfigure}
    \hfill
    \begin{subfigure}[c]{0.15\textwidth}
        \centering
        \includegraphics[width=\textwidth]{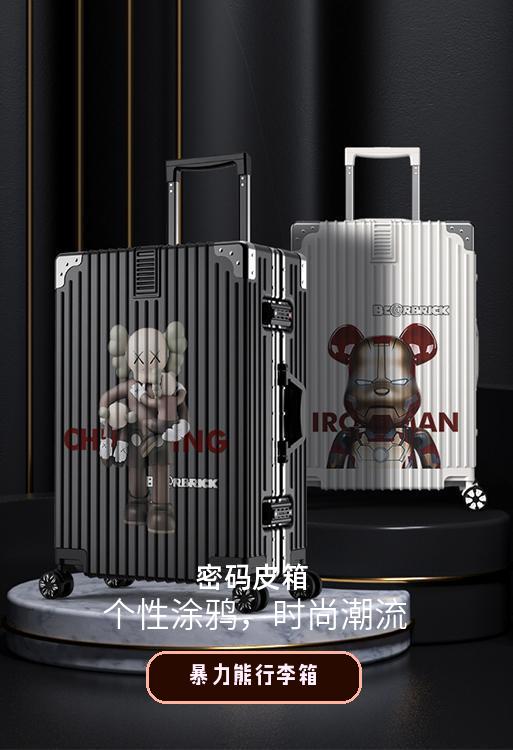}
    \end{subfigure}
    
    \begin{minipage}[c]{\textwidth}
    \centering
       Synthesized Taglines: [Password Suitcase; Personalized Graffiti Fashion Trend; Violent Bear Luggage.]
    \end{minipage}
    
    \begin{minipage}[c]{0.2\textwidth}
        \begin{minipage}[c]{0.05\textwidth}
            \rotatebox{90}{\small Instruction}
        \end{minipage}
        \hfill
        \begin{subfigure}[c]{0.92\textwidth}
            \centering
            \includegraphics[width=\textwidth]{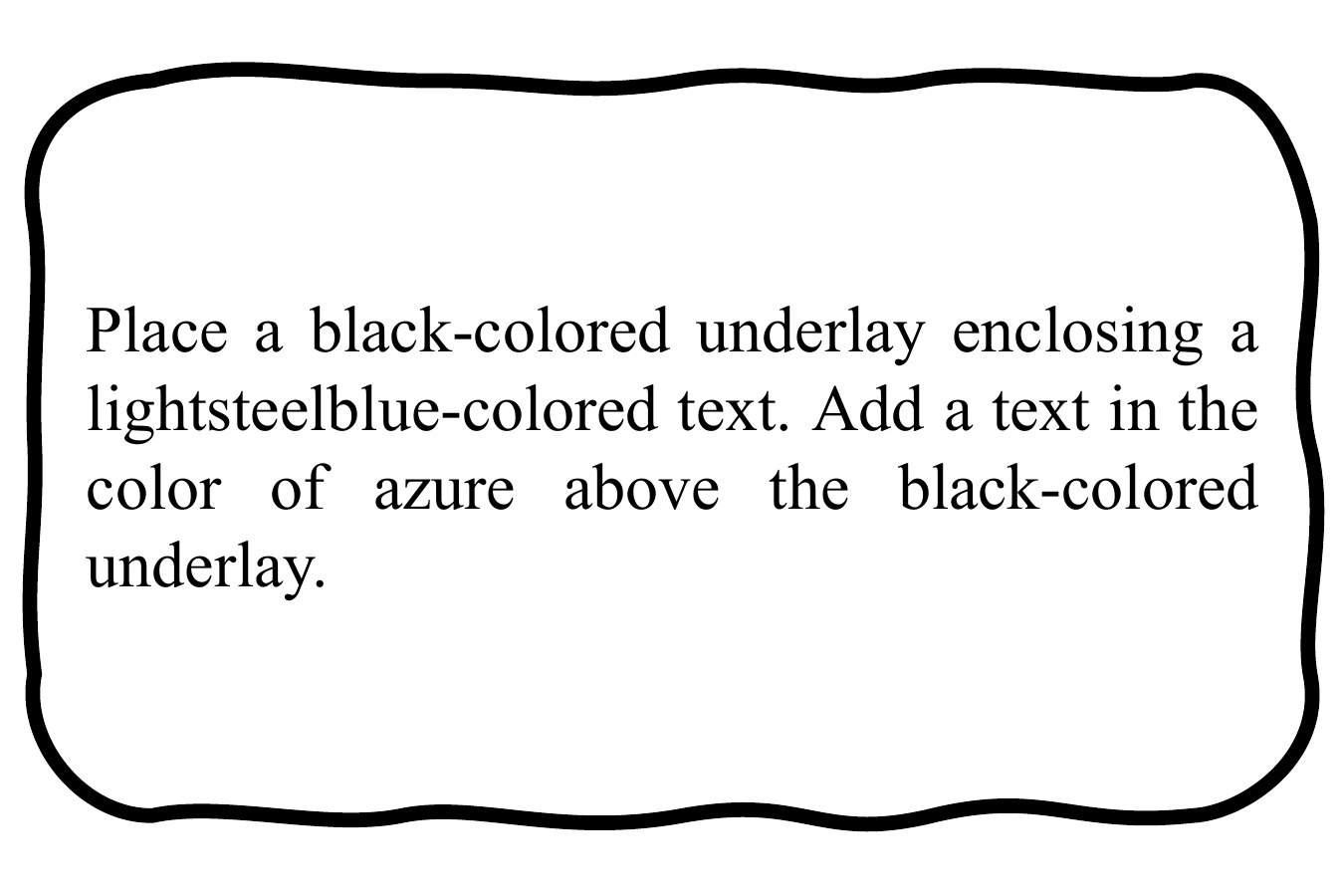}
        \end{subfigure}
        \begin{minipage}[c]{0.05\textwidth}
            \rotatebox{90}{\small Information}
        \end{minipage}
        \hfill
        \begin{subfigure}[c]{0.92\textwidth}
            \centering
            \includegraphics[width=\textwidth]{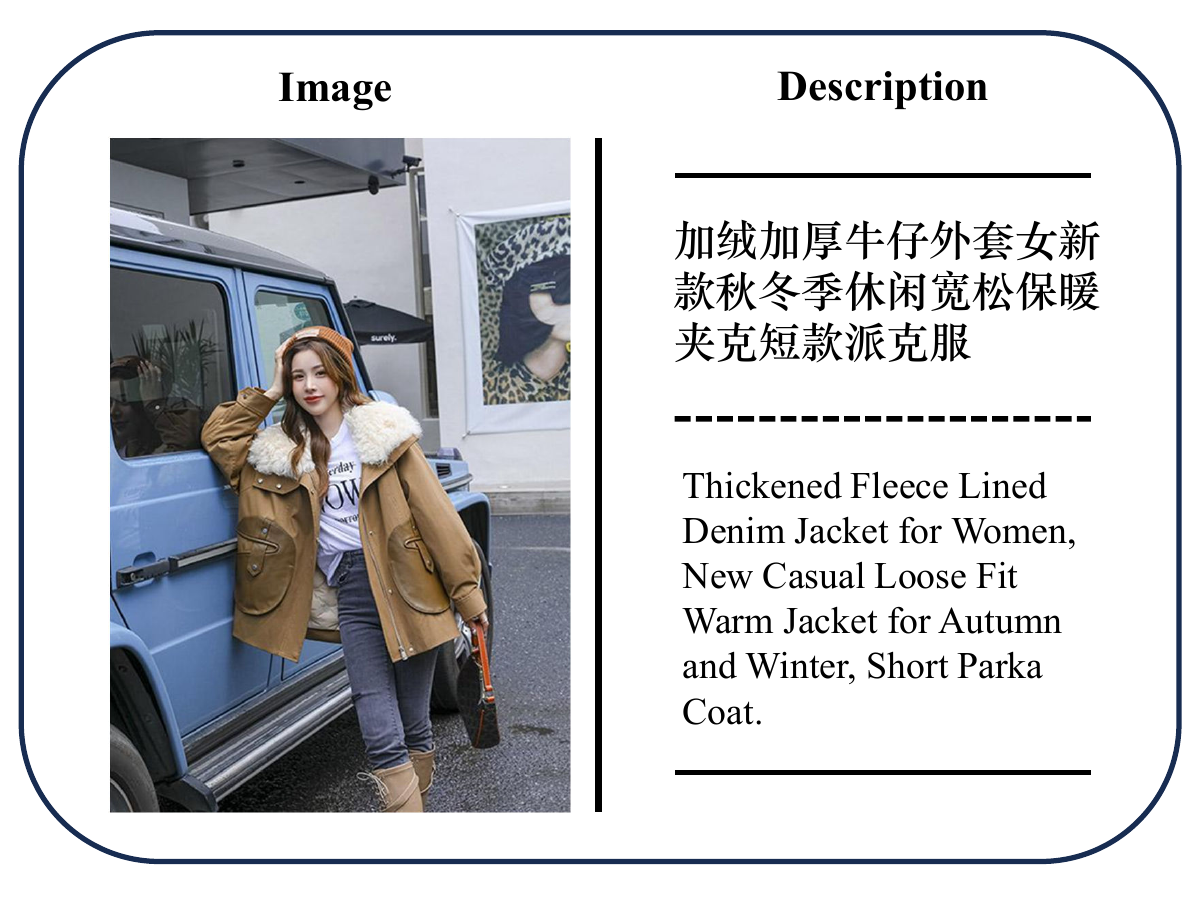}
        \end{subfigure}
    \end{minipage}
    \hfill
    \begin{subfigure}[c]{0.15\textwidth}
        \centering
        \includegraphics[width=\textwidth]{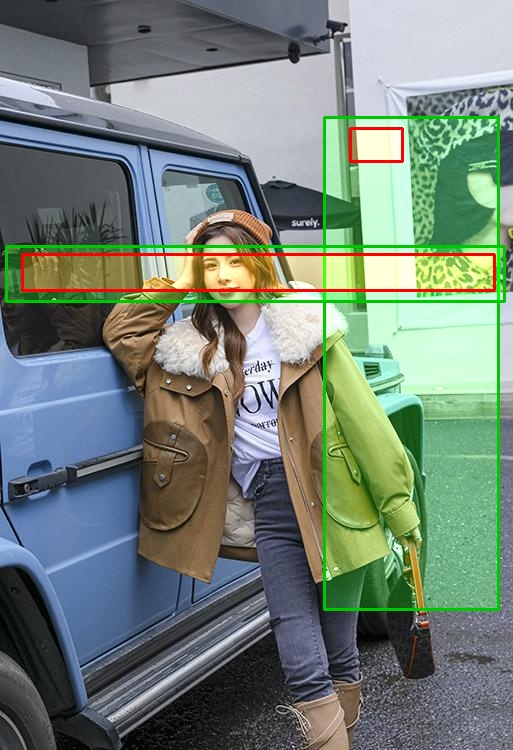}
    \end{subfigure}
    \hfill
    \begin{subfigure}[c]{0.15\textwidth}
        \centering
        \includegraphics[width=\textwidth]{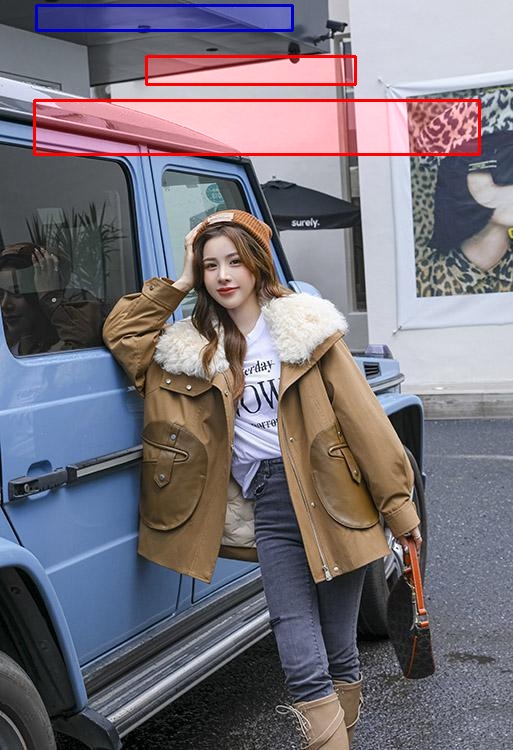}
    \end{subfigure}
    \hfill
    \begin{subfigure}[c]{0.15\textwidth}
        \centering
        \includegraphics[width=\textwidth]{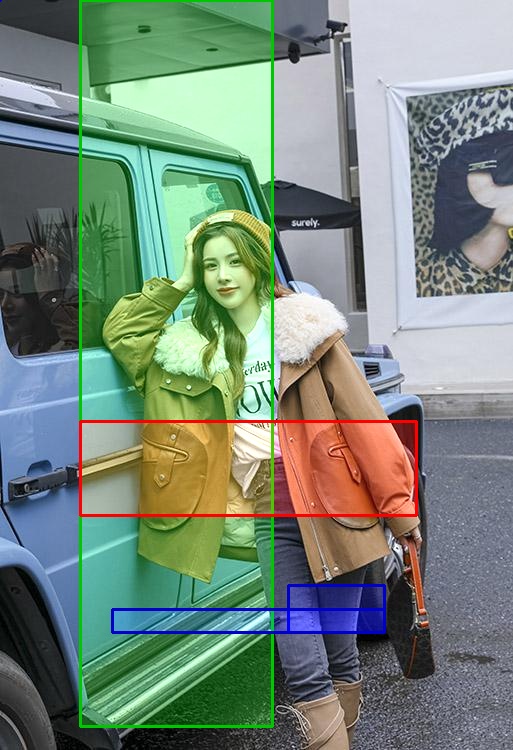}
    \end{subfigure}
    \hfill
    \begin{subfigure}[c]{0.15\textwidth}
        \centering
        \includegraphics[width=\textwidth]{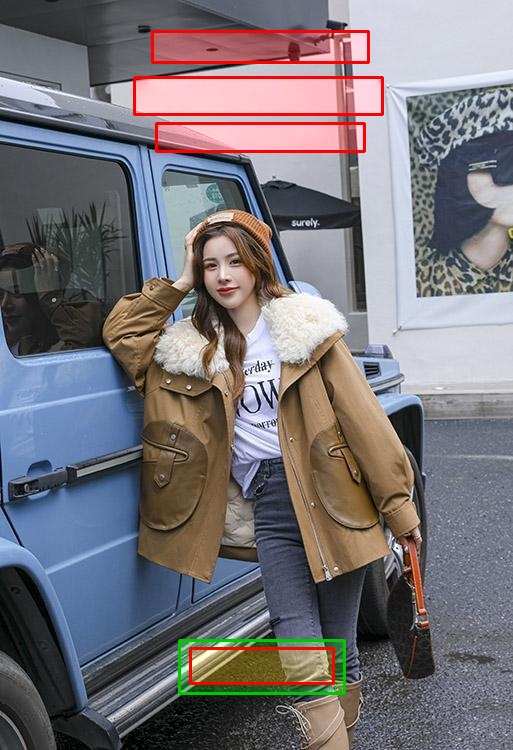}
    \end{subfigure}
    \hfill
    \begin{subfigure}[c]{0.15\textwidth}
        \centering
        \includegraphics[width=\textwidth]{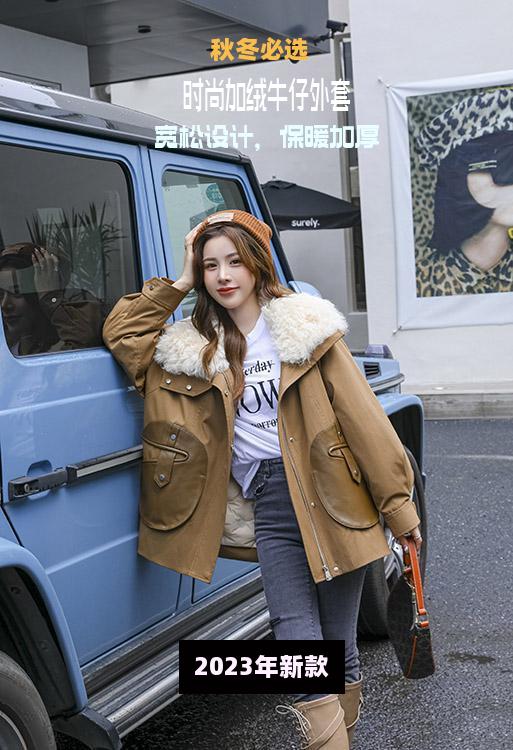}
    \end{subfigure}
    
    \begin{minipage}[c]{\textwidth}
    \centering
       Synthesized Taglines: [Autumn and winter essential; Fashionable fleece-lined denim jacket;\\ Loose design for warmth and thickness; 2023 New Style.]
    \end{minipage}
    
    \caption{Visualizations for instruction-drive synthesized 2D posters by LayoutTrans~\citep{Gupta_2021_ICCV}, LayoutVAE~\citep{Jyothi_2019_ICCV}, LayoutDM~\citep{Inoue_2023_CVPR} and our method. Note that posters are rendered according to synthesized graphic and textual features.}
    \label{fig:2d_vis_3}
\end{figure*}

Comparisons of our synthesized 2D posters with other state-of-the-art methods are shown in Figure~\ref{fig:2d_vis_1}, \ref{fig:2d_vis_2}, and \ref{fig:2d_vis_3}, demonstrating the superiority of our method compared to previous state-of-the-art approaches in terms of better mimicking the behavior of human designers and following provided instructions.

\section{Discussions}\label{apx:disc}
\textbf{Out-of-distribution Inputs.}
For out-of-distribution (OOD) inputs, the proposed system can still produce reasonable layouts, though they may not align closely with the input semantics.
For instance, given instructions like ``a black cat'' or ``a football playground'', the generated 3D indoor scenes tend to adopt black tones or football-themed decorations, while still maintaining a plausible bedroom layout, rather than generating a cat or an outdoor playground.
Similarly, in 2D poster generation, when provided with natural images, the system interprets them as poster backgrounds and places plausible text bounding boxes accordingly.
This generalization ability stems from the pretrained CLIP text and image encoders, which are adapted to extract semantically meaningful features from diverse inputs.

\begin{figure*}[htbp]
    \centering
    \begin{subfigure}[b]{\textwidth}
        \centering
        \begin{subfigure}[b]{0.3\textwidth}
            \includegraphics[width=\textwidth]{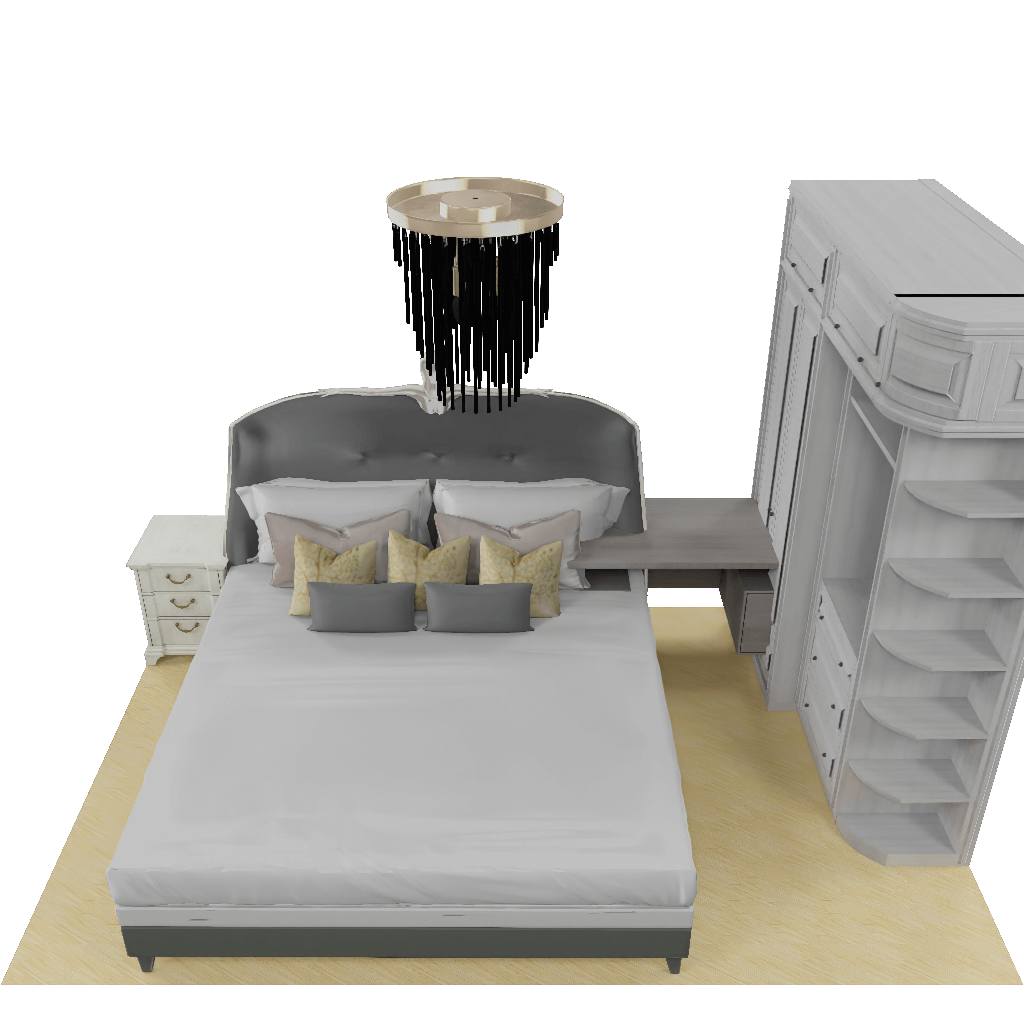}
        \end{subfigure}
        \hfill
        \begin{subfigure}[b]{0.3\textwidth}
            \includegraphics[width=\textwidth]{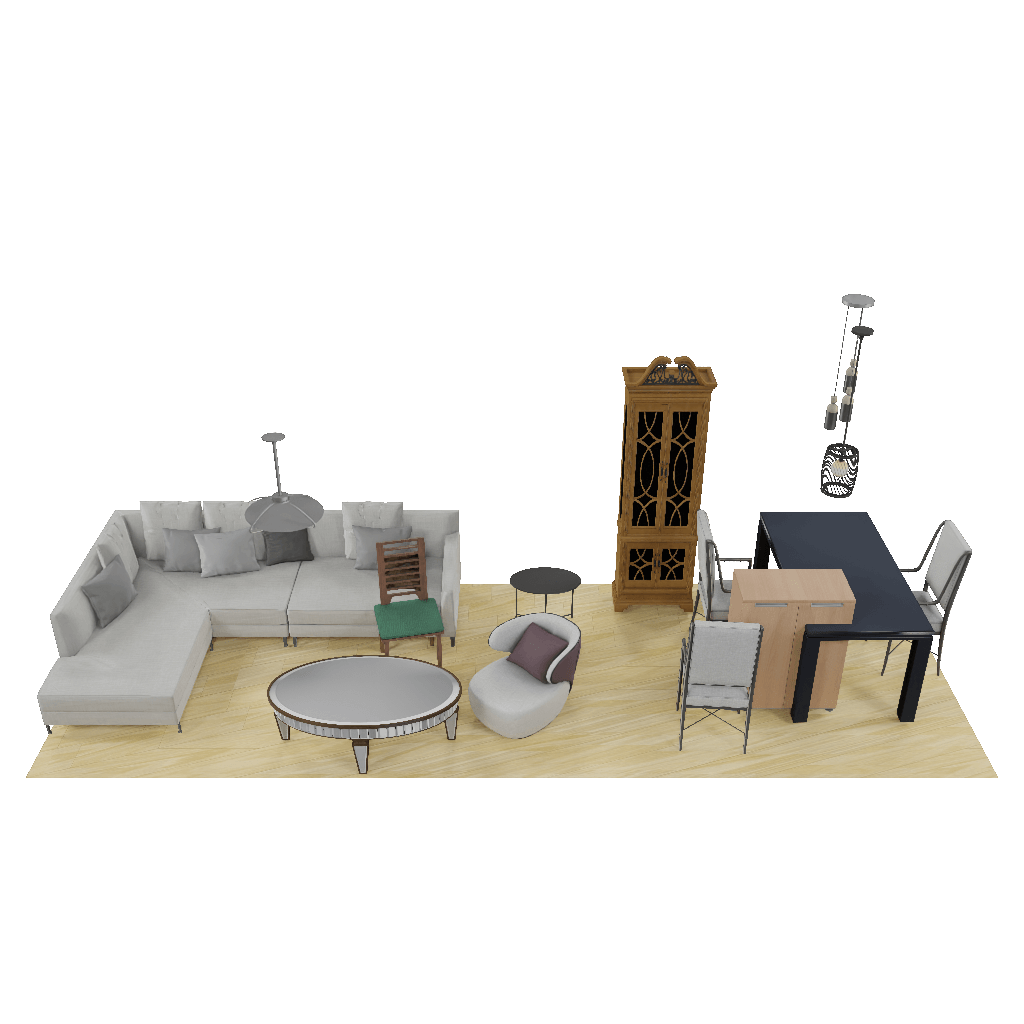}
        \end{subfigure}
        \hfill
        \begin{subfigure}[b]{0.3\textwidth}
            \includegraphics[width=\textwidth]{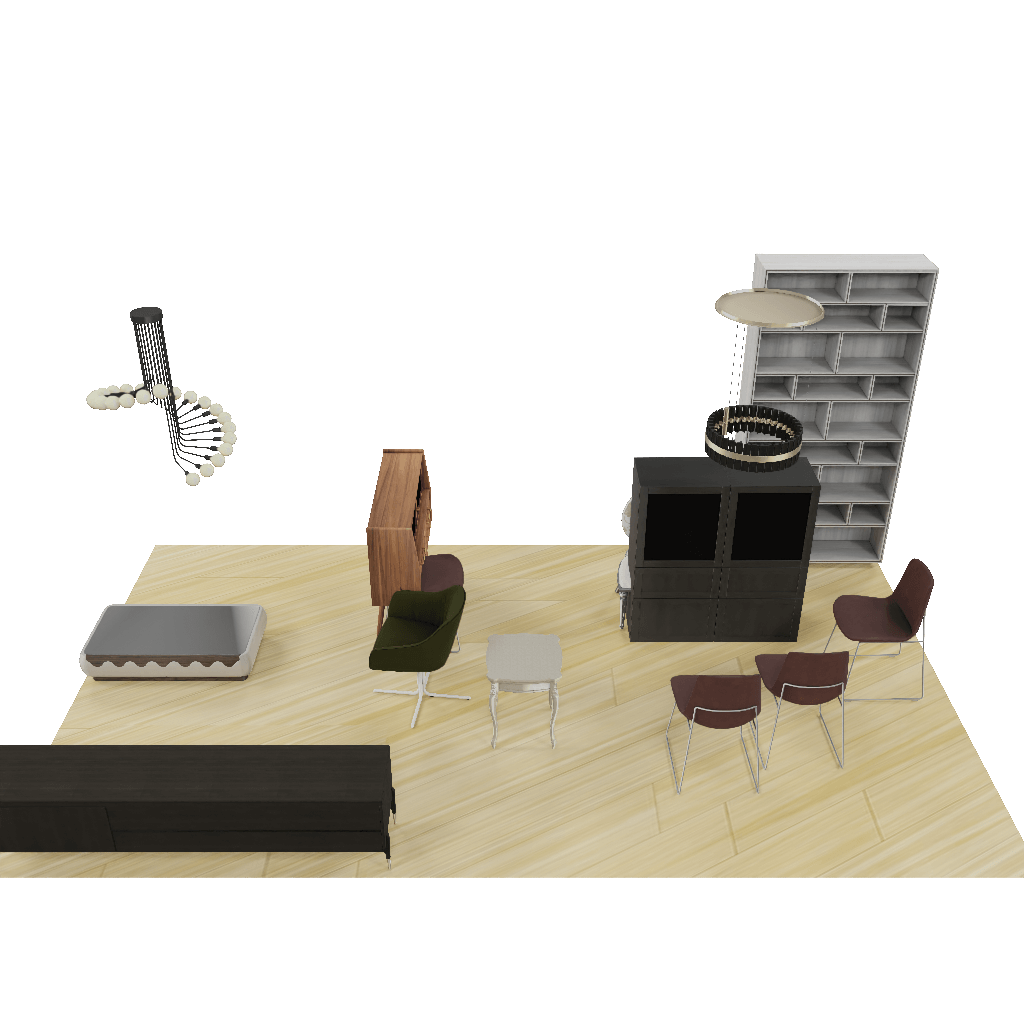}
        \end{subfigure}
        \caption*{(a) Failure cases of 3D scene synthesis: object collisions resulting from limited modeling of inter-object relationships and noisy dataset annotations.}
    \end{subfigure}
    
    \vspace{0.5em} 

    \begin{subfigure}[b]{\textwidth}
        \centering
        \begin{subfigure}[b]{0.3\textwidth}
            \includegraphics[width=\textwidth]{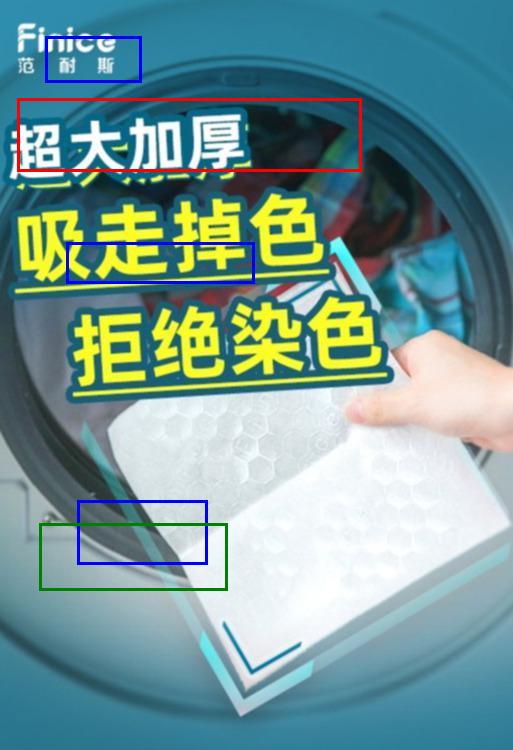}
        \end{subfigure}
        \hfill
        \begin{subfigure}[b]{0.3\textwidth}
            \includegraphics[width=\textwidth]{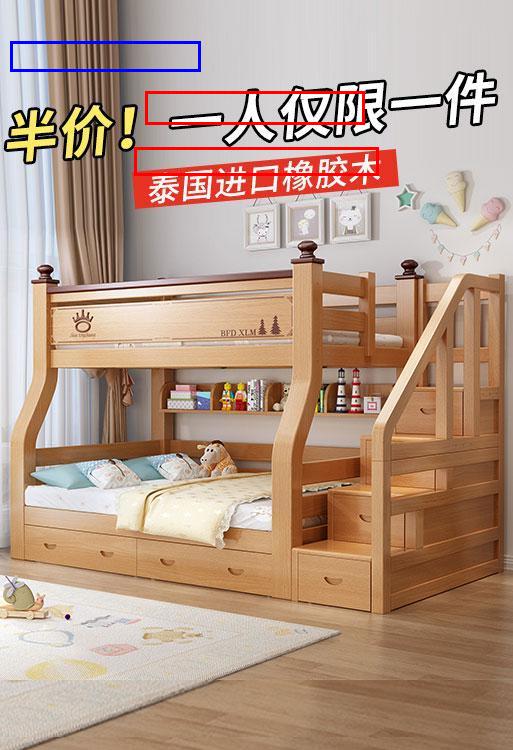}
        \end{subfigure}
        \hfill
        \begin{subfigure}[b]{0.3\textwidth}
            \includegraphics[width=\textwidth]{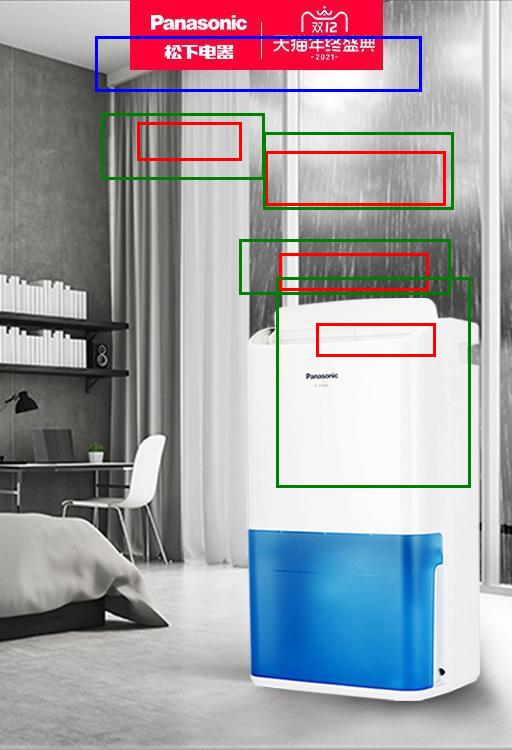}
        \end{subfigure}
        \caption*{(b) Failure cases of 2D poster synthesis: redundant bounding boxes and overlaps with existing background text due to insufficient content awareness.}
    \end{subfigure}
    
    \caption{Failure cases of \textsc{InstructLayout}.}
    \label{fig:failures}
\end{figure*}

\textbf{Failure Cases.}
Although the proposed generative framework demonstrates strong performance and robustness to OOD inputs across both 3D indoor scenes and 2D posters, it still exhibits several failure cases.
For example, the predicted 3D bounding boxes do not fully account for the complex relationships between meshes, leading to occasional object collisions in generated 3D scenes, which is also partly due to noisy annotations in the 3D-FRONT dataset.
In the 2D poster generation task, the model sometimes produces redundant bounding boxes, especially when the background images already contain text.
This can result in visually cluttered designs, as the generated boxes may unintentionally overlap with existing visual elements.
Visualization results for these failure cases are provided in Figure~\ref{fig:failures}.

\textbf{Limitation and Future Work.}
Another limitation of our work lies in its insufficient generalization across categories for both 3D indoor scene and 2D poster generation.
The current approach relies on training a small, specialized generative model for each data type, whereas recent trends favor unified models trained on large-scale, diverse datasets.
In light of the rapid development of large vision language models (VLMs), integrating a VLM into our instruction-driven pipeline holds strong potential for improving both generalization and controllability in future work~\citep{sun2025layoutvlm,pang2025paper2poster}. 
Meanwhile, designing a unified framework that can generate either a 3D scene~\citep{lin2025partcrafter} or a 2D poster~\citep{chen2025postercraft} in a single framework is also a promising direction, as it could simplify the pipeline design and accelerate inference.

\begin{table*}[ht]
    \centering
    \caption{Comparison of the types of elements contained in various poster datasets.}
    \label{tab:2d_dataset}
    \renewcommand\arraystretch{1.3}
    \resizebox{\textwidth}{!}{
    \begin{tabular}{l|c|cc|cccc|c}
        \toprule[1.2pt]
        \multirow{2}{*}{\diagbox{Dataset}{Data Type}} & \multirow{2}{*}{Instruction} & \multicolumn{2}{c|}{Product Information} & \multicolumn{4}{c|}{Graphic Element Annotation} & \multirow{2}{*}{Poster} \\ \cline{3-8}
        & & Image & Description & Category & Bounding Box & Attribute & Tagline & \\ \midrule[1.2pt]
        CGL-GAN~\citep{zhou2022composition} & & & & \checkmark & \checkmark & & & \checkmark \\
        PosterLayout~\citep{hsu2023posterlayout} &  & \checkmark & & \checkmark & \checkmark & & & \checkmark \\
        CapOnImage2M \citep{gao2022caponimage} & & & \checkmark & \checkmark & \checkmark & & \checkmark & \checkmark \\
        AutoPoster\citep{lin2023autoposter} & & & & \checkmark & \checkmark & \checkmark & \checkmark & \checkmark \\
        \hline
        \cellcolor{mygray}Ours &\cellcolor{mygray} \checkmark & \cellcolor{mygray}\checkmark & \cellcolor{mygray}\checkmark & \cellcolor{mygray}\checkmark &\cellcolor{mygray} \checkmark & \cellcolor{mygray}\checkmark & \cellcolor{mygray}\checkmark & \cellcolor{mygray}\checkmark \\ \bottomrule[1.2pt]
    \end{tabular}
    }
\end{table*}

\begin{table*}[th]
    \begin{minipage}[t]{0.49\textwidth}
        \centering
        \caption{Rules to determine the spatial relationships between objects in a 3D scene.}
        \label{tab:rules_3d}
        \renewcommand\arraystretch{1.22}
        \resizebox{\textwidth}{!}{
        \begin{tabular}{c|c}
            \toprule[1.2pt]
    
            \textbf{Relationship} &
            \textbf{Rule} \\
    
            \midrule[1.2pt]
    
            Left of &
            ($\theta_{so}\geq\frac{3\pi}{4}$ or $\theta_{so}<-\frac{3\pi}{4}$)
            and $1<d(s,o)\leq3$ \\
            Right of &
            $-\frac{\pi}{4}\leq\theta_{so}<\frac{\pi}{4}$
            and $1<d(s,o)\leq3$ \\
            In front of &
            $\frac{\pi}{4}\leq\theta_{so}<\frac{3\pi}{4}$
            and $1<d(s,o)\leq3$ \\
            Behind &
            $-\frac{3\pi}{4}\leq\theta_{so}<-\frac{\pi}{4}$
            and $1<d(s,o)\leq3$ \\
            Closely left of &
            ($\theta_{so}\geq\frac{3\pi}{4}$ or $\theta_{so}<-\frac{3\pi}{4}$)
            and $d(s,o)\leq1$ \\
            Closely right of &
            $-\frac{\pi}{4}\leq\theta_{so}<\frac{\pi}{4}$
            and $d(s,o)\leq1$ \\
            Closely in front of &
            $\frac{\pi}{4}\leq\theta_{so}<\frac{3\pi}{4}$
            and $d(s,o)\leq1$ \\
            Closely behind &
            $-\frac{3\pi}{4}\leq\theta_{so}<-\frac{\pi}{4}$
            and $d(s,o)\leq1$ \\
            Above &
            $(\text{\texttt{Center}}_{Z_s} - \text{\texttt{Center}}_{Z_o})>(\text{\texttt{Height}}_s+\text{\texttt{Height}}_o)/2$
            \\ &
            and ($\text{Inside}(s,o)$ or $\text{Inside}(o,s)$)\\
            Below &
            $(\text{\texttt{Center}}_{Z_o} - \text{\texttt{Center}}_{Z_s})>(\text{\texttt{Height}}_s+\text{\texttt{Height}}_o)/2$
            \\ &
            and ($\text{Inside}(s,o)$ or $\text{Inside}(o,s)$) \\
            None & $d(s,o)>3$ \\
    
            \bottomrule[1.2pt]
        \end{tabular}
        }
    \end{minipage}
    \hfill
    \begin{minipage}[t]{0.49\textwidth}
        \centering
        \caption{Rules to determine the spatial relationships between objects in a 2D poster.}
        \label{tab:rules_2d}
        \renewcommand\arraystretch{1.25}
        \resizebox{\textwidth}{!}{
        \begin{tabular}{c|c}
            \toprule[1.2pt]
    
            \textbf{Relationship} &
            \textbf{Rule} \\
    
            \midrule[1.2pt]
    
            Left of &
            ($\theta_{so}\geq\frac{3\pi}{4}$ or $\theta_{so}<-\frac{3\pi}{4}$)
            and $80<d(s,o)\leq300$ \\
            Right of &
            $-\frac{\pi}{4}\leq\theta_{so}<\frac{\pi}{4}$
            and $80<d(s,o)\leq300$ \\
            above &
            $\frac{\pi}{4}\leq\theta_{so}<\frac{3\pi}{4}$
            and $80<d(s,o)\leq300$ \\
            below &
            $-\frac{3\pi}{4}\leq\theta_{so}<-\frac{\pi}{4}$
            and $80<d(s,o)\leq300$ \\
            Closely left of &
            ($\theta_{so}\geq\frac{3\pi}{4}$ or $\theta_{so}<-\frac{3\pi}{4}$)
            and $d(s,o)\leq80$ \\
            Closely right of &
            $-\frac{\pi}{4}\leq\theta_{so}<\frac{\pi}{4}$
            and $d(s,o)\leq80$ \\
            Closely above &
            $\frac{\pi}{4}\leq\theta_{so}<\frac{3\pi}{4}$
            and $d(s,o)\leq80$ \\
            Closely below &
            $-\frac{3\pi}{4}\leq\theta_{so}<-\frac{\pi}{4}$
            and $d(s,o)\leq80$ \\
            Enclosing & Inside($o$,$s$) \\
            Enclosed with & Inside($s$,$o$) \\
            None & $d(s,o)>300$ \\
    
            \bottomrule[1.2pt]
        \end{tabular}
        }
    \end{minipage}
\end{table*}

\begin{table*}[th]
    \centering
    \caption{Prompt for ChatGPT to refine raw 3D object descriptions.}
    \label{tab:chatgpt}
    \renewcommand\arraystretch{1.2}
    \resizebox{\textwidth}{!}{
        \begin{tabular}{p{15cm}}
            \toprule[1.2pt]

            Given a description of furniture from a captioning model and its ground-truth category, please combine their information and generate a new short description in one line. The provided category must be the descriptive subject of the new description. The new description should be as short and concise as possible, encoded in ASCII. Do not describe the background and counting numbers. Do not describe size like `small', `large', etc. Do not include descriptions like `a 3D model', `a 3D image', `a 3D printed', etc. Descriptions such as color, shape and material are very important, you should include them. If the old description is already good enough, you can just copy it. If the old description is meaningless, you can just only include the category. For example: Given `a 3D image of a brown sofa with four wooden legs' and `multi-seat sofa', you should return: a brown multi-seat sofa with wooden legs. Given `a pendant lamp with six hanging balls on the white background' and `pendant lamp', you should return: a pendant lamp with hanging balls. Given `a black and brown chair with a floral pattern' and `armchair', you should return: a black and brown floral armchair. The above examples indicate that you should delete the redundant words in the old description, such as `3D image', `four', `six' and `white background', and you must include the category name as the subject in the new description. The old descriptions is `\texttt{\{BLIP caption\}}', its category is `\texttt{\{ground-truth category\}}', the new descriptions should be: \\

            \bottomrule[1.2pt]
        \end{tabular}
    }
\end{table*}

\end{document}